%% file: main.tex
\documentclass[10pt,twocolumn,letterpaper]{article}

\usepackage{cvpr}              
\usepackage{graphicx}
\usepackage{amsmath}
\usepackage{amssymb}
\usepackage{booktabs}

\usepackage{float}
\usepackage{amssymb}

\usepackage{stfloats} 

\usepackage{dsfont}
\usepackage{subcaption}
\usepackage{amsmath} 
 
\usepackage{tikz}
\usepackage{pifont}

\usepackage{caption}
\usepackage{nicefrac}
\usepackage{mathrsfs}

\usepackage{enumitem}
\usepackage{colortbl}
\usepackage{multirow}

\usepackage{marvosym}
\usepackage[export]{adjustbox}

\newcommand{\cmark}{\ding{51}}%
\newcommand{\xmark}{\ding{55}}%

\definecolor{sh_gray}{rgb}{0.84,0.84,0.84}
\definecolor{sh_gray2}{rgb}{1,0.89,0.75}
\definecolor{color3}{rgb}{0.95,0.95,0.95}
\definecolor{color4}{rgb}{0.94,0.94,1}
\definecolor{color5}{rgb}{1,0.96,0.88}

\definecolor{secondbest}{rgb}{0.8, 0.88, 1} 
\definecolor{best}{rgb}{1, 0.88, 0.88} 


\definecolor{cvprblue}{rgb}{0.21,0.49,0.74}
\usepackage[pagebackref,breaklinks,colorlinks,allcolors=cvprblue]{hyperref}
\usepackage{colortbl} 
\usepackage{xcolor} 

\definecolor{best}{rgb}{1, 0.88, 0.88} 
\definecolor{secondbest}{rgb}{0.8, 0.88, 1} 
\definecolor{best}{rgb}{1, 0.88, 0.88} 
\definecolor{secondbest}{rgb}{0.8, 0.88, 1} 

\newlength \g
\newlength\savewidth\newcommand\shline{\noalign{\global\savewidth\arrayrulewidth\global\arrayrulewidth 1pt}\hline\noalign{\global\arrayrulewidth\savewidth}}

\def\paperID{12129} 
\def\confName{CVPR}
\def\confYear{2025}

\title{TSFormer: A Robust Framework for Efficient UHD Image Restoration}

\author{
	Xin Su\\
	Fuzhou University\\
	{\tt\small suxin4726@gmail.com}
	\and
	Chen Wu\\
	University of Science and Technology of China\\
	{\tt\small wuchen5X@mail.ustc.edu.cn}
	\and
	Zhuoran Zheng\thanks{Corresponding author.}\\ 
	Sun Yat-sen University\\
	{\tt\small zhengzr@njust.edu.cn}
}

\begin{document}
\maketitle
\input{sec/0_abstract}    
\input{sec/1_intro}
\input{sec/2_formatting}
\input{sec/3_finalcopy}

\input{sec/4_Experiments_and_Analysis}
{
    \small
    
    \bibliographystyle{ieeenat_fullname}
    \bibliography{main}
}


\end{document}

%% file: sec/0_abstract.tex
\begin{abstract}
Ultra-high-definition (UHD) image restoration is vital for applications demanding exceptional visual fidelity, yet existing methods often face a trade-off between restoration quality and efficiency, limiting their practical deployment. 
In this paper, we propose \textbf{TSFormer}, an all-in-one framework that integrates \textbf{T}rusted learning with \textbf{S}parsification to boost both generalization capability and computational efficiency in UHD image restoration.
The key is that only a small amount of token movement is allowed within the model.
To efficiently filter tokens, we use Min-$p$ with random matrix theory to quantify the uncertainty of tokens, thereby improving the robustness of the model.  
Our model can run a 4K image in real time (40fps) with 3.38 M parameters.
Extensive experiments demonstrate that TSFormer achieves state-of-the-art restoration quality while enhancing generalization and reducing computational demands. 
In addition, our token filtering method can be applied to other image restoration models to effectively accelerate inference and maintain performance.
\end{abstract}

%% file: sec/1_intro.tex
\section{Introduction}

%
Ultra-high-definition (UHD) image restoration is essential for various applications that require high-resolution image quality, including medical imaging, video streaming, and digital surveillance \cite{wang2024ultra, yang2024advanced, liu2020high, chen2021ultrahd}.
Since UHD images have millions of pixels, processing them with limited resources can be a huge challenge.

Currently, there are some deep learning frameworks~\cite{Wave-Mamba, OUR-GAN} that can process UHD images on consumer-grade GPUs. However, these methods directly or indirectly downsample the input image, and may lose some tokens that are important for image restoration. Therefore, this results in a loss of detail in the image, which is particularly important for UHD images.
\begin{figure}[!t]
\begin{center}
\scalebox{0.75}{ %
\begin{tabular}[b]{c@{ } c@{ } c@{ } c@{ }}
      
    \includegraphics[width=3.5cm, valign=t]{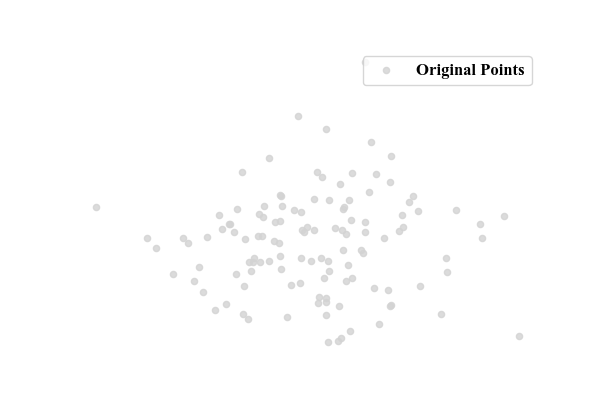}&
    \includegraphics[width=3.5cm, valign=t]{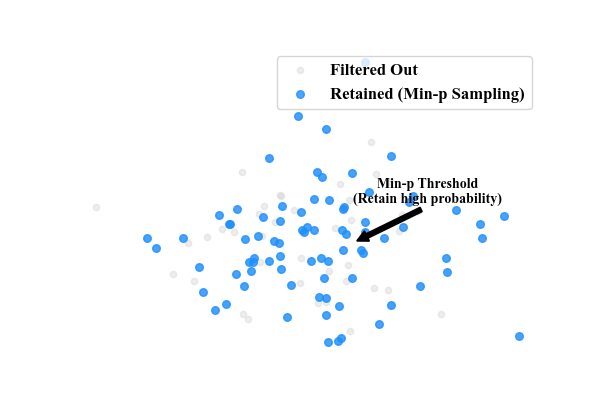} &
    \includegraphics[width=3.5cm, valign=t]{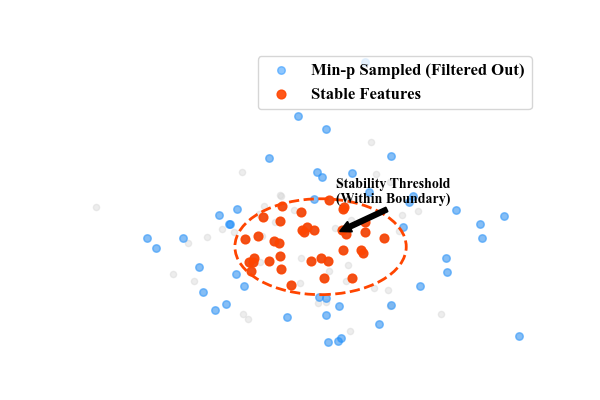}  \\
    
    \includegraphics[width=3.5cm,height= 2.35cm, valign=t]{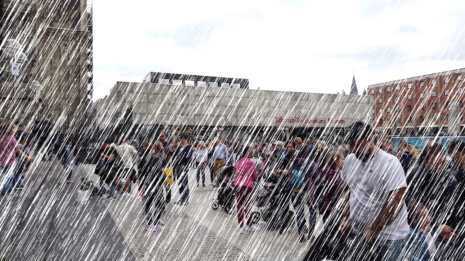}&
    \includegraphics[width=3.5cm, valign=t]{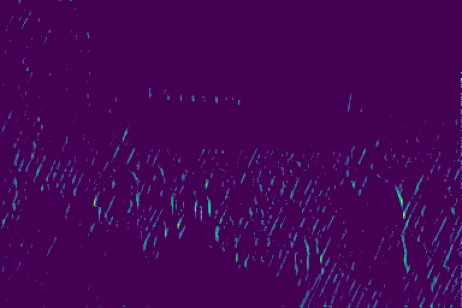} &
    \includegraphics[width=3.5cm, valign=t]{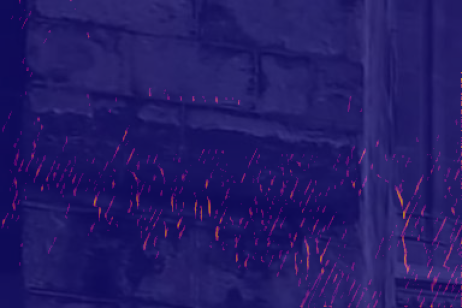}  \\
    
     \small~(a) Input Features & \small~(b) After Min-$p$ Sampling & \small~(c) After Stability Filtering  \\
\end{tabular}}
\end{center}
\vspace*{-6mm}
\caption{\textbf{Illustration of the Min-$p$ sampling and stability filtering process.}(a) The original input image used as a reference. (b) Result after Min-$p$ sampling, where high-probability regions are retained and highlighted in a distinct color, indicating sparsified yet significant areas. (c) Result after stability filtering, where only the stable, high-confidence features are preserved, with stable regions marked in red and unstable regions suppressed.}\vspace{-6mm}
\label{fig:f1}
\end{figure}
To address this problem, we propose \textbf{TSFormer}, a lightweight and trusted framework that combines sparsification with random matrix theory. 
TSFormer is designed to retain only the most informative features, thereby significantly reducing computational overhead while maintaining high restoration quality and robustness.

Indeed, the first key component of TSFormer is \textbf{Min-$p$ sampling}, a probability-driven sparsification technique inspired by recent advances in probabilistic sparsification theory \cite{bellec2023estimating}. 
Unlike conventional Top-$k$ filtering methods \cite{liu2021swin}, Min-$p$ sampling selectively retains high-confidence features based on a probabilistic threshold, allowing dynamic feature selection. 
This ``less is more`` approach recognizes that not all features contribute equally to the final output, especially in high-dimensional data like UHD images. 
%
However, while Min-$p$ sampling effectively reduces the model's computational demands, it may introduce instability due to noise in the large number of tokens. To address this issue, TSFormer incorporates a \textbf{trusted} mechanism grounded in random matrix theory \cite{edelman1988eigenvalues, tao2012topics}. This trusted filtering involves analyzing the eigenvalues of feature matrices to ensure that only robust, high-confidence features are retained. As shown in Figure \ref{fig:f1}(a), Min-$p$ sampling applies a probability-based threshold to retain high-confidence features while discarding those of lower importance, achieving effective sparsification. In Figure \ref{fig:f1}(b), trusted filtering based on random matrix theory refines the remaining features further by excluding points outside a trusted threshold (dashed circle). 
This trusted-driven feature selection improves generalization, allowing the model to perform reliably across varied degraded images \cite{huang2023robust, yang2024uncertainty}.
So far, we have constructed a Min-$p$ Sparse Attention (MSA) by enforcing a Min-$p$ with a trusted mechanism, which efficiently and robustly generates an attention map.
MSA is integrated into each block of the TSFormer, and each block also contains frequency domain learning and multi-scale learning components.
TSFormer shows promising performance in multiple UHD image tasks and can run a UHD image in real time on resource-constrained devices (a single 3090 GPU shader with 24G RAM).
In summary, our contributions are as follows:
\begin{itemize}
    \item 
    We develop a novel UHD image restoration model (TSFormer) that can run a 4K resolution image in real-time on a single GPU with strong generalization capabilities.
    \item We design a token filtering method with a trusted mechanism, which is integrated into the TSFormer to generate high-quality attention maps.
    \item The token filtering method with a trusted mechanism can be used in any Transformer-based image restoration framework to improve the efficiency of the model. Extensive experimental results show the effectiveness of our method.
\end{itemize}
\vspace{-4mm}

%% file: sec/2_formatting.tex
\section{Related Work}

\subsection{UHD Image Restoration}
Ultra-high-definition (UHD) image restoration is essential for applications in medical imaging, video streaming, and digital surveillance \cite{wang2024ultra, yang2024advanced}.
%
%
%
Recently, deep learning methods have efficiently reconstructed UHD images' details and colors through sampling and parallelization techniques. 
Zheng et al.~\cite{zheng2021multi} introduced a multi-guided bilateral upsampling model for UHD image dehazing, enhancing clarity through multiple guidance inputs. Deng et al.~\cite{deng2021separable} developed a separable-patch integration network for UHD video deblurring, employing a multi-scale integration scheme to mitigate motion and blur artifacts. Wang et al.~\cite{wang2023llformer} proposed LLFormer, a transformer-based method for low-light enhancement utilizing axis-based multi-head self-attention and cross-layer attention fusion blocks to improve illumination and contrast. 
%
In addition, there are some methods~\cite{OUR-GAN,Wave-Mamba} to reconstruct clear UHD images in real time by building lightweight models and looking up tables.

Although these methods can enhance a UHD image in real time, the sampling and table lookup methods are not supervised by a trusted mechanism, which limits the generalisation ability of the model.
In contrast, our proposed TSFormer incorporates Min-$p$ sampling for adaptive sparsification and employs trusted filtering based on random matrix theory~\cite{bun2017random, hachem2007analysis}, enhancing both feature reliability and restoration quality while maintaining low computational costs.

\subsection{Token Sampling Technology}
Currently, large language models (LLM) use some token sampling techniques to speed up inference.
Traditional methods like Top-$k$ filtering \cite{liu2021swin, fan2018hierarchical} prioritize features based on magnitude but use fixed thresholds that may not adapt to varying data distributions, potentially discarding valuable information.
Probabilistic sparsification methods, such as Min-$p$ sampling \cite{bellec2023estimating, zhou2020self}, introduce dynamic, probability-based thresholds that better adapt to data distributions, allowing flexible feature retention. 
Inspired by this, we introduced the token technology to build an efficient model. On this basis, while ensuring real-time performance, a trusted mechanism (random matrix theory) is introduced to accurately sample tokens~\cite{couillet2018random}.

\subsection{Random Matrix Theory}
Trusted filtering of tokens is difficult, and it can lead to a significant slowdown in the speed of model inference.
Compared to other approaches to trusted modeling such as Bayesian, variational inference, and labeled distributions, random matrix theory trades off the speed and accuracy of inference.
%
Random matrix theory (RMT) offers a framework for analyzing and enhancing feature stability in high-dimensional data \cite{edelman1988eigenvalues, tao2012topics}. By examining eigenvalue distributions, RMT-based methods can identify and retain the most stable and salient features, improving robustness and generalization \cite{huang2023robust, yang2024uncertainty}. However, integrating RMT into deep learning for UHD restoration remains underexplored, presenting an opportunity to enhance feature reliability without significantly increasing computational costs.

%% file: sec/3_finalcopy.tex
\section{Method}

\begin{figure*}
    \centering
    \includegraphics[width=1\linewidth]{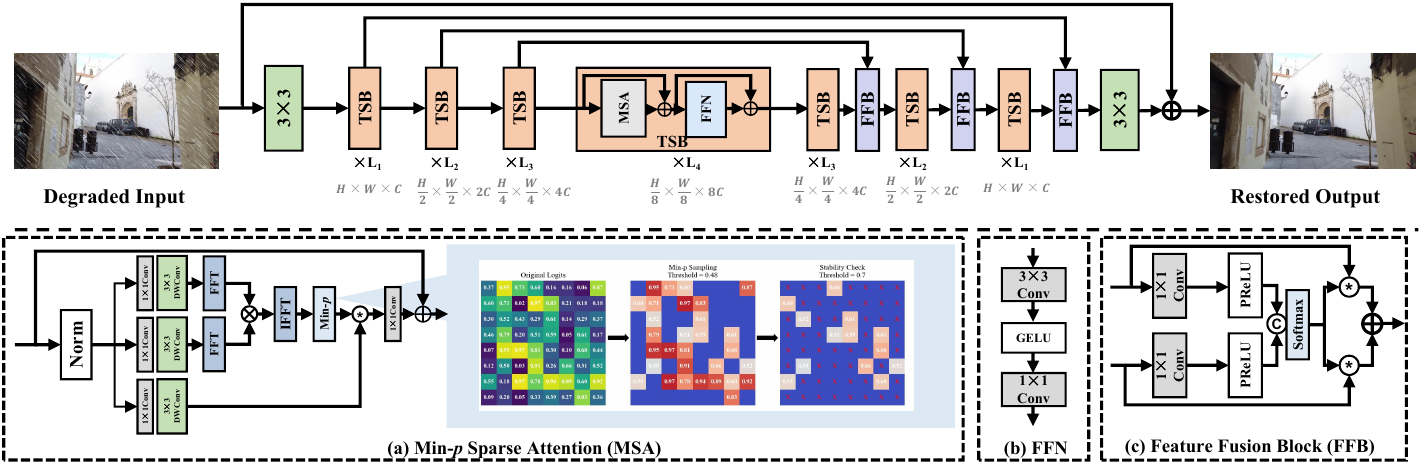}
    \caption{The overall architecture of the proposed TSFormer for UHD image restoration, which main consists of Trusted Sparse Blocks (TSB) and a Feed-forward Network (FFN). 
    Our core is Min-p Sapre Attention (MSA), which incorporates the ability of dynamic token filtering and sparse representation, effectively reducing the burden of running a UHD image. In addition, MSA is also a plug-and-play algorithm that can be used in any image restoration network based on the Transformer architecture.}
    \label{fig: model}
    \vspace{-4mm}
\end{figure*}

In this section, we introduce TSFormer, an efficient Transformer network for UHD image restoration. 
It is a symmetrical encoder and decoder structure, as shown in Figure~\ref{fig: model}.
%

\subsection{Preliminaries}

\noindent\textbf{Min-$p$ Sampling.} Min-$p$ sampling is a novel sampling method originally designed for large language model tasks, where it dynamically adjusts the sampling threshold to retain high-confidence tokens \cite{nguyen2024minp}. 
%
%
Specifically, given a set of logits $\mathbf{L}$ within a patch, Min-$p$ sampling applies a threshold defined as a proportion $p_\text{base}$ of the maximum value in the feature patch:
\begin{equation}
    \text{Threshold} = p_\text{base} \cdot \max(\mathbf{L}),
\end{equation}
where $p_\text{base} \in [0,1]$ is a hyperparameter that controls sparsity. This threshold filters out low-confidence features by setting elements in $\mathbf{L}$ below the threshold to zero:
\begin{equation}
    \mathbf{L}_{\text{sampled}} = 
    \begin{cases}
        \mathbf{L}, & \text{if} \quad \mathbf{L} \geq \text{Threshold}, \\
        0, & \text{otherwise}.
    \end{cases}
\end{equation}
This method acts on the sparsification of the attention map to filter out unimportant features and noise.


\noindent\textbf{Fast Fourier Transform.} To further optimize the computational efficiency of our attention mechanism, we employ Fast Fourier Transform (FFT) for attention computation in the frequency domain. FFT significantly reduces the complexity of calculating attention scores, particularly for high-resolution images.

The forward Fourier transform of a 1D signal \( x(t) \) is given by:
\begin{equation}
X(f) = \int_{-\infty}^{\infty} x(t) e^{-i 2 \pi f t} \, dt
\end{equation}

The inverse Fourier transform is given by:
\begin{equation}
x(t) = \int_{-\infty}^{\infty} X(f) e^{i 2 \pi f t} \, df
\end{equation}

By leveraging the FFT, we can perform operations such as attention more efficiently, especially when dealing with high-resolution images, by processing data in the frequency domain.

\subsection{Overall Pipeline}
%

Given a degraded image $\mathbf{I}_{\text{degraded}} \in \mathbb{R}^{H \times W \times 3}$, where $H \times W$ represents the spatial resolution, it starts by being tokenized by a $3 \times 3$ convolution. 
Then, this tokenized feature map is fed into a encode-decode structure.
%
%

In the Encoder, we progressively down-sample the feature map using a sequence of Trusted Sparse Blocks (TSBs) across multiple hierarchical levels, with each level having a specific number of TSBs, denoted by $N_i \in \{1, 2, 2, 4\}$. 
Each TSB integrates frequency domain transformation and Min-$p$ Sparse Attention (MSA).
Each TSB also contains a Feed-Forward Network (FFN) with a series of depthwise ($3 \times 3$) and pointwise ($1 \times 1$) convolutions to enrich feature representation across multiple scales.
%


A decoder is a mirrored operation of a decoder to reconstruct the details and colours of a high-resolution image.
Feature Fusion Blocks (FFBs) are introduced between corresponding encoder and decoder levels to merge features from different resolutions. Each FFB applies a lightweight convolutional block with PReLU activation, facilitating smooth transition and refinement of features across scales, ultimately improving model performance.

Finally, we employ a residual connection around the network to obtain the restored image as:
\begin{equation}
    \mathbf{I}_{\text{restored}} = \text{F}(\mathbf{I}_{\text{degraded}}) + \mathbf{I}_{\text{degraded}},
\end{equation}
where $\text{F}(\cdot)$ denotes the network's transformation. The model is trained by minimizing the $L_1$ loss between the restored output and the ground-truth image $\mathbf{I}_{\text{gt}}$:
\begin{equation}
    \mathcal{L} = \|\mathbf{I}_{\text{restored}} - \mathbf{I}_{\text{gt}}\|_1,
\end{equation}
where $\|\cdot\|_1$ represents the $L_1$-norm.


\subsection{Trusted Sparse Block (TSB)}

Self-attention mechanisms are not only computationally expensive, but also prone to noise, especially in high-frequency, detail-rich UHD images. 
To address this, we design a Trusted Sparse Block (TSB) as a feature extraction unit, integrating Min-$p$ Sampling and trusted learning. 

Specifically, given input features $\mathbf{X}^{(l-1)}$ from the $(l-1)$-th block, the encoding procedure in TSB can be defined as follows:
\begin{align}
    \mathbf{X}'^{(l)} &= \mathbf{X}^{(l-1)} + \text{MSA}(\text{LN}(\mathbf{X}^{(l-1)})), \\
    \mathbf{X}^{(l)} &= \mathbf{X}'^{(l)} + \text{FFN}(\text{LN}(\mathbf{X}'^{(l)})),
\end{align}
where $\text{LN}$ denotes layer normalization, $\text{MSA}$ represents the Min-$p$ Sparse Attention mechanism, and $\text{FFN}$ is the Feed-Forward Network. Here, $\mathbf{X}'^{(l)}$ and $\mathbf{X}^{(l)}$ denote the outputs of the attention and feed-forward layers, respectively.

\noindent \textbf{Min-$p$ Sparse Attention (MSA).}
Our MSA leverages Min-$p$ Sampling to dynamically retain only high-probability features, using Fourier transforms to efficiently compute attention in the frequency domain.

Given an input feature map $\mathbf{X} \in \mathbb{R}^{B \times C \times H \times W}$, where $B$, $C$, $H$, and $W$ denote the batch, channel, height, and width dimensions, respectively, we first obtain query $\mathbf{Q}$, key $\mathbf{K}$, and value $\mathbf{V}$ representations through convolutional transformations:
\begin{equation}
    \mathbf{Q}, \mathbf{K}, \mathbf{V} = \text{Conv}_{3 \times 3}(\mathbf{X}).
\end{equation}
To capture local feature interactions, we divide $\mathbf{Q}$ and $\mathbf{K}$ into patches and apply Fast Fourier Transform (FFT) to convert each patch to the frequency domain:
\begin{equation}
    \mathbf{Q}_{\text{FFT}} = \text{FFT}(\mathbf{Q}_{\text{patch}}), \quad \mathbf{K}_{\text{FFT}} = \text{FFT}(\mathbf{K}_{\text{patch}}).
\end{equation}
In the frequency domain, the attention scores are computed as element-wise multiplication of $\mathbf{Q}_{\text{FFT}}$ and $\mathbf{K}_{\text{FFT}}$, which significantly reduces computational complexity compared to spatial domain operations. The inverse FFT (IFFT) is then applied to return the result to the spatial domain:
\begin{equation}
    \mathbf{M} = \text{IFFT}(\mathbf{Q}_{\text{FFT}} \cdot \mathbf{K}_{\text{FFT}}^\top).
\end{equation}
Next, Min-$p$ Sampling is applied to $\mathbf{M}$ to filter out low-confidence attention scores. A threshold is defined as a proportion $p_{\text{base}}$ of the maximum value in each row:
\begin{equation}
    \text{Threshold} = p_{\text{base}} \cdot \max(\mathbf{M}).
\end{equation}
The Min-$p$ operation retains only elements in $\mathbf{M}$ that exceed this threshold, sparsifying the attention map by setting lower scores to zero:
\begin{equation}
    [\text{Min-$p$}(\mathbf{M})]_{ij} = 
    \begin{cases}
        \mathbf{M}_{ij}, & \text{if } \mathbf{M}_{ij} \geq \text{Threshold}, \\
        0, & \text{otherwise}.
    \end{cases}
\end{equation}

\noindent \textbf{Trusted Learning with Random Matrix Theory.}
To establish a trusted element filtering method for the $\mathbf{M}$, we treat the $\mathbf{M}$ as a random matrix for research.
First, we compute the spectral density $\rho$ of the random matrix $\mathbf{M}$. Then, the spectral density $\rho$ is used to generate a probability weight scalar $\hat{\rho}$ through a mean pooling layer and a sigmoid function. Finally, $\hat{\rho}$ is multiplied by Threshold to dynamically adjust Threshold.
This can be written as follows:
\begin{equation}
    \hat{\text{Threshold}} = \hat{\rho} \cdot \text{Threshold}.
\end{equation}
If $\mathbf{M}$ is more random, it may contain more noise and therefore requires a higher threshold $\hat{\text{Threshold}}$.
Note that the trusted mechanism can lead to more computing power overhead. To alleviate this problem, $\mathbf{M}$ is downsampled (bilinear interpolation) before calculating the spectral density.
%
%
%
%
The final sparse attention output is calculated as:
\begin{equation}
    \text{MSA}(\mathbf{Q}, \mathbf{K}, \mathbf{V}) = \text{Min-$p$}(\mathbf{M}) \cdot \mathbf{V}.
\end{equation}

\noindent \textbf{Feed-Forward Network (FFN).}
Following the sparse attention mechanism, the Feed-Forward Network (FFN) refines the features. This module consists of a series of convolutions to capture and enhance multi-scale information. For the input $\mathbf{X}'^{(l)}$, FFN is defined as:
\begin{equation}
    \text{FFN}(\mathbf{X}'^{(l)}) = \sigma(\text{Conv}_{3\times3}(\mathbf{X}'^{(l)})) \cdot \text{Conv}_{1\times1}(\mathbf{X}'^{(l)}),
\end{equation}
where $\sigma$ is the GELU activation, $\text{Conv}_{3\times3}$ represents a depthwise convolution, and $\text{Conv}_{1\times1}$ is a pointwise convolution. This gating mechanism selectively refines feature representations, ensuring that only the most informative elements are propagated.




\subsection{Feature Fusion Blocks (FFB)}

Feature Fusion Blocks (FFB) are designed to integrate multi-scale features from different levels of the encoder and decoder. By dynamically adjusting the contribution of each feature map, FFB enables the model to capture both fine-grained details and global context effectively.

Given two input feature maps, $\mathbf{X}$ and $\mathbf{Y}$, from different stages of the network, FFB performs the following operations:
\begin{equation}
   \mathbf{X}_f = \text{PReLU}(\text{Conv}_{1 \times 1}(\mathbf{X})),  
\end{equation}
\begin{equation}
    \mathbf{Y}_f = \text{PReLU}(\text{Conv}_{1 \times 1}(\mathbf{Y})),
\end{equation}
\begin{equation}
    \mathbf{M_{att}} = \text{softmax}(\text{Conv}_{1 \times 1}([\mathbf{X}_f, \mathbf{Y}_f])),
\end{equation}
where $[\mathbf{X}_f, \mathbf{Y}_f]$ denotes the concatenation of $\mathbf{X}_f$ and $\mathbf{Y}_f$.
\begin{equation}
\mathbf{F}_{\text{fused}} = \mathbf{M_{att}}_{\mathbf{X}} \cdot \mathbf{X} + \mathbf{M_{att}}_{\mathbf{Y}} \cdot \mathbf{Y},
\end{equation}
where $\mathbf{M_{att}}_{\mathbf{X}}$ and $\mathbf{M_{att}}_{\mathbf{Y}}$ represent the channel-wise weights applied to $\mathbf{X}$ and $\mathbf{Y}$, respectively. This selective fusion allows FFB to prioritize information from either the encoder or decoder based on the spatial and contextual demands of the task.

%% file: sec/4_Experiments_and_Analysis.tex
\section{Experiments and Analysis}
\label{sec:experiments}
We show performance comparisons with state-of-the-art approaches on $5$ UHD image restoration tasks, including low-light enhancement, dehazing, deblurring, desnowing, and deraining.
\subsection{Experimental settings.}
\noindent \textbf{Datasets.}
For UHD low-light image enhancement, we employ the UHD-LL dataset \cite{li2023embedding} and UHD-LOL4K \cite{wang2023ultra}. To evaluate deblurring capabilities, we use the UHD-Blur dataset \cite{wang2024correlation}. For dehazing evaluations, we adopt the UHD-Haze dataset \cite{wang2024correlation}. These selections align with methodologies employed in prior research \cite{li2023embedding, wang2024correlation}. Additionally, to evaluate our UHD image desnowing and deraining ability, we utilize the UHD-Snow and UHD-Rain datasets introduced by Wang et al. \cite{wang2024ultra}.
%
We adopt PSNR \cite{huynh2008scope} and SSIM \cite{wang2004image} as the evaluation metrics for the above benchmarks.
 
\noindent \textbf{Compared Methods.} In our study, we benchmark our method against eight general image restoration (IR) techniques: SwinIR \cite{liang2021swinir}, Uformer \cite{wang2022uformer}, Restormer \cite{zamir2022restormer}, DehazeFormer \cite{song2023dehazeforemer}, Stripformer \cite{tsai2022stripformer}, FFTformer \cite{kong2023fftformer}, and SFNet \cite{wang2024sfnet}. Additionally, we include four ultra-high-definition image restoration (UHDIR) methods: LLFormer \cite{wang2023llformer}, UHD-Four \cite{li2023uhdfour}, UHD \cite{zheng2021uhd}, and UHDformer \cite{wang2024uhdformer}, and UHDDIP~\cite{wang2024uhdrestoration}. For a fair comparison, we retrain these models using their official implementations and evaluate them with the same number of iterations as our proposed method.

\noindent \textbf{Training details.} In our model, the initial channel $\mathbf{C}$ is $32$ and the expanding ratio is set to $2$, and the channel expansion factor r in FFN is set to $2.0$. During training, we use AdamW optimizer with batch size of $6$ and patch size of $512$ for a total of $300,000$ iterations. The initial learning rate is fixed as $2\times10^{-4}$. For data augmentation, vertical and horizontal flips are randomly applied. The entire framework is performed on the PyTorch with 2 NVIDIA RTX 3090 GPUs. 
\subsection{Main Results}
\begin{figure*}[!t]
    \centering
    \begin{adjustbox}{max width=\textwidth}
    \begin{tabular}{cccccccccc}
        \includegraphics[width=0.153\textwidth]{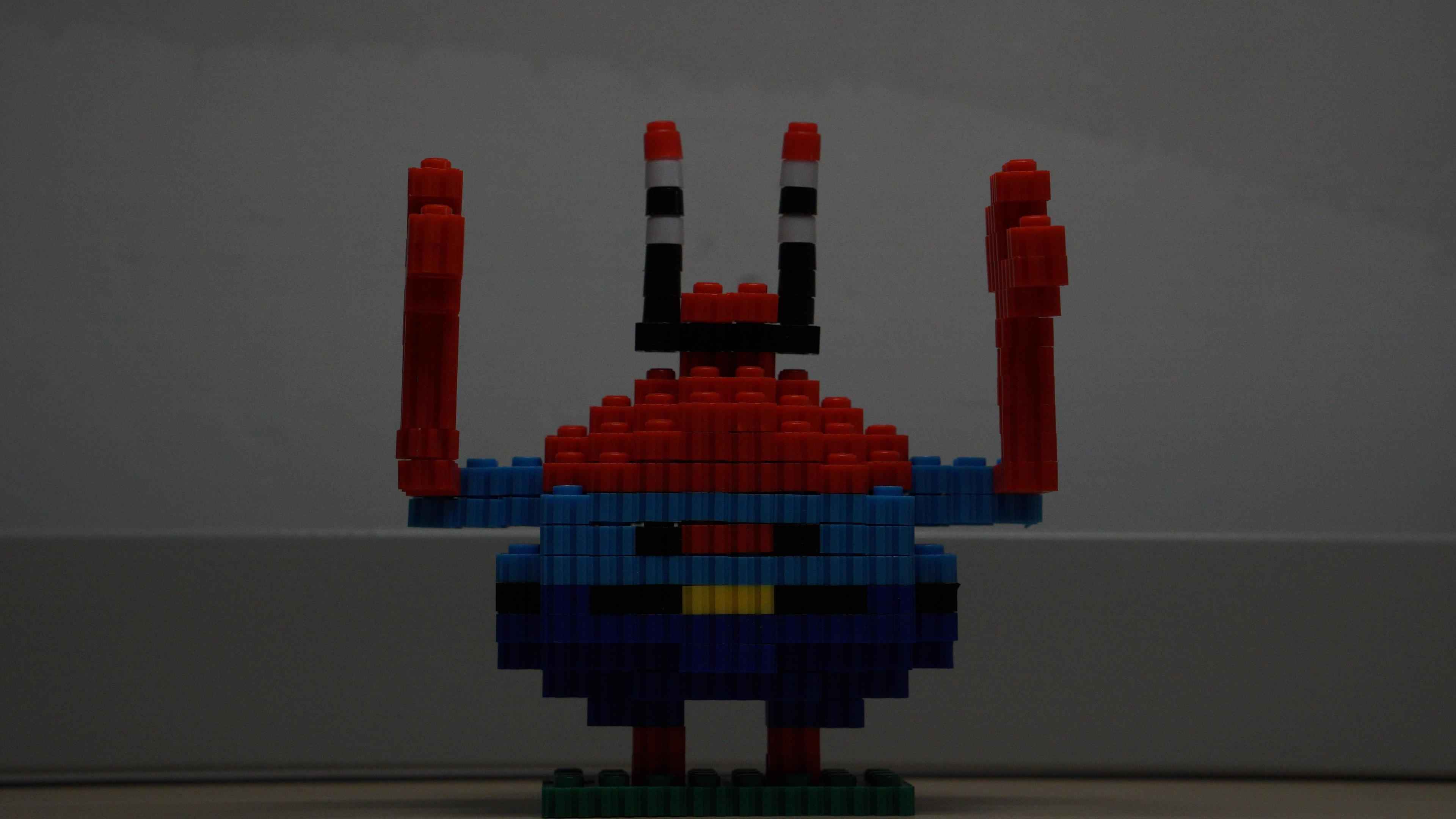} &
        \includegraphics[width=0.153\textwidth]{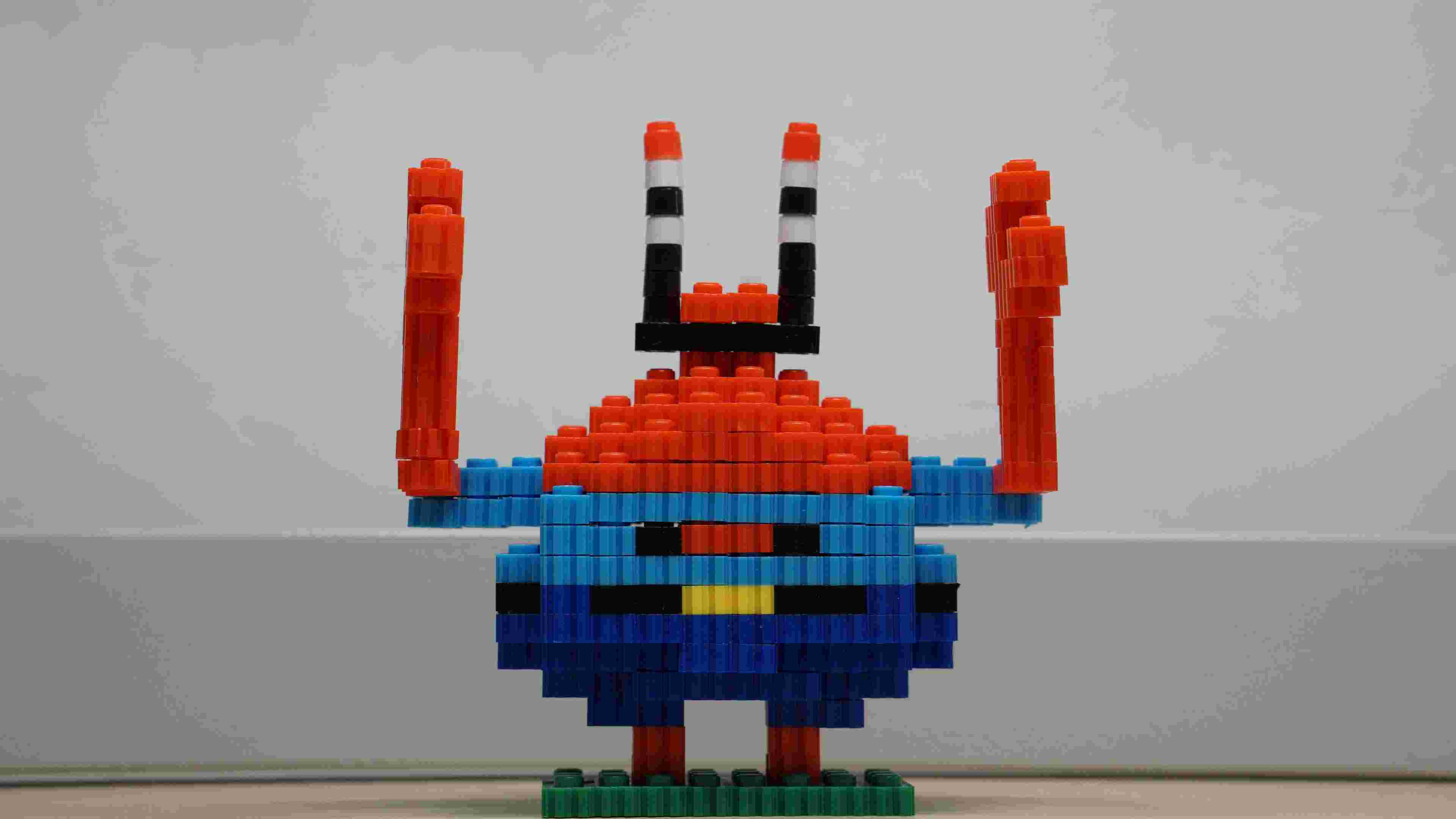} &
        \includegraphics[width=0.153\textwidth]{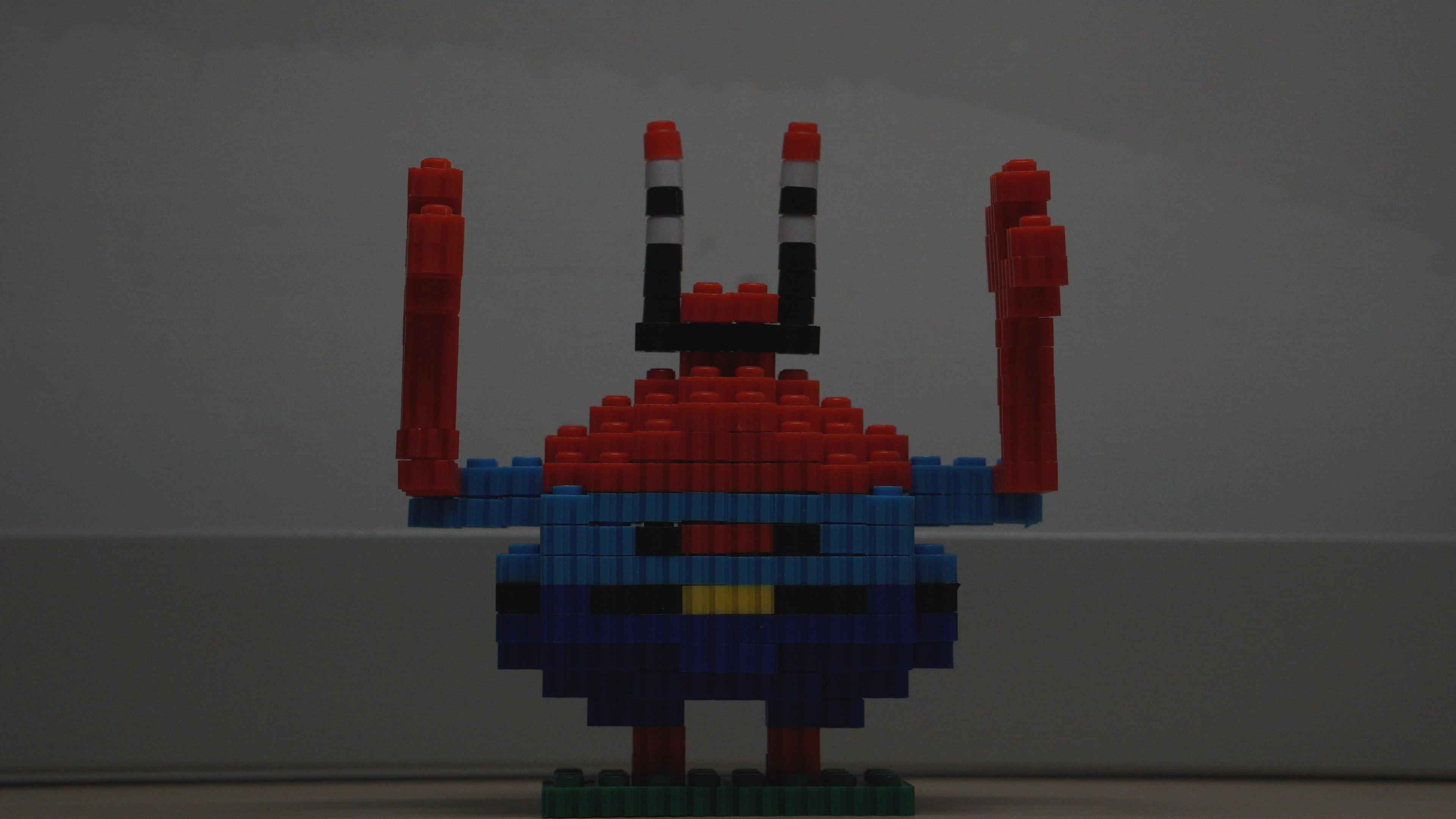} &
        \includegraphics[width=0.153\textwidth]{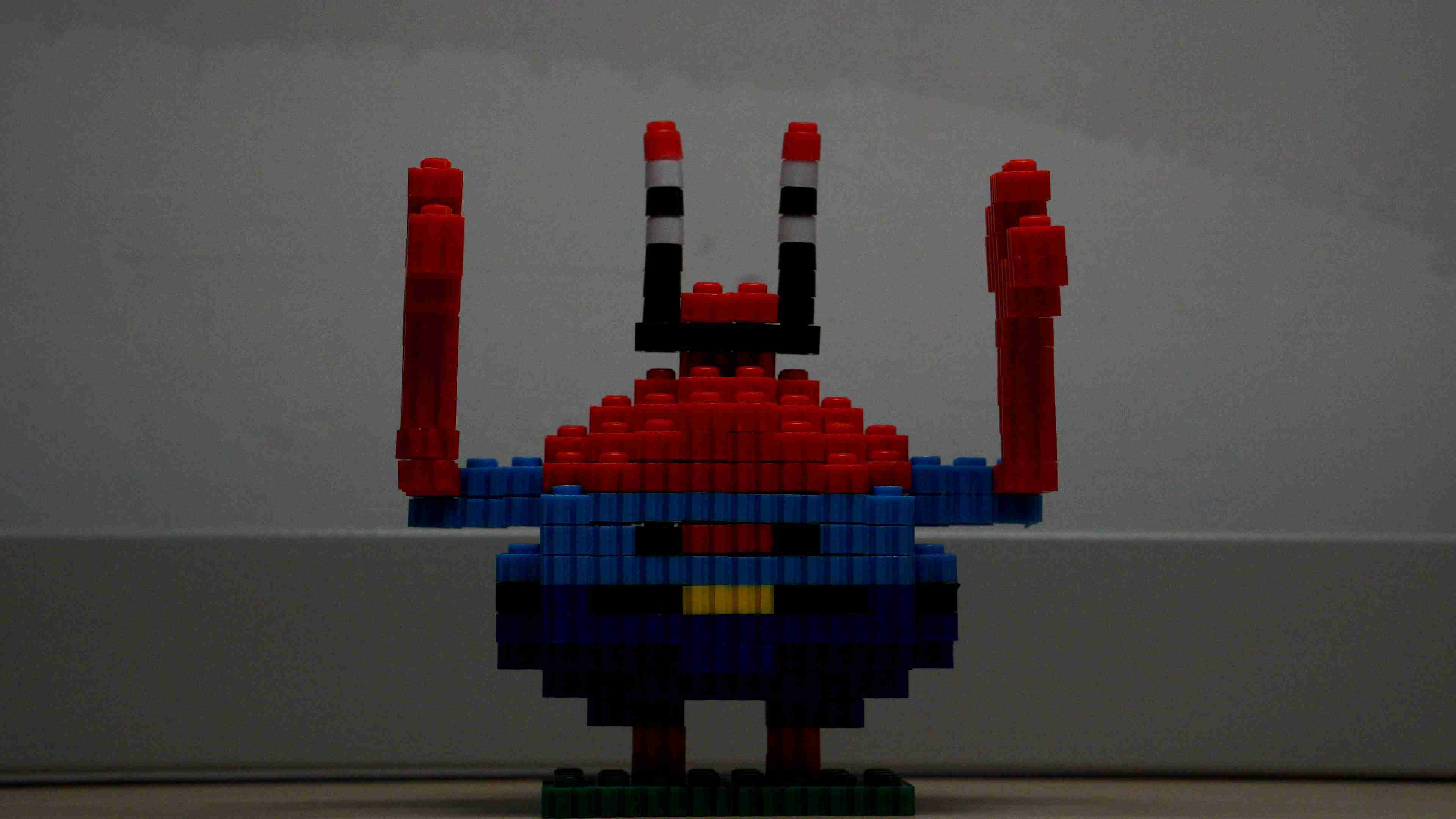} &
        \includegraphics[width=0.153\textwidth]{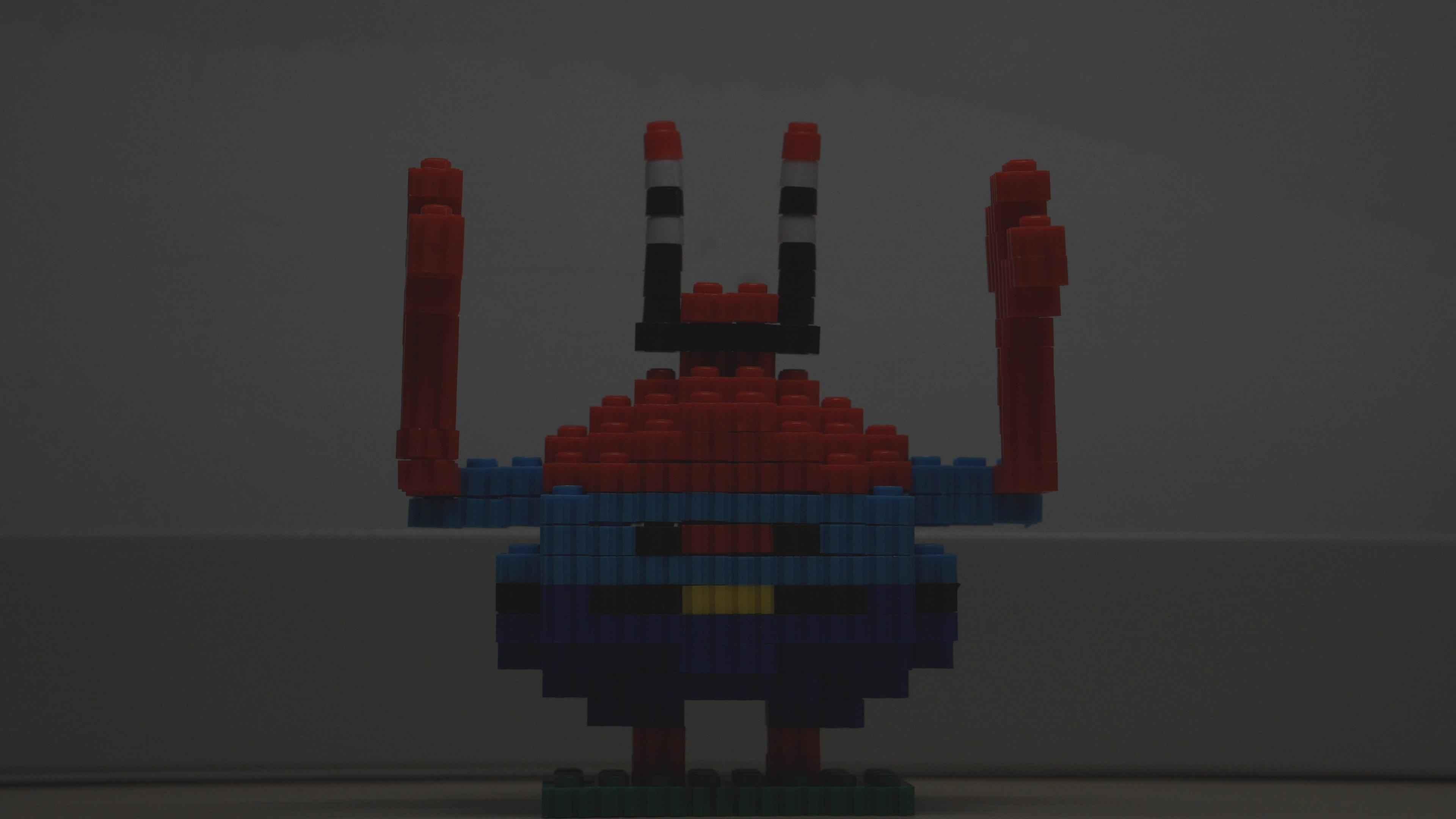} &
        \includegraphics[width=0.153\textwidth]{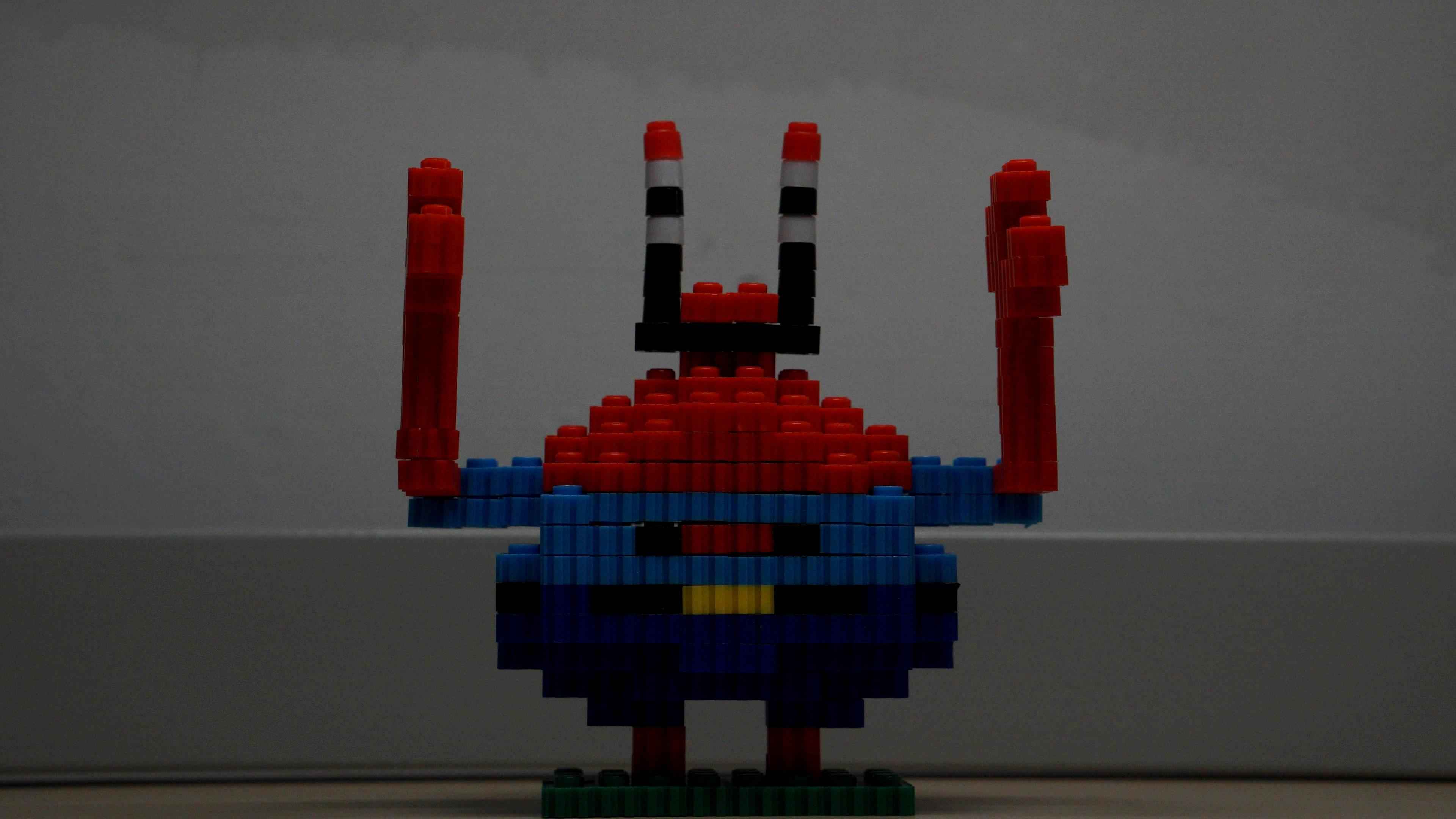} &
        \includegraphics[width=0.153\textwidth]{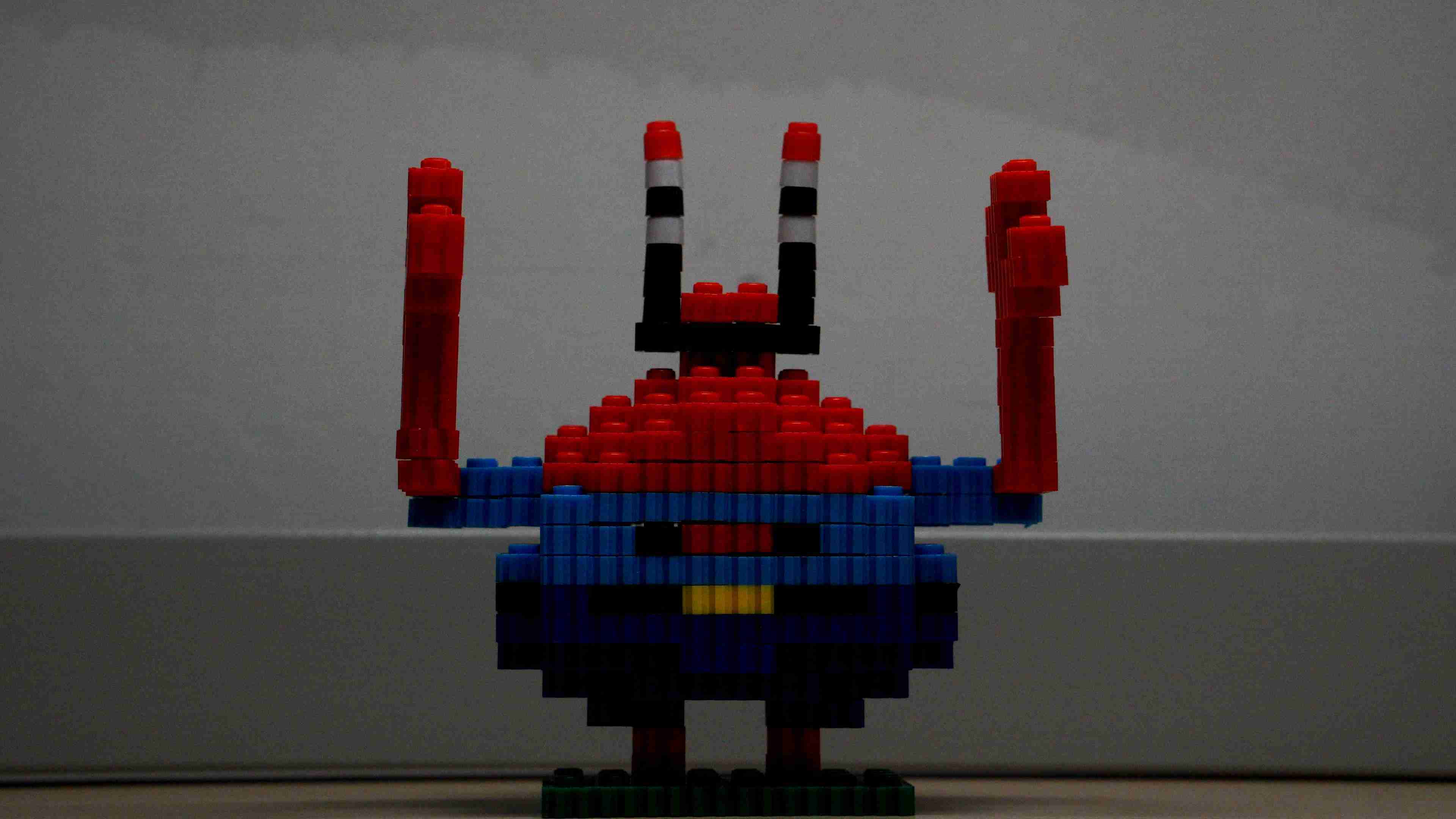} &
        \includegraphics[width=0.153\textwidth]{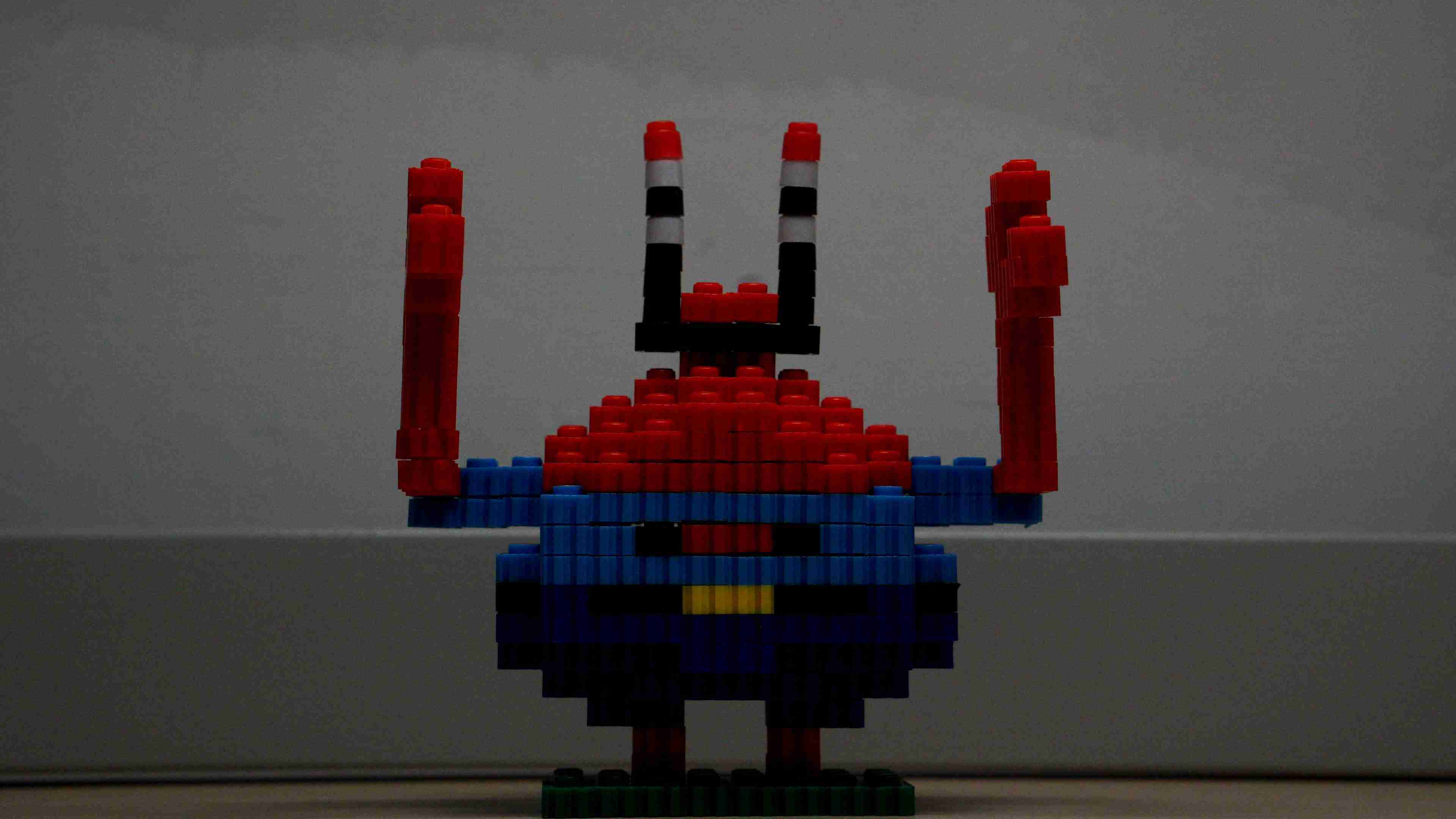} &
        \includegraphics[width=0.153\textwidth]{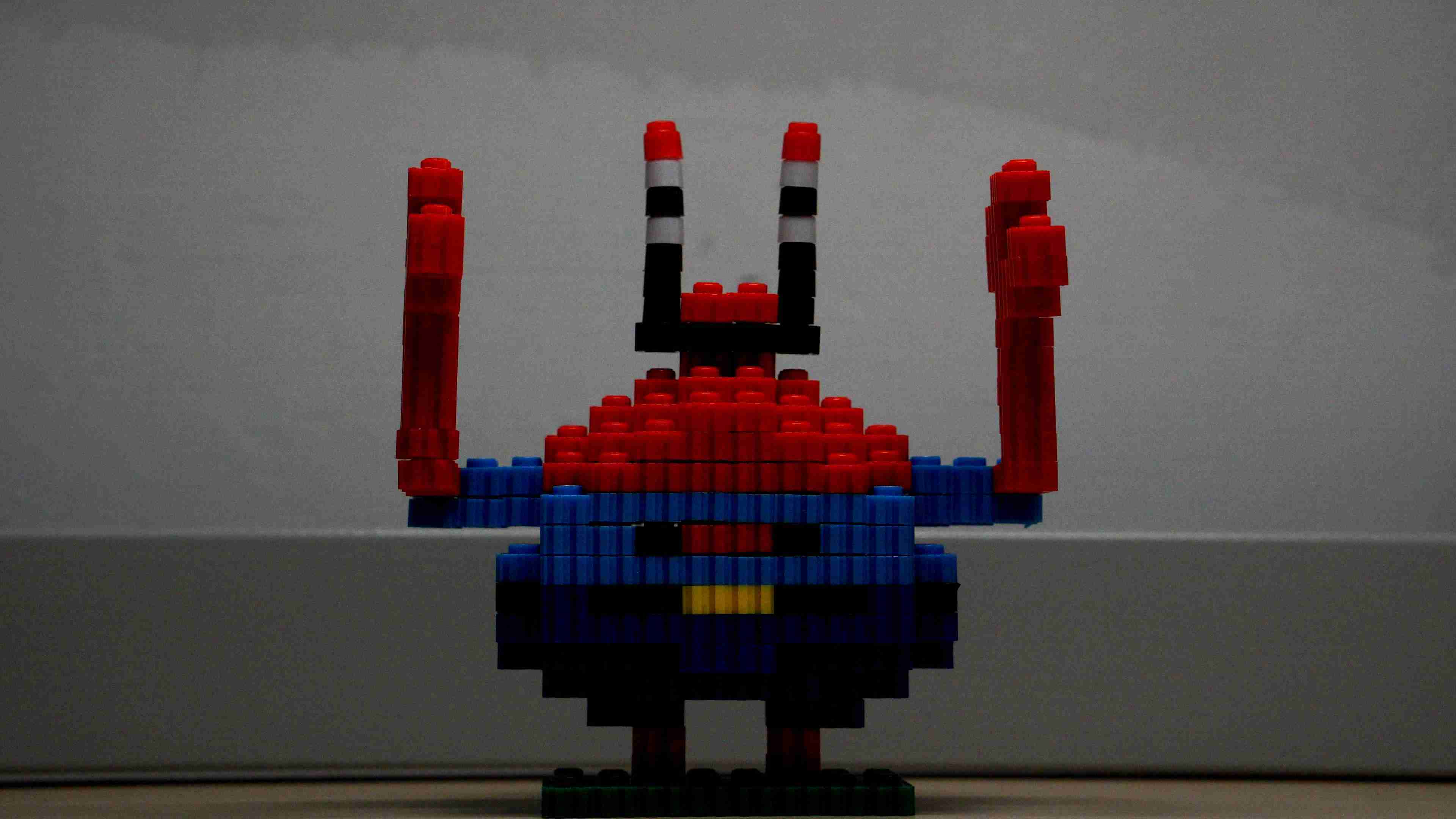} &
        \includegraphics[width=0.153\textwidth]{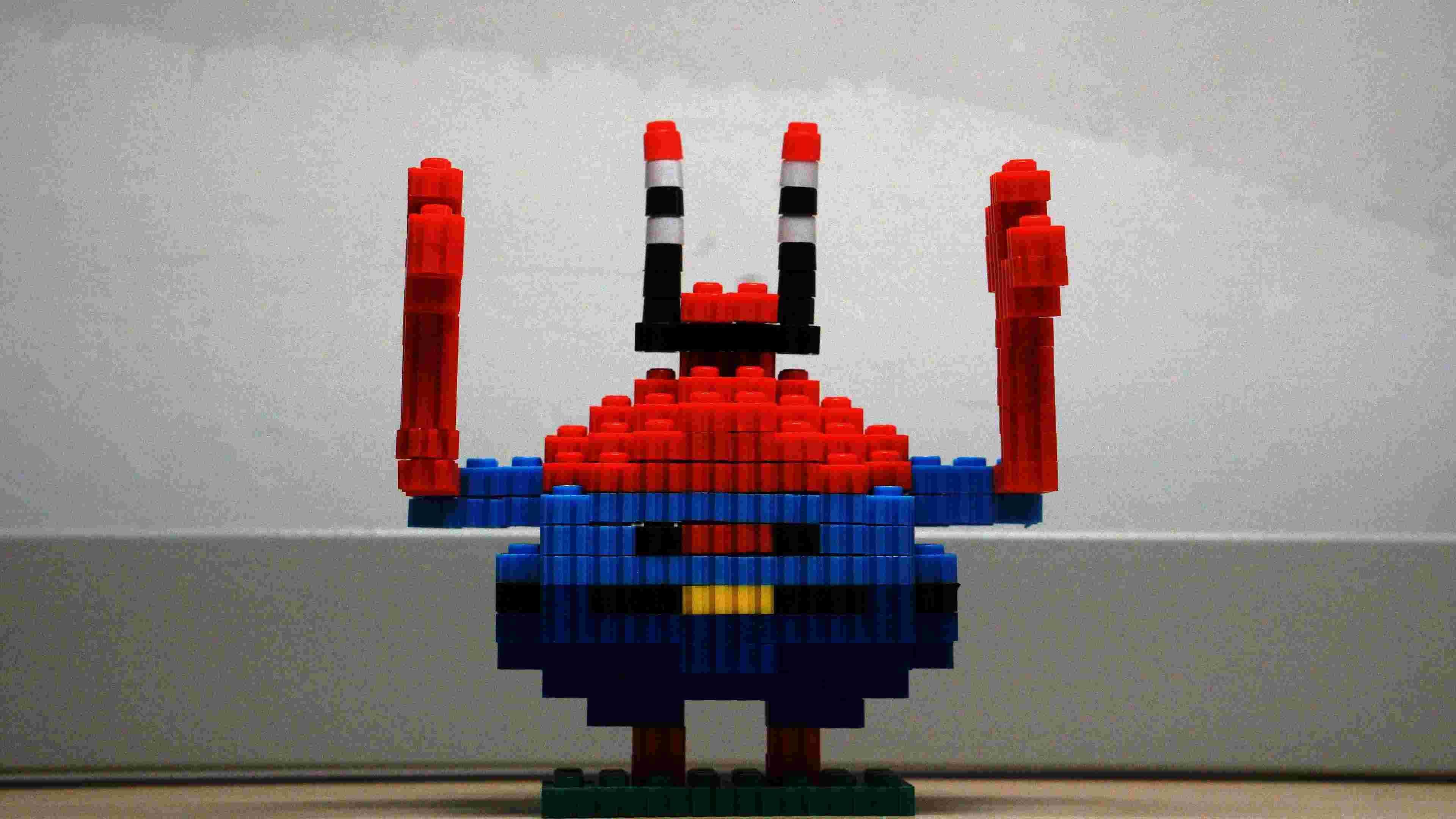} \\
        \includegraphics[width=0.153\textwidth]{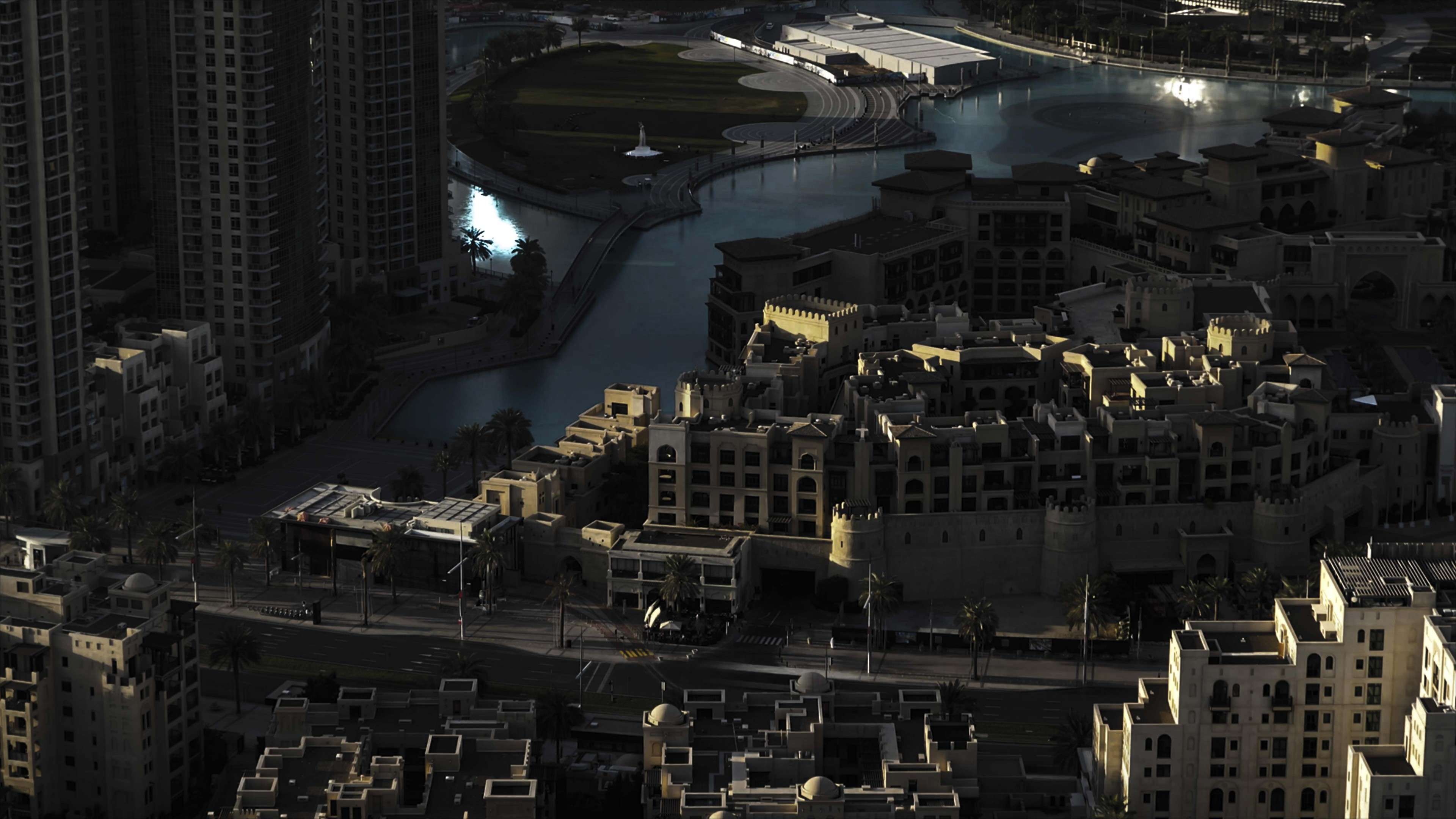} &
        \includegraphics[width=0.153\textwidth]{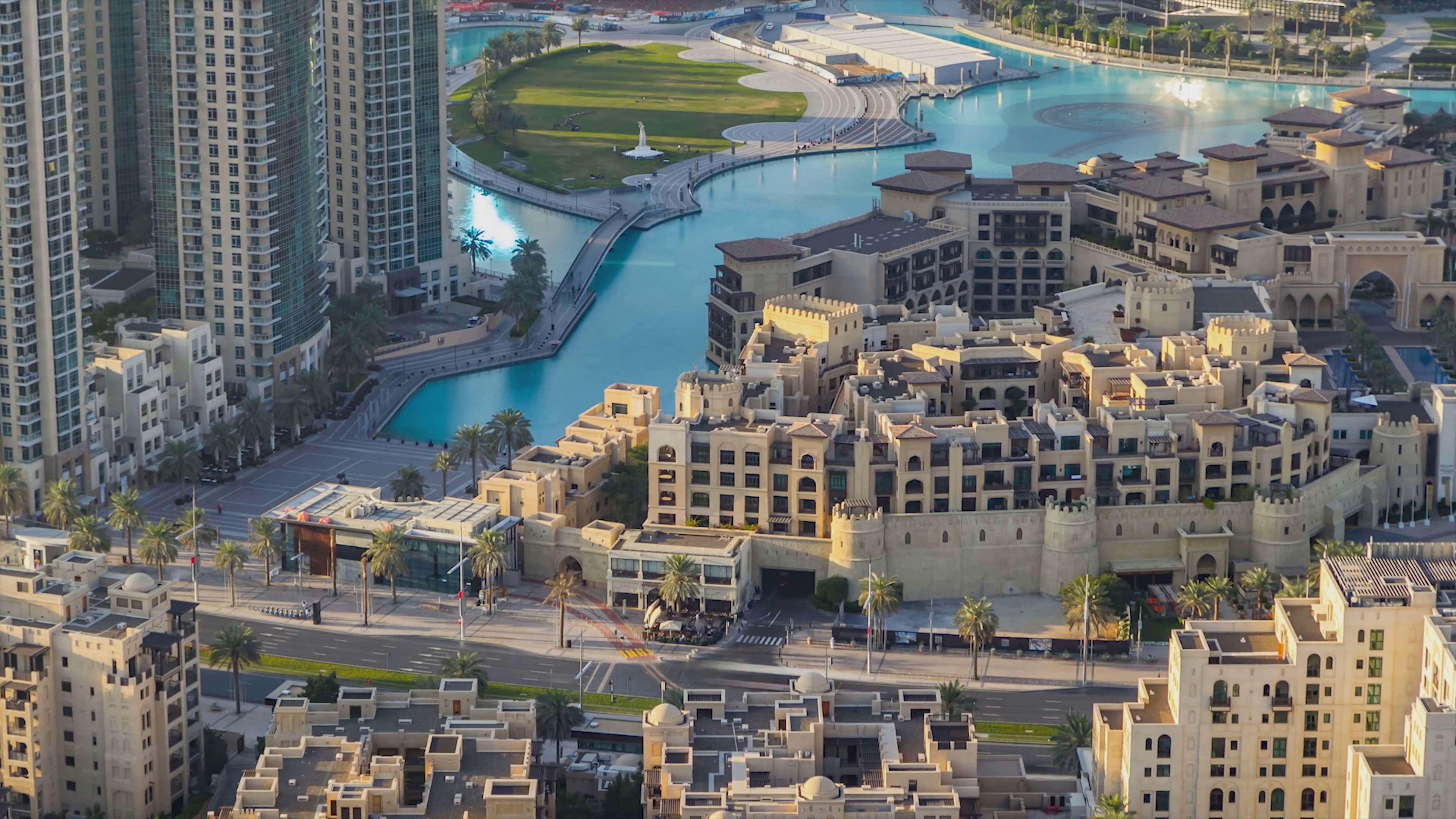} &
        \includegraphics[width=0.153\textwidth]{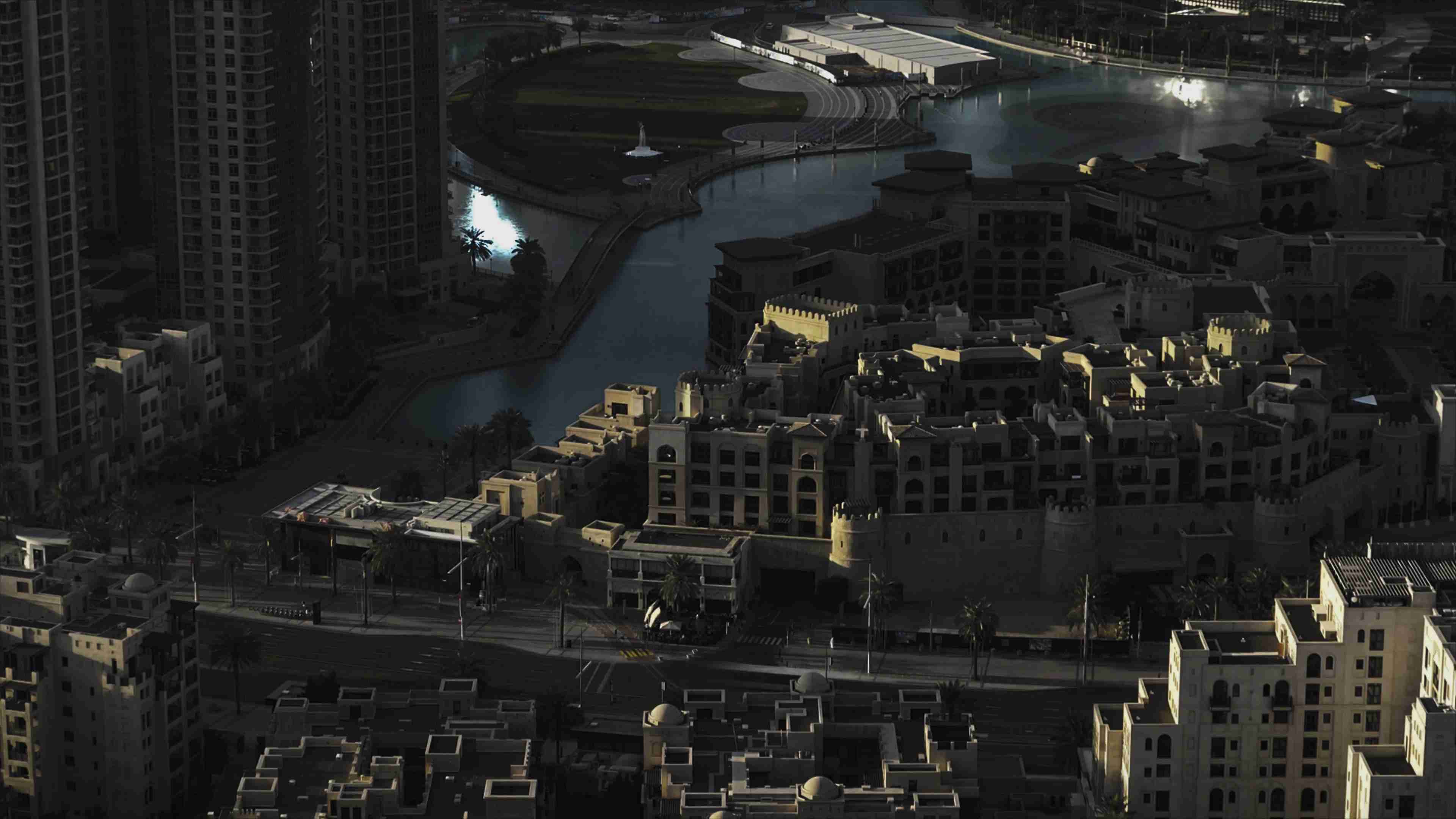} &
        \includegraphics[width=0.153\textwidth]{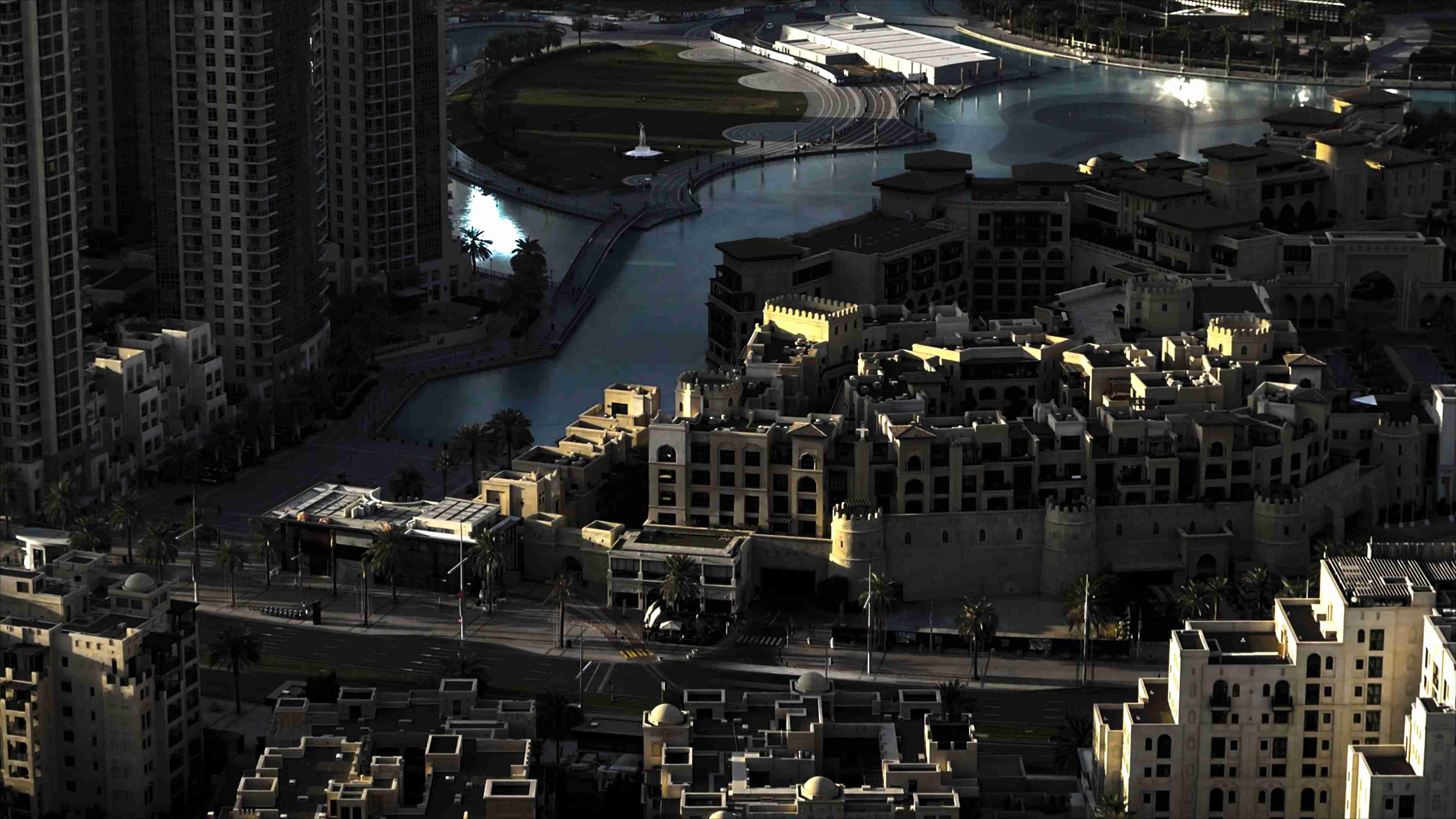} &
        \includegraphics[width=0.153\textwidth]{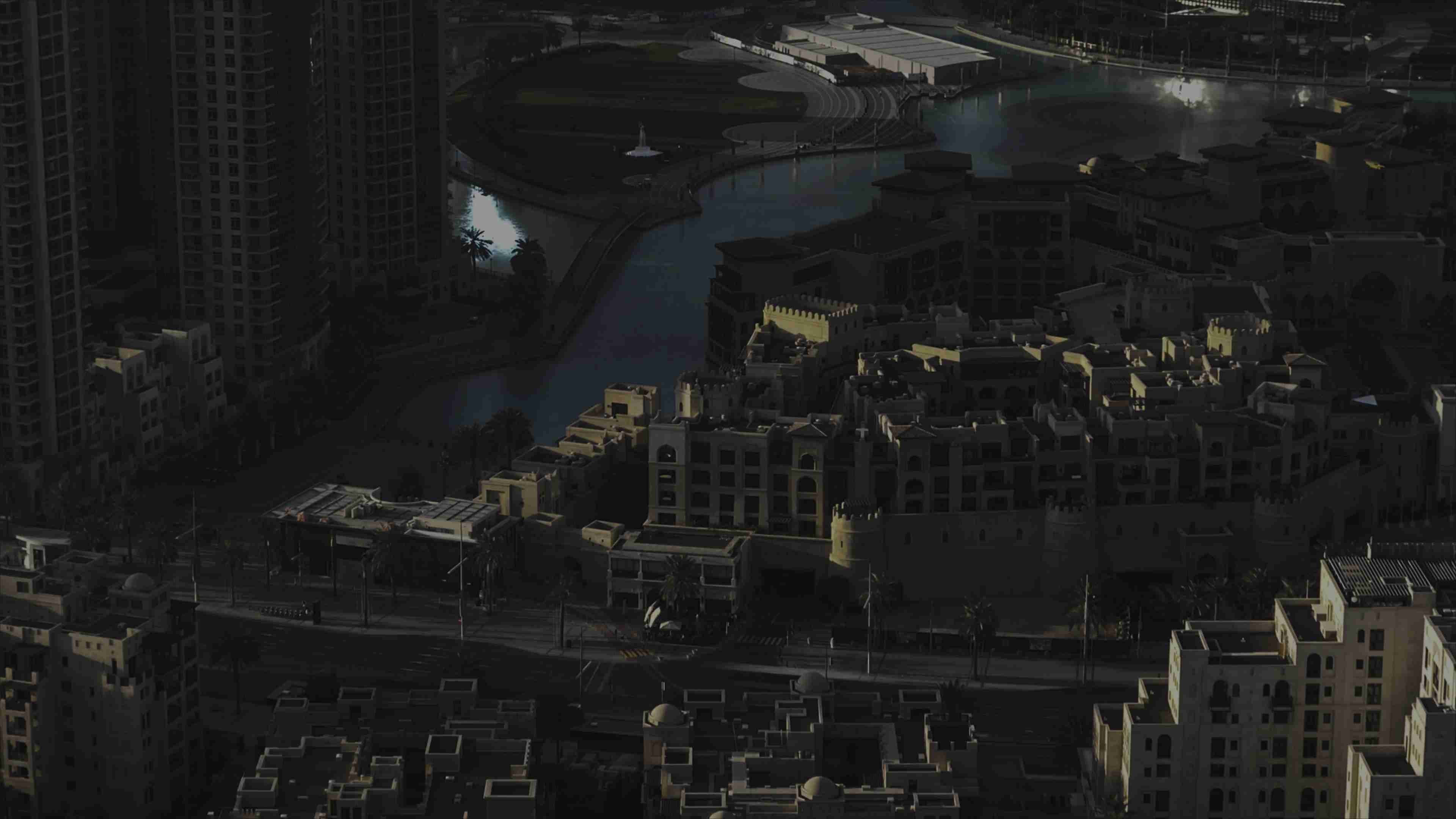} &
        \includegraphics[width=0.153\textwidth]{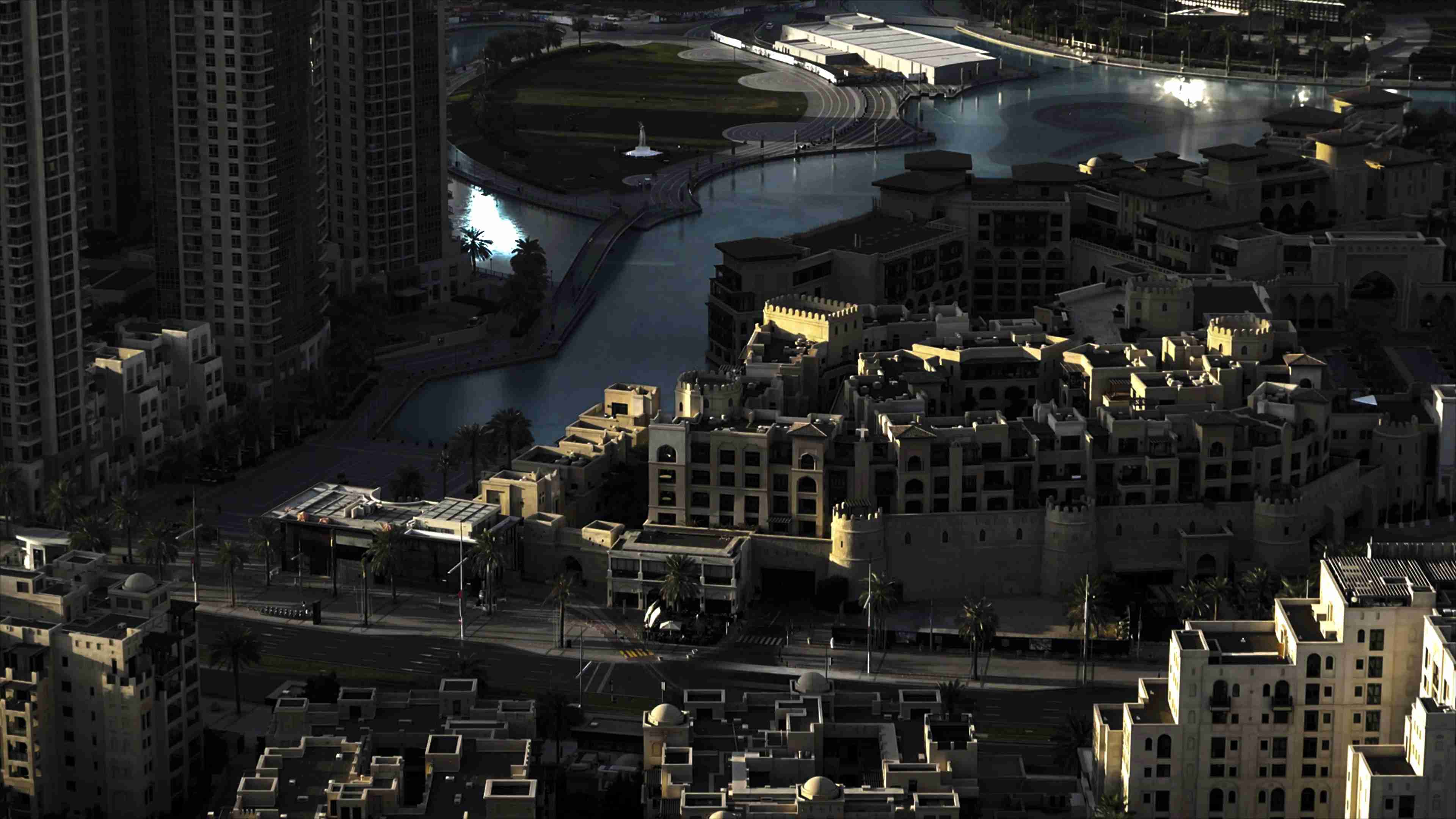} &
        \includegraphics[width=0.153\textwidth]{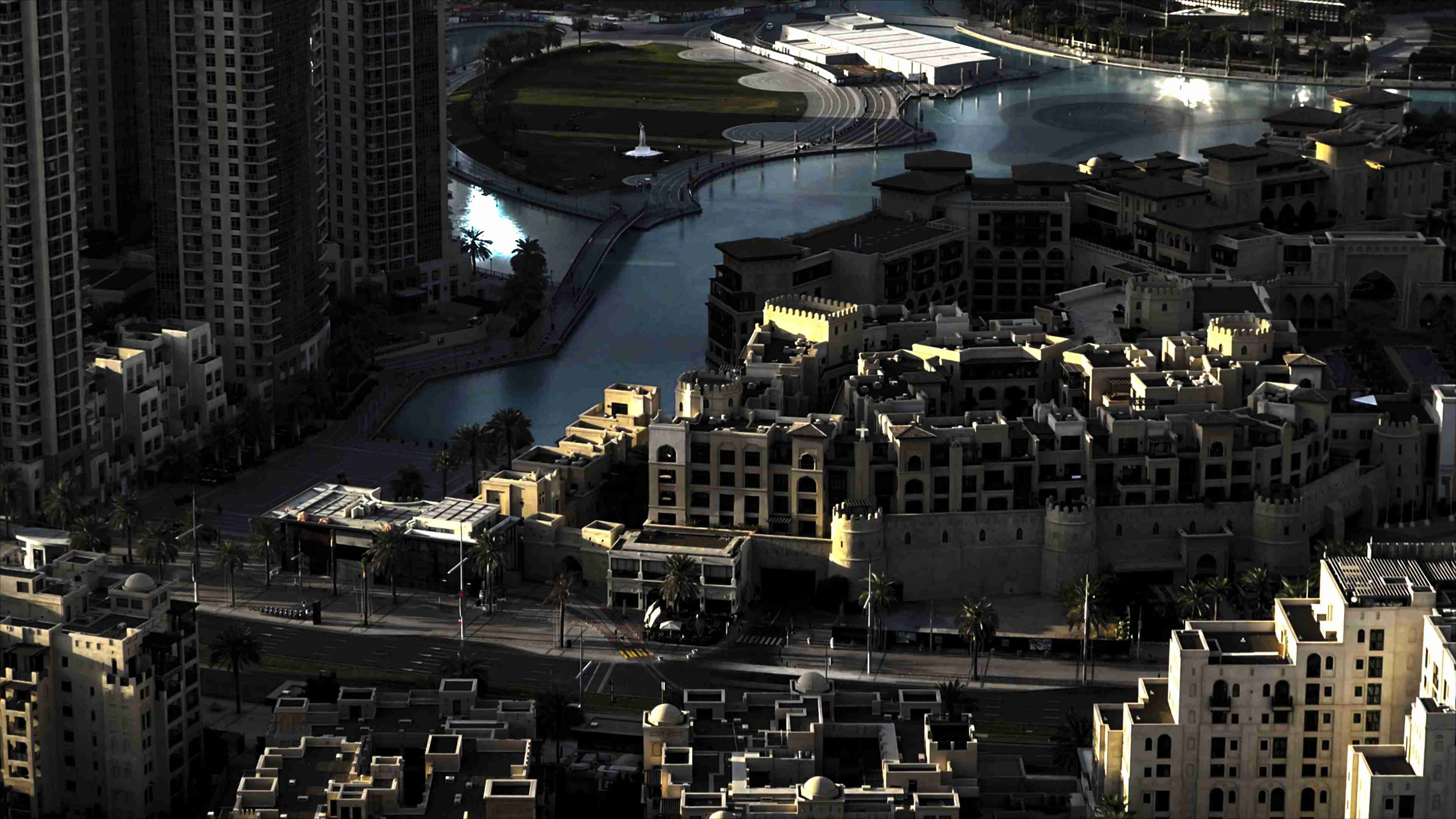} &
        \includegraphics[width=0.153\textwidth]{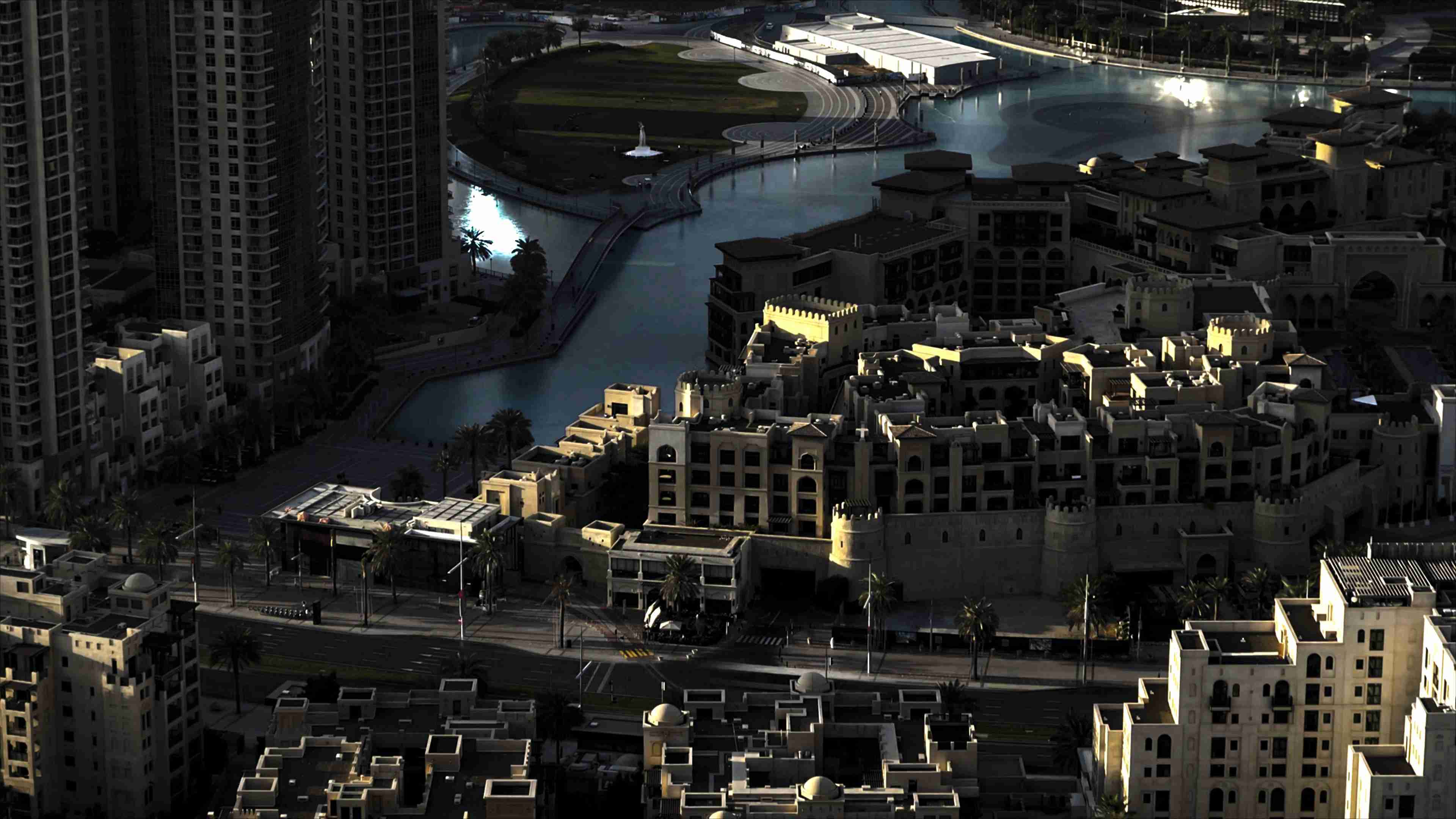} &
        \includegraphics[width=0.153\textwidth]{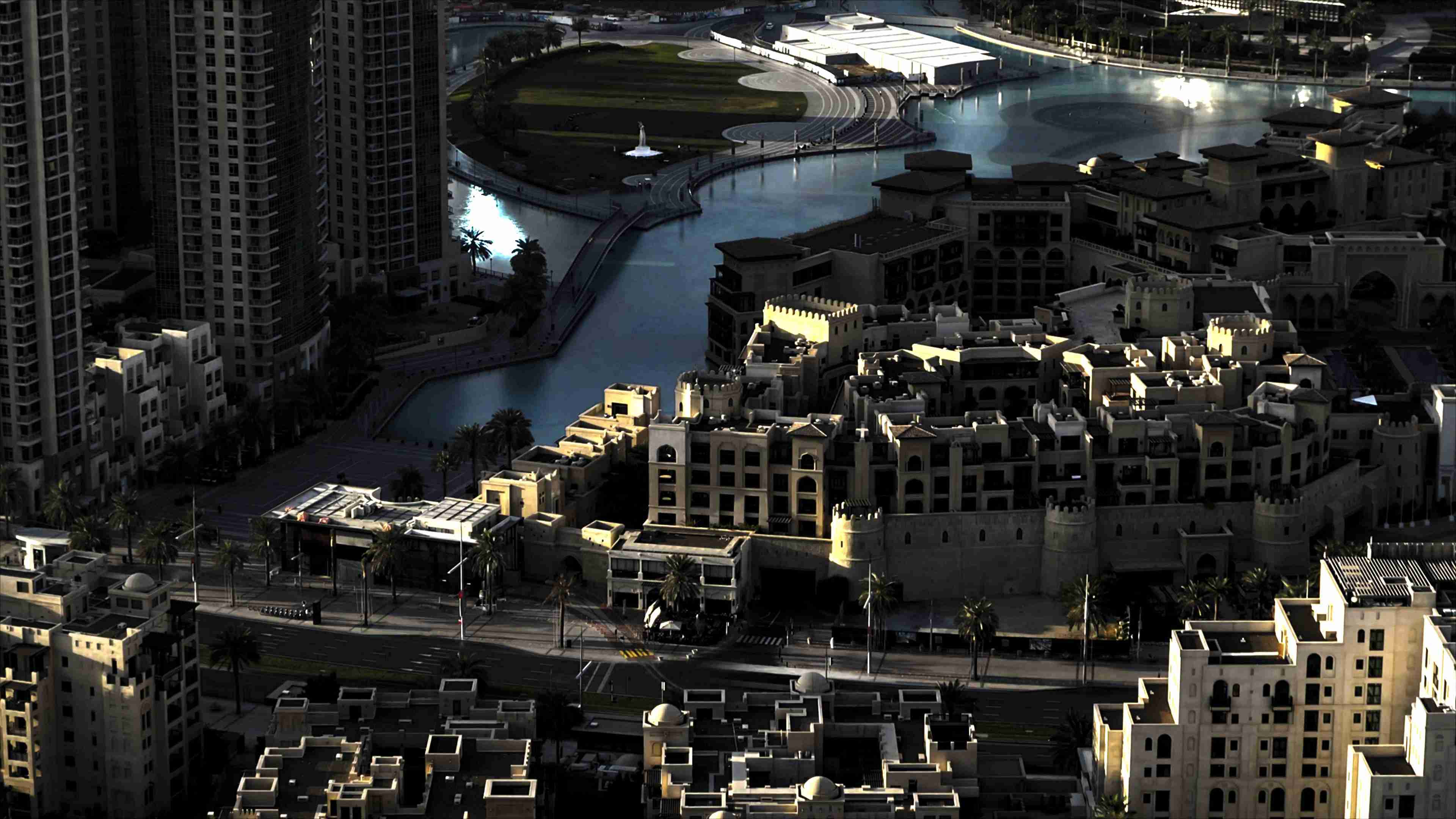} &
        \includegraphics[width=0.153\textwidth]{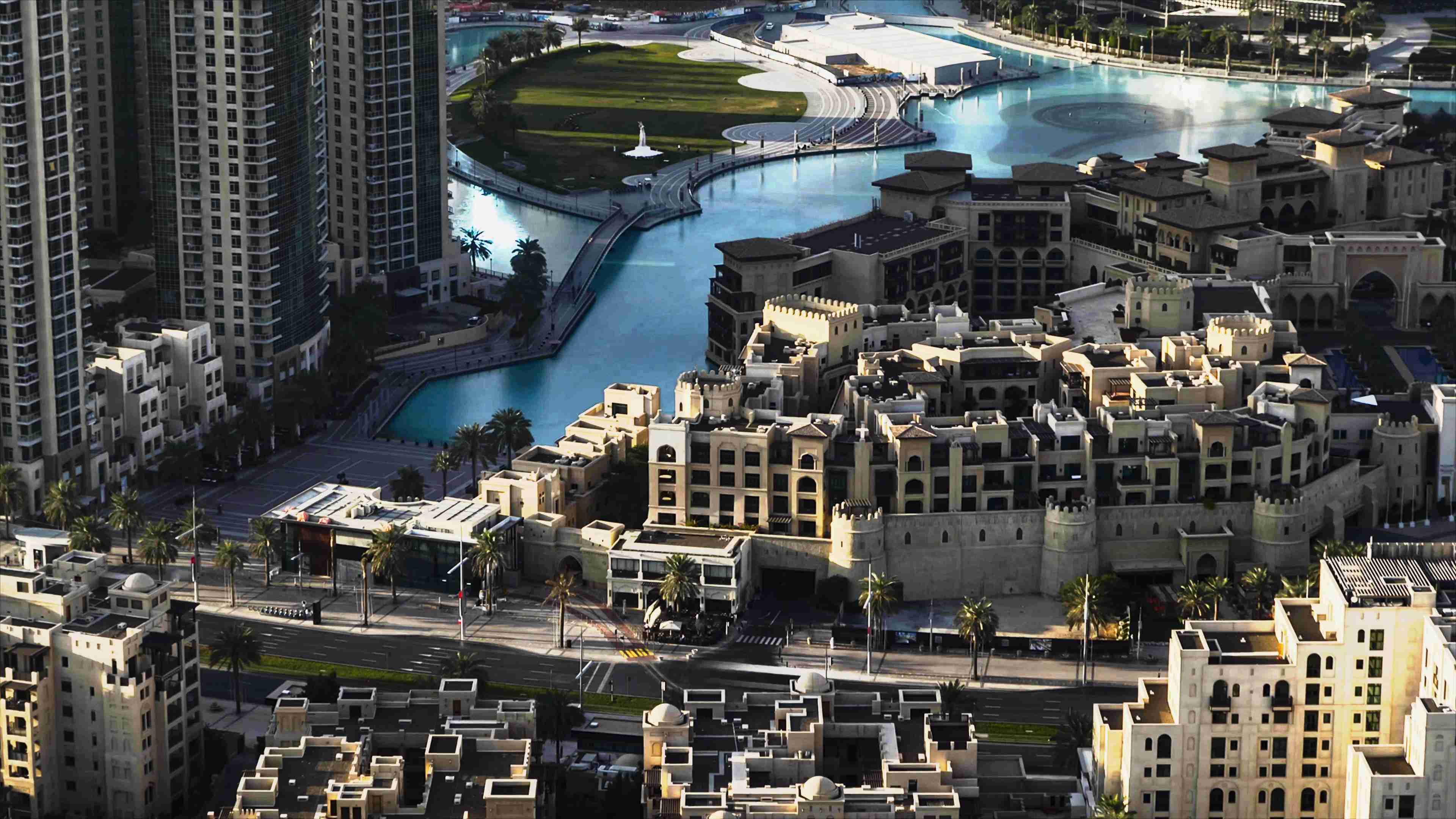} \\
        \LARGE Input &\LARGE GT &\LARGE SwinIR &\LARGE Restormer &\LARGE RUAS &\LARGE Uformer &\LARGE LLFormer &\LARGE UHDFour &\LARGE UHDformer &\LARGE Ours \\
    \end{tabular}
    \end{adjustbox}
    \vspace{-4mm}
    \caption{Comparison of Low-light Enhancement Methods. The top row is from the UHD-LL dataset, and the bottom is from the UHD-LOL dataset.}
    \vspace{-4mm}
    \label{fig: Low-light image enhancement}
\end{figure*}

\textbf{Low-Light Image Enhancement Results.}
We evaluate UHD low-light image enhancement results on UHD-LL with two training dataset sets, including UHD-LOL4k~\cite{wang2023llformer} and UHD-LL~\cite{Li2023ICLR_uhdfour}.
In Table \ref{tab:Low-light image enhancement.},  TSFormer achieves state-of-the-art results for low-light image enhancement on both UHD-LOL4K and UHD-LL datasets, with the highest PSNR and SSIM values across both datasets. 
Despite its strong performance, TSFormer maintains a lightweight architecture with only 3.38M parameters, making it significantly more efficient than other high-performing models such as Restormer and UHDFour. This balance of accuracy and efficiency suggests TSFormer’s suitability for real-time applications.
Figure \ref{fig: Low-light image enhancement} illustrates the visual improvements, with TSFormer effectively reducing noise and enhancing details under challenging low-light conditions, outperforming prior methods. 
%
%
\begin{table}[!t]\footnotesize
\setlength{\tabcolsep}{3pt} 
\renewcommand{\arraystretch}{1.2} 
\caption{Low-light image enhancement results on UHD-LOL4K and UHD-LL datasets. We highlight the \colorbox{best}{best} and \colorbox{secondbest}{secondbest} values for each metric. Our method, TSFormer, achieves state-of-the-art performance with competitive parameter efficiency.}
\begin{tabular}{l|c|cc|cc|l}
\shline
\textbf{Method} & \textbf{Venue} & \multicolumn{2}{c|}{\textbf{UHD-LOL4K}} & \multicolumn{2}{c|}{\textbf{UHD-LL}} &\textbf{Param}\\
 & & \textbf{PSNR} & \textbf{SSIM} & \textbf{PSNR} & \textbf{SSIM} &  \\
\shline
SwinIR & ICCVW'21 & 35.92 & 0.961 &  17.90 & 0.738 & 11.5M  \\
Restormer & CVPR'22 & 36.91  & 0.988  & 21.54 & 0.844 & 26.1M \\
RUAS & CVPR’22 & 14.68 & 0.758  & 13.56  &0.749 & 0.35M \\
Uformer & CVPR'22 & 29.98 & 0.980 & 21.30 & 0.823 & 20.6M \\
LLFormer & AAAI'23 & \cellcolor{secondbest}37.33 & \cellcolor{secondbest}0.989  & 24.07 & 0.858 & 13.2M \\
UHDFour & ICLR'23 & 36.12 & 0.990  & 26.23 & 0.900 & 17.5M \\
UHDformer & AAAI'24 & 36.11 & 0.971  & \cellcolor{secondbest}27.11 & \cellcolor{secondbest}0.927 & 0.34M \\
TSFormer (Ours) & - & \cellcolor{best}\textbf{37.48} & \cellcolor{best}\textbf{0.993}  & \cellcolor{best}\textbf{27.40} & \cellcolor{best}\textbf{0.933} & 3.38M \\
\shline
\end{tabular}
\label{tab:Low-light image enhancement.}
\vspace{-4mm}
\end{table}

\noindent \textbf{Image Deblurring Results.} 
We evaluate image deblurring on UHD-Blur dataset. As shown in Table \ref{tab: image deblurring.}, the quantitative results of image deblurring on the UHD-Blur dataset. TSFormer achieves excellent performance with significant improvements across key metrics. 
Specifically, TSFormer achieves a PSNR of 29.52 dB, surpassing UHDformer’s 28.82 dB and outperforming other general-purpose models like Restormer, Uformer, and Stripformer, which fall in the 25–25.4 dB. This highlights TSFormer’s capability to restore high-resolution details in UHD images.
We also evaluate the performance of GoPro~\cite{nah2017deep} datasets and report the results in Table~\ref{tab:gopro}.

\begin{figure*}[!t]
	\centering
	\begin{adjustbox}{max width=\textwidth}
		\begin{tabular}{ccccccccc}
			\includegraphics[width=0.24\textwidth]{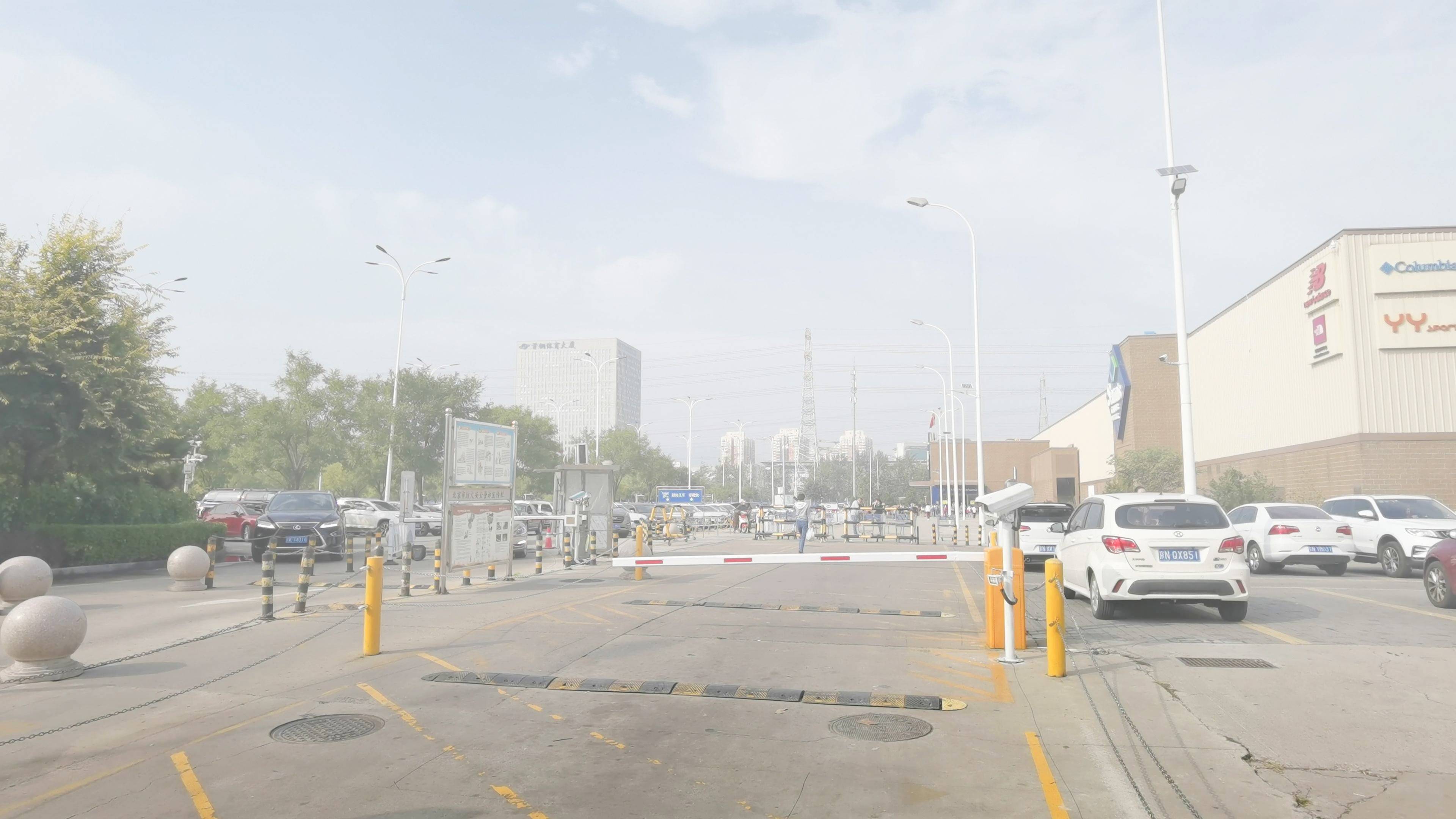} &
			\includegraphics[width=0.24\textwidth]{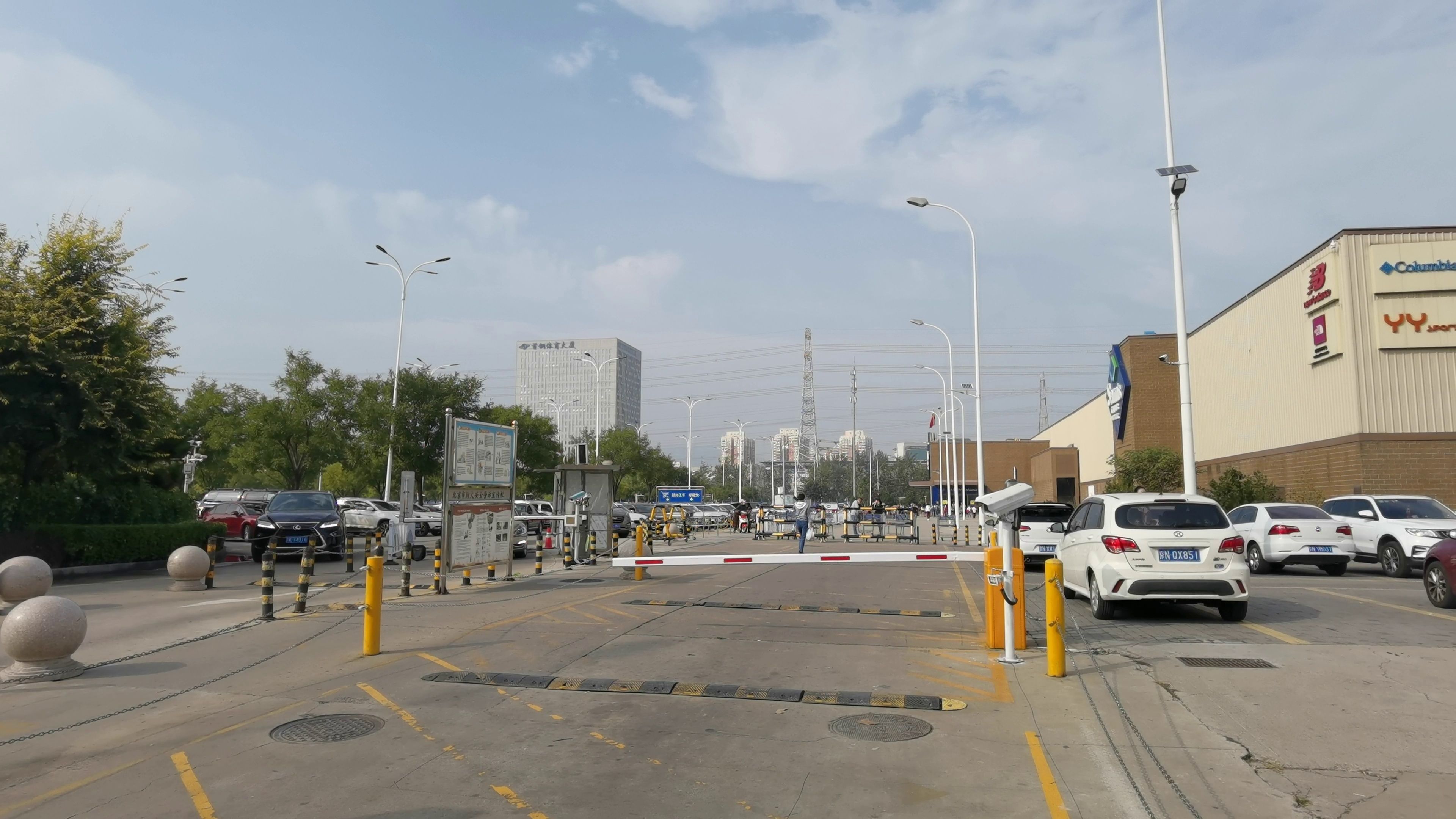} &
			\includegraphics[width=0.24\textwidth]{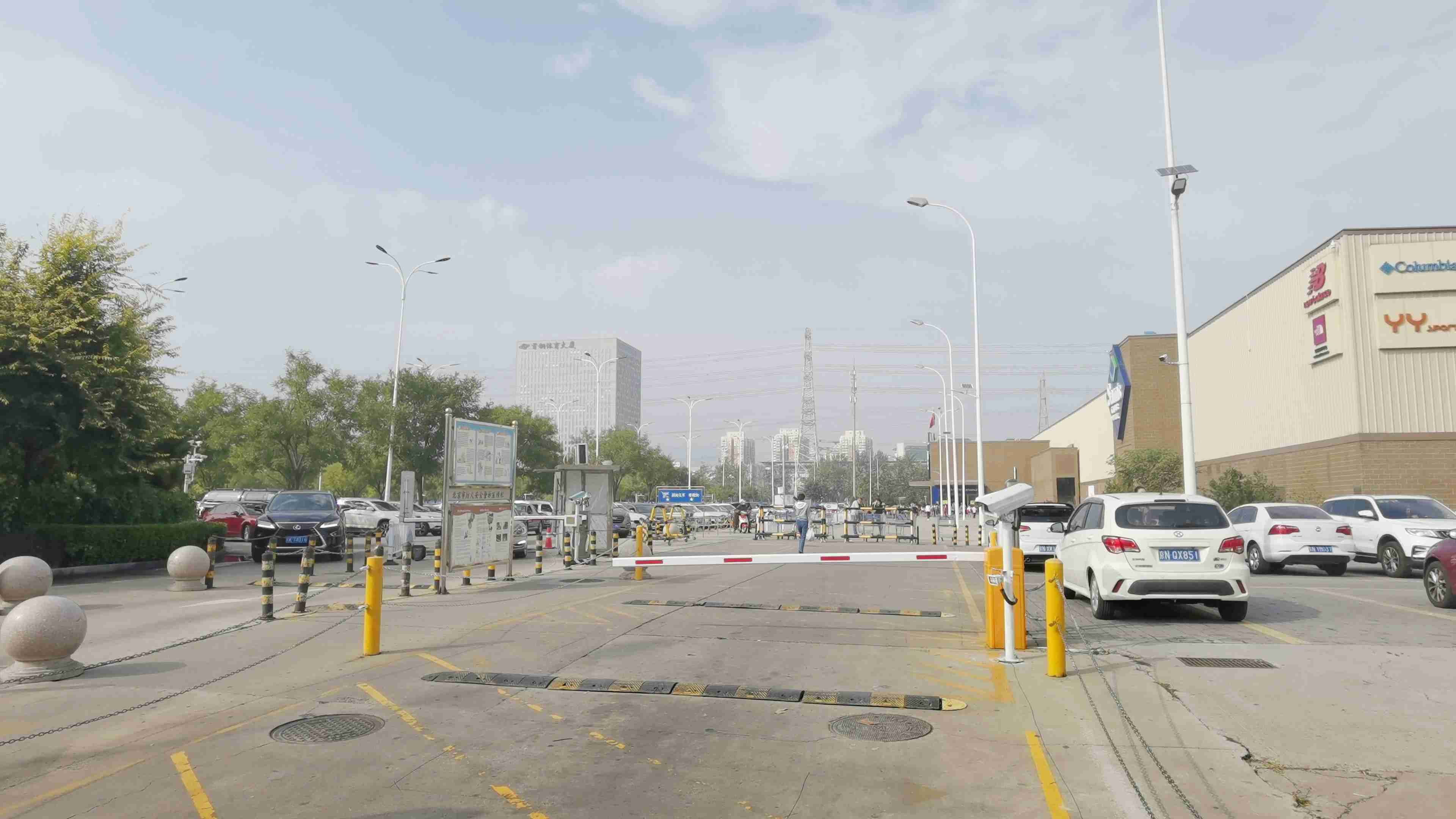} &
			\includegraphics[width=0.24\textwidth]{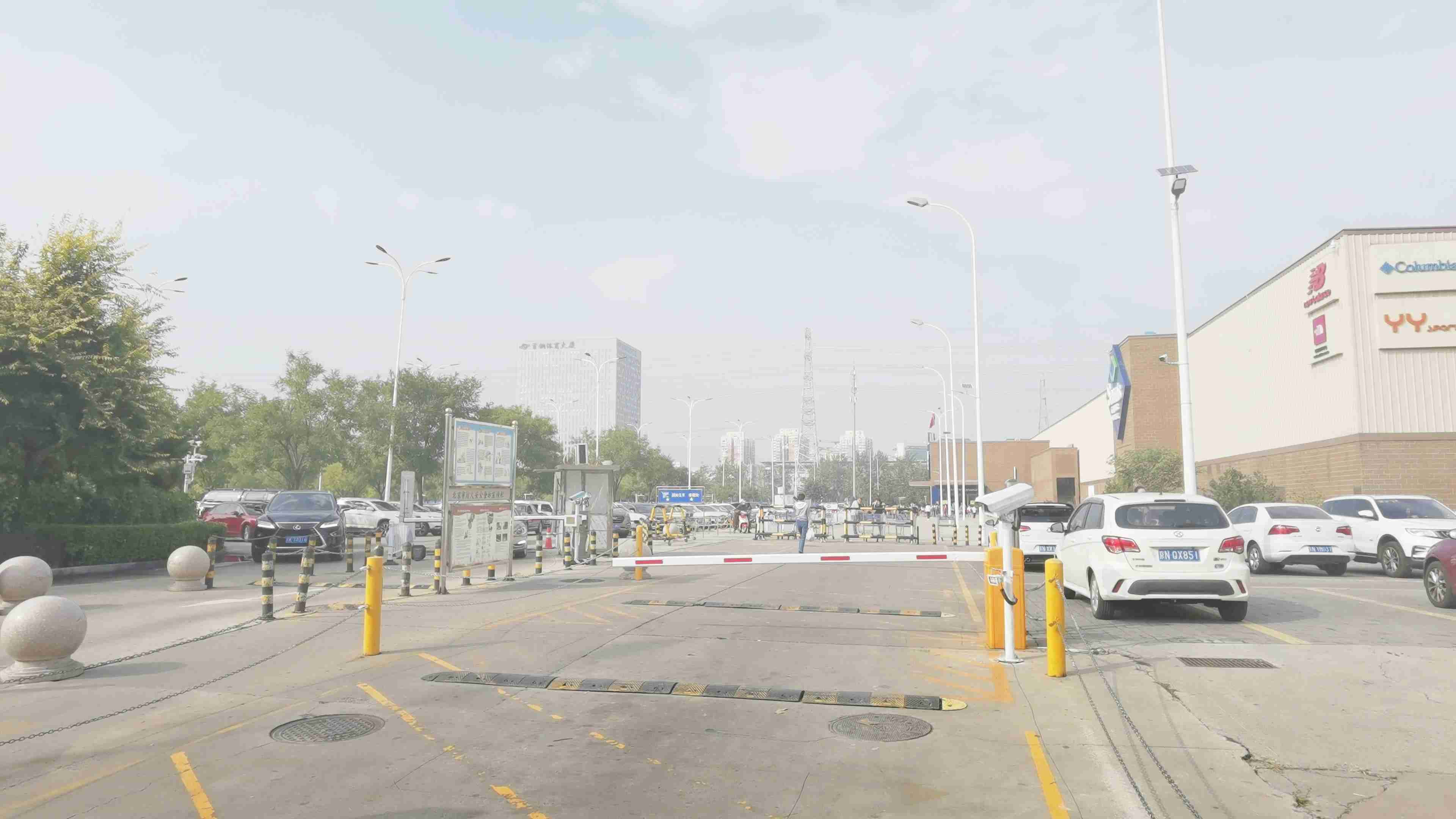} &
			\includegraphics[width=0.24\textwidth]{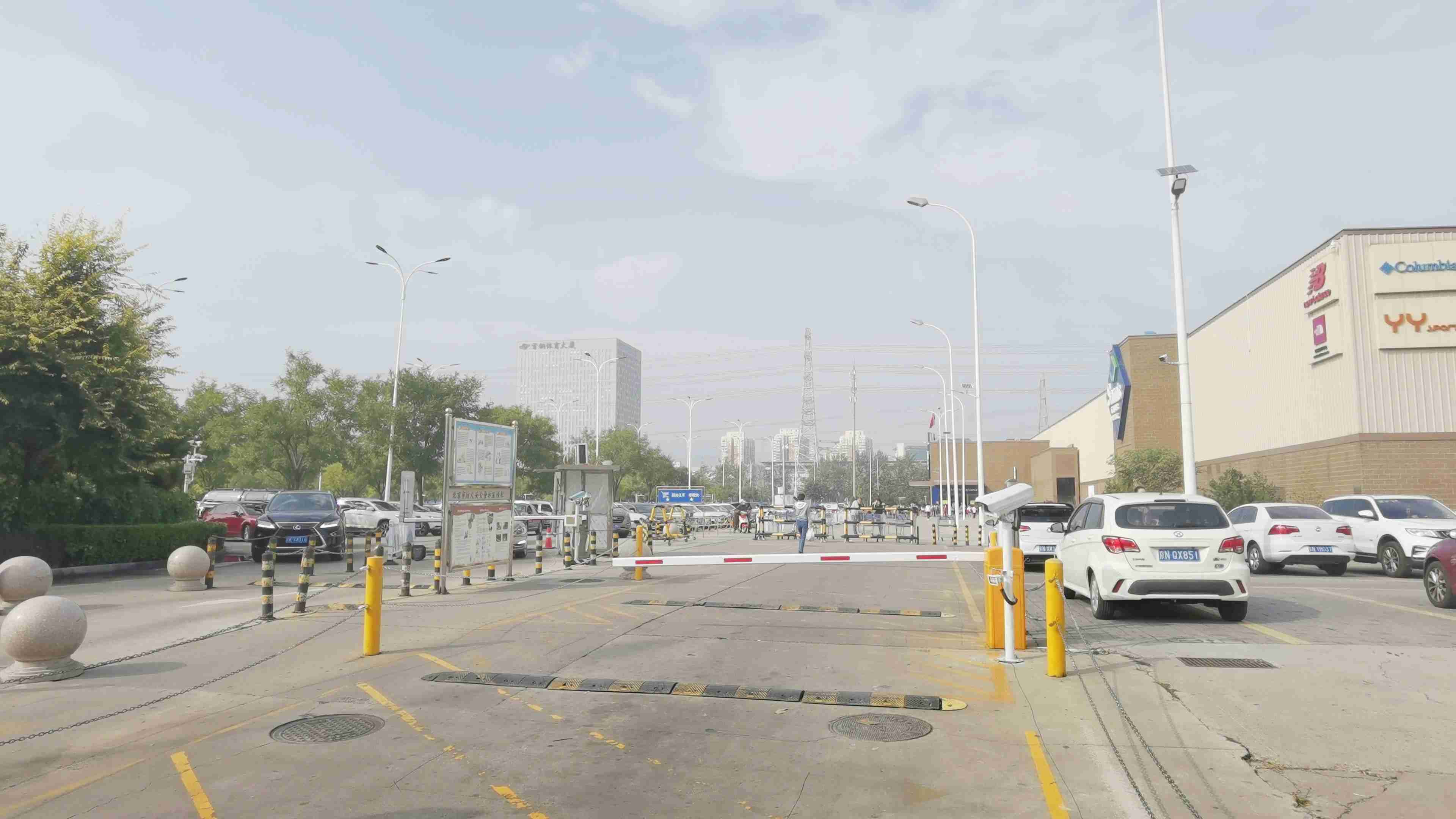} &
			\includegraphics[width=0.24\textwidth]{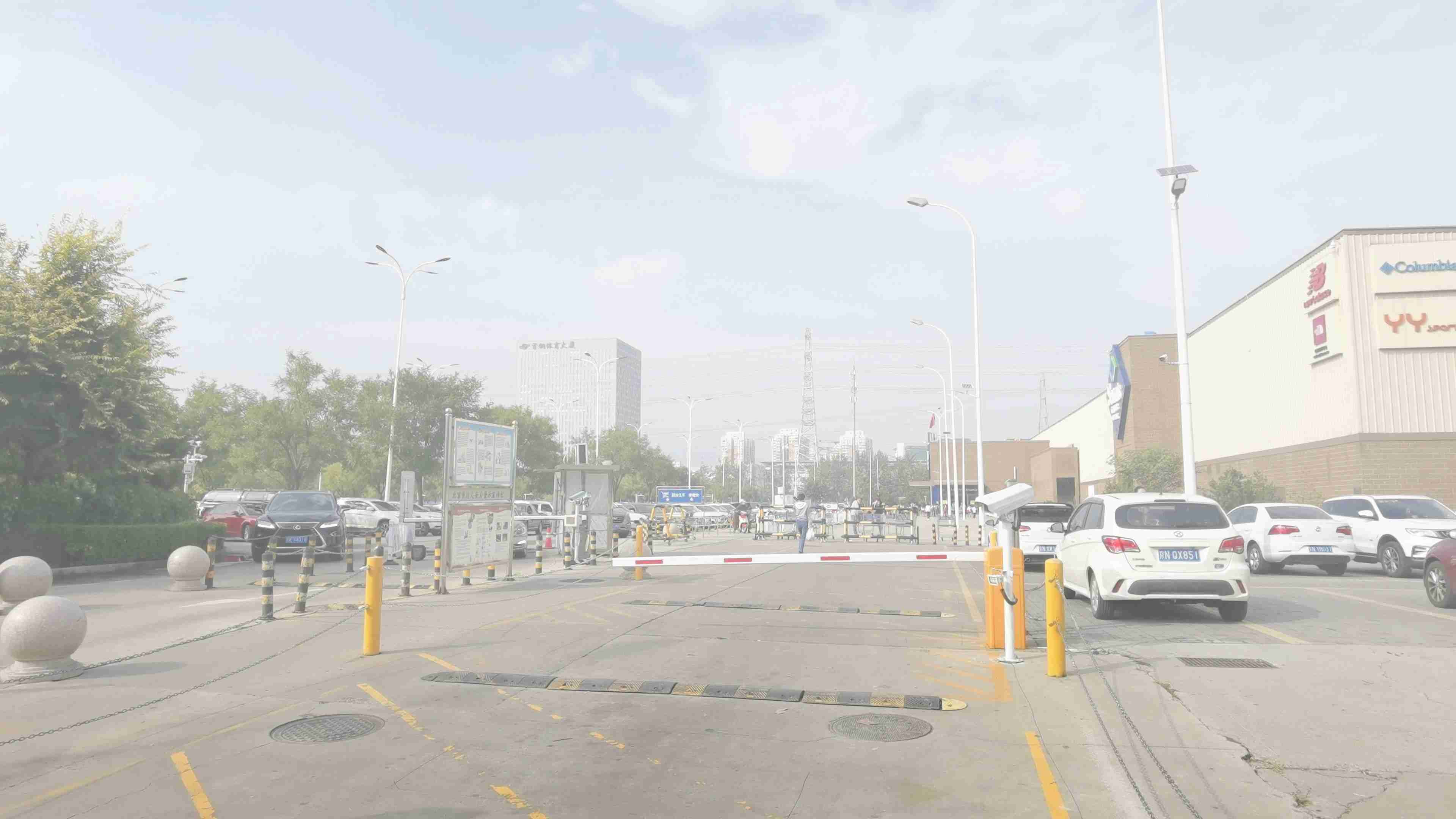} &
			\includegraphics[width=0.24\textwidth]{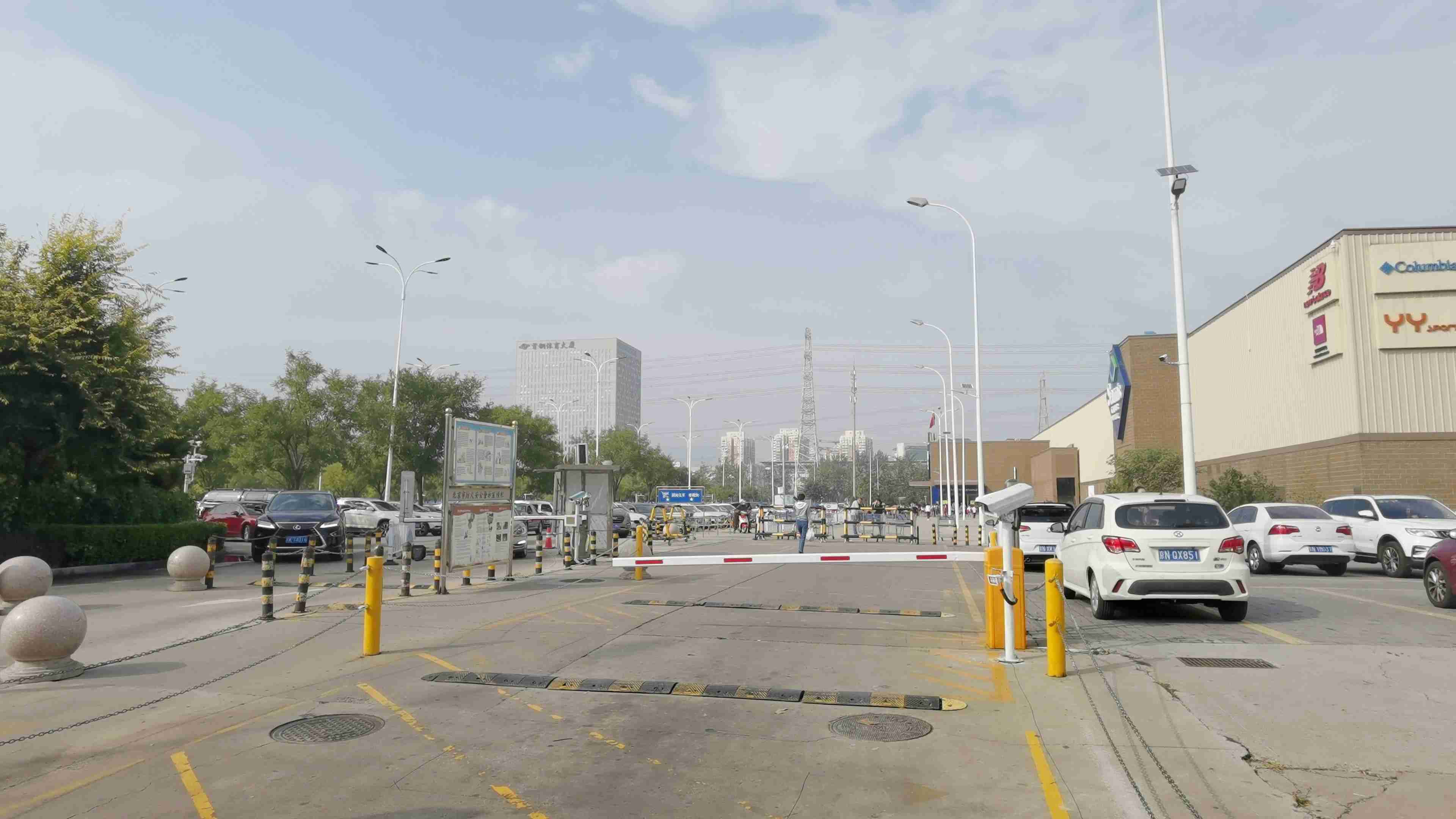} &
			\includegraphics[width=0.24\textwidth]{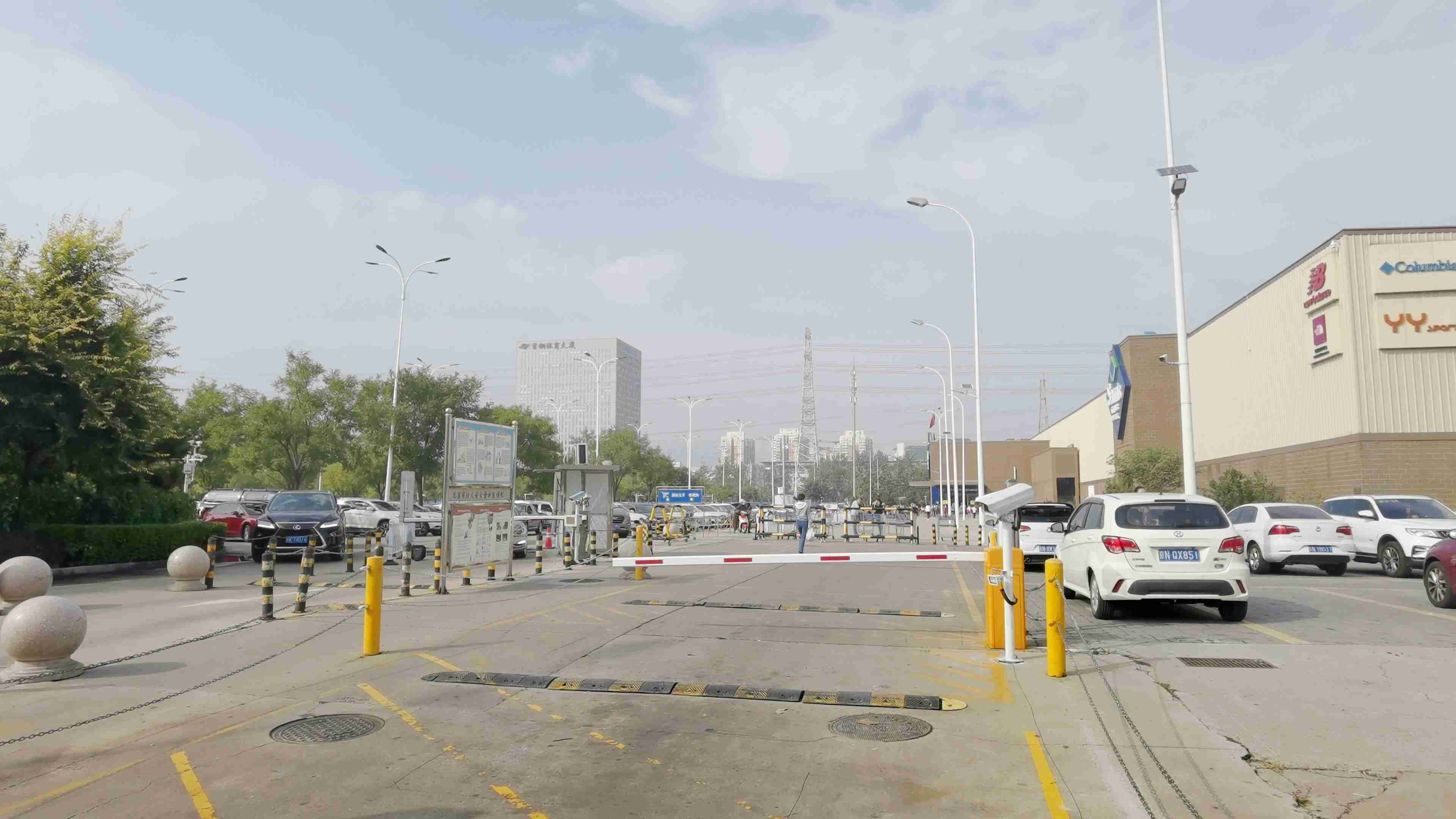}&
			\includegraphics[width=0.24\textwidth]{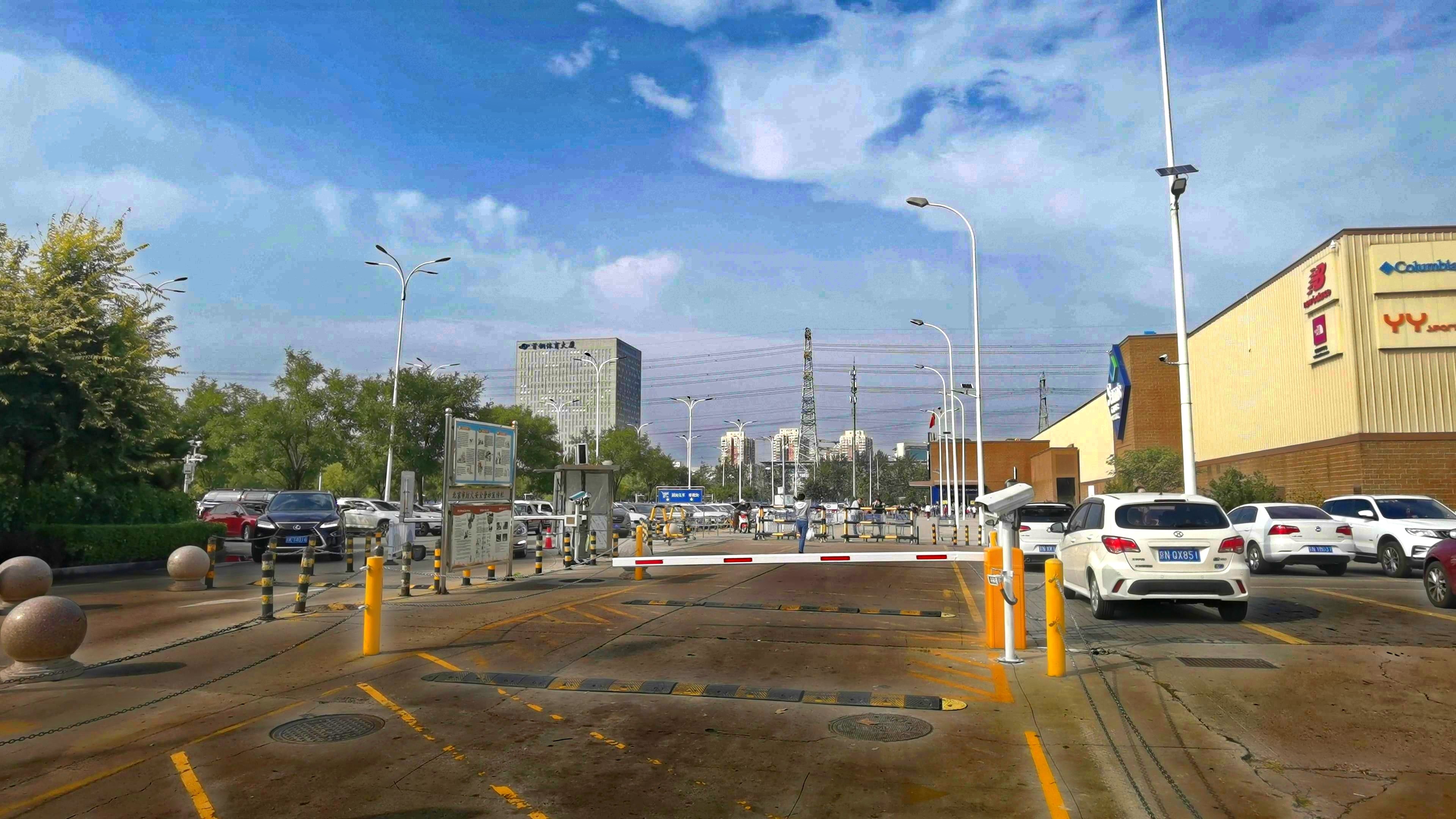} \\
			\includegraphics[width=0.24\textwidth]{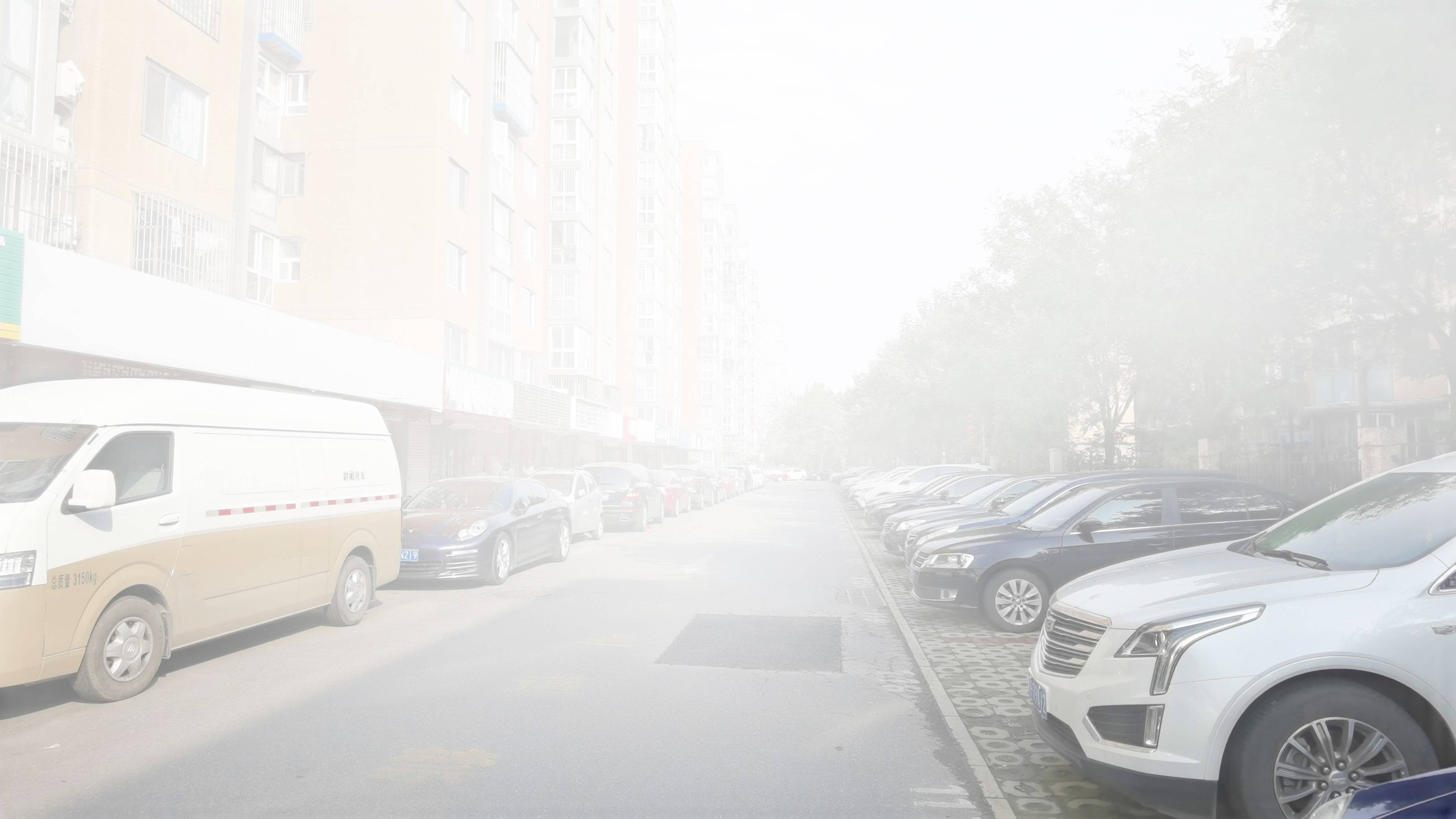} &
			\includegraphics[width=0.24\textwidth]{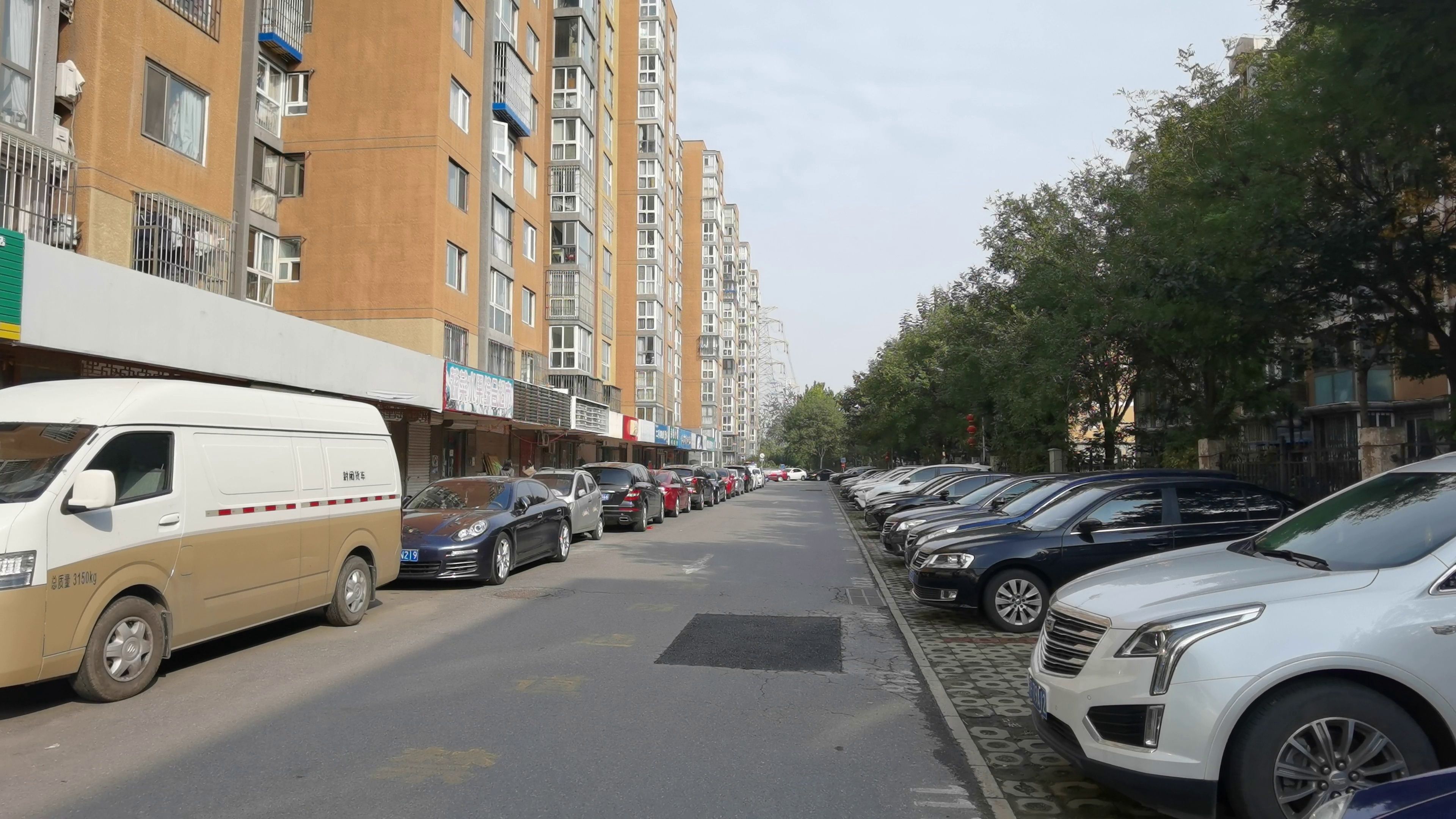} &
			\includegraphics[width=0.24\textwidth]{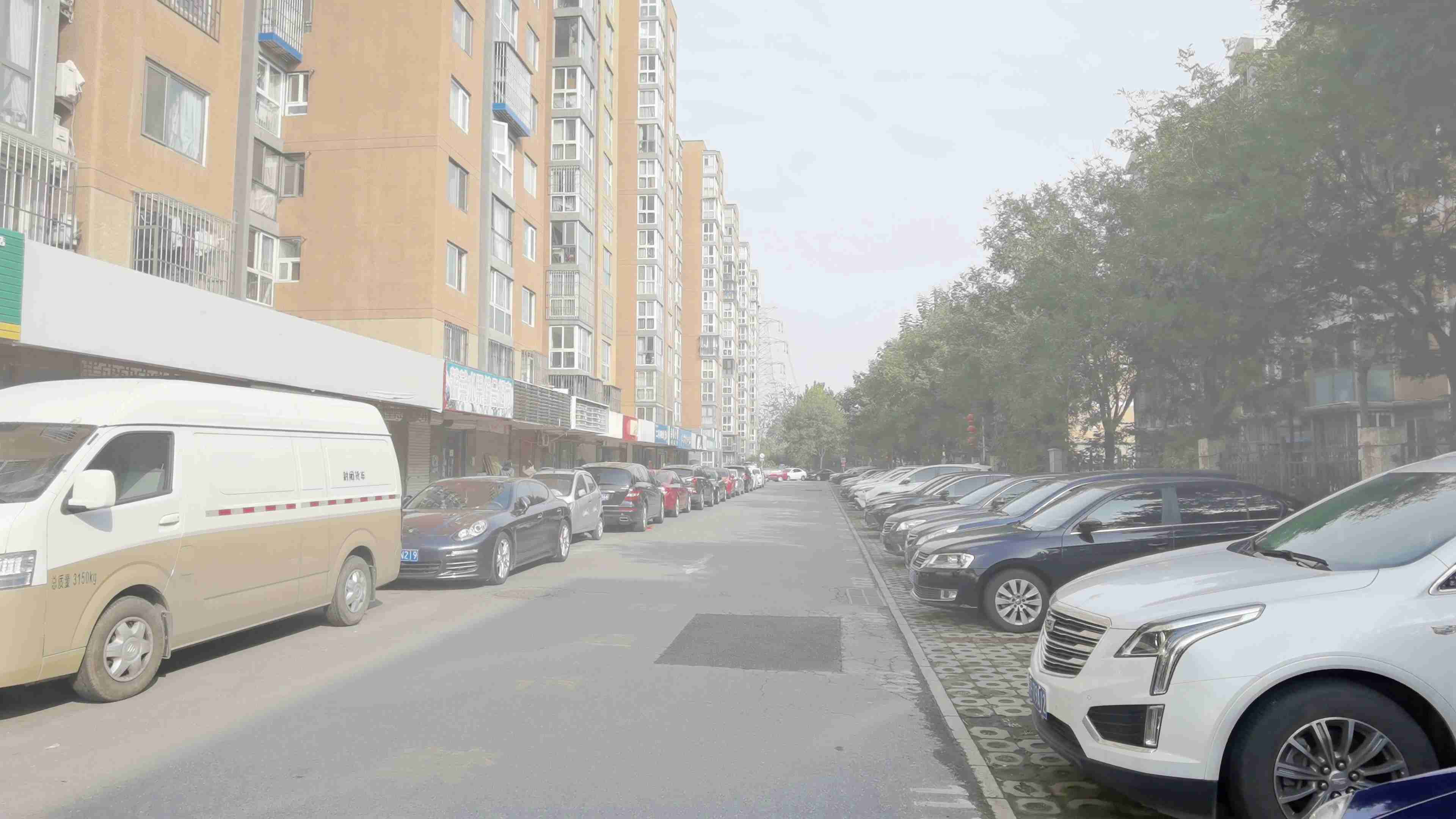} &
			\includegraphics[width=0.24\textwidth]{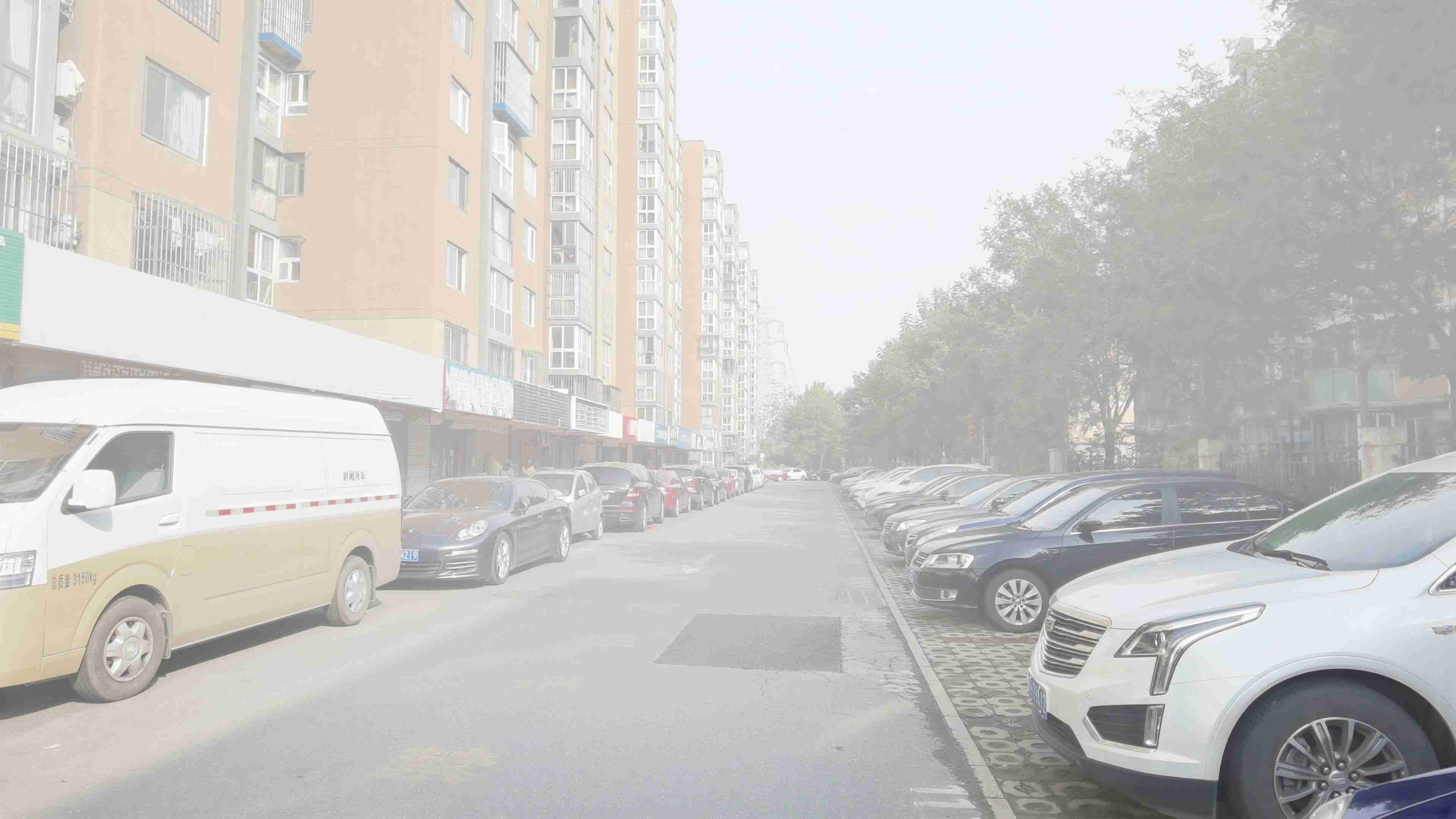} &
			\includegraphics[width=0.24\textwidth]{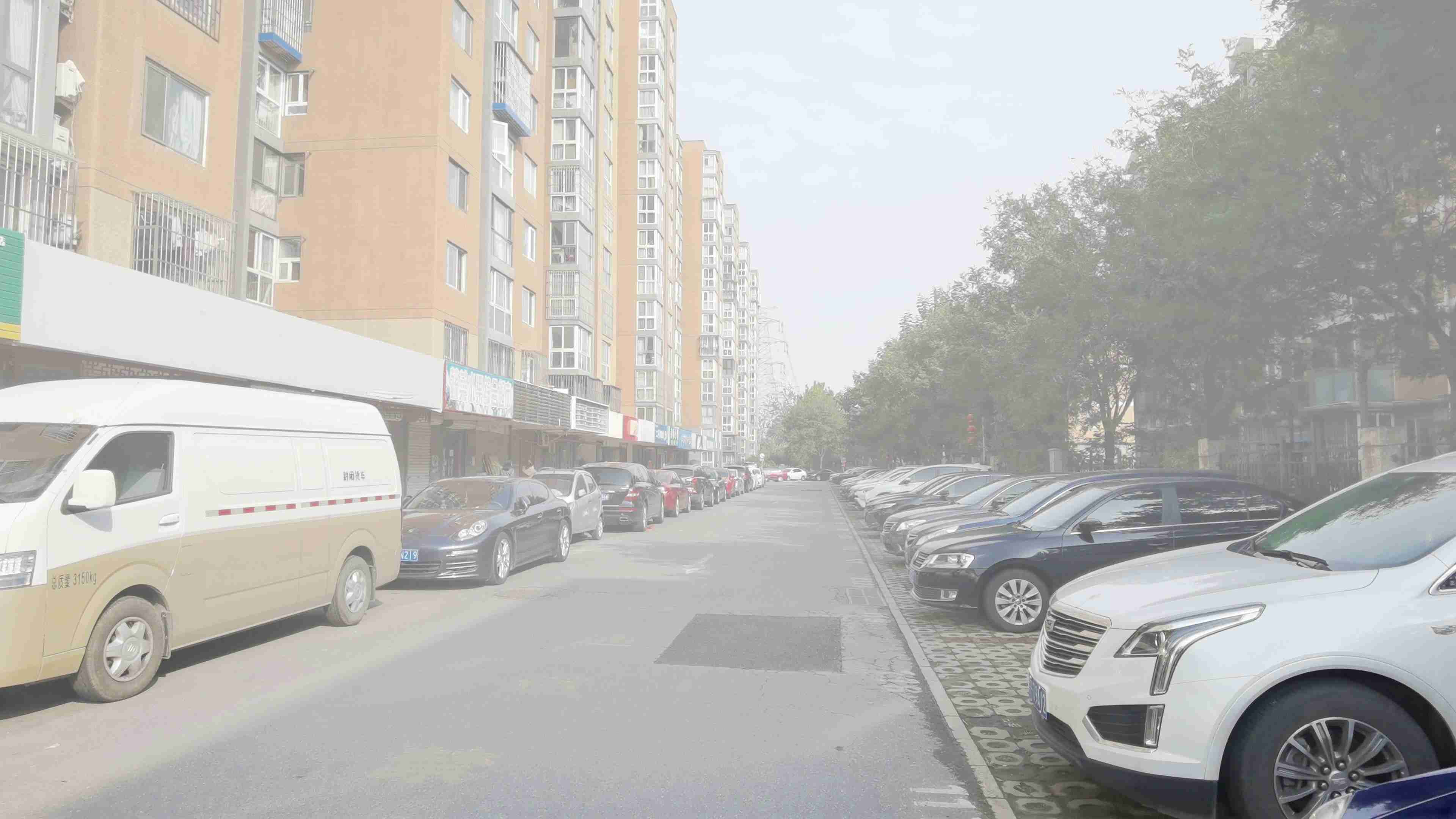} &
			\includegraphics[width=0.24\textwidth]{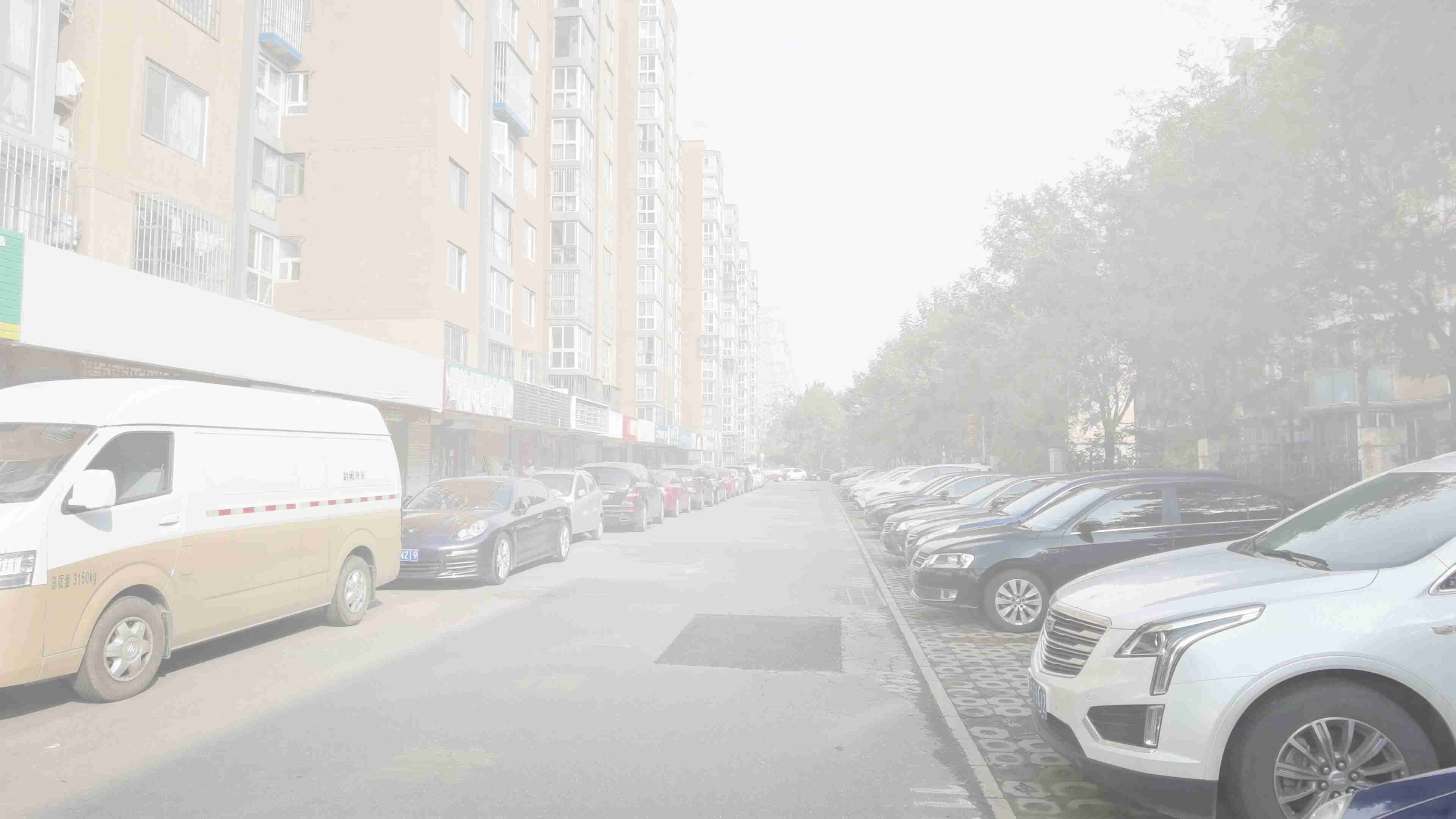} &
			\includegraphics[width=0.24\textwidth]{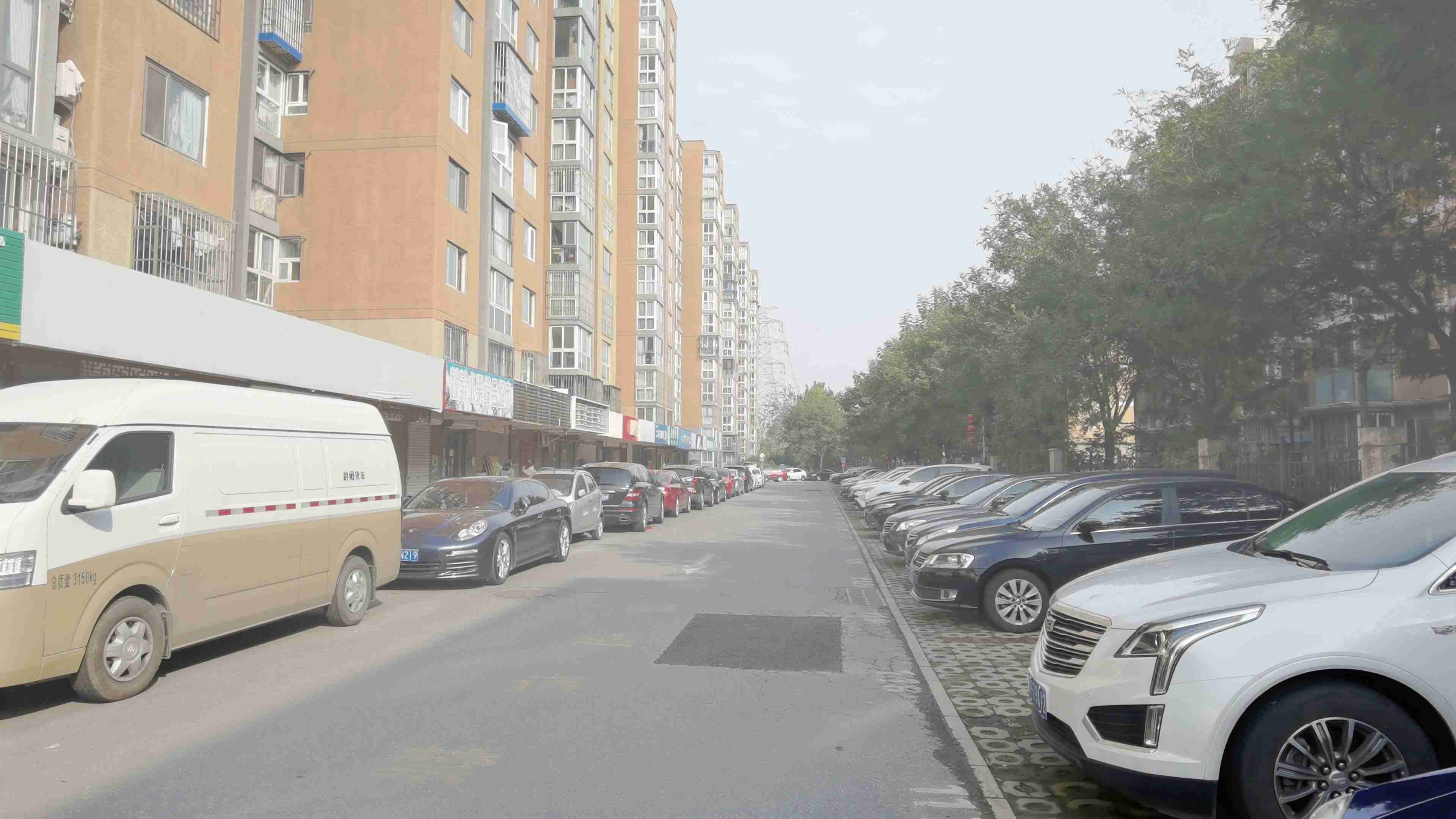} &
			\includegraphics[width=0.24\textwidth]{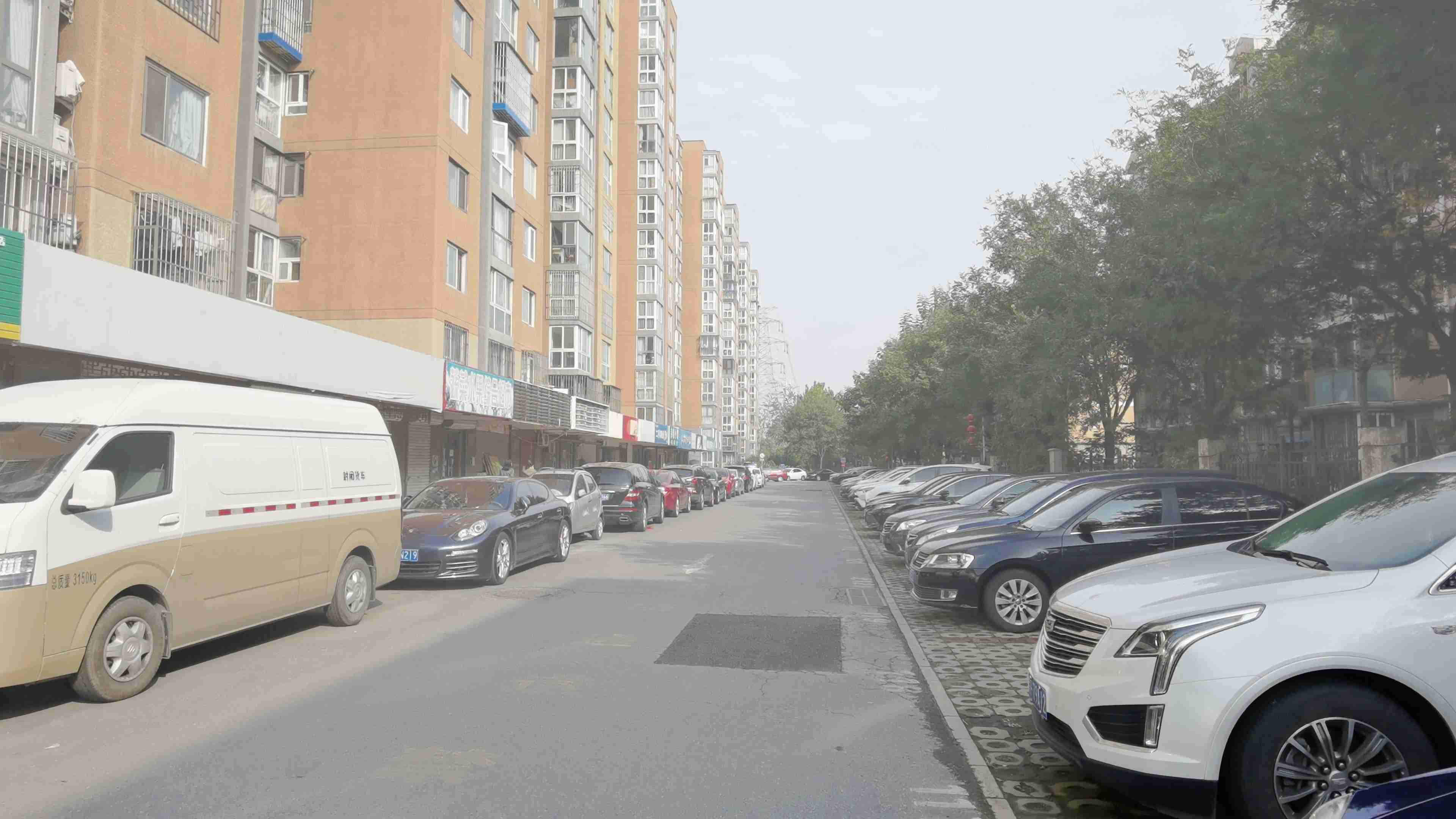}&
			\includegraphics[width=0.24\textwidth]{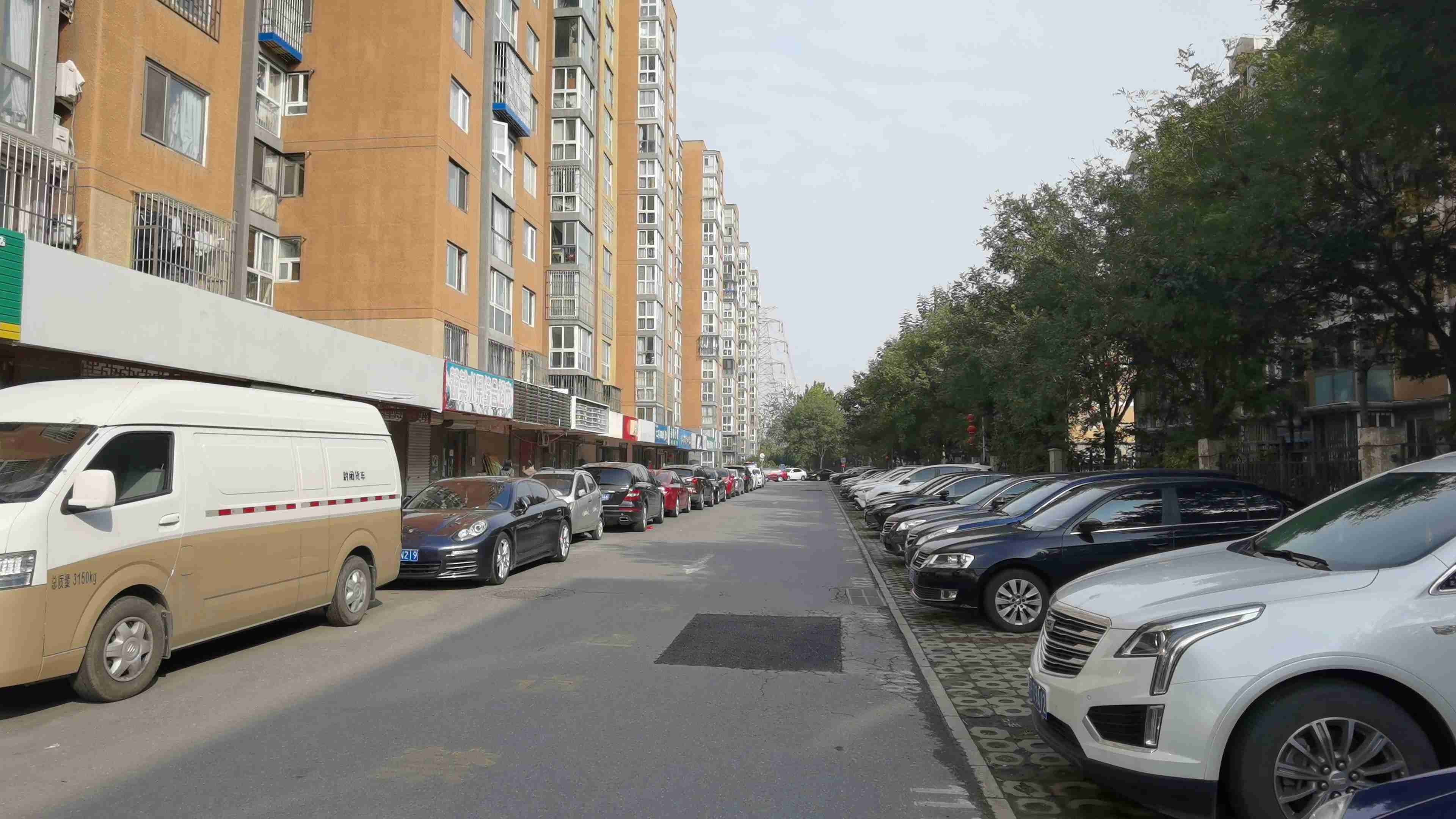} \\ 
			\huge Input &\huge GT & \huge DehazeFormer &\huge Restormer & \huge Uformer & \huge UHD & \huge UHDformer & \huge UHDDIP & \huge Ours \\
		\end{tabular}
	\end{adjustbox}
	\caption{Image dehazing on UHD-Haze. TSFormer is capable of producing clearer results.}
	\label{fig: dehaze}\vspace{-5mm}
\end{figure*}

\noindent \textbf{Image Dehazing Results.} 
Table~\ref{tab: image dehaze.} presents quantitative results on UHD-Haze using models trained on UHD-Haze dataset. TSFormer achieves the highest PSNR and SSIM scores, with a 0.77 dB gain over UHDformer on the UHD-Haze dataset, and consistently outperforms other methods in visual quality. TSFormer also provides a significant reduction in LPIPS while maintaining a competitive parameter count compared to prior models, showing that it balances efficiency and performance effectively. Figure~\ref{fig: dehaze} illustrates qualitative results, where TSFormer visibly produces the clearest output among all methods, showcasing its ability to effectively remove haze and recover fine details that are typically obscured by haze in other approaches. 

\begin{figure*}[!t]
    \centering
    \begin{adjustbox}{max width=\textwidth}
    \begin{tabular}{ccccccccc}
        \includegraphics[width=0.24\textwidth]{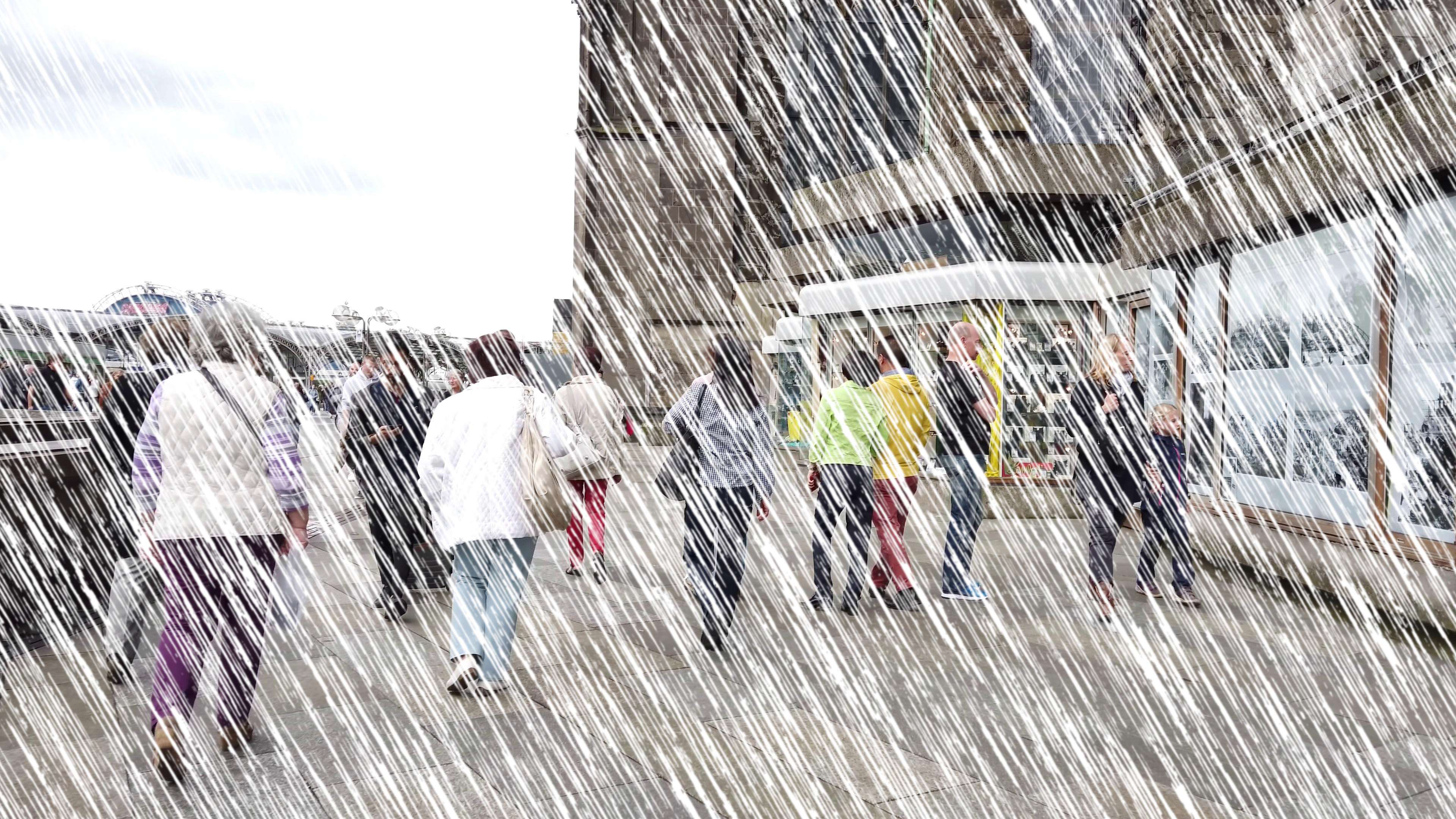} &
        \includegraphics[width=0.24\textwidth]{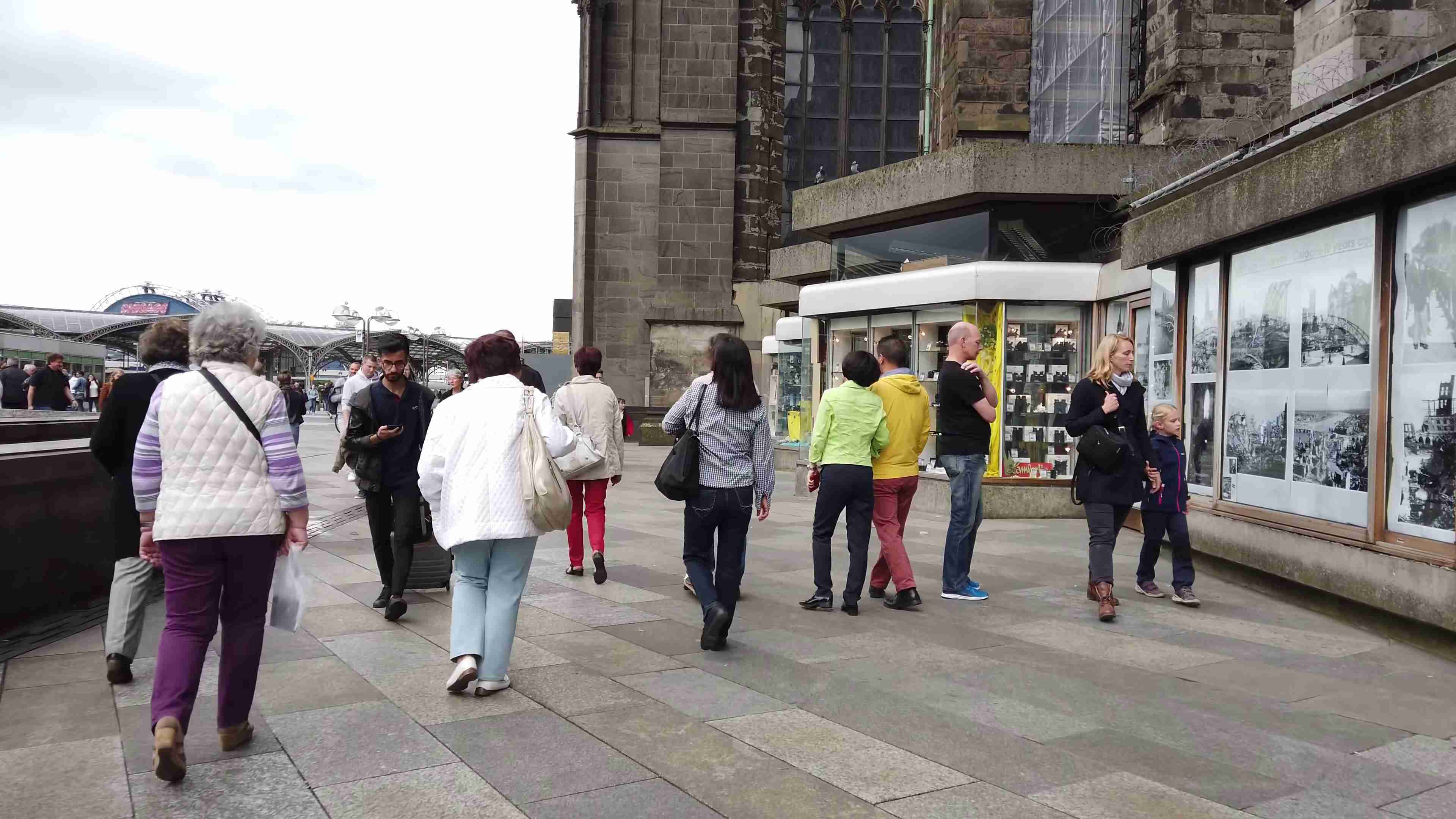} &
        \includegraphics[width=0.24\textwidth]{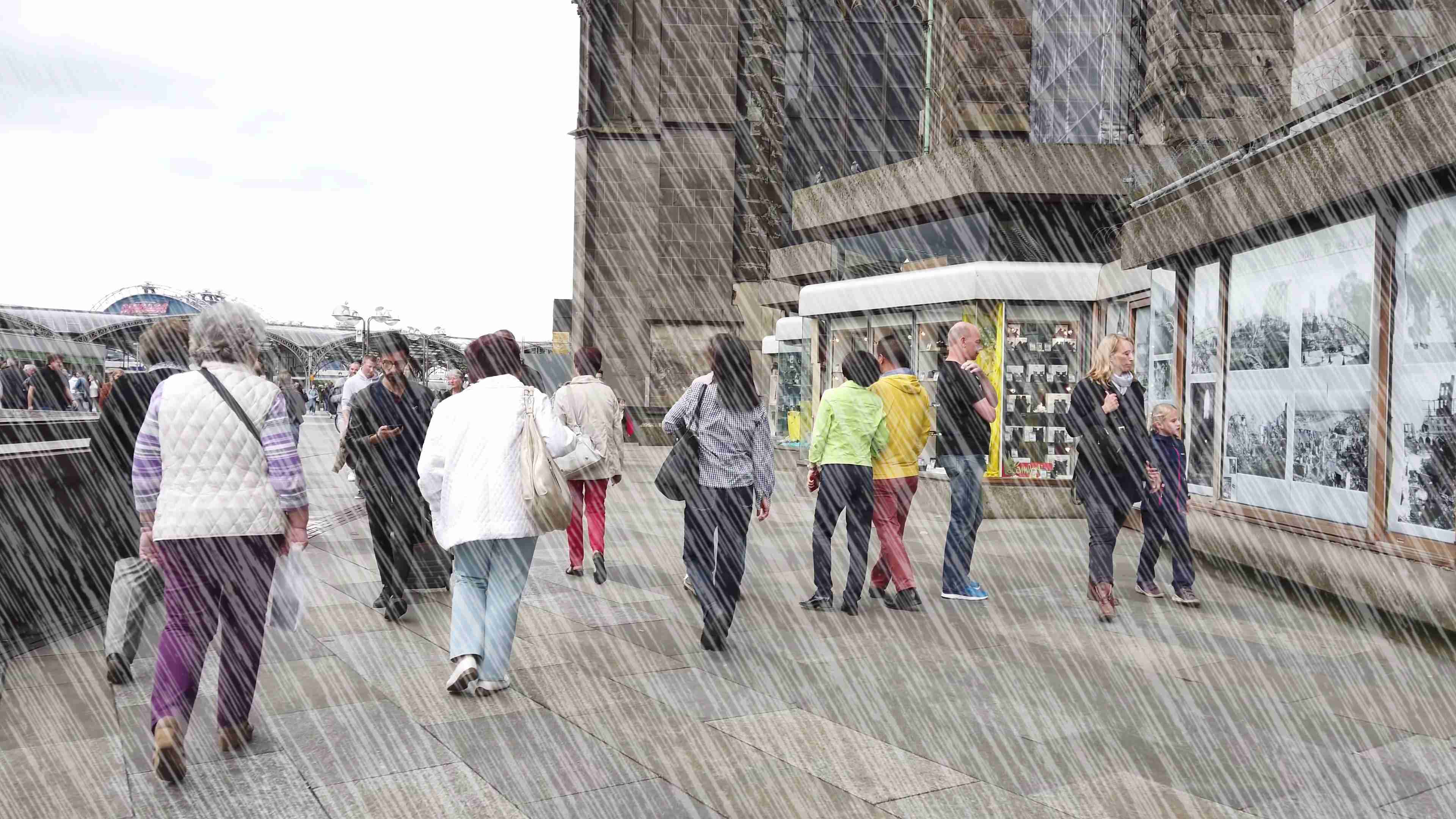} &
        \includegraphics[width=0.24\textwidth]{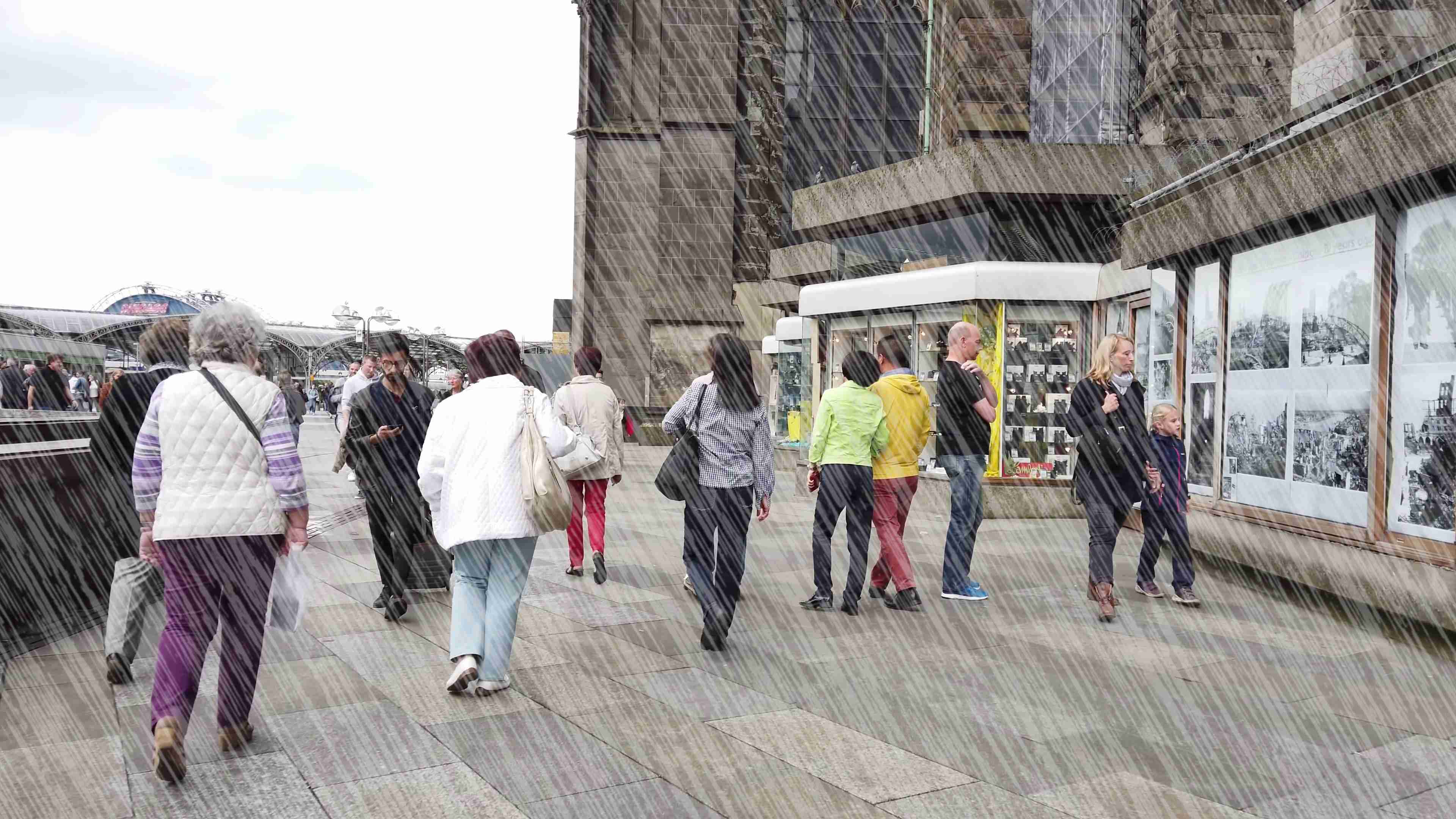} &
        \includegraphics[width=0.24\textwidth]{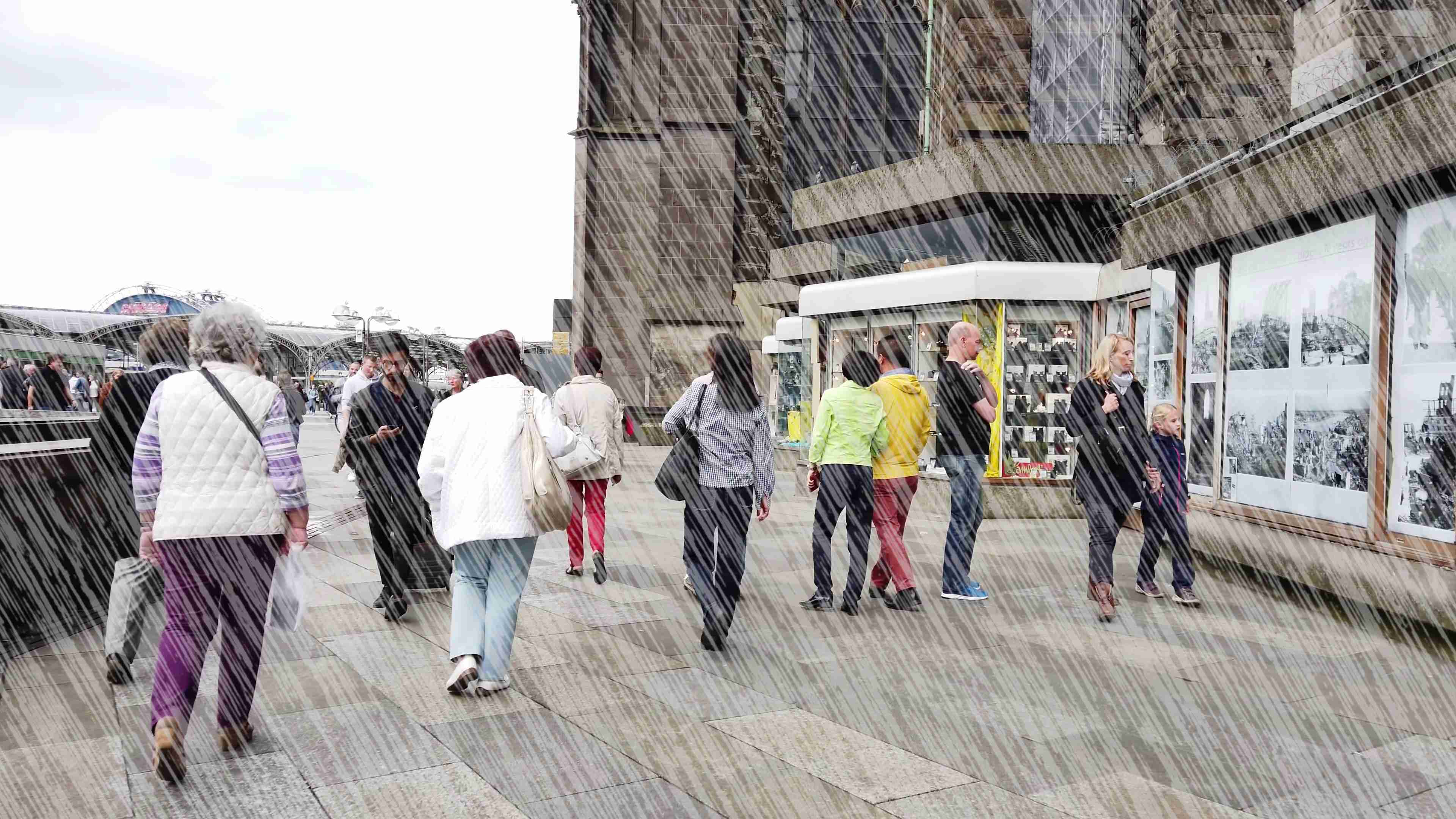} &
        \includegraphics[width=0.24\textwidth]{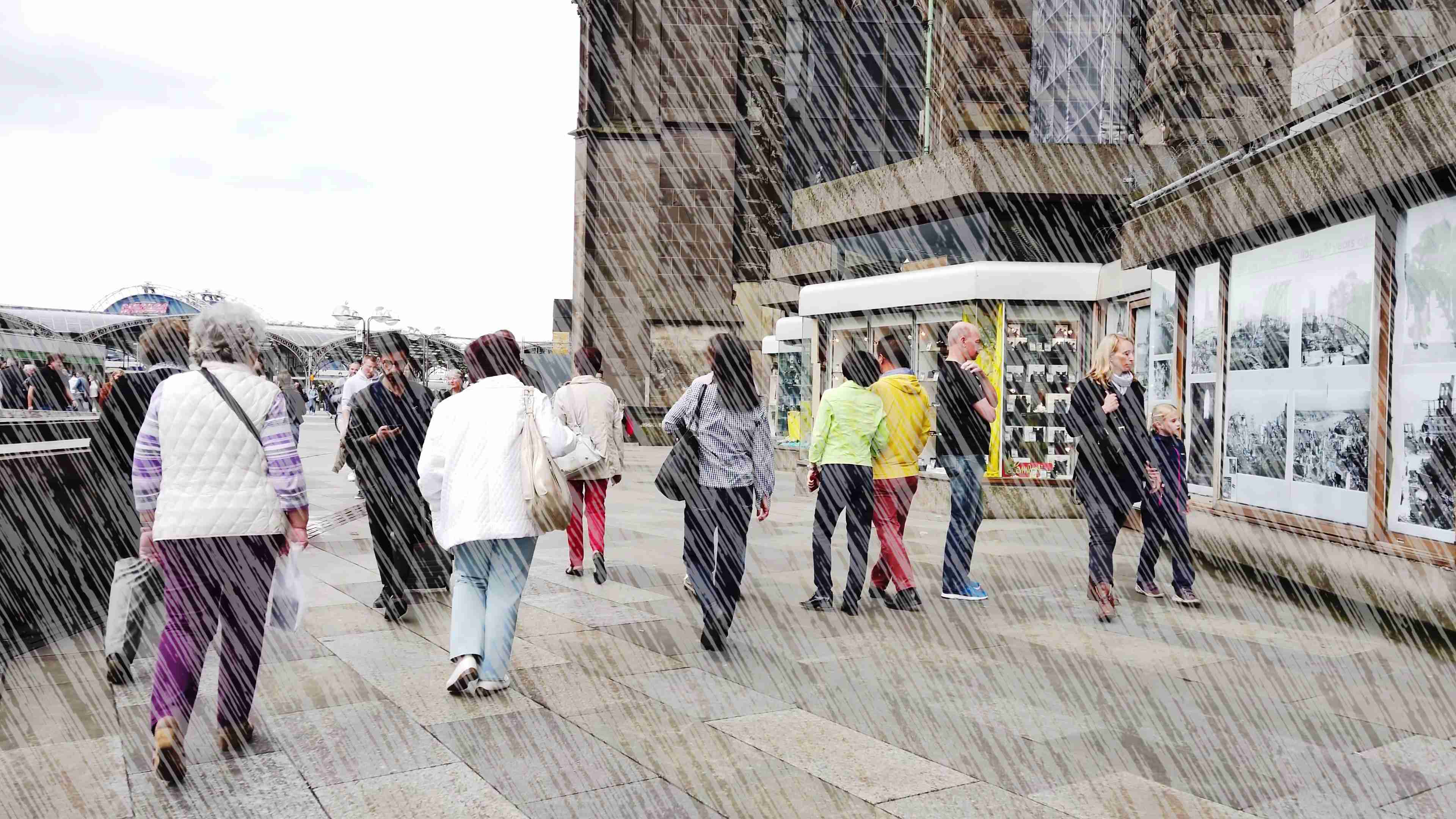} &
        \includegraphics[width=0.24\textwidth]{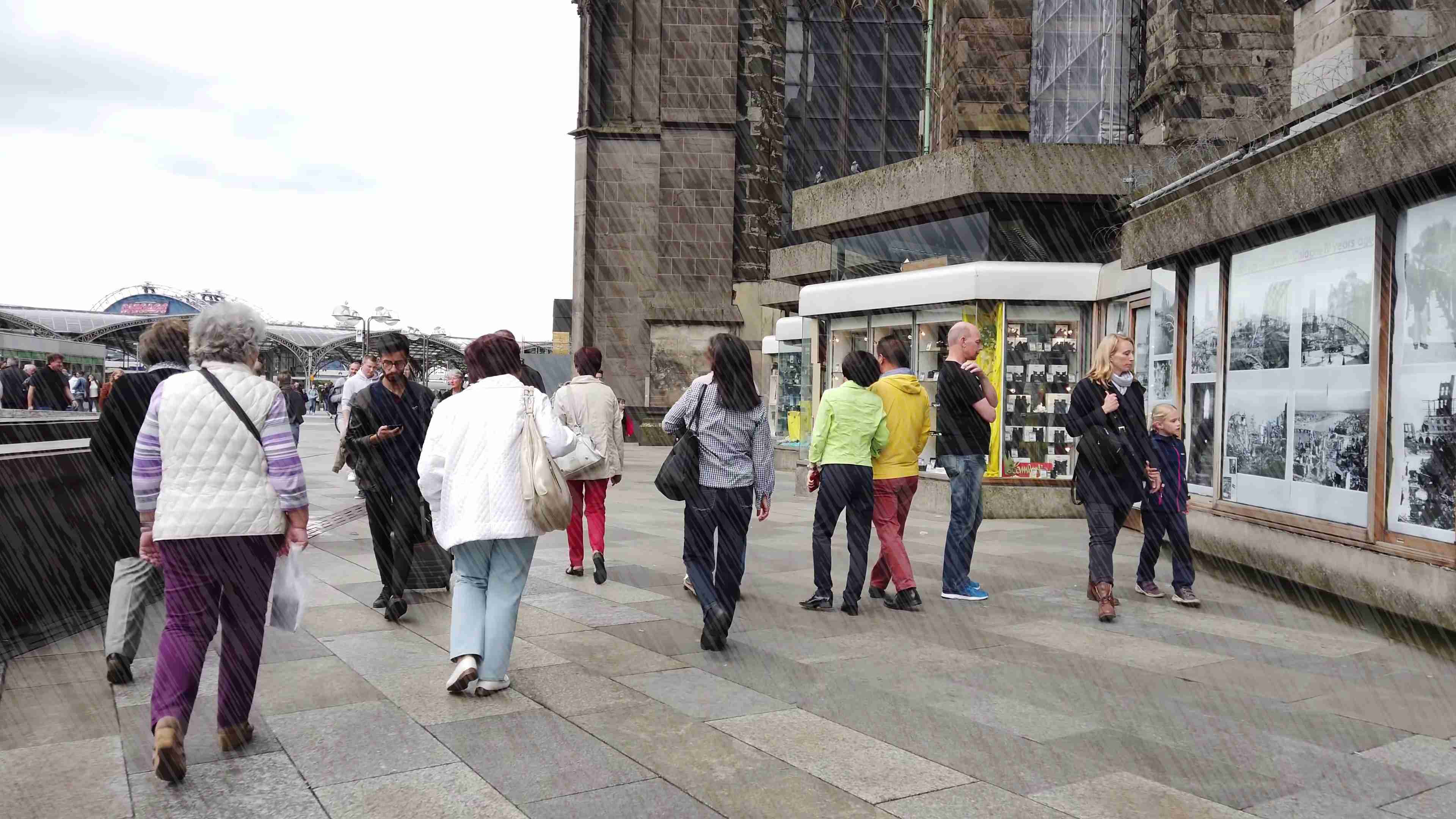} &
        \includegraphics[width=0.24\textwidth]{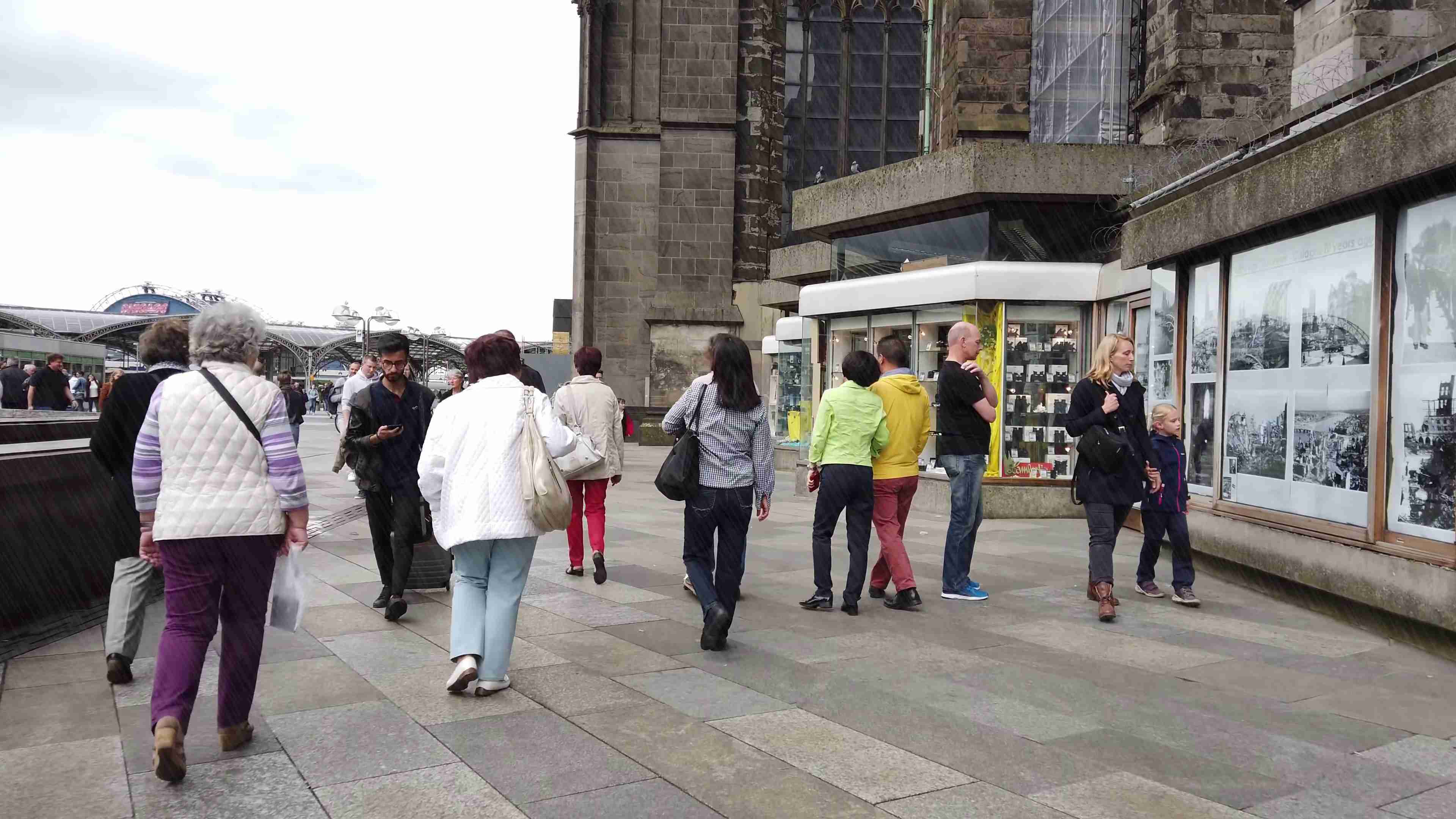}&
        \includegraphics[width=0.24\textwidth]{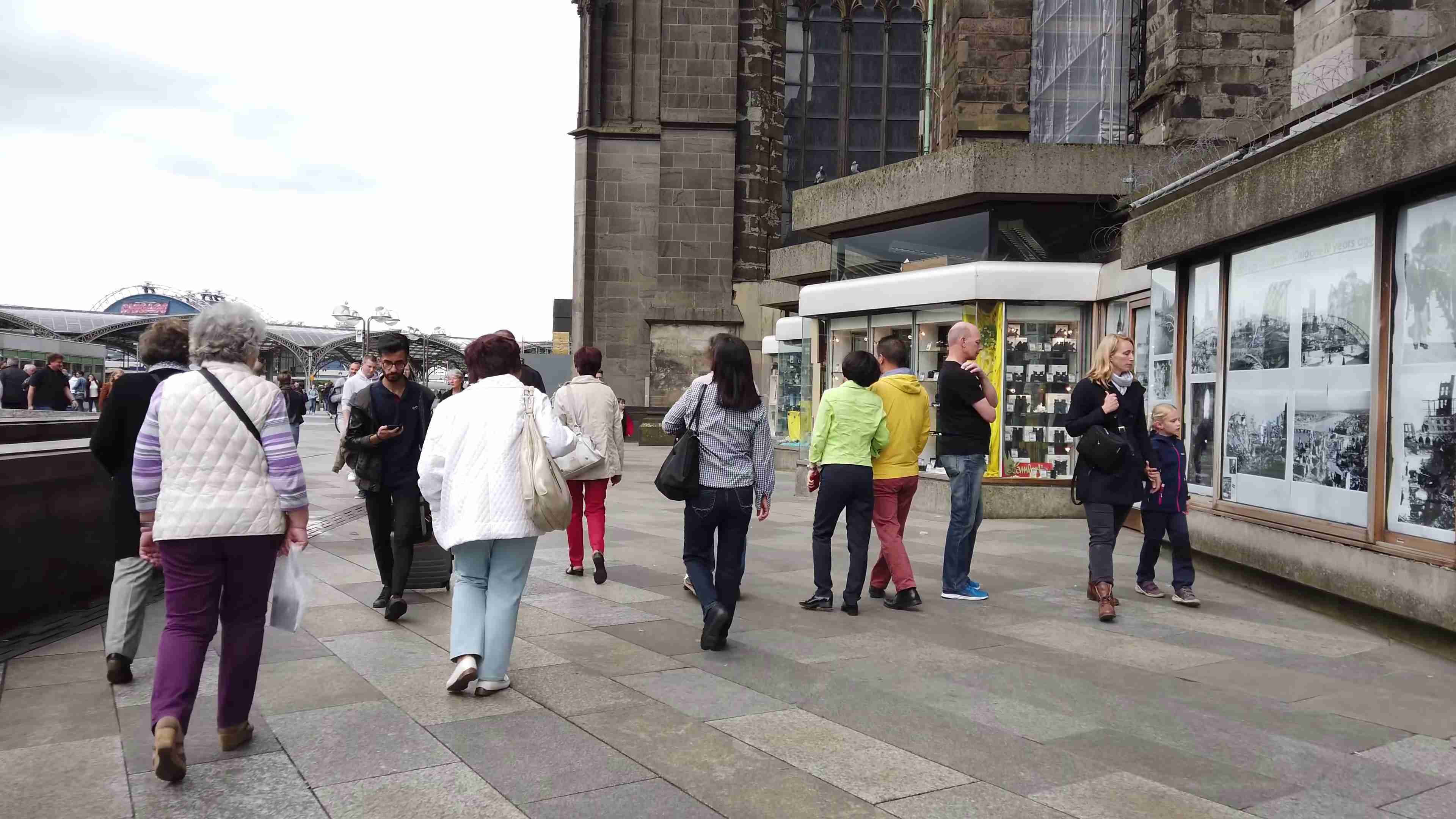} \\
        \includegraphics[width=0.24\textwidth]{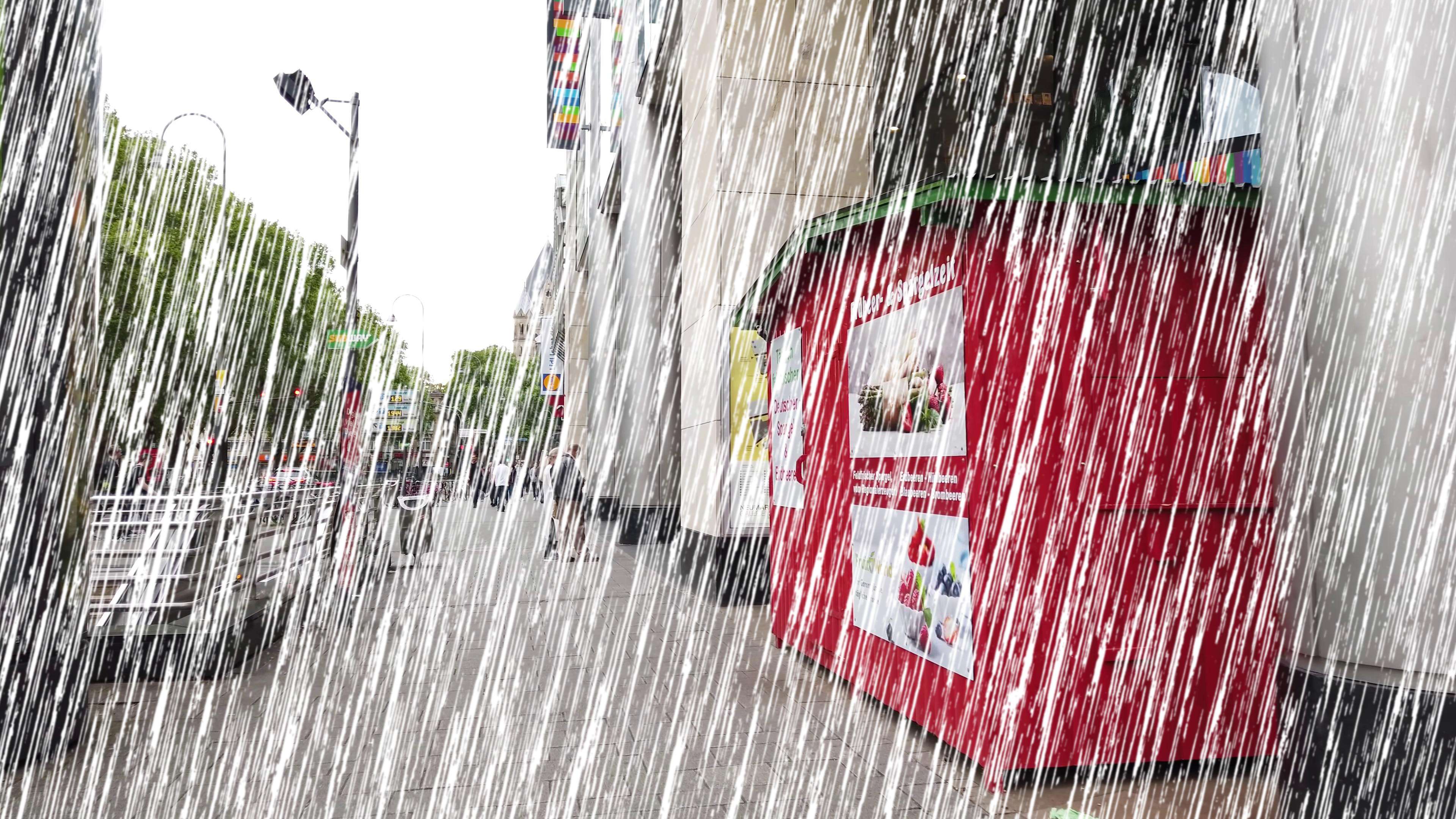} &
        \includegraphics[width=0.24\textwidth]{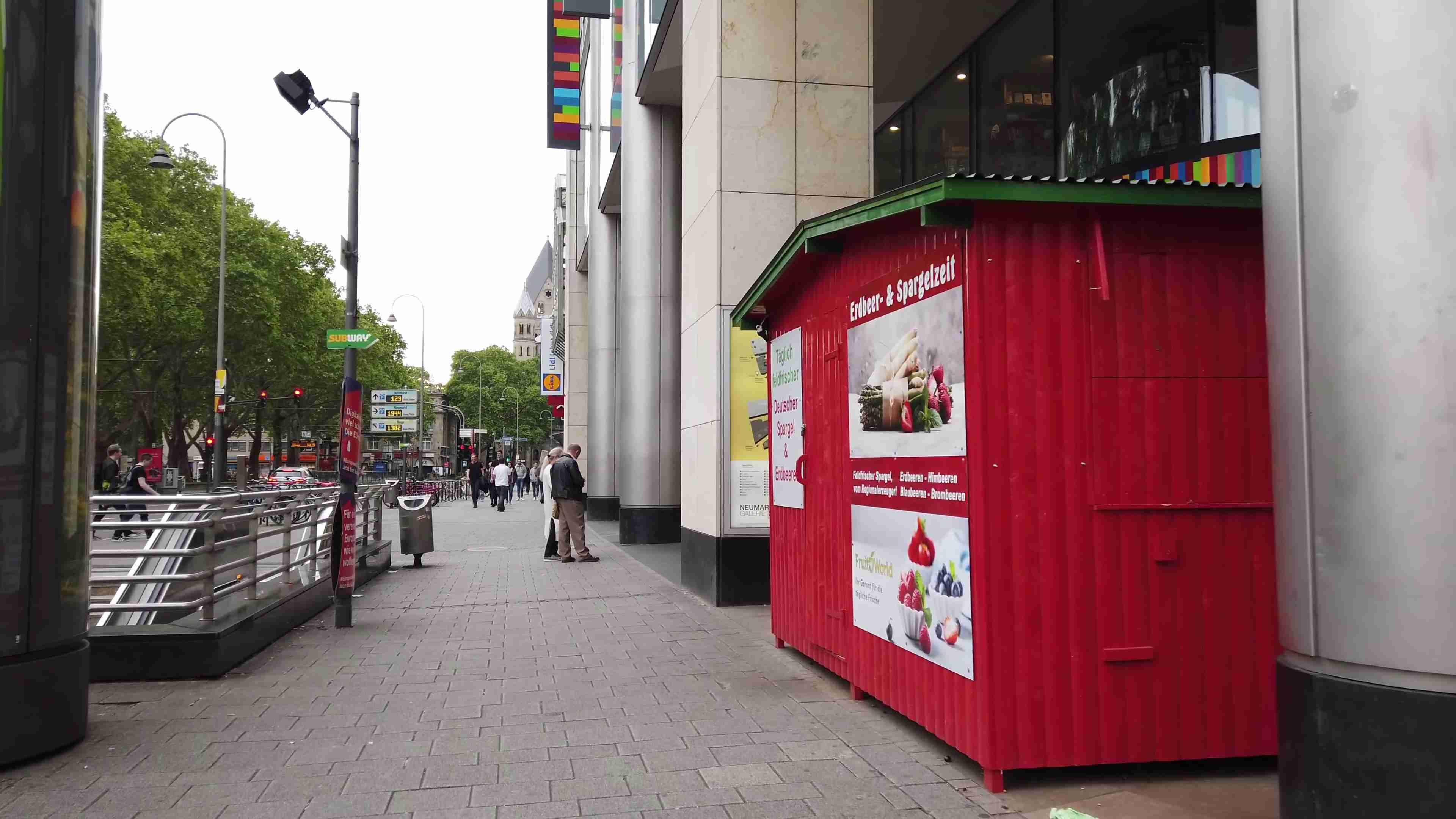} &
        \includegraphics[width=0.24\textwidth]{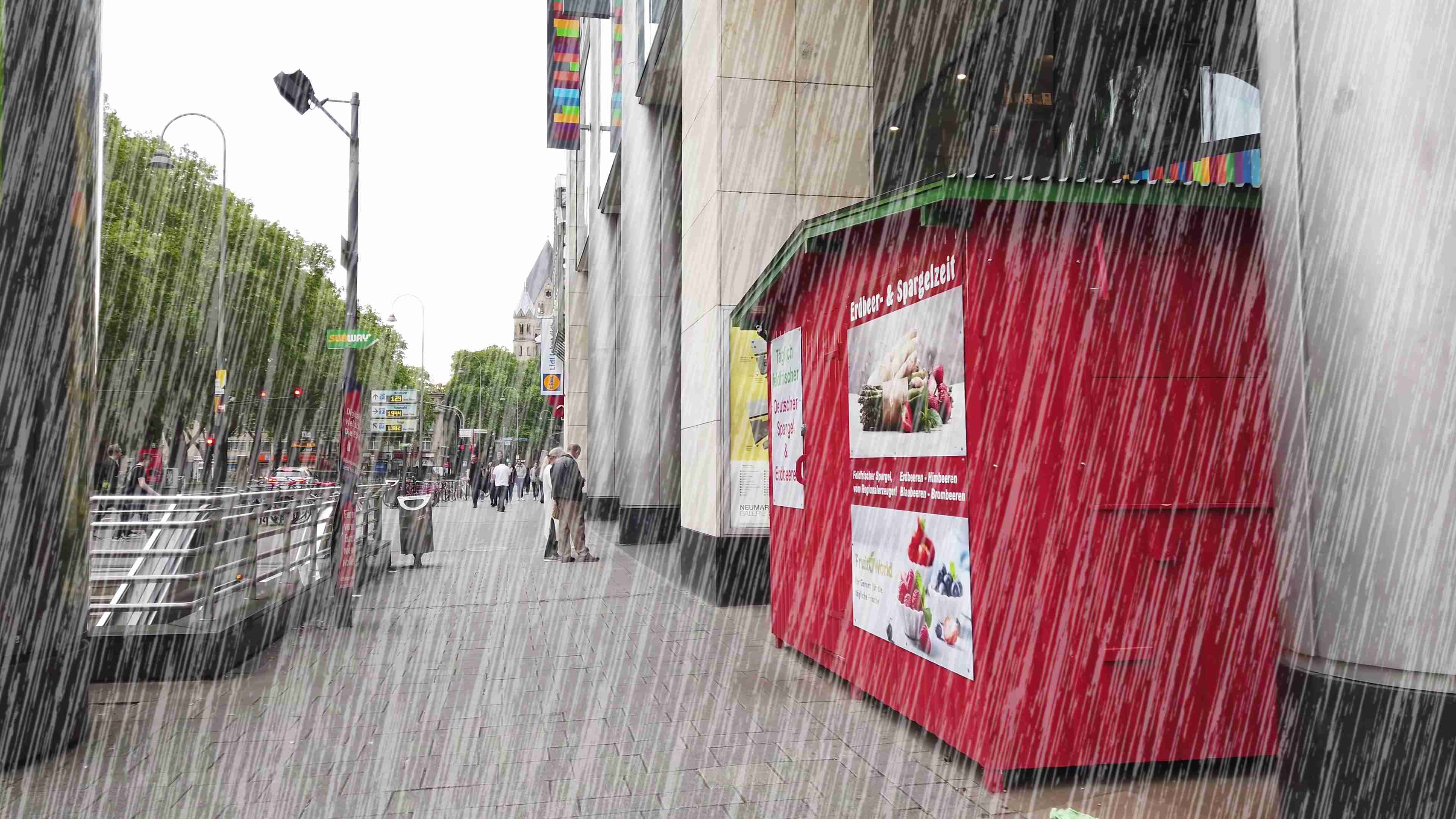} &
        \includegraphics[width=0.24\textwidth]{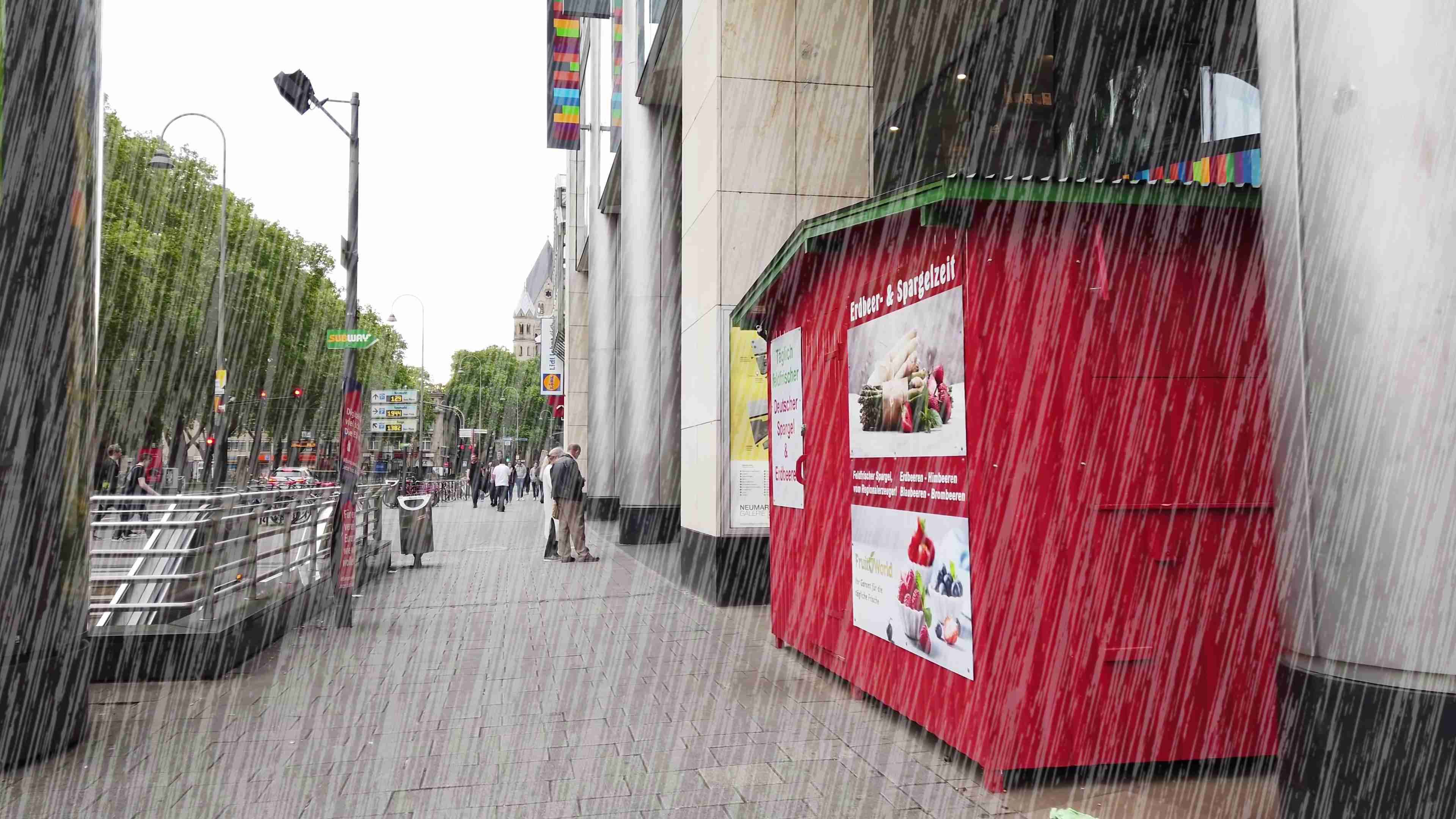} &
        \includegraphics[width=0.24\textwidth]{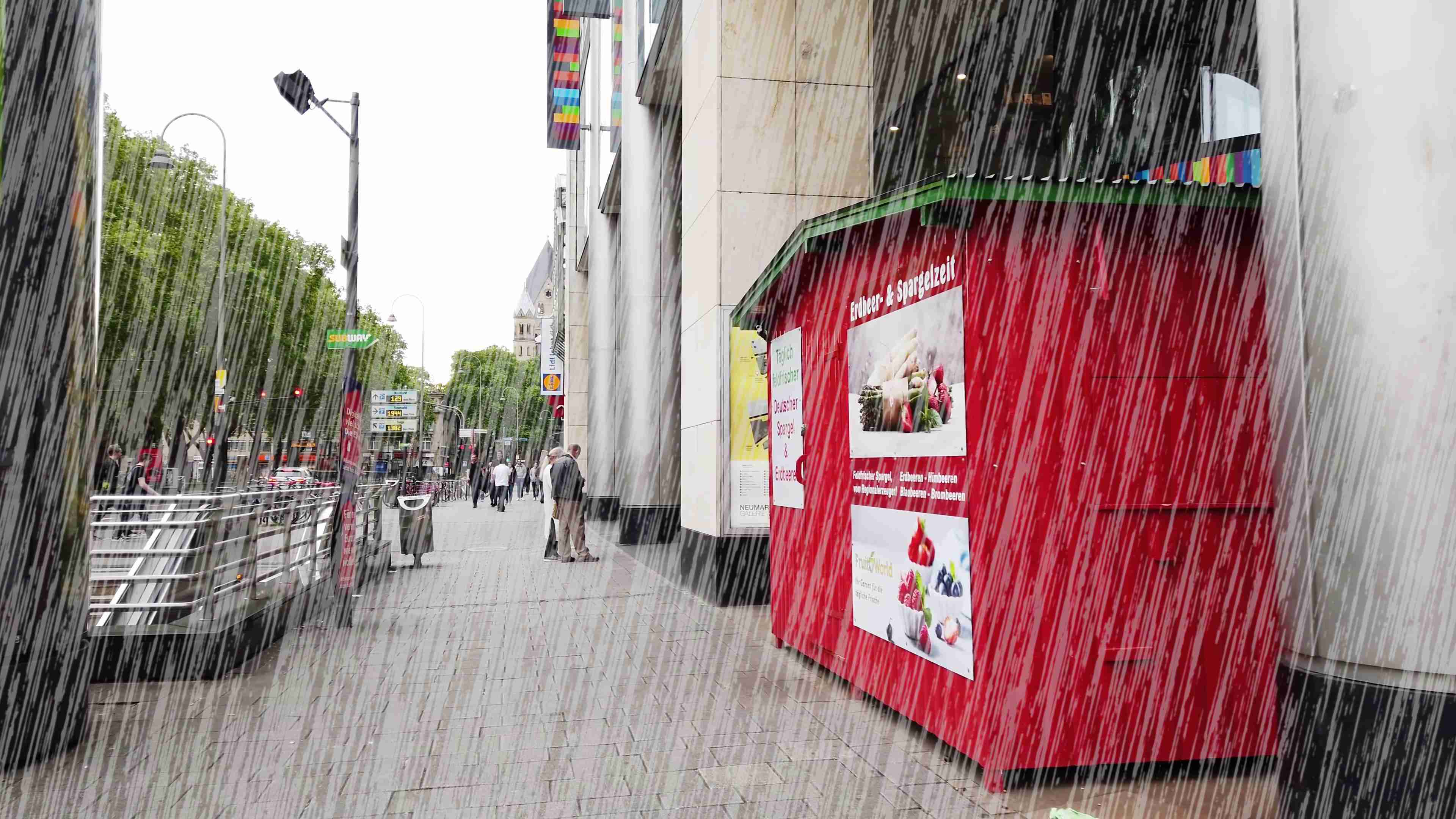} &
        \includegraphics[width=0.24\textwidth]{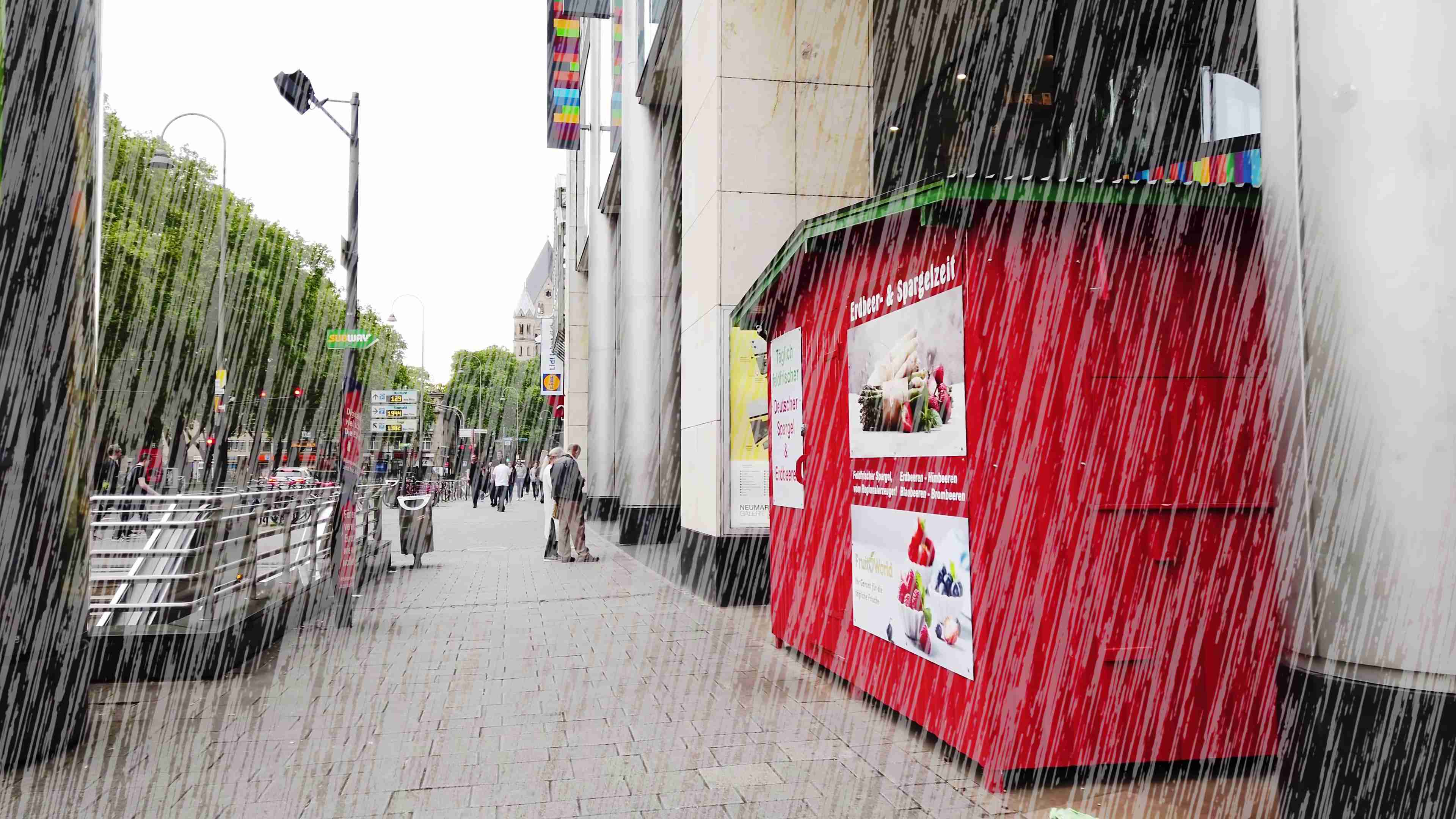} &
        \includegraphics[width=0.24\textwidth]{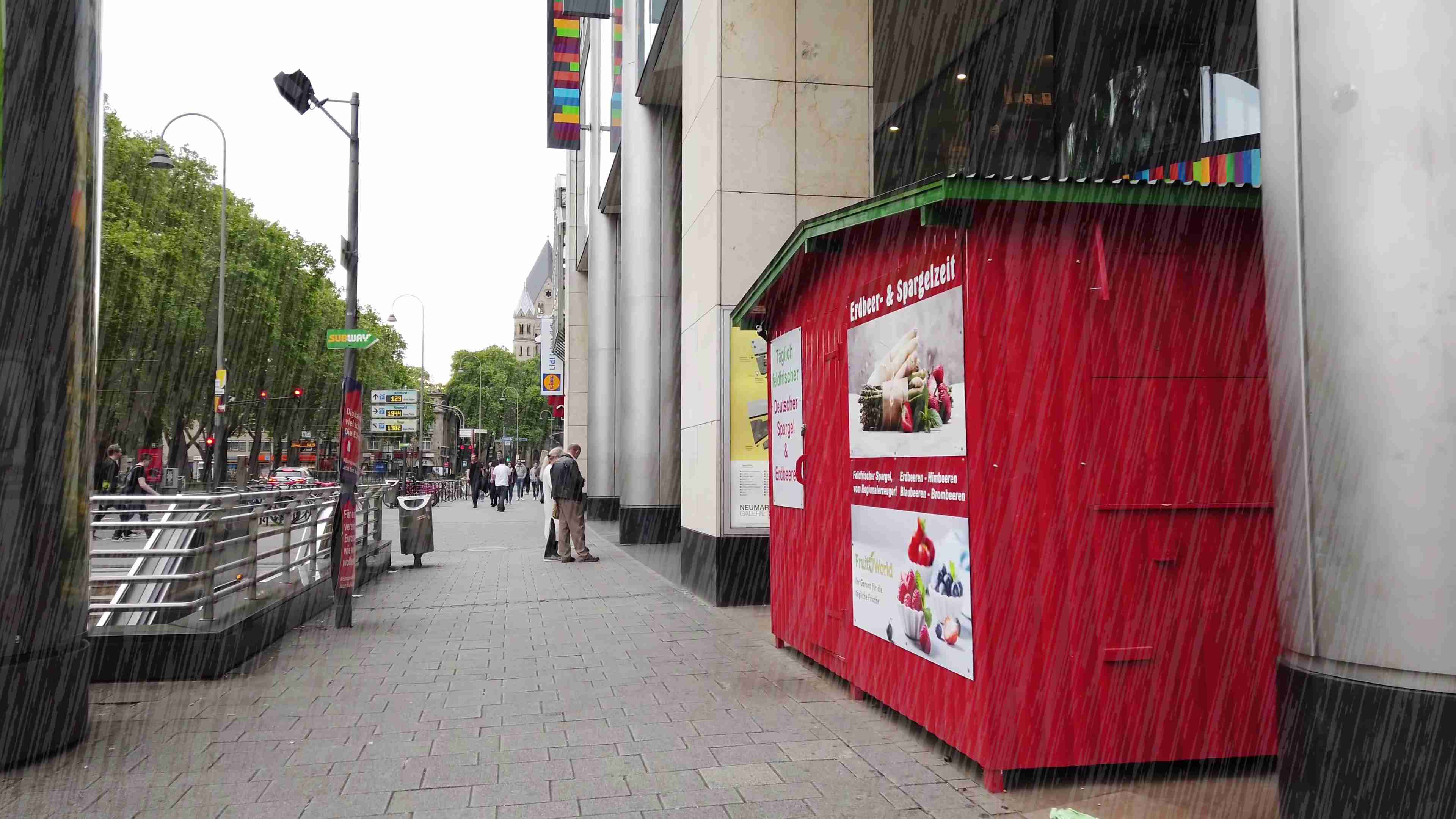} &
        \includegraphics[width=0.24\textwidth]{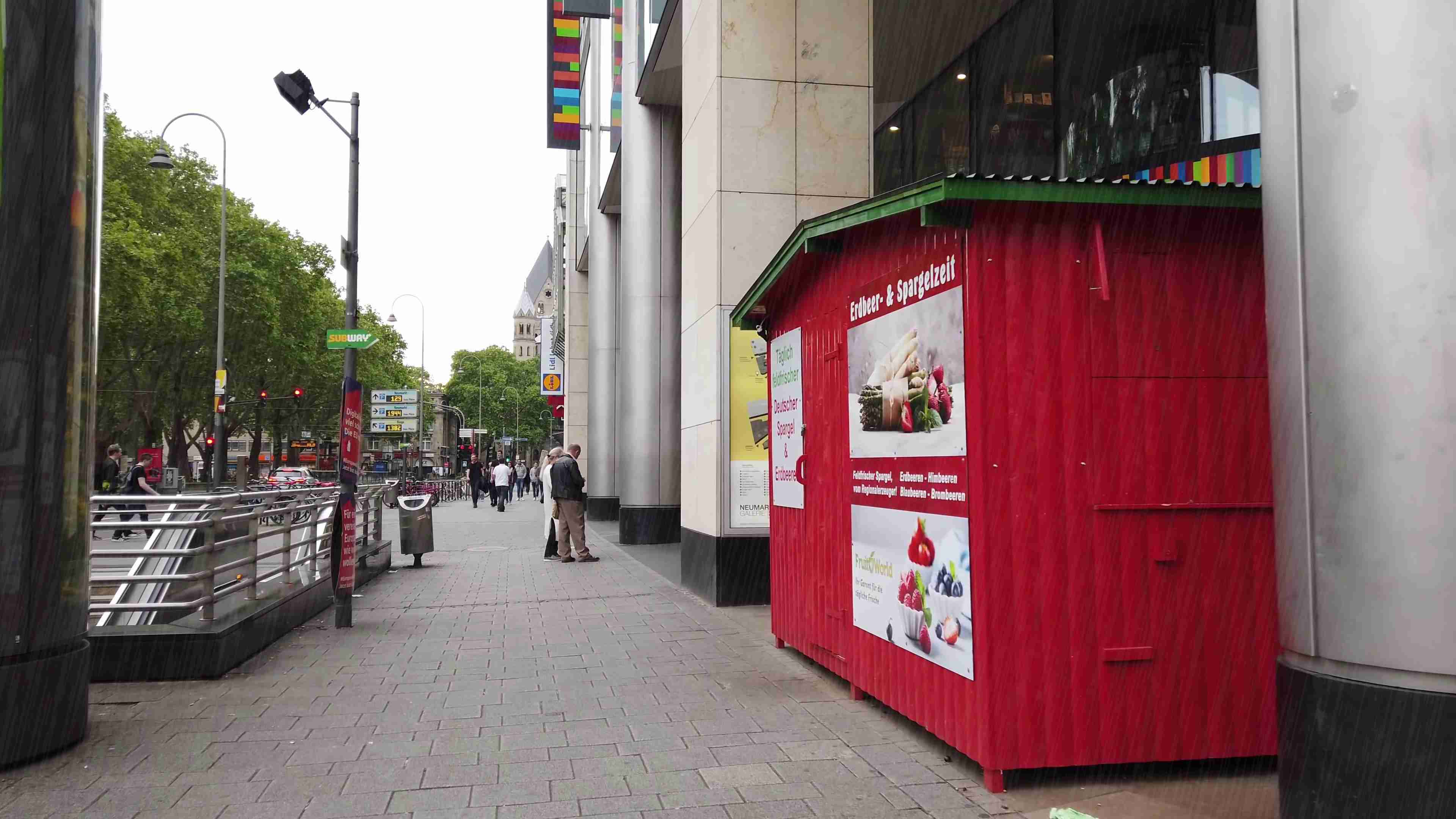}&
        \includegraphics[width=0.24\textwidth]{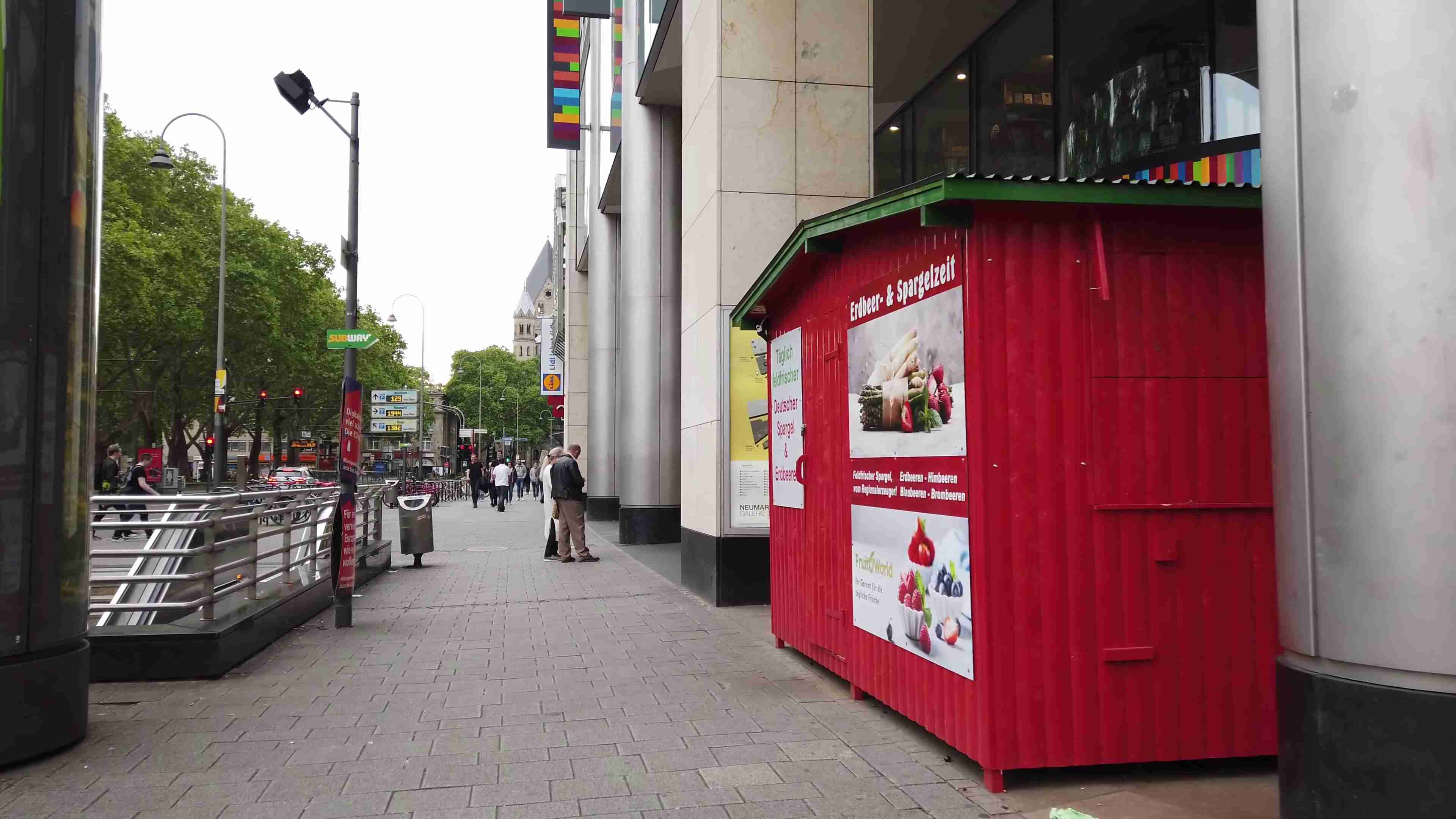} \\ 
       \huge Input &\huge GT & \huge SFNet &\huge Restormer & \huge Uformer & \huge UHD & \huge UHDformer & \huge UHDDIP & \huge Ours \\
    \end{tabular}
    \end{adjustbox}\vspace{-4mm}
    \caption{Image deraining on UHD-Rain. Visual comparisons of different methods on the UHD-Rain dataset. Our proposed TSFormer effectively removes rain streaks and haze, producing clearer images with enhanced details and minimal artifacts. }
    \label{fig: derain}\vspace{-4mm}
\end{figure*}

\noindent \textbf{Image Deraining and Desnowing Results.} We evaluate UHD image deraining with the constructed UHD-Rain dataset. The results are reported in Table \ref{tab: image deraining}. TSFormer achieves state-of-the-art performance, significantly outperforming existing methods across all key metrics.  Compared to previous models like UHDformer, Restormer, and UHDDIP, TSFormer achieves significantly higher fidelity and perceptual quality, even while handling UHD resolutions effectively.
Figure~\ref{fig: derain} illustrates visual comparisons, showing that TSFormer is more effective in removing rain streaks while preserving finer details. 
 
We implement UHD desnowing experiments on the UHD-Snow dataset, with results summarized in Table~\ref{tab:image_desnowing}, TSFormer achieves superior performance.
\begin{table}[!t]\footnotesize
\setlength{\tabcolsep}{3pt} %
\renewcommand{\arraystretch}{1.2} 
\caption{Image deblurring results on the UHD-Blur dataset. TSFormer achieves state-of-the-art performance across PSNR, SSIM, and LPIPS metrics, demonstrating its effectiveness and efficiency for UHD deblurring tasks.}
\vspace{-4mm}
\begin{center}
\begin{tabular}{l|c|ccc|l}
\shline
\textbf{Method} & \textbf{Venue} & \multicolumn{3}{c|}{\textbf{UHD-Blur}} & \textbf{Param} \\
 & & \textbf{PSNR} $\uparrow$ & \textbf{SSIM} $\uparrow$ & \textbf{LPIPS} $\downarrow$ \\
\shline
Restormer & CVPR'22 & 25.21 & 0.752 & 0.370 & 26.10M \\
Uformer & CVPR'22 & 25.27 & 0.752 & 0.385 & 20.60M \\
Stripformer & ECCV'22 & 25.05 & 0.750 & 0.374 & 19.70M \\
FFTformer & CVPR'23 & 25.41 & 0.757 & 0.371 & 16.60M \\
UHDformer & AAAI'24 & 28.82 & 0.844 & 0.235 & 0.34M \\
UHDDIP & arxiv'24 & \cellcolor{secondbest}{29.51} & \cellcolor{secondbest}{0.859} & \cellcolor{secondbest}{0.213} & 0.81M \\
TSFormer (Ours) & - & \cellcolor{best}\textbf{29.52} & \cellcolor{best}\textbf{0.861} & \cellcolor{best}\textbf{0.203} & 3.38M \\
\shline
\end{tabular}
\end{center}
\label{tab: image deblurring.}
\vspace{-6mm}
\end{table}
\begin{table}[!t]\footnotesize
\setlength{\tabcolsep}{3pt} %
\renewcommand{\arraystretch}{1.2} 
\caption{Quantitative evaluations on the GoPro dataset.}
\vspace{-6mm}
\begin{center}
\begin{tabular}{l|c|cc}
\shline
\textbf{Method} & \textbf{Venue} & \multicolumn{2}{c}{\textbf{GoPro}} \\
 & & \textbf{PSNR} $\uparrow$ & \textbf{SSIM} $\uparrow$ \\
\shline
MIMO-Unet++&ICCV'21&32.45 & 0.956  \\
MPRNet&CVPR'21&32.66 & 0.959  \\
Restormer&CVPR'22& 32.92 & 0.961 \\
Stripformer&ECCV'22&33.08 & 0.962 \\
FFTformer&CVPR'23& \cellcolor{secondbest}34.21 & \cellcolor{secondbest}0.969 \\
TSFormer (Ours) & - & \cellcolor{best}\textbf{34.37} & \cellcolor{best}\textbf{0.977} \\
\shline
\end{tabular}
\end{center}
\label{tab:gopro}
\vspace{-7mm}
\end{table}
\begin{figure*}[!t]
    \centering
    \begin{adjustbox}{max width=\textwidth}
    \begin{tabular}{ccccccccc}
        \includegraphics[width=0.24\textwidth]{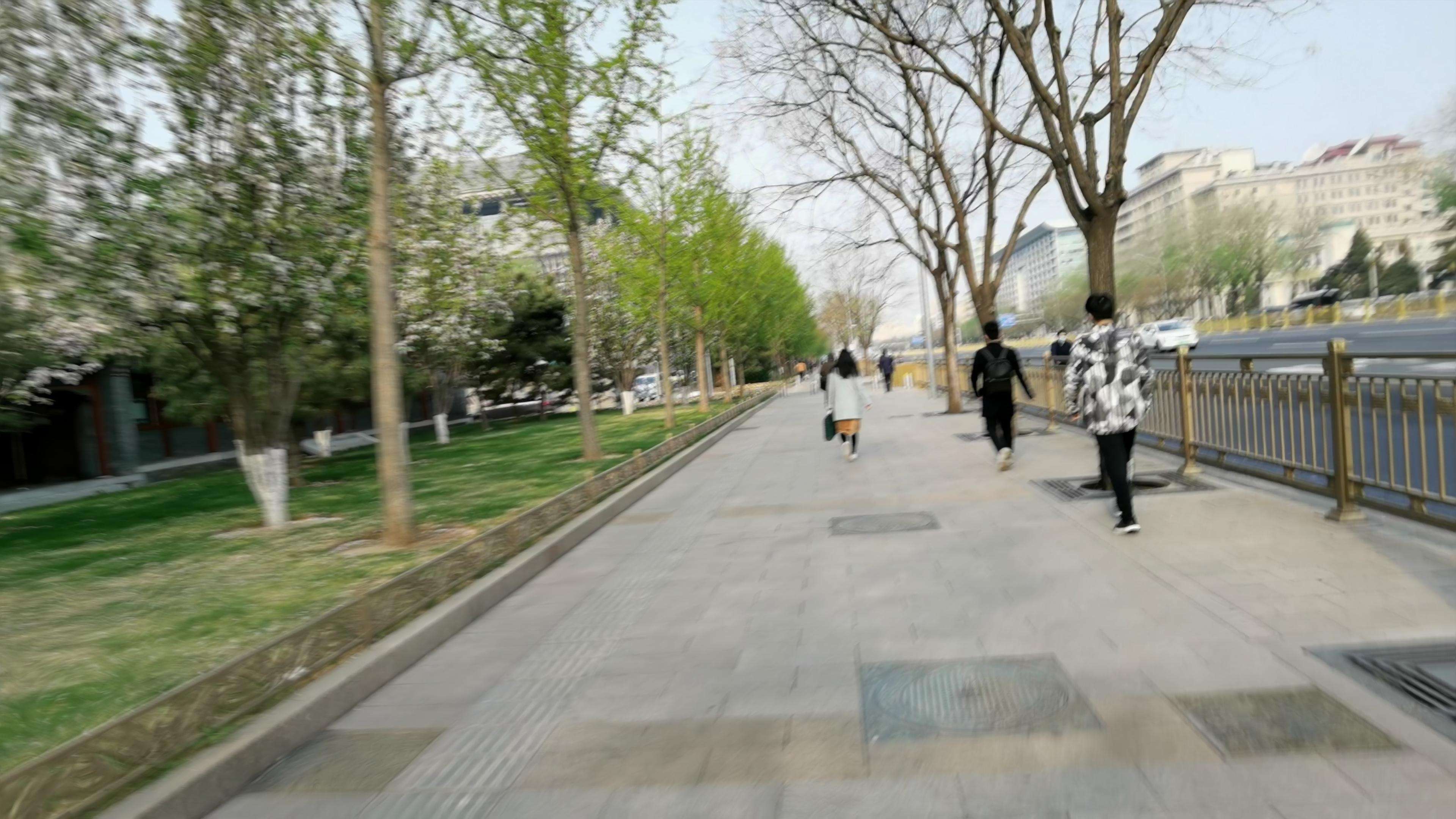} &
        \includegraphics[width=0.24\textwidth]{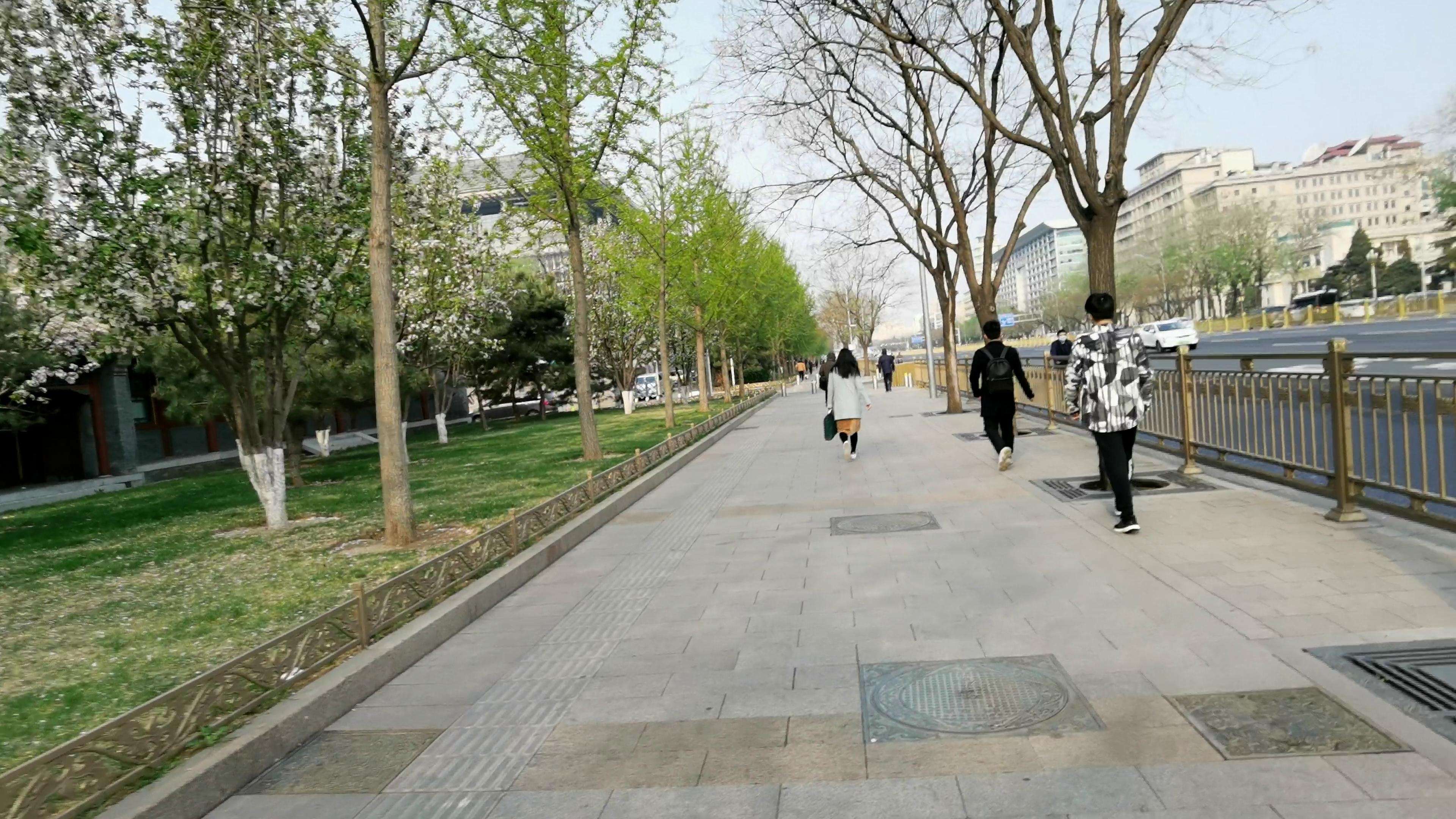} &
        \includegraphics[width=0.24\textwidth]{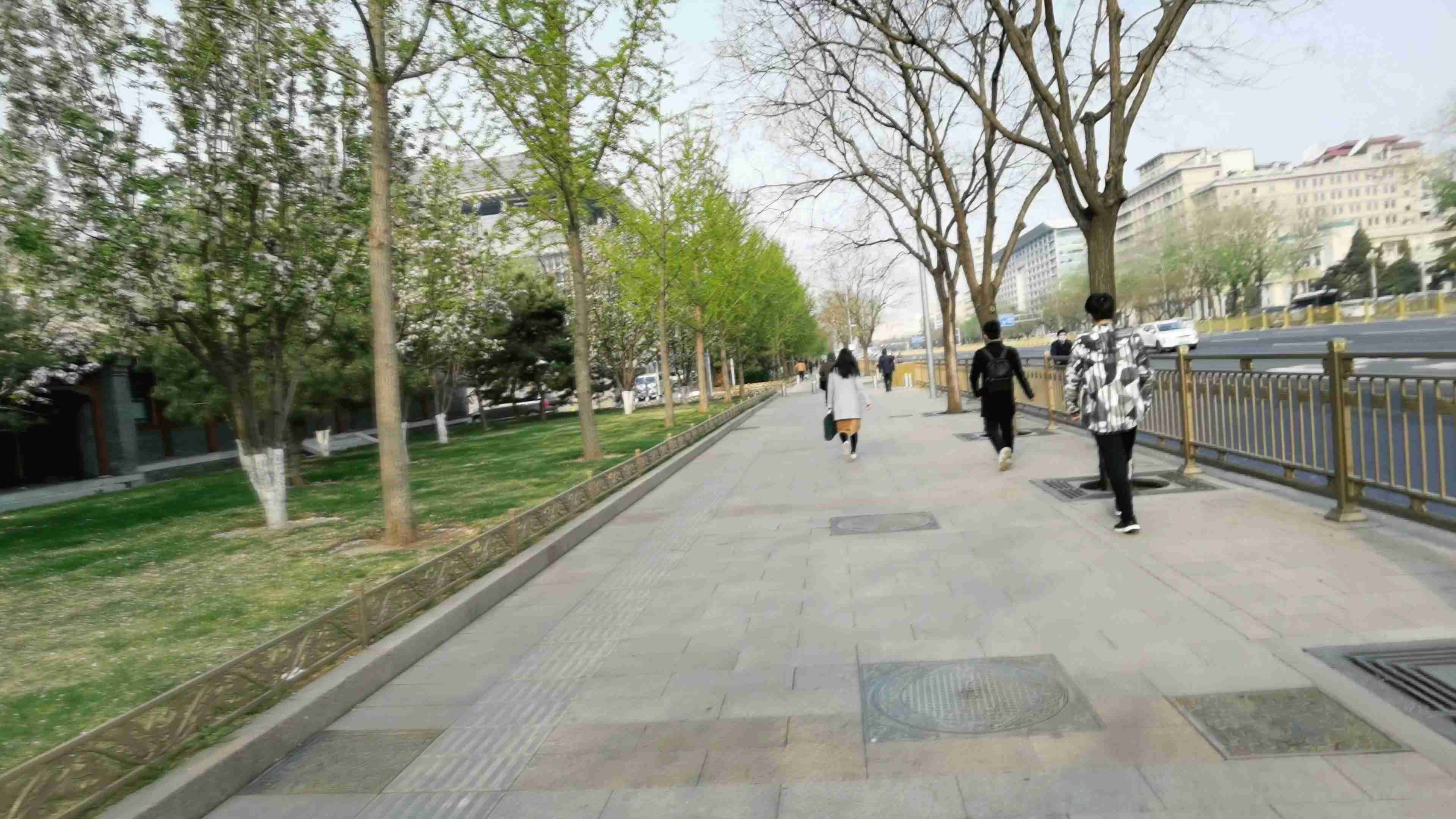} &
        \includegraphics[width=0.24\textwidth]{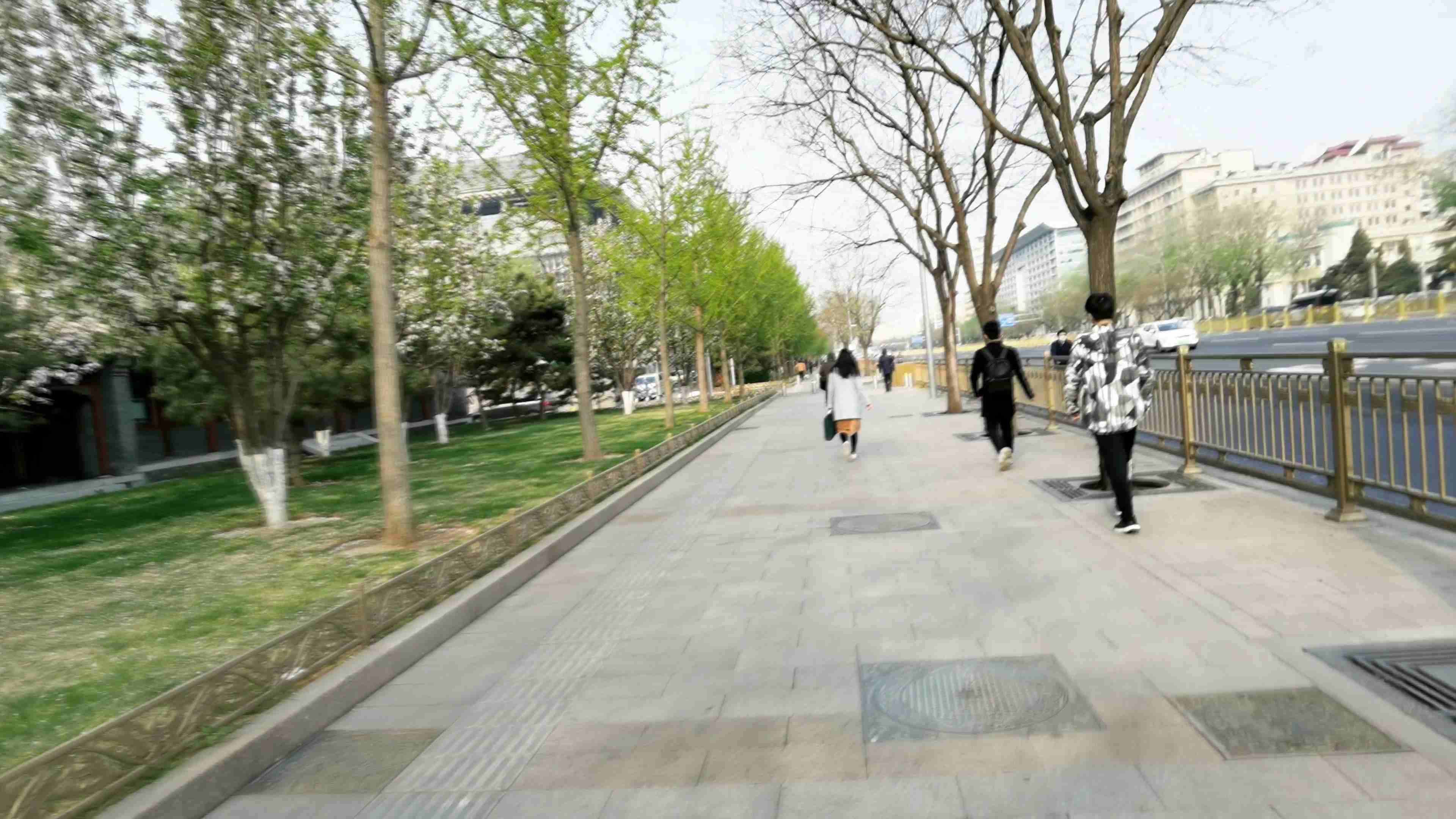} &
        \includegraphics[width=0.24\textwidth]{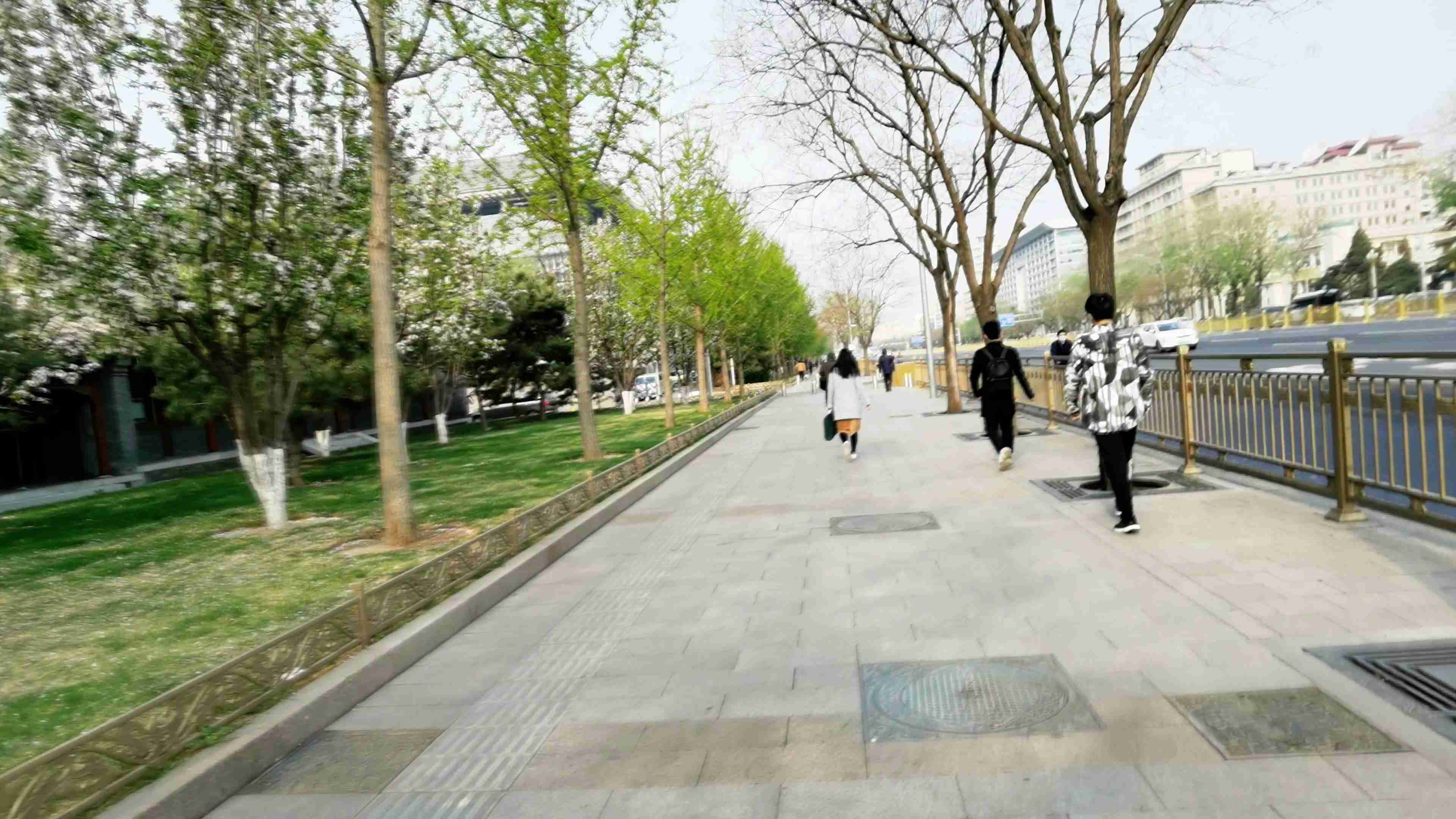} &
        \includegraphics[width=0.24\textwidth]{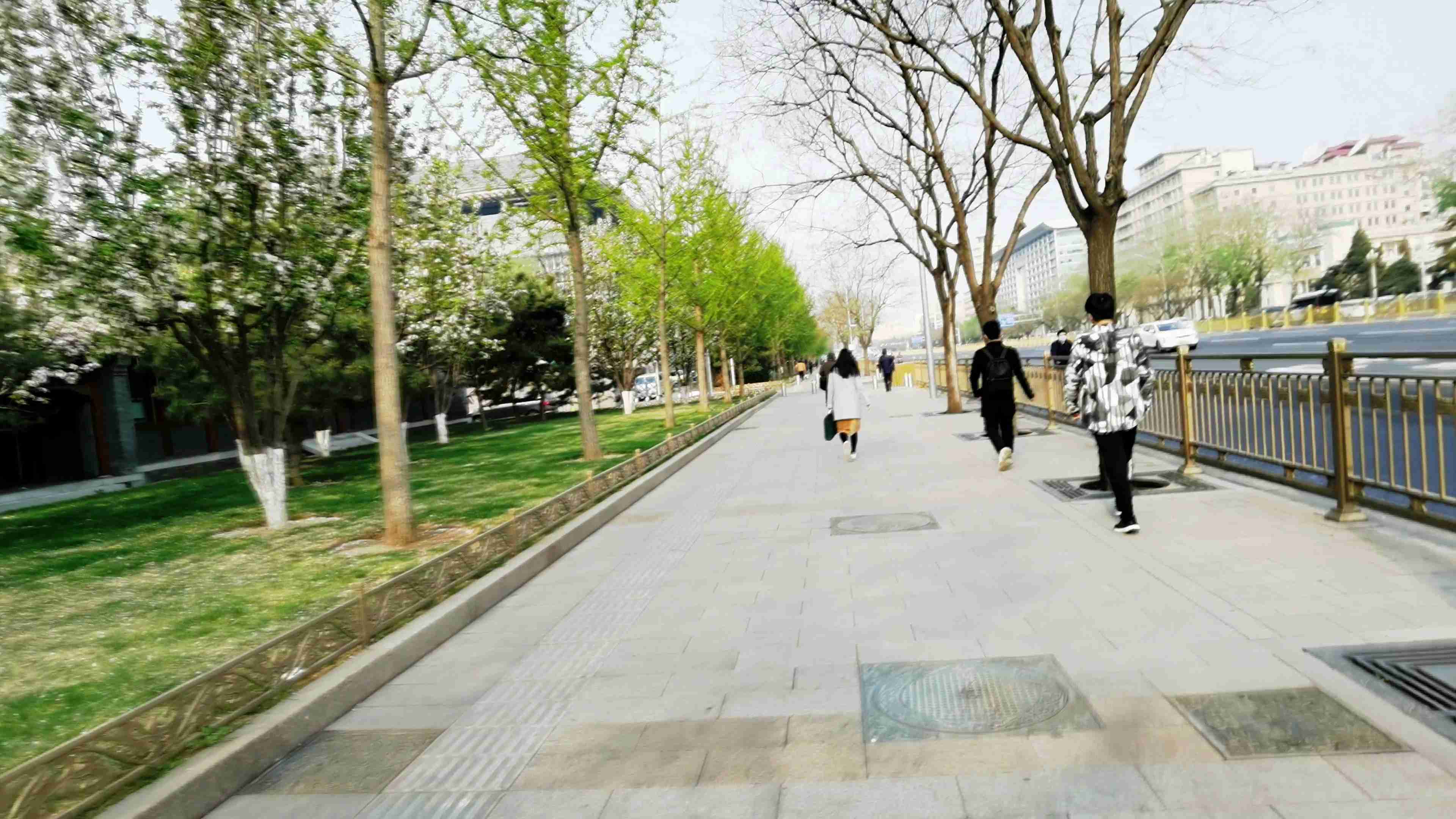} &
        \includegraphics[width=0.24\textwidth]{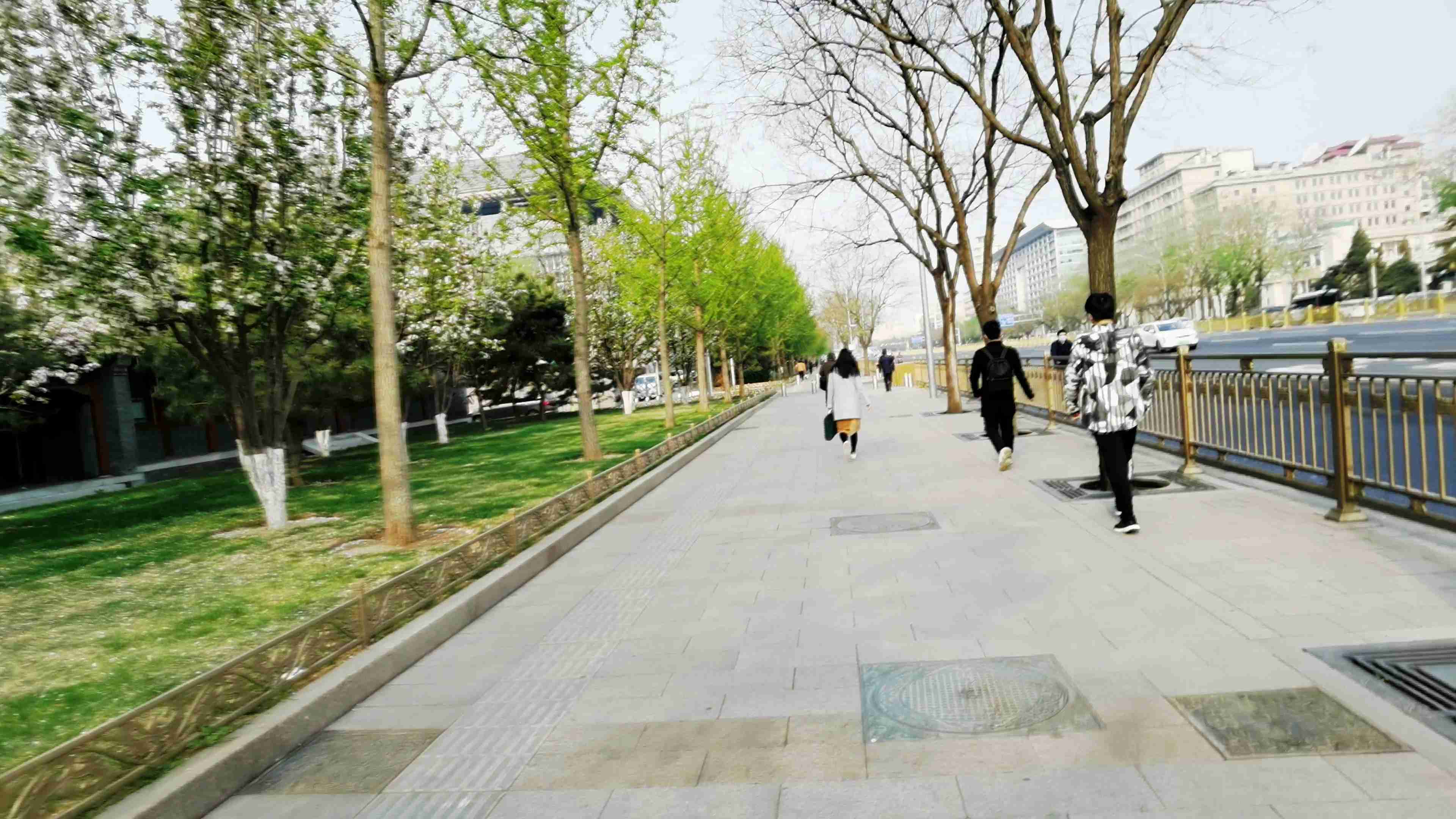} &
        \includegraphics[width=0.24\textwidth]{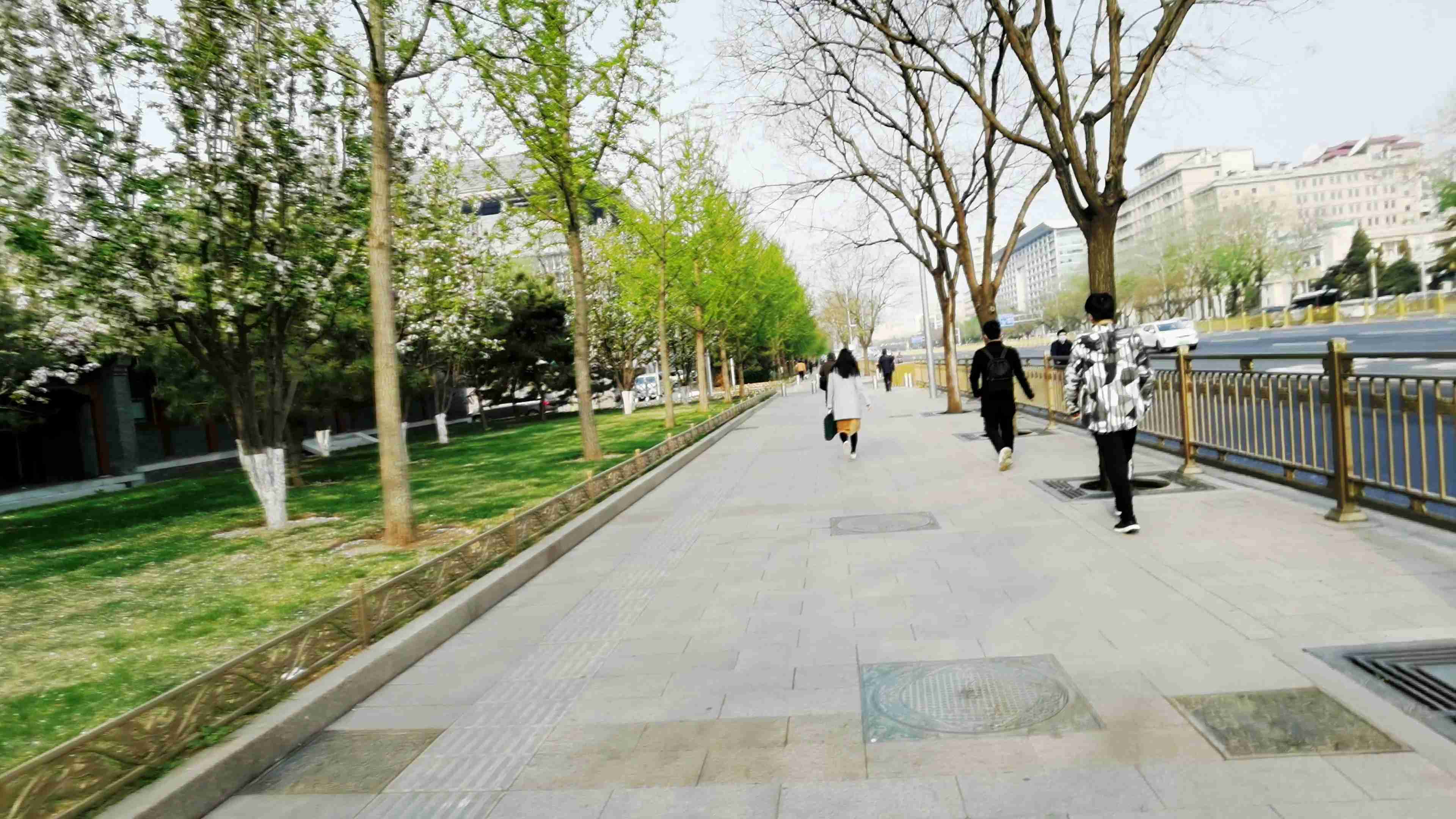}&
        \includegraphics[width=0.24\textwidth]{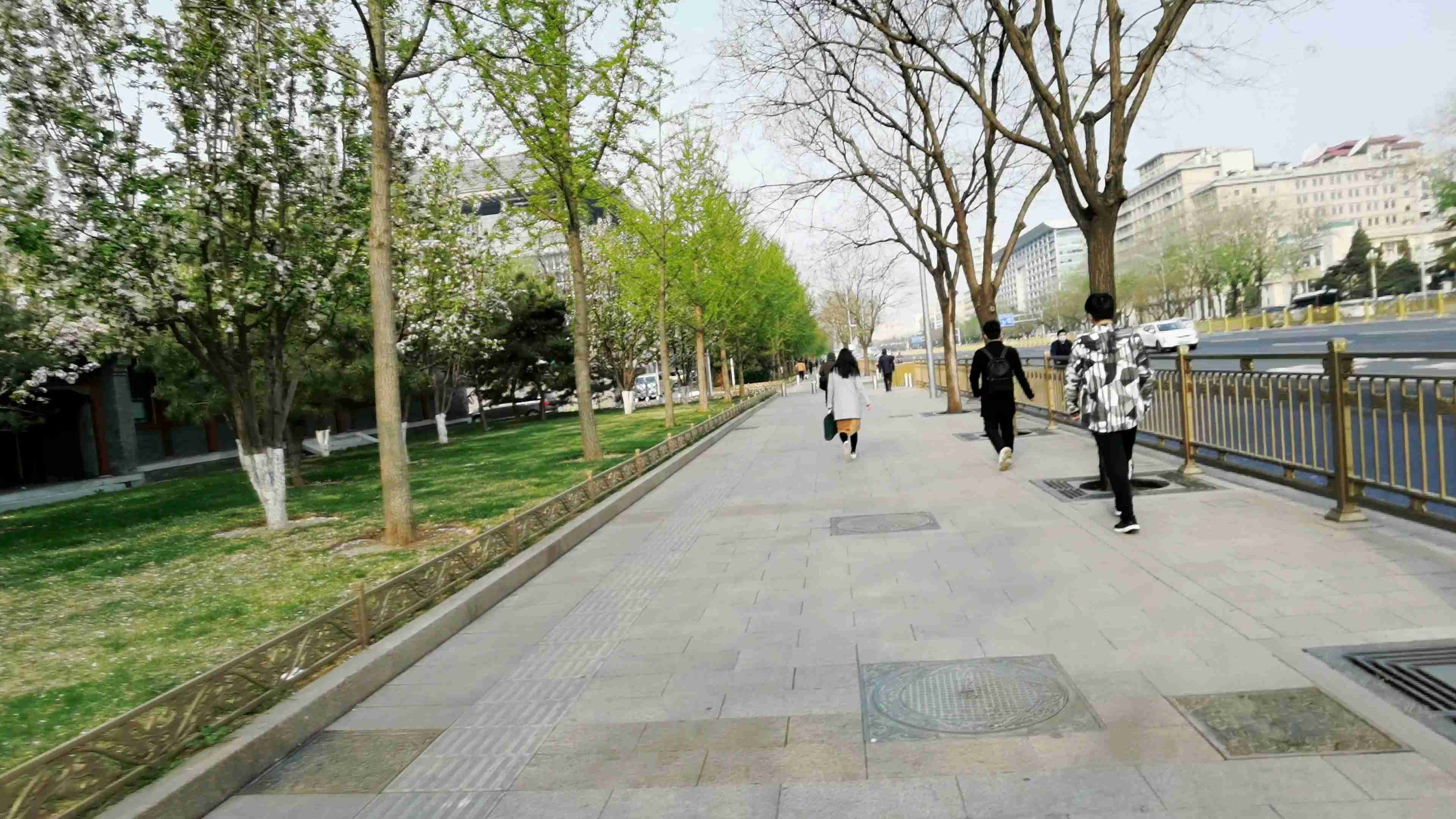} \\
        \includegraphics[width=0.24\textwidth]{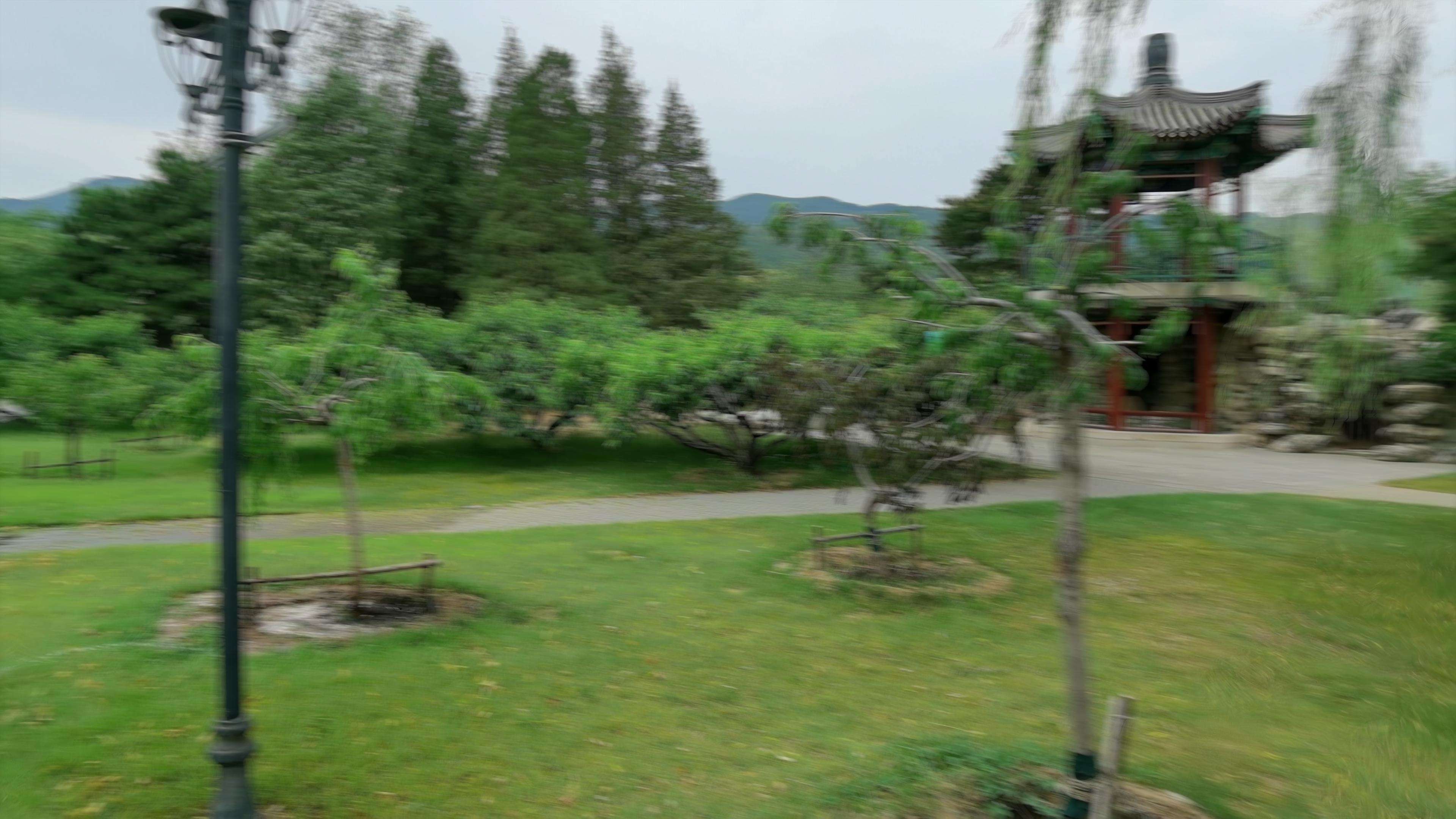} &
        \includegraphics[width=0.24\textwidth]{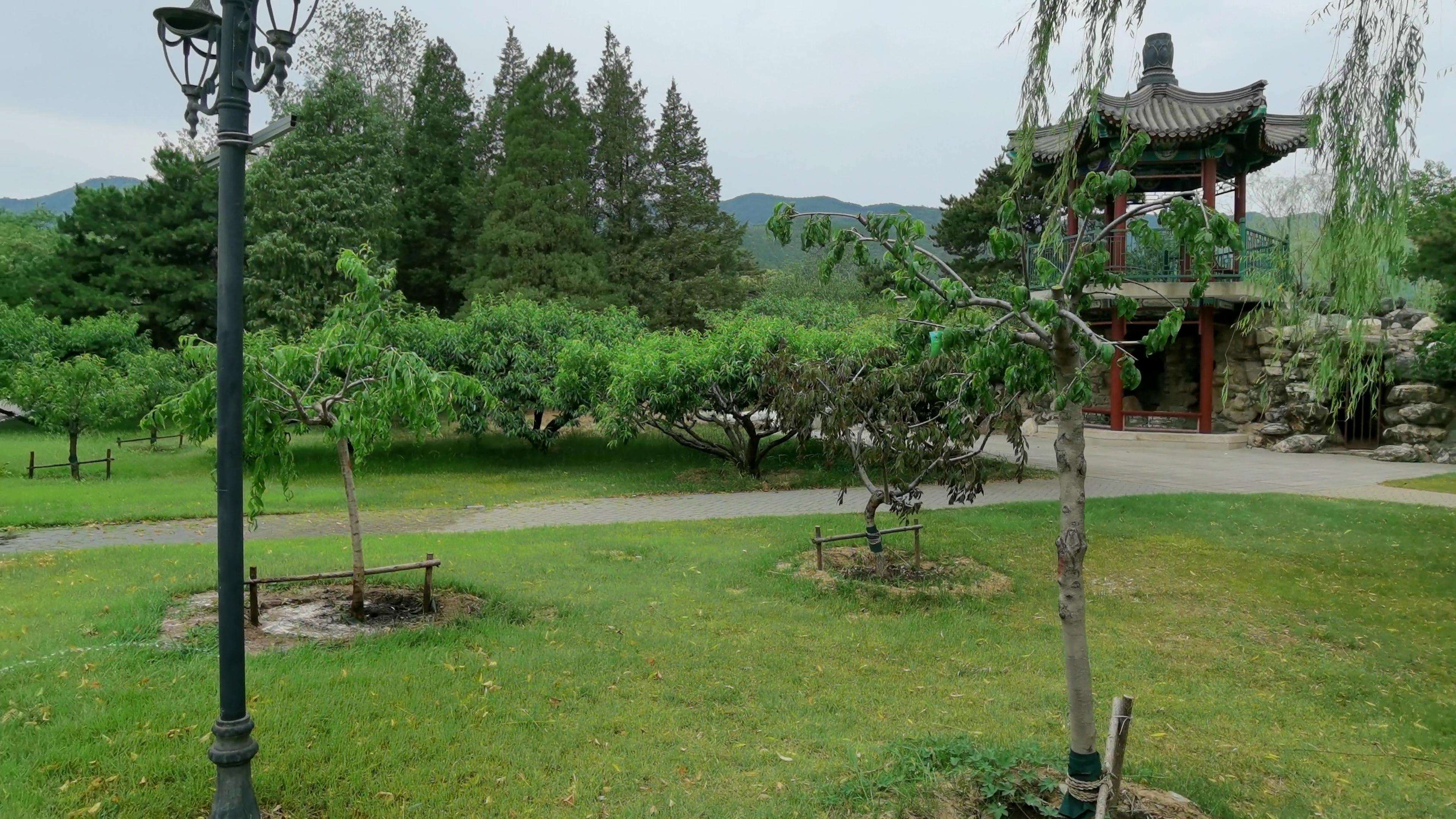} &
        \includegraphics[width=0.24\textwidth]{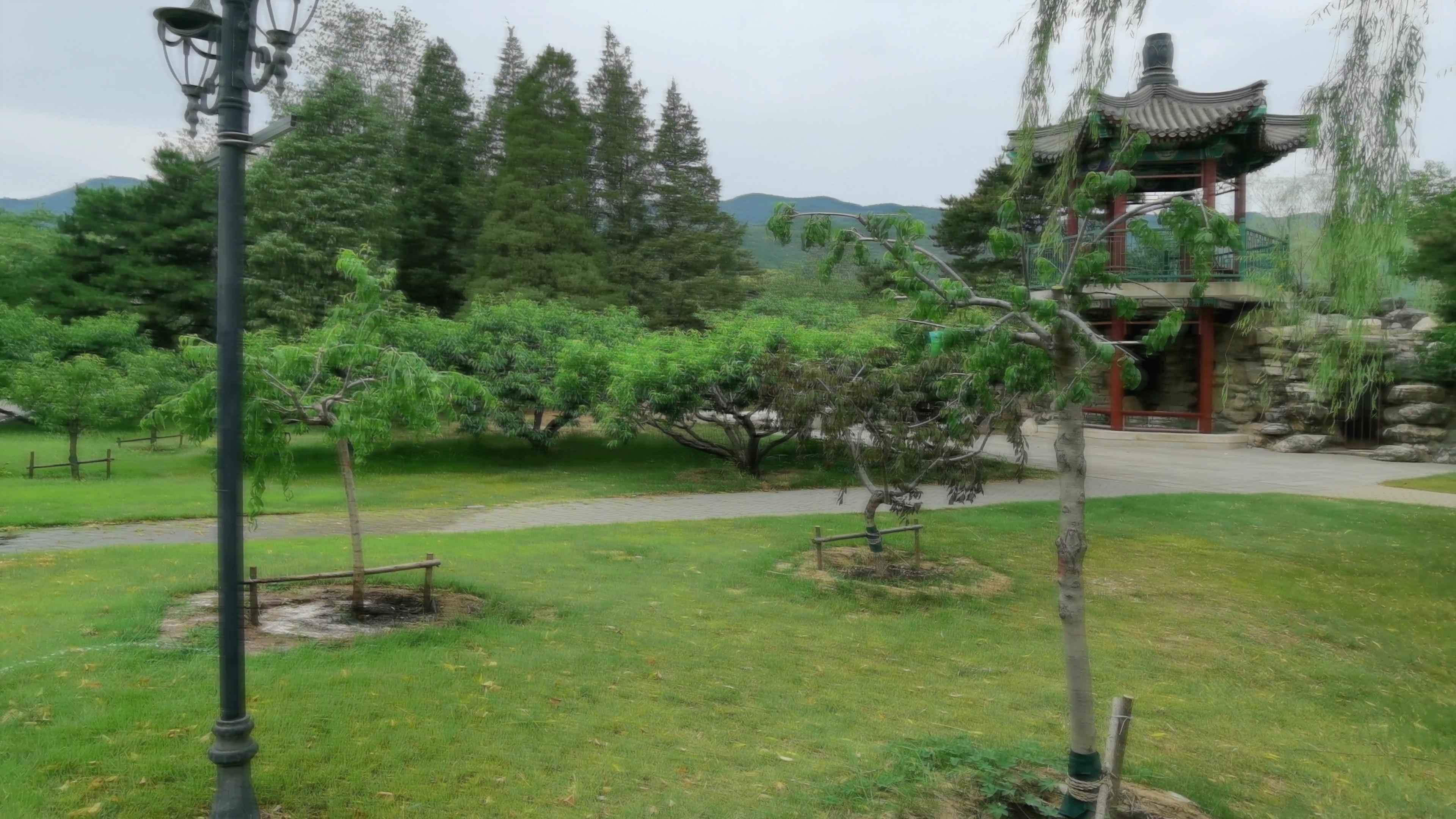} &
        \includegraphics[width=0.24\textwidth]{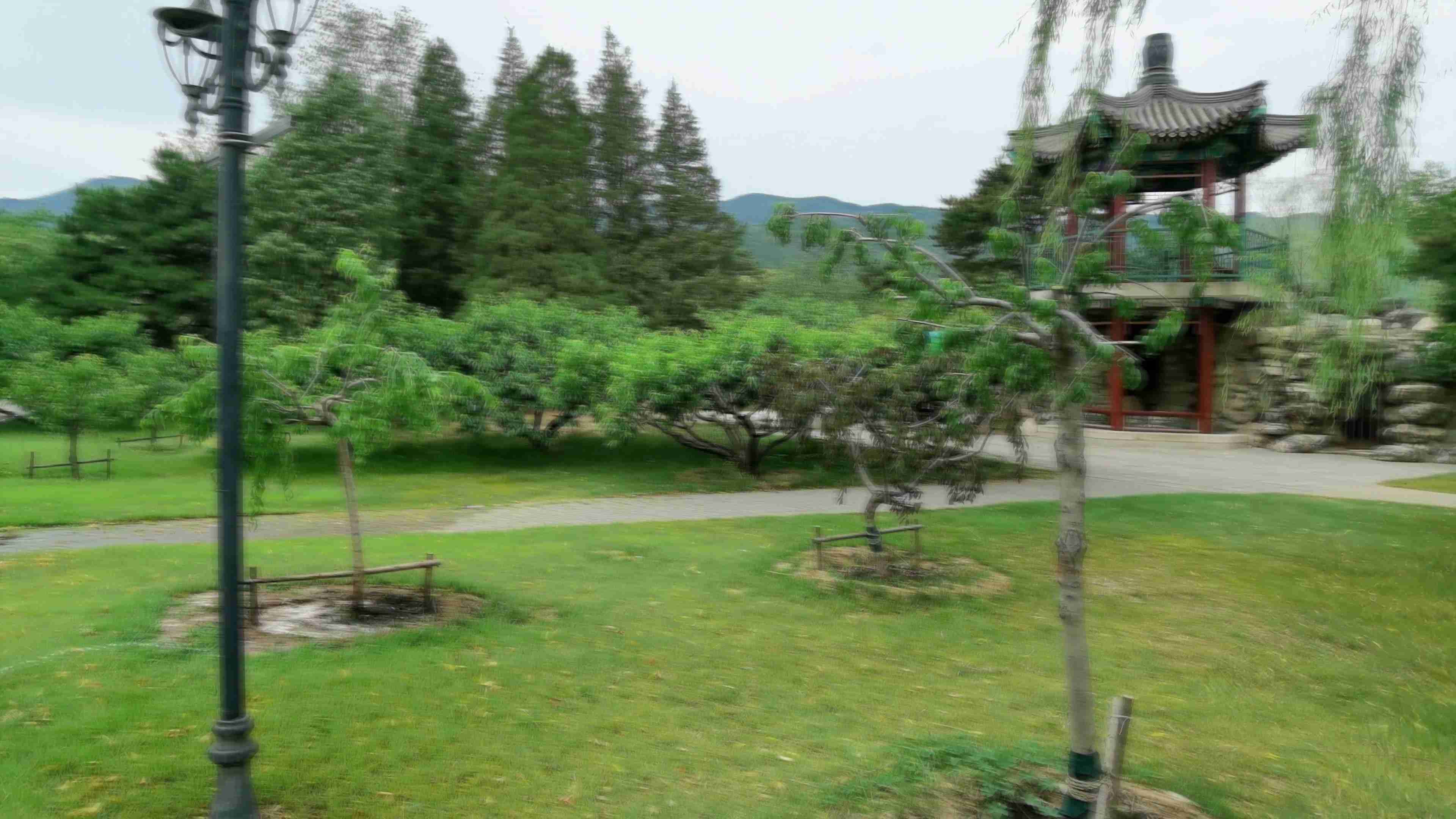} &
        \includegraphics[width=0.24\textwidth]{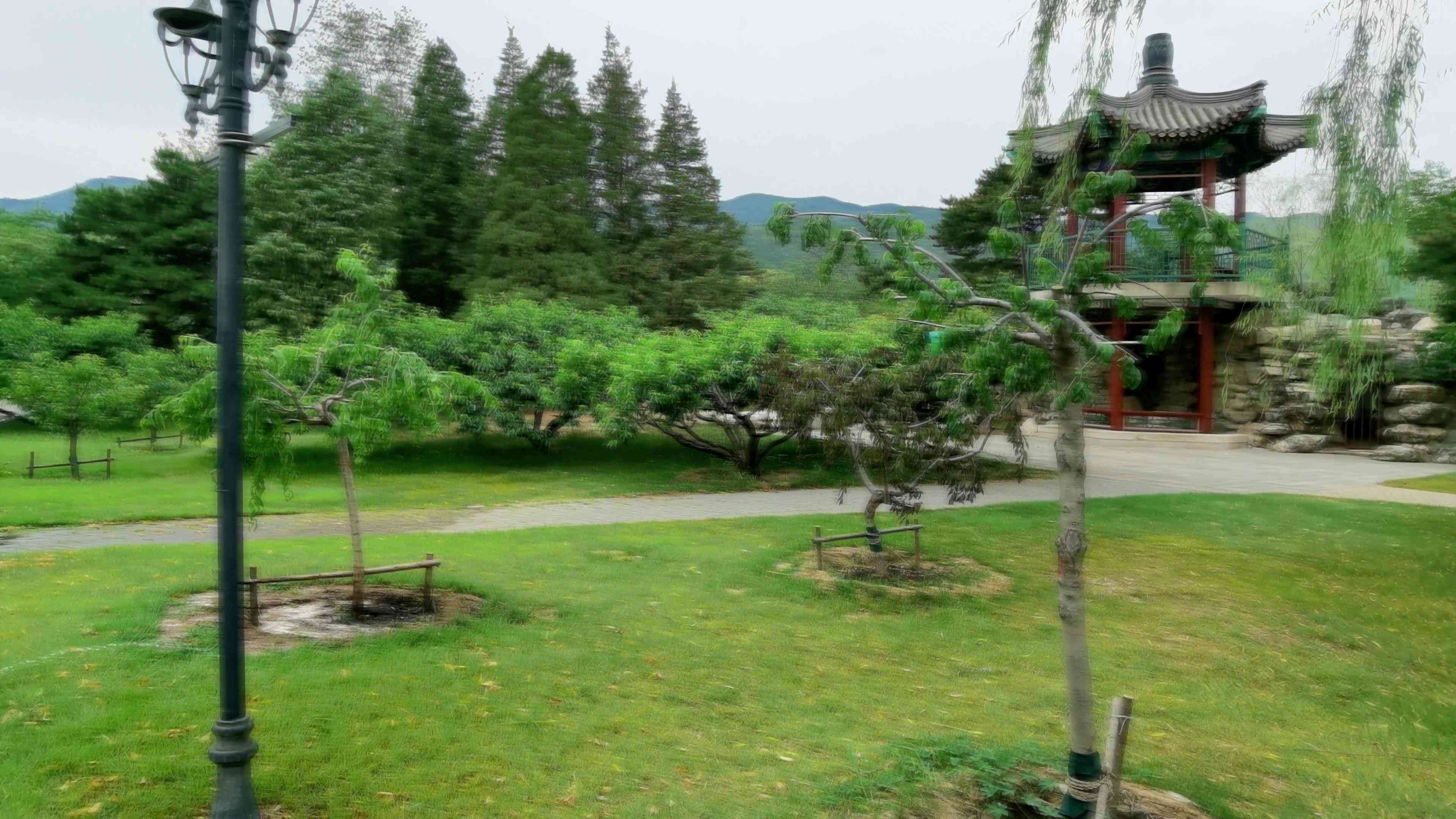} &
        \includegraphics[width=0.24\textwidth]{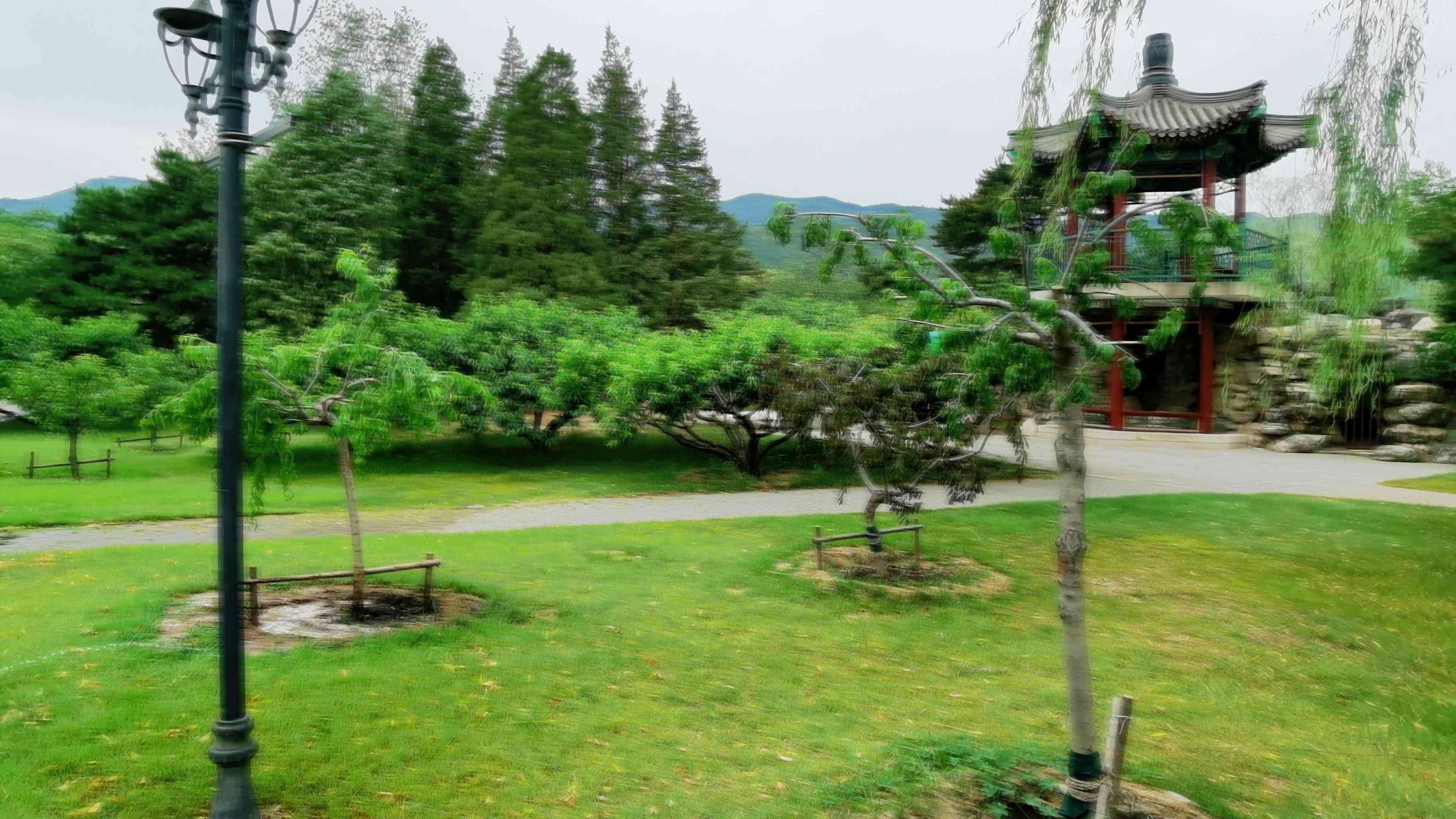} &
        \includegraphics[width=0.24\textwidth]{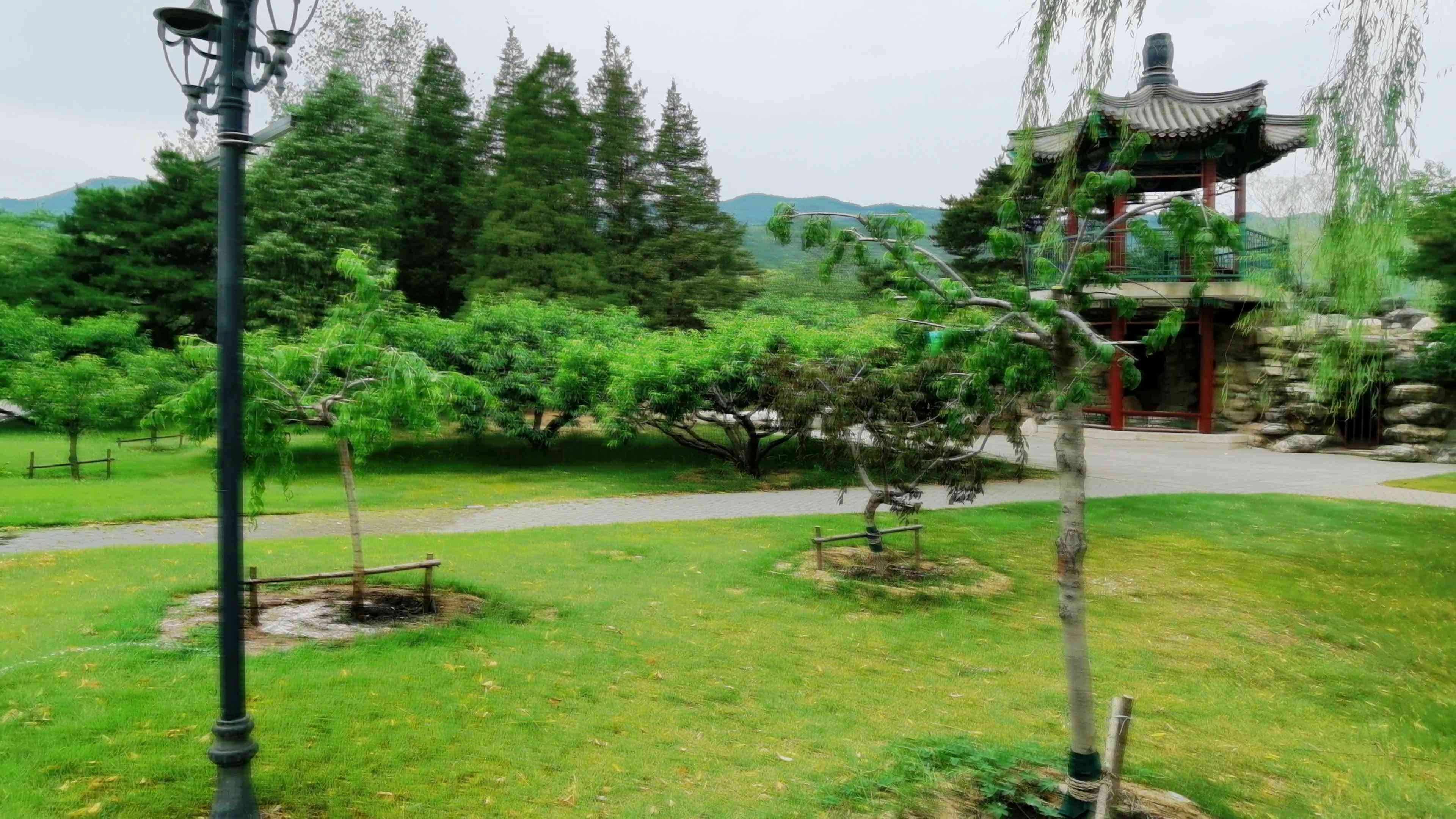} &
        \includegraphics[width=0.24\textwidth]{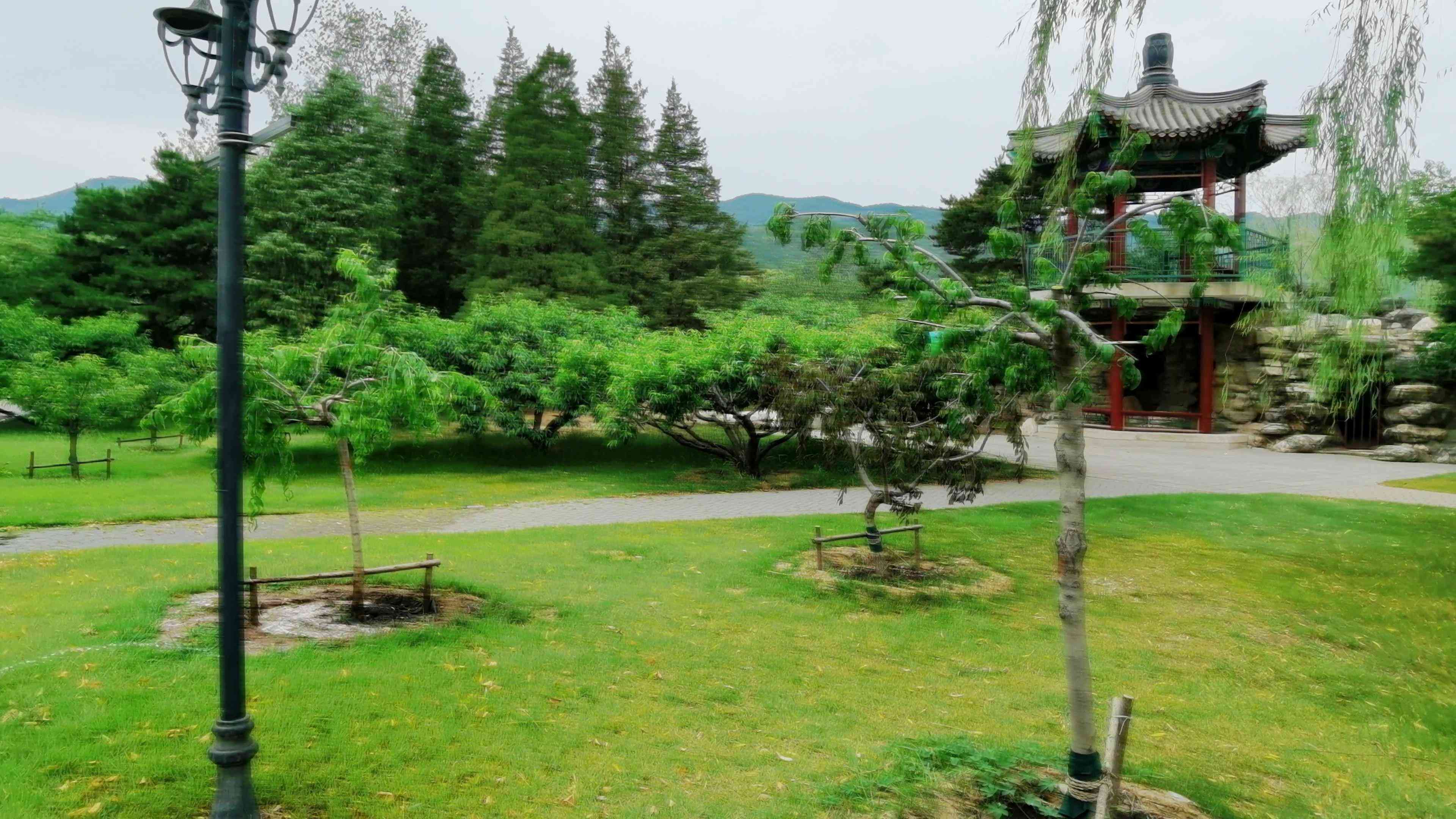}&
        \includegraphics[width=0.24\textwidth]{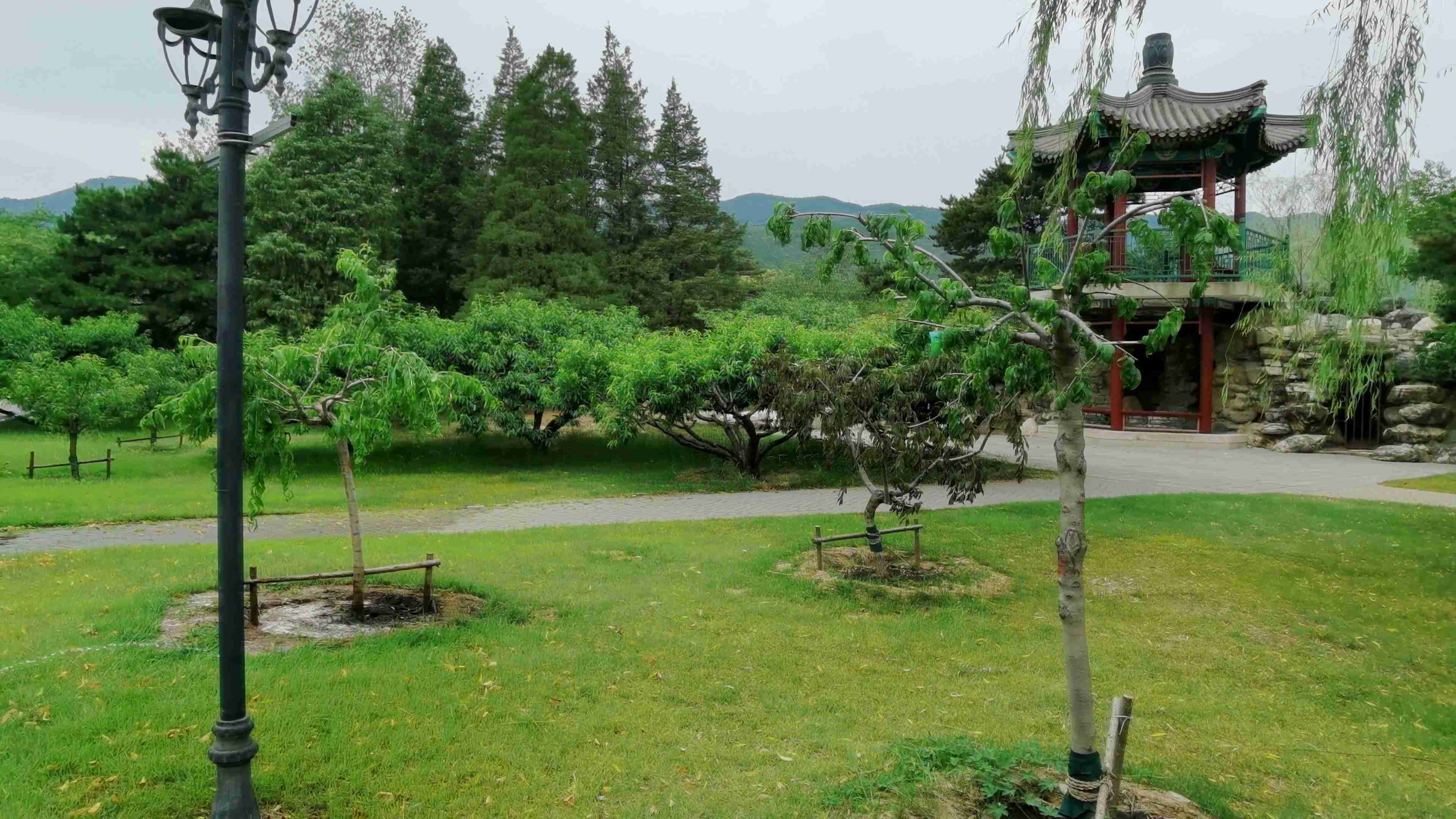} \\ 
       \huge Input &\huge GT & \huge Stripformer &\huge Restormer & \huge Uformer & \huge FFTformer & \huge UHDformer & \huge UHDDIP & \huge Ours \\
    \end{tabular}
    \end{adjustbox}
    \vspace{-4mm}
    \caption{Image deblurring on UHD-Blur. TSFormer is able to generate deblurring results with sharper structures.}
    \label{fig: deblur}
    \vspace{-4mm}
\end{figure*}

\begin{table}[!t]\footnotesize
\setlength{\tabcolsep}{3pt} 
\renewcommand{\arraystretch}{1.2} 
\caption{Image dehazing results on UHD-Haze dataset. We highlight the \colorbox{best}{best} values for each metric. TSFormer achieves state-of-the-art performance with competitive parameter efficiency.}
\vspace{-5mm}
\begin{center}

\begin{tabular}{l|c|ccc|l}
\shline
\textbf{Method} & \textbf{Venue} & \multicolumn{3}{c|}{\textbf{UHD-Haze}} &\textbf{Param}\\
 &  & \textbf{PSNR} $\uparrow$ & \textbf{SSIM} $\uparrow$ & \textbf{LPIPS} $\downarrow$ \\
\shline
UHD & ICCV'21&18.05& 0.811 &0.359& 34.50M\\
Restormer&CVPR'22&13.88 & 0.641 &0.440& 26.10M\\
Uformer &CVPR'22 &19.83 &0.737& 0.422&20.60M\\
DehazeFormer & TIP'23 & 15.37 & 0.725&0.399 & 2.50M \\
UHDformer & AAAI'24 & 22.59& 0.942 & 0.120&0.34M \\
UHDDIP & arxiv'24 &\cellcolor{secondbest}24.69 &\cellcolor{secondbest}0.952& \cellcolor{secondbest}0.104 & 0.81M \\
TSFormer (Ours) & - & \cellcolor{best}\textbf{24.88} & \cellcolor{best}\textbf{0.953} & \cellcolor{best}\textbf{0.092}& 3.38M \\
\shline
\end{tabular}
\end{center}
\label{tab: image dehaze.}
\vspace{-6mm}
\end{table}
\begin{figure*}[!t]
    \centering
    \begin{subfigure}{0.32\textwidth}
        \includegraphics[width=\linewidth]{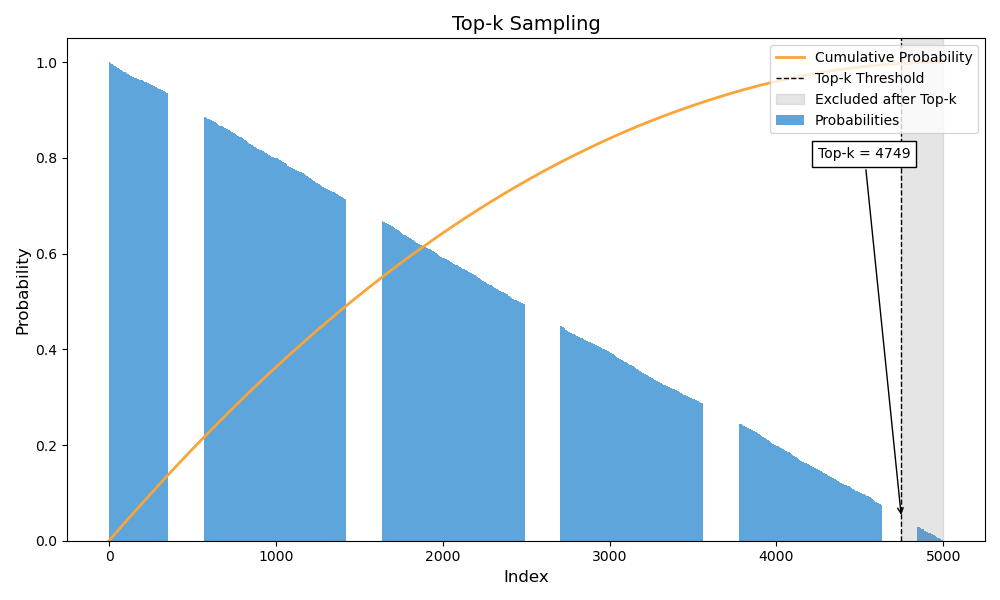}
        \caption{Top-k sampling}
        \label{fig:topk_sampling_distribution}
    \end{subfigure}
    \hfill
    \begin{subfigure}{0.32\textwidth}
        \includegraphics[width=\linewidth]{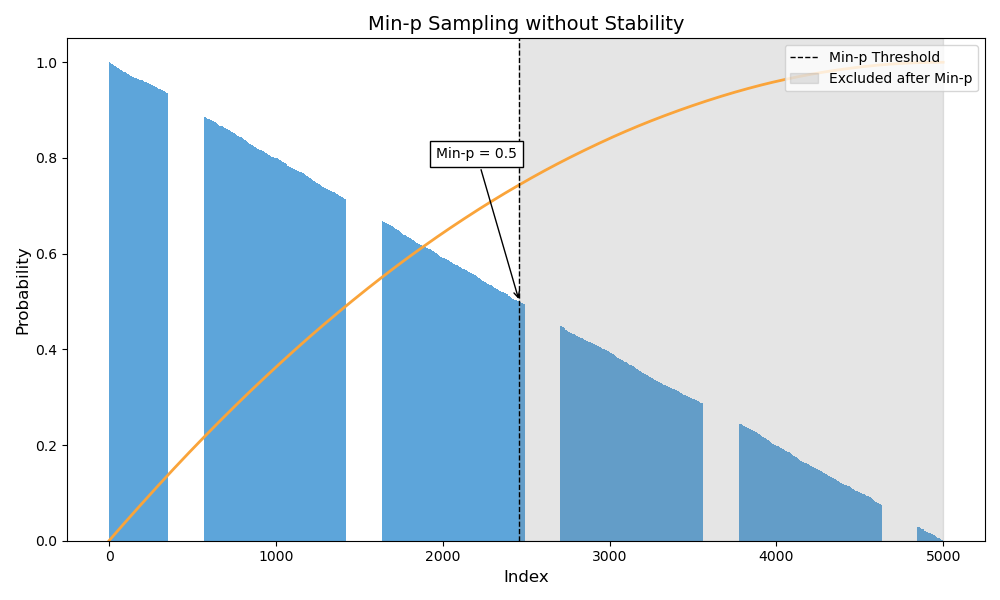}
        \caption{Min-$p$ sampling without trusted mechanism}
        \label{fig:minp_no_stability_distribution}
    \end{subfigure}
    \hfill
    \begin{subfigure}{0.32\textwidth}
        \includegraphics[width=\linewidth]{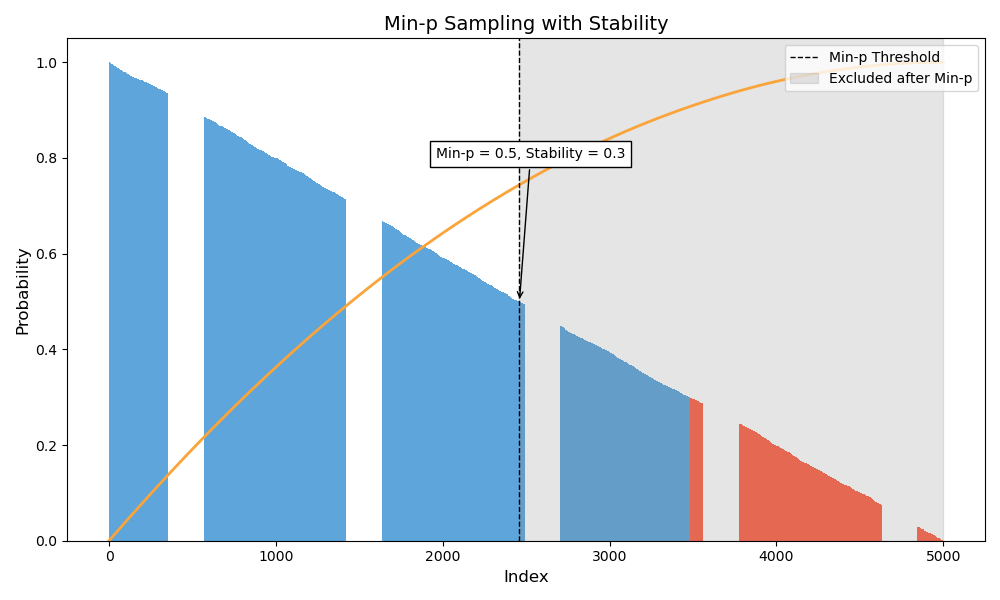}
        \caption{Ours}
        \label{fig:minp_with_stability_distribution}
    \end{subfigure}\vspace{-4mm}
    \caption{Cumulative probability distribution plots for different sampling methods. Min-p Sampling with trusted mechanism (c) demonstrates effective filtering, preserving essential features while excluding less relevant ones.}\vspace{-4mm}
    \label{fig:sampling_distribution_comparison}
\end{figure*}

\begin{table}[!t]\footnotesize
\setlength{\tabcolsep}{3pt} 
\renewcommand{\arraystretch}{1.2} 
\caption{Image deraining results on the UHD-Rain dataset. The $\uparrow$ means the higher is better and the $\downarrow$ means the lower is better.}
\vspace{-4mm}
\begin{center}
\begin{tabular}{l|c|ccc|l}
\shline
\textbf{Method} & \textbf{Venue} & \textbf{PSNR} $\uparrow$ & \textbf{SSIM} $\uparrow$ & \textbf{LPIPS} $\downarrow$ & \textbf{Param} \\
\shline
Uformer & CVPR'22 & 19.50 & 0.716 & 0.460 & 20.60M \\
Restormer & CVPR'22 & 19.41 & 0.711 & 0.478 & 26.10M \\
SFNet & ICLR'23 & 20.09 & 0.709 & 0.477 & 34.50M \\
UHD & ICCV'21 & 26.18 & 0.863 & 0.289 & 34.50M \\
UHDformer & AAAI'24 & 37.35 & 0.975 & 0.055 & 0.34M \\
UHDDIP & arxiv'24 & \cellcolor{secondbest}40.18 & \cellcolor{secondbest}0.982 & \cellcolor{secondbest}0.030 & 0.81M \\
TSFormer (Ours) & - & \cellcolor{best}\textbf{40.40} & \cellcolor{best}\textbf{0.983} & \cellcolor{best}\textbf{0.028} & 3.38M \\
\shline
\end{tabular}
\end{center}
\label{tab: image deraining}
\vspace{-6mm}
\end{table}

\begin{figure*}[!t]
    \centering
    \begin{adjustbox}{max width=\textwidth}
    \begin{tabular}{ccccccccc}
        \includegraphics[width=0.24\textwidth]{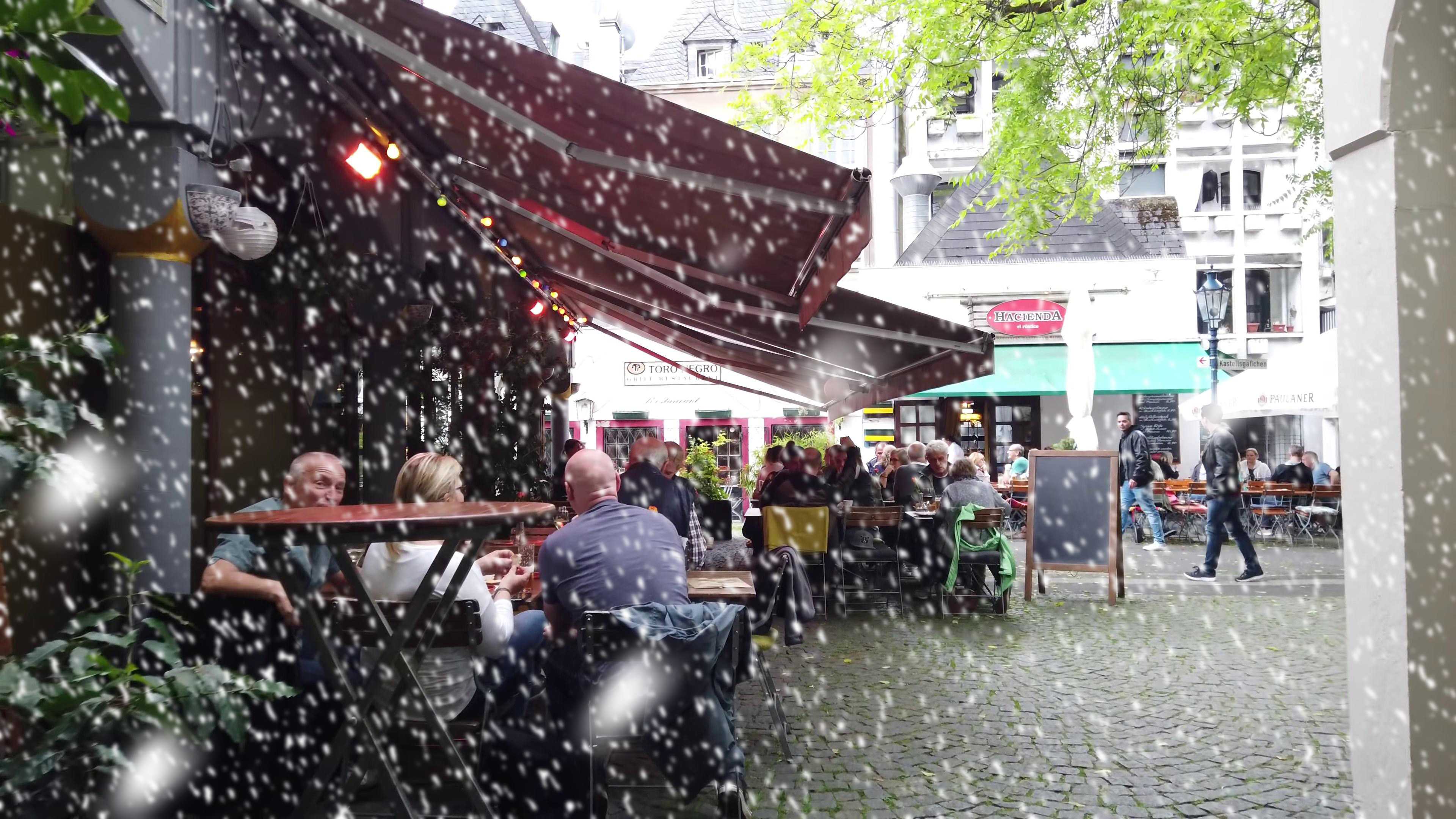} &
        \includegraphics[width=0.24\textwidth]{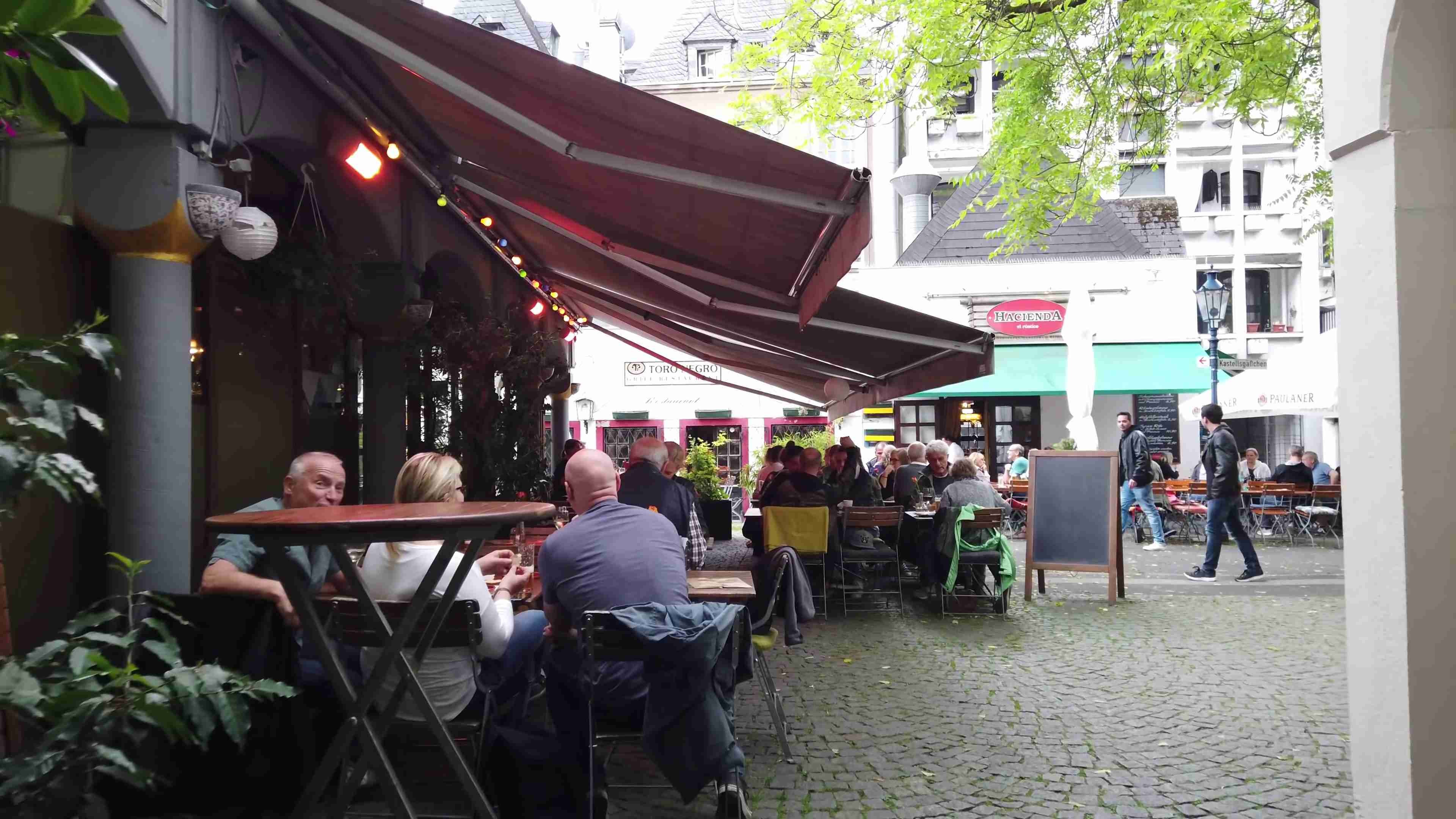} &
        \includegraphics[width=0.24\textwidth]{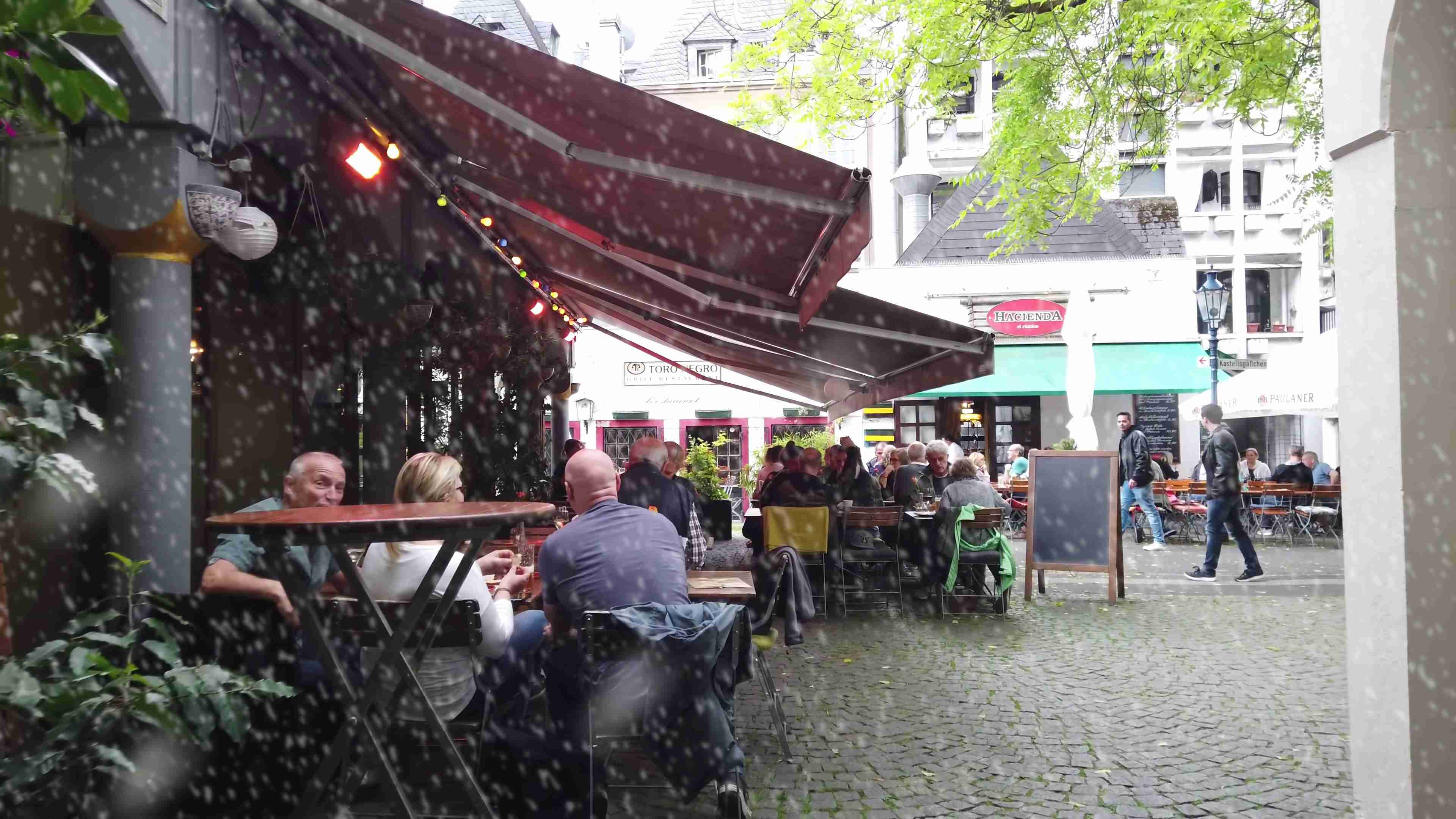} &
        \includegraphics[width=0.24\textwidth]{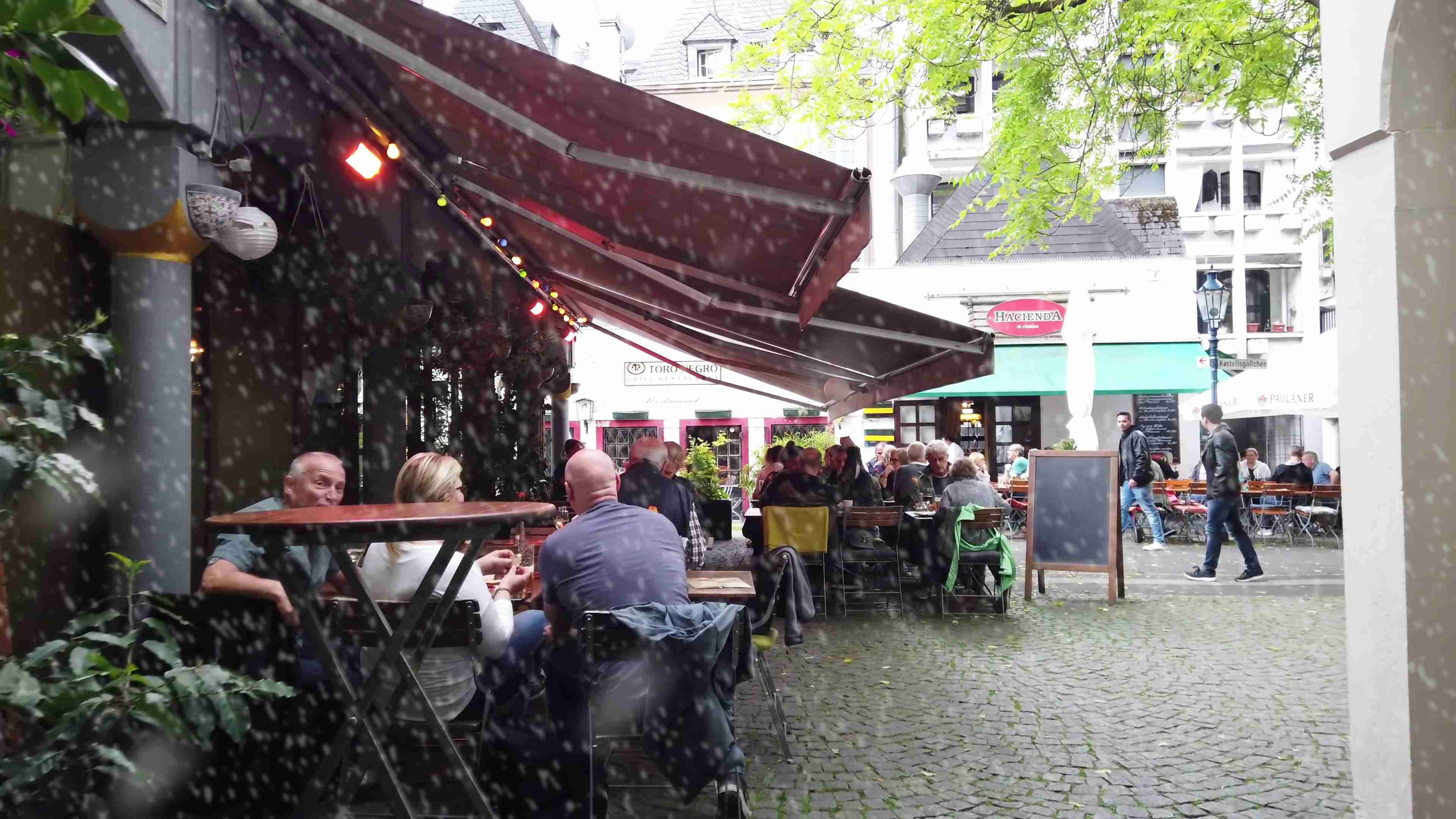} &
        \includegraphics[width=0.24\textwidth]{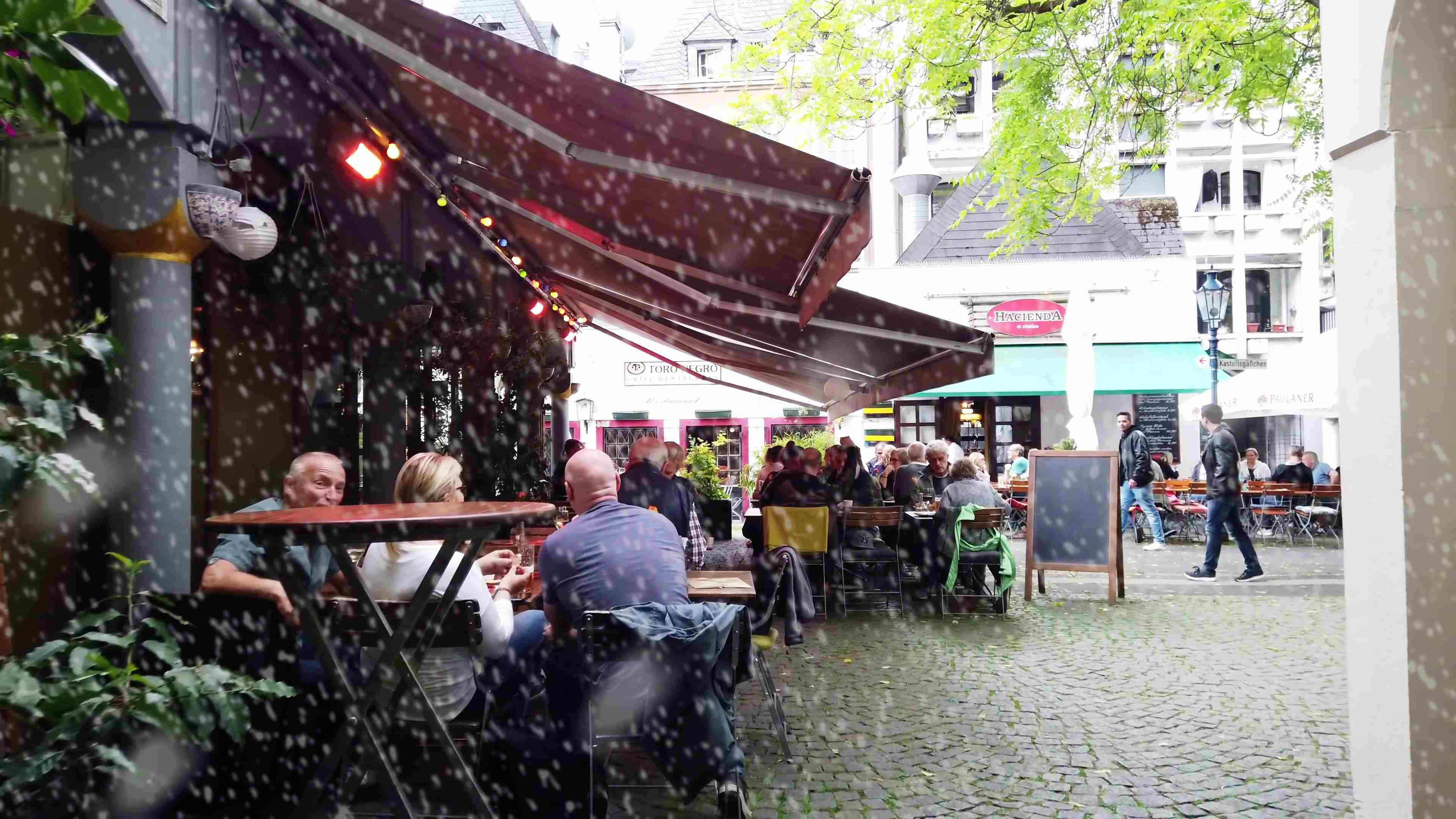} &
        \includegraphics[width=0.24\textwidth]{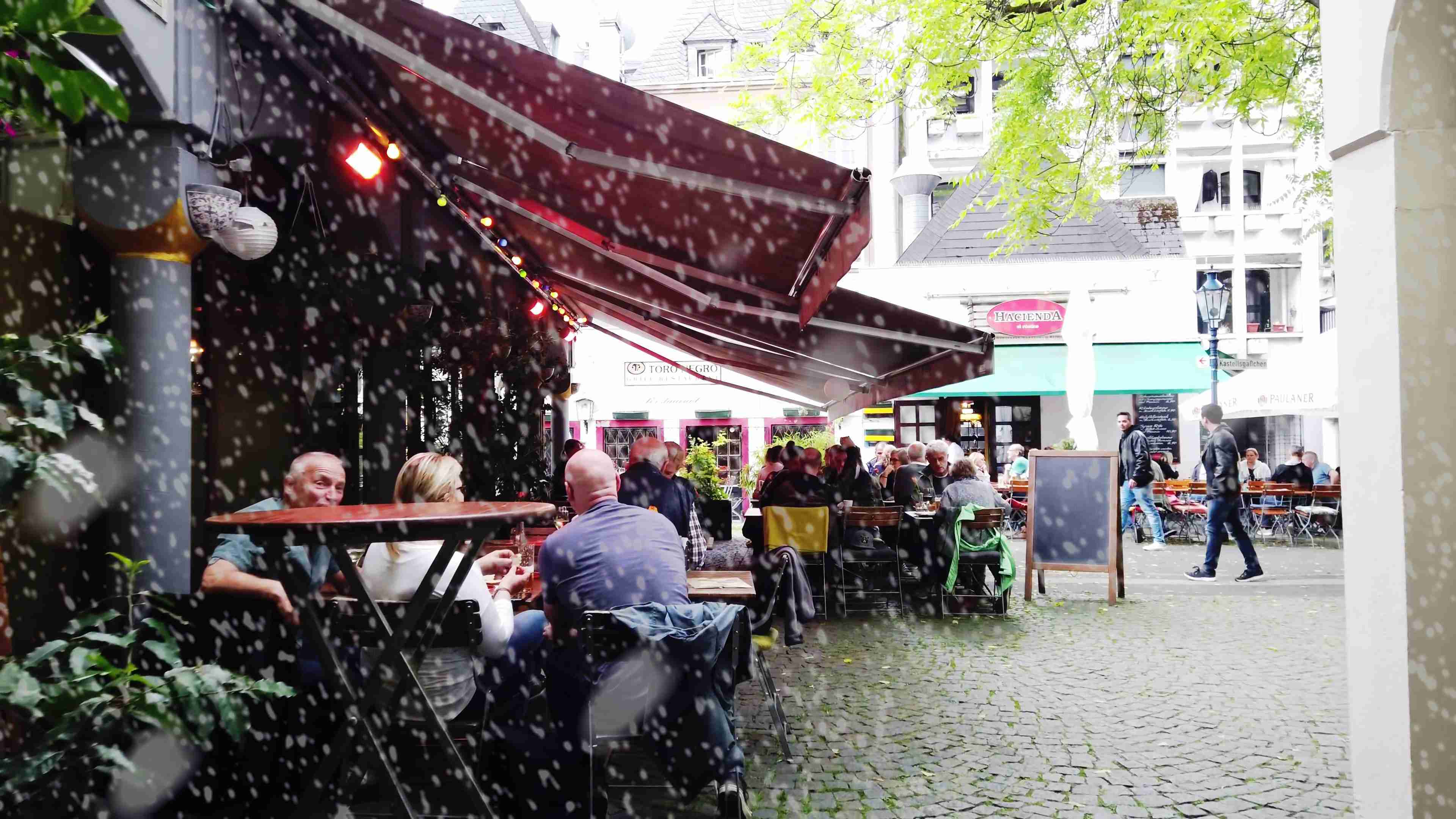} &
        \includegraphics[width=0.24\textwidth]{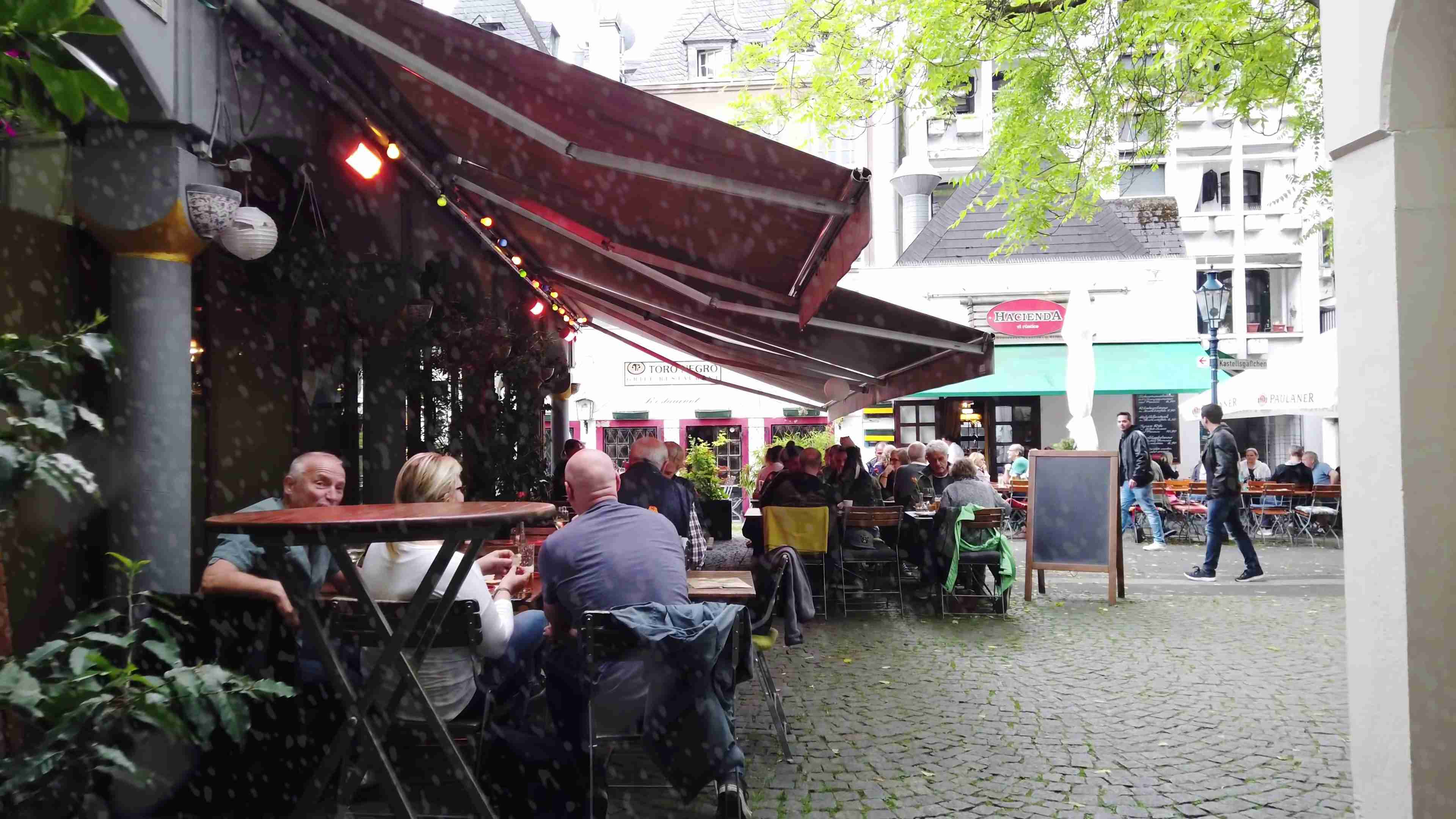} &
        \includegraphics[width=0.24\textwidth]{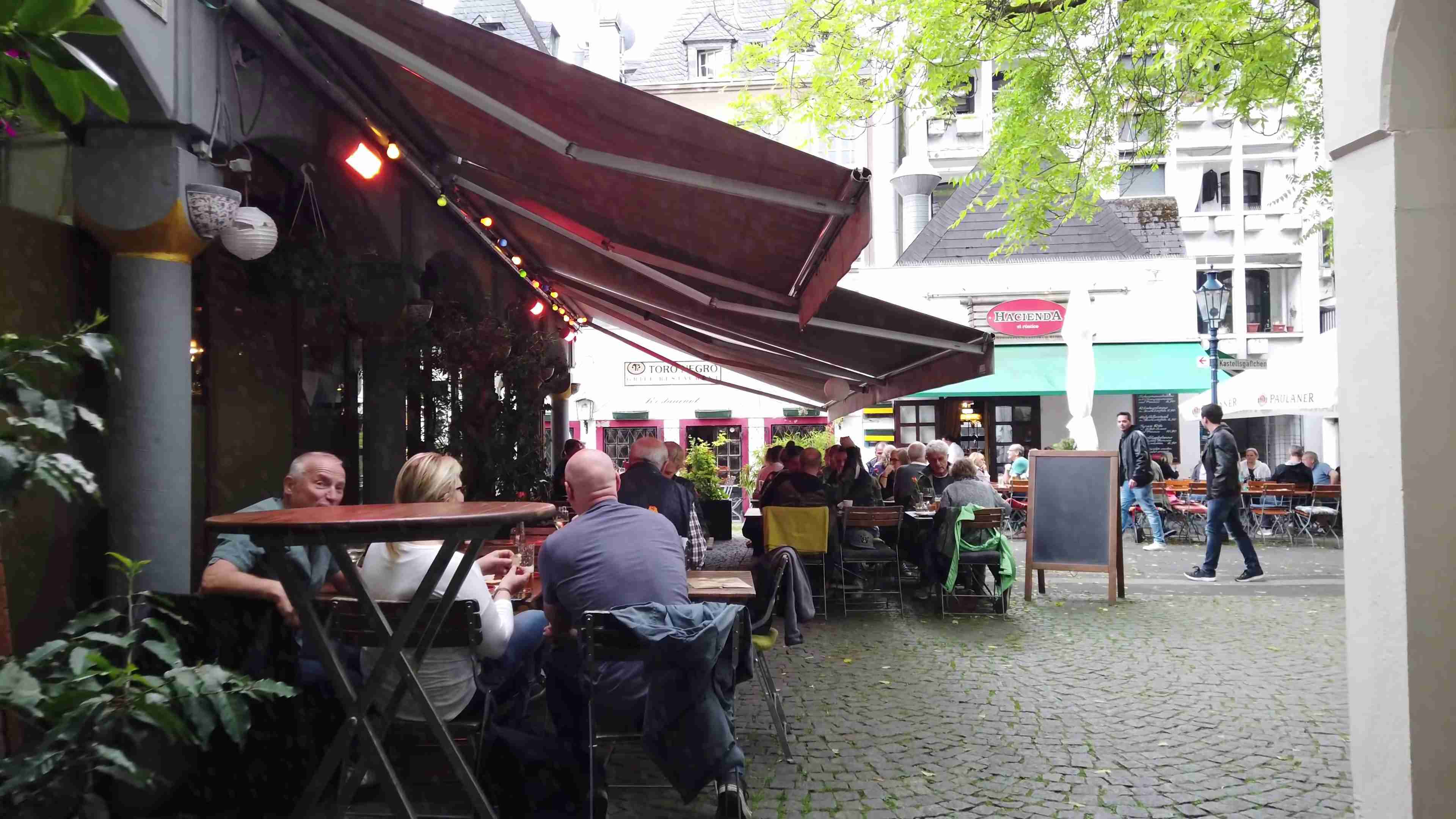}&
        \includegraphics[width=0.24\textwidth]{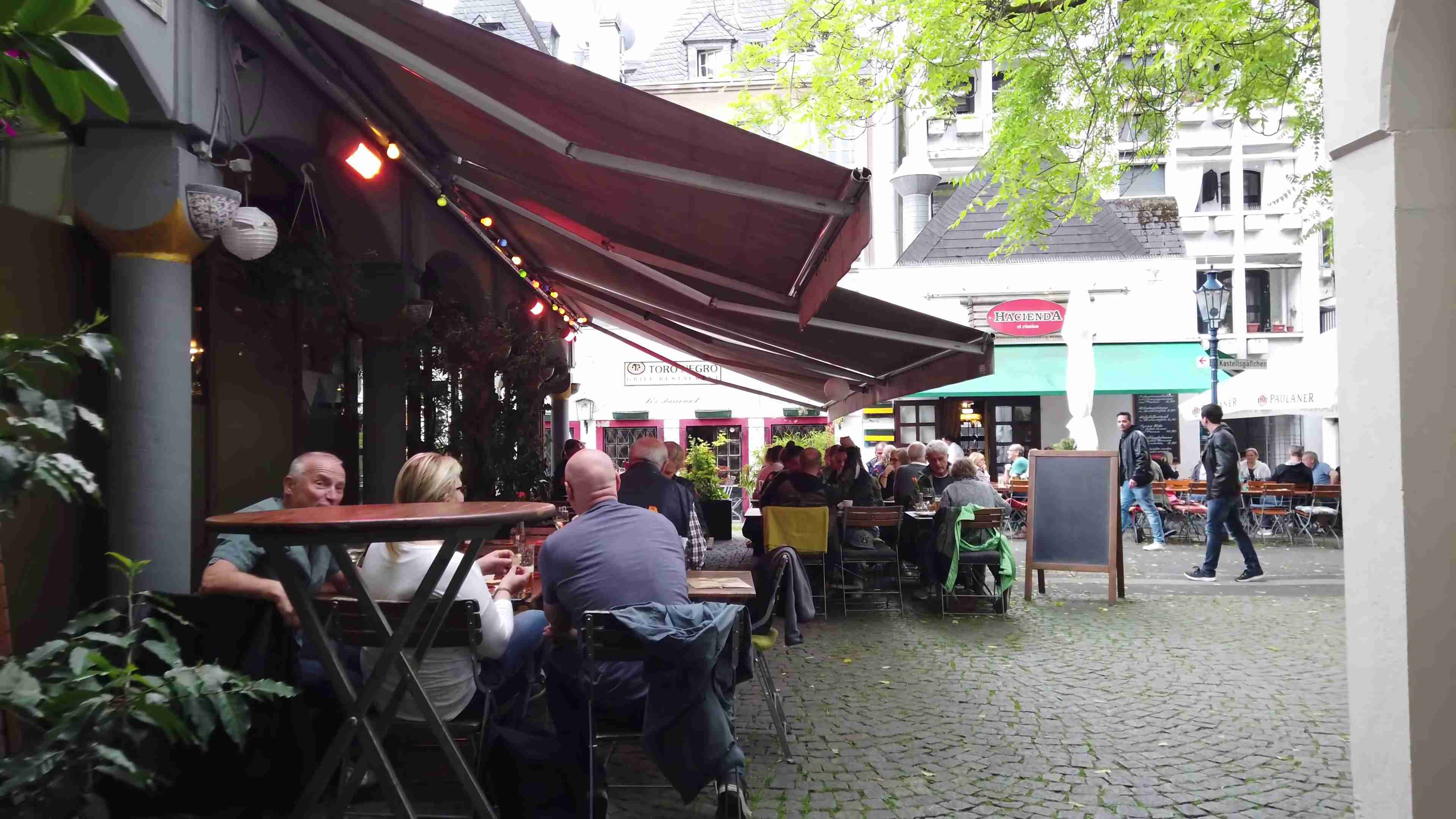} \\
        \includegraphics[width=0.24\textwidth]{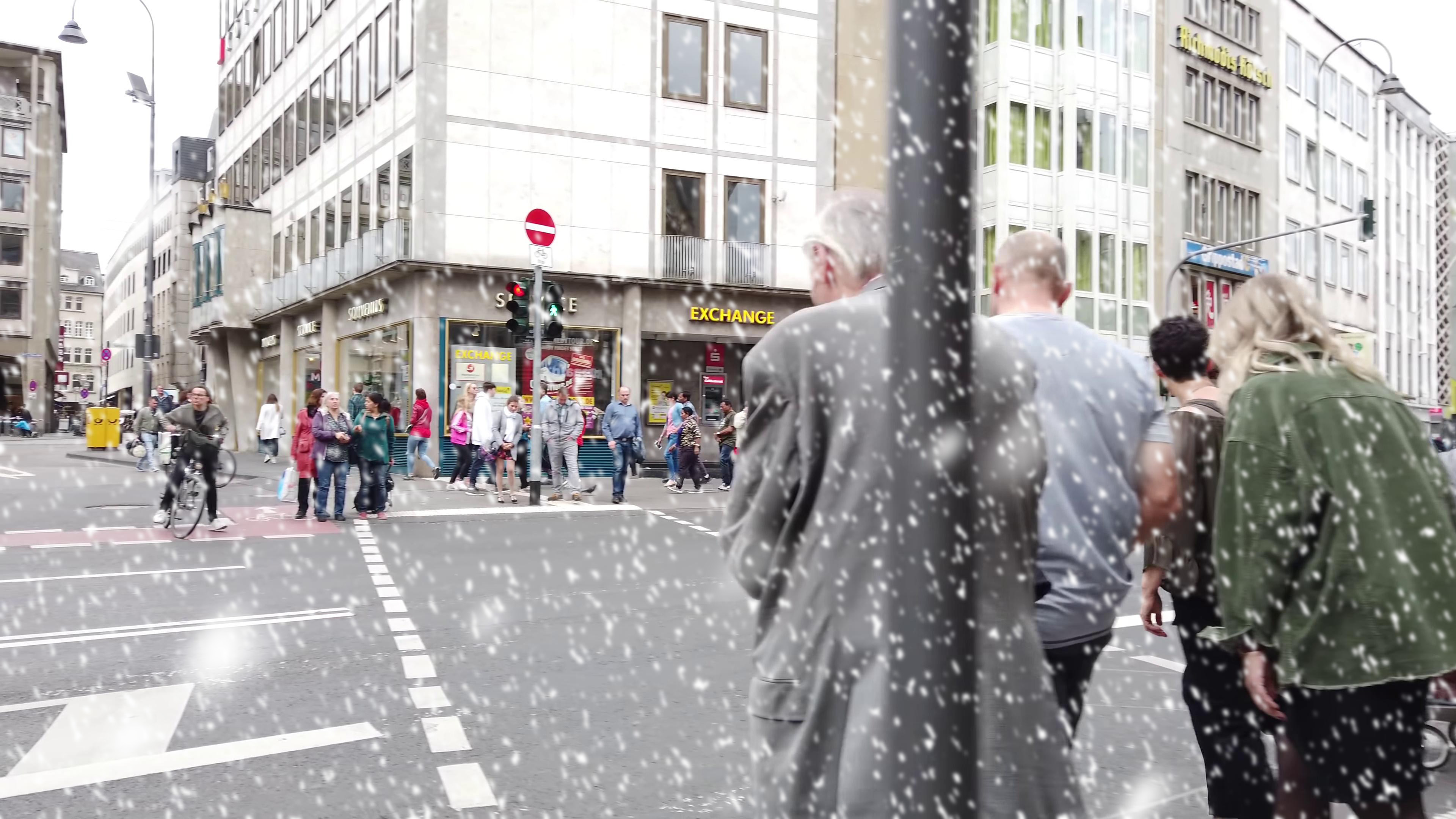} &
        \includegraphics[width=0.24\textwidth]{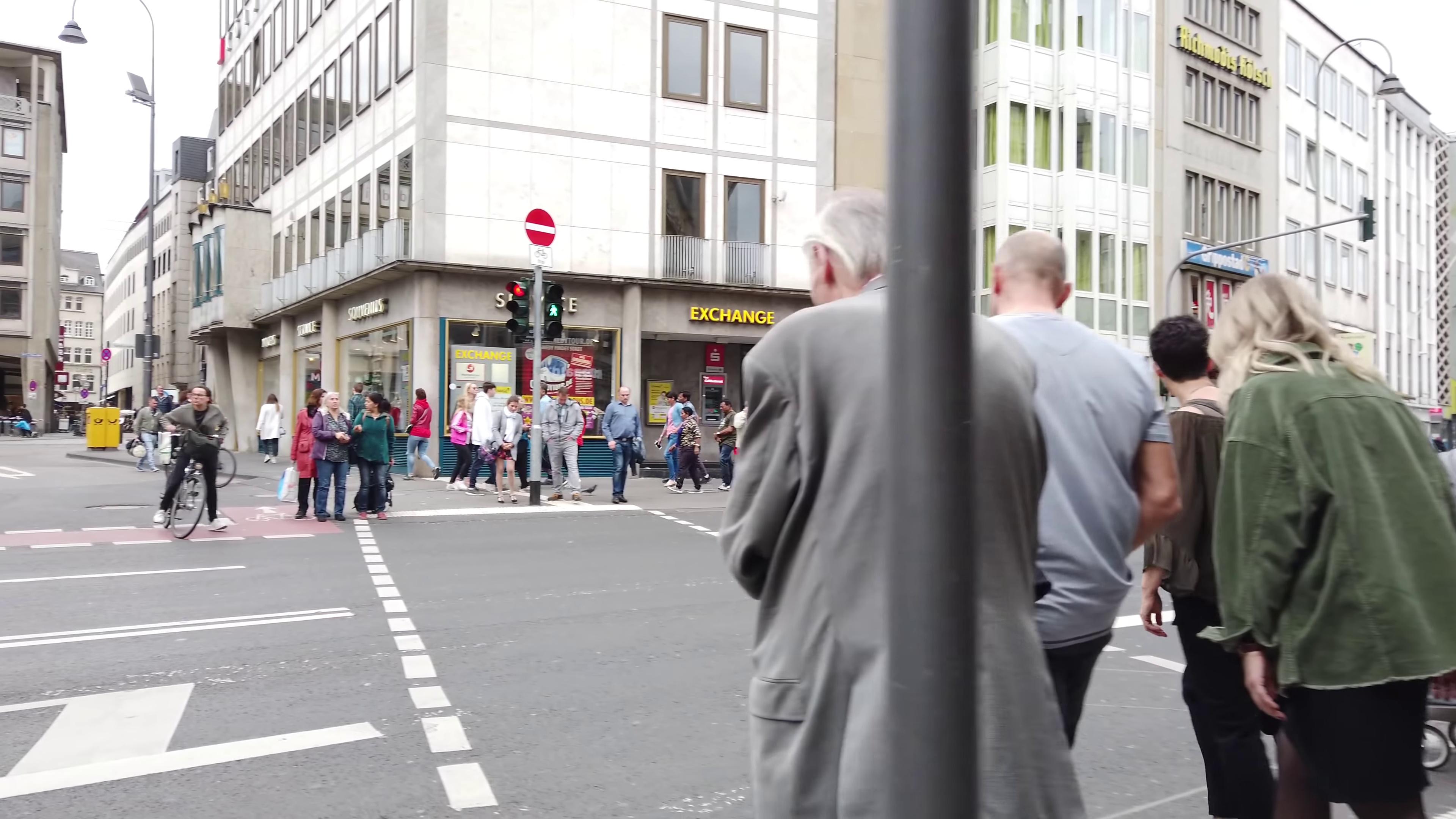} &
        \includegraphics[width=0.24\textwidth]{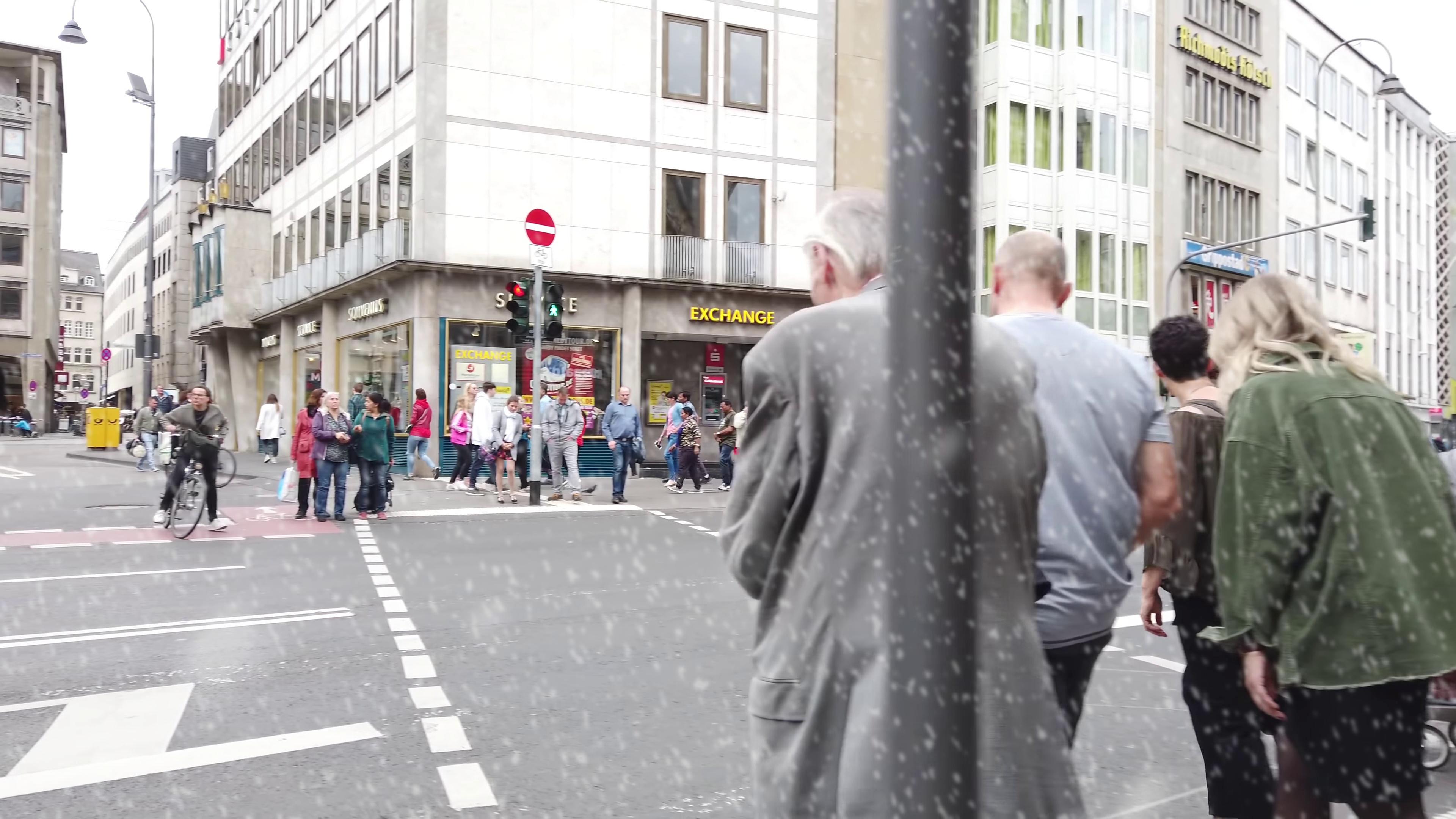} &
        \includegraphics[width=0.24\textwidth]{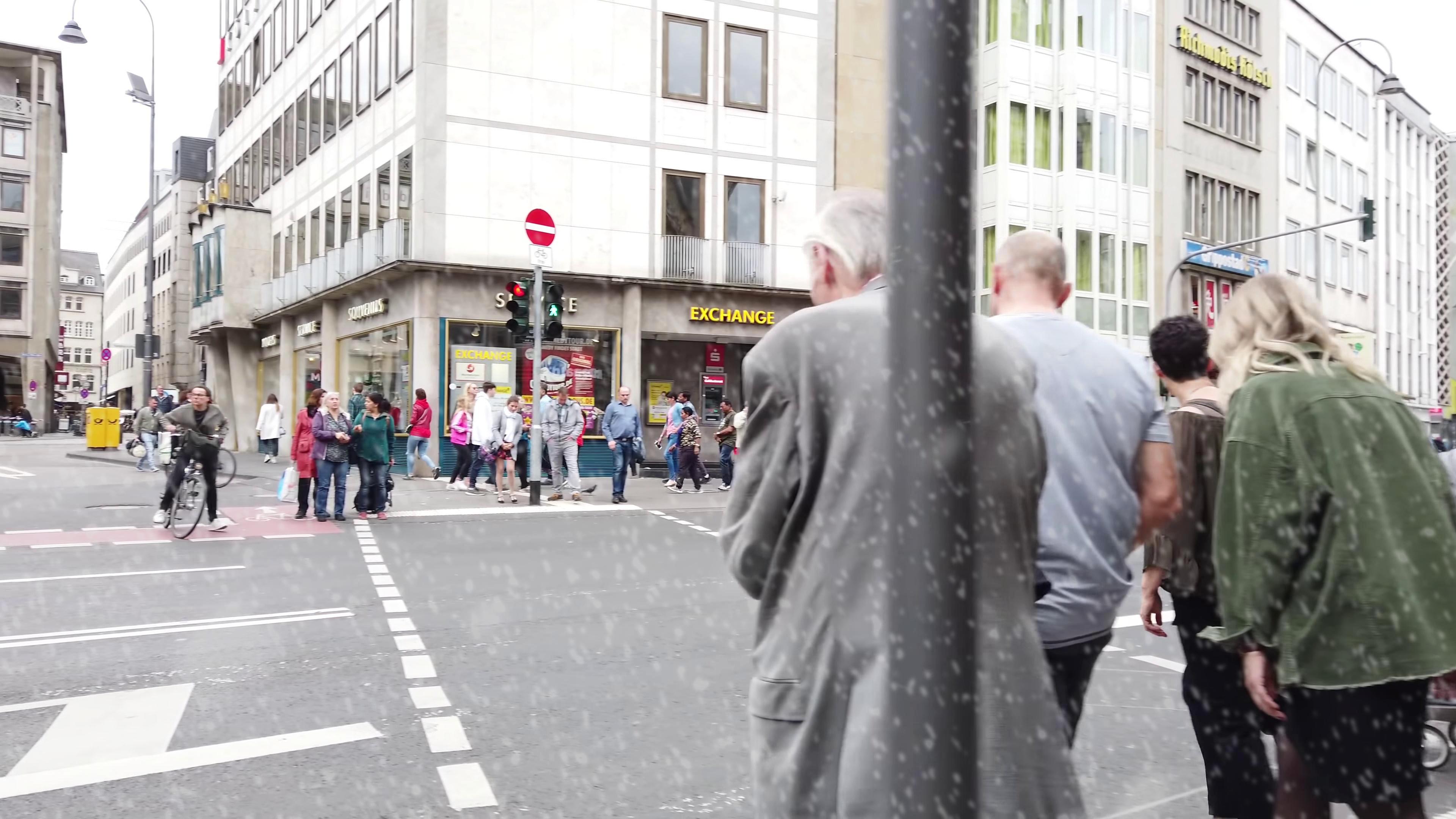} &
        \includegraphics[width=0.24\textwidth]{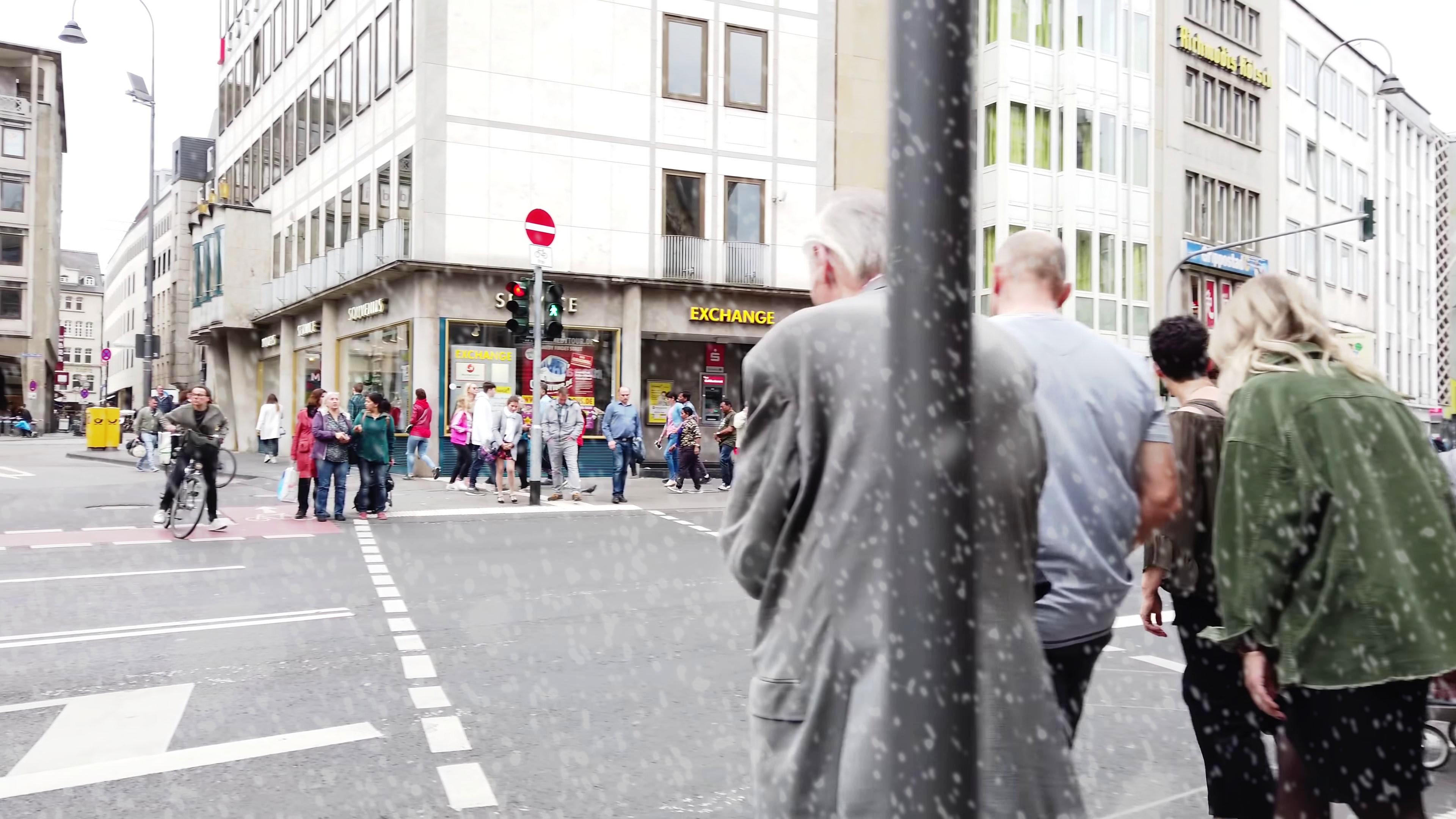} &
        \includegraphics[width=0.24\textwidth]{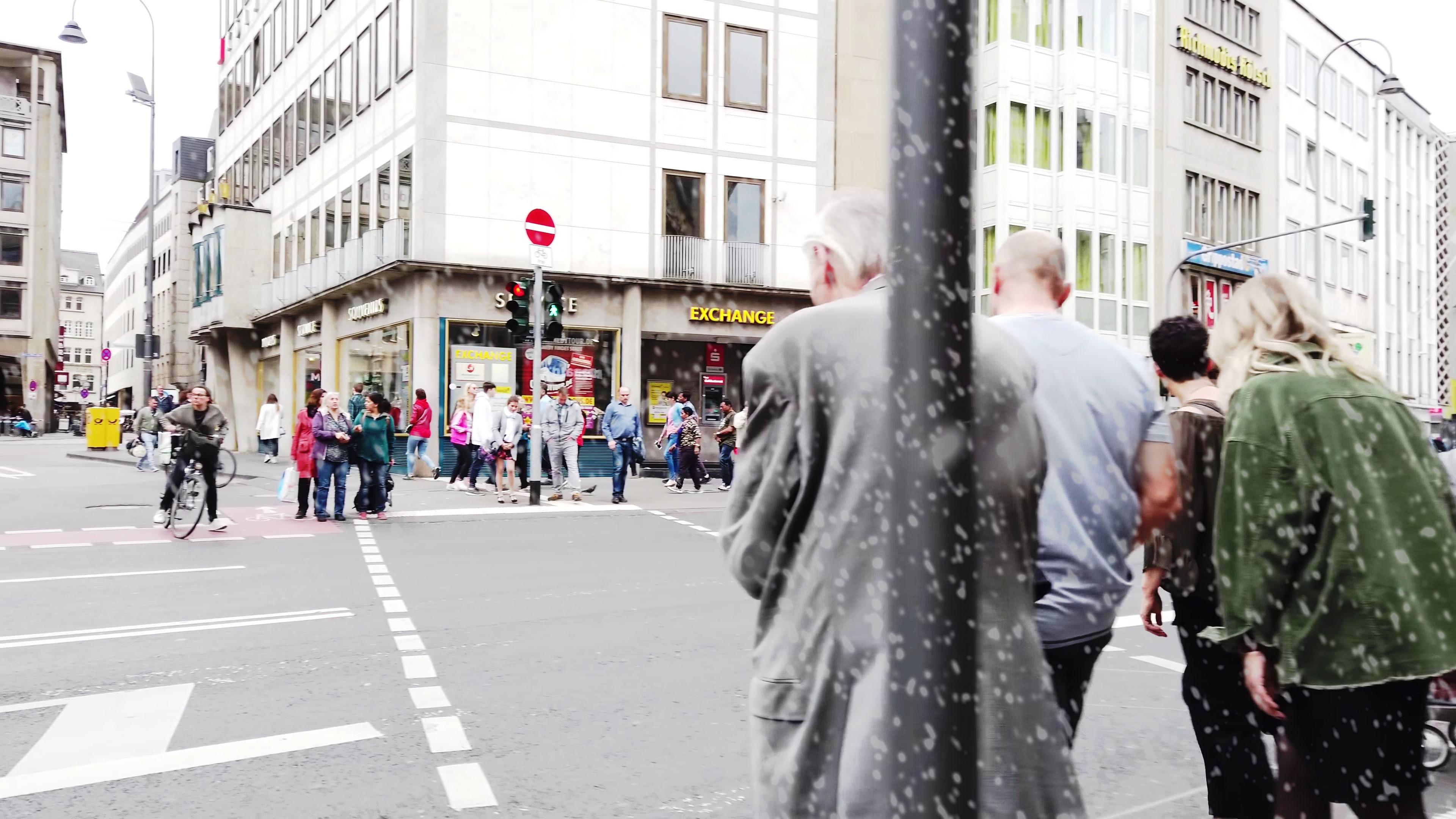} &
        \includegraphics[width=0.24\textwidth]{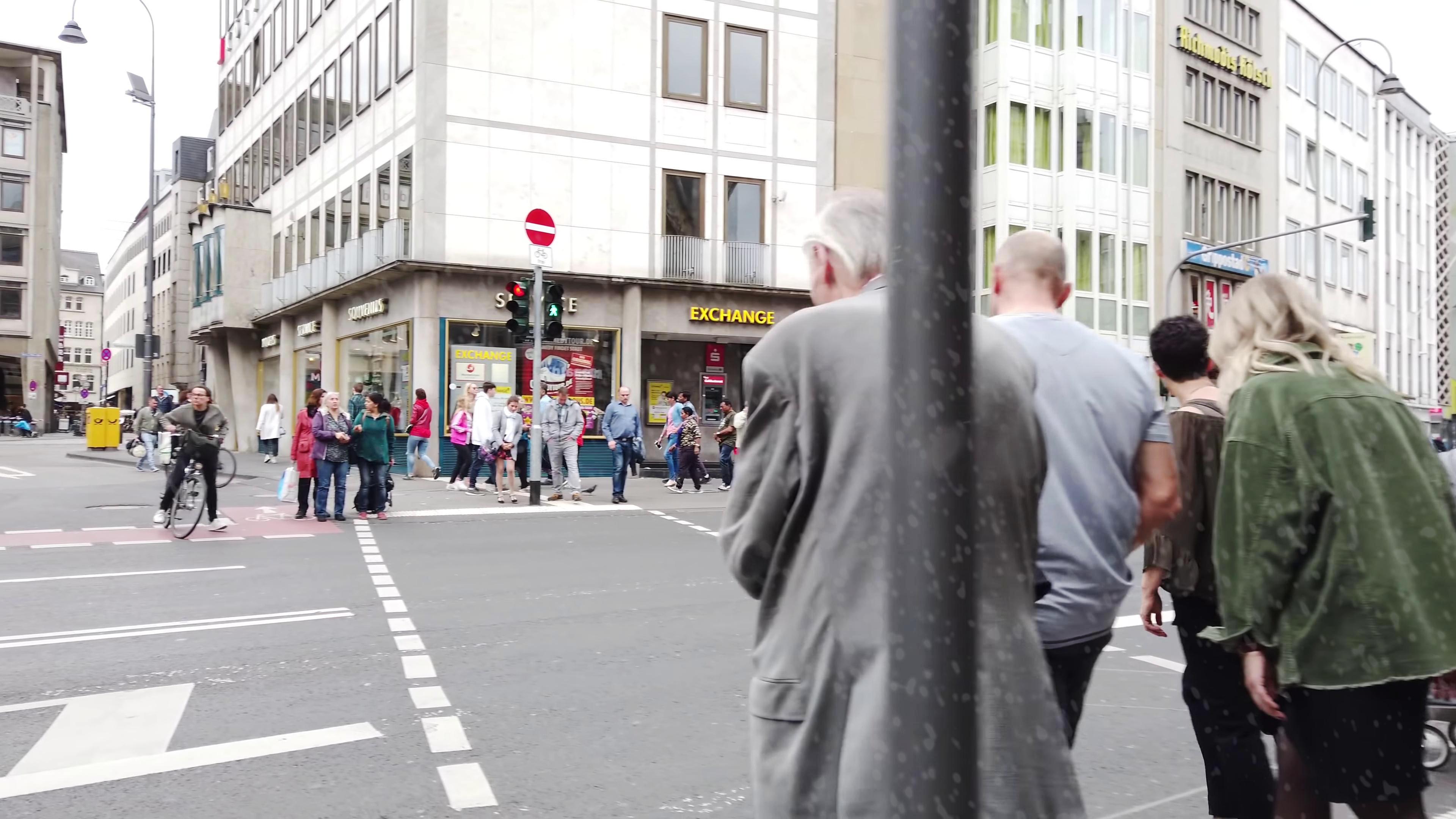} &
        \includegraphics[width=0.24\textwidth]{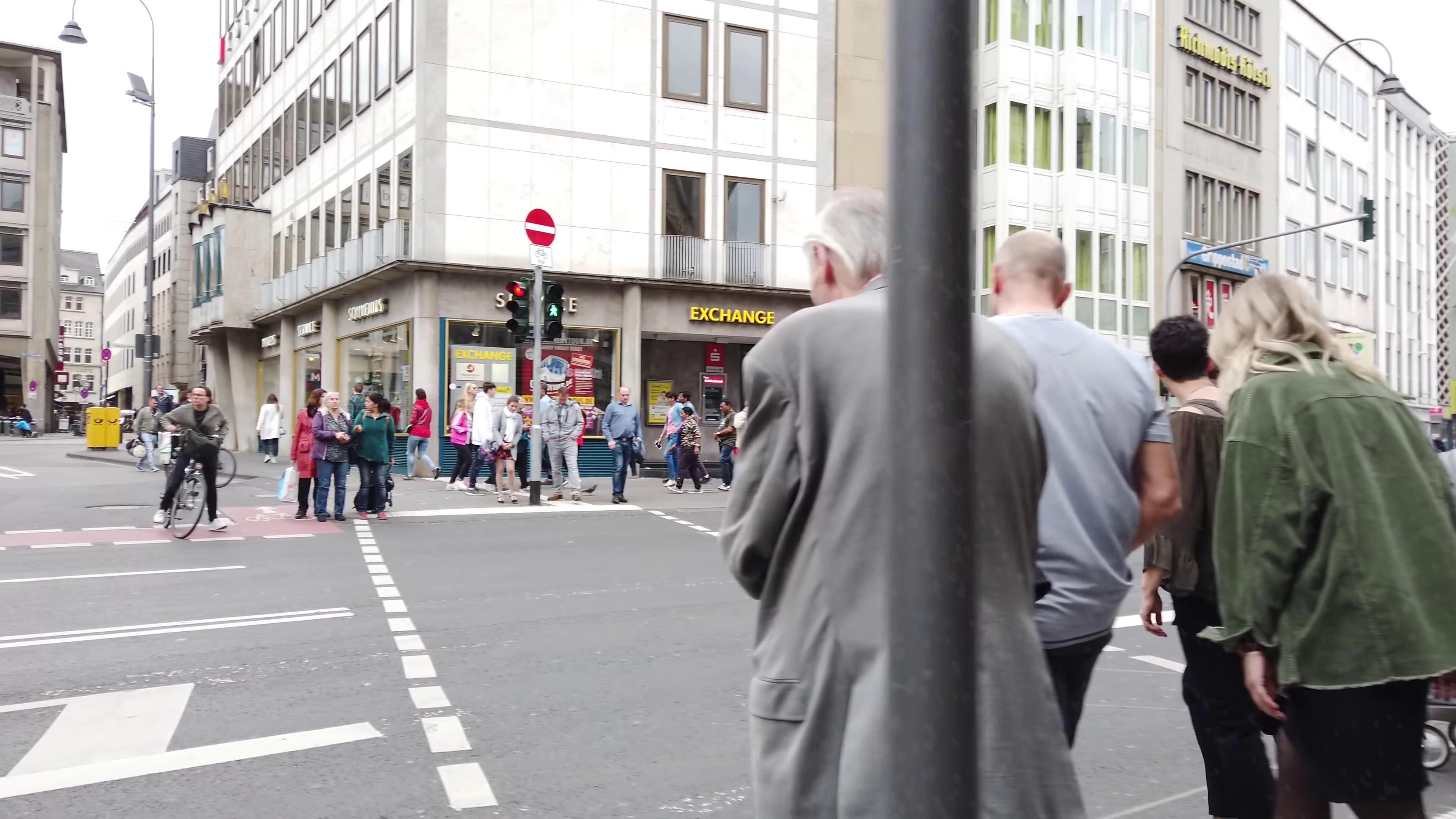}&
        \includegraphics[width=0.24\textwidth]{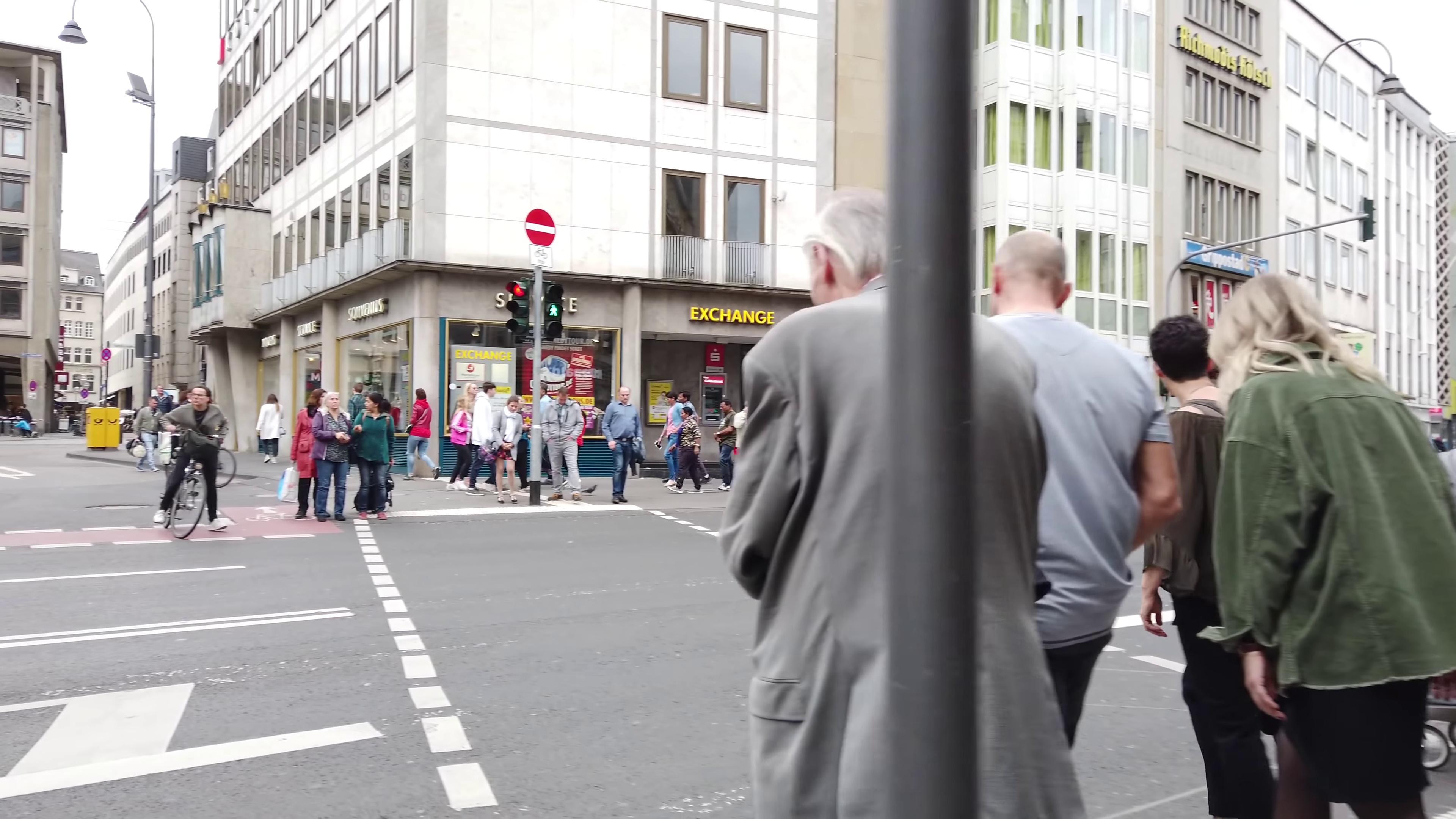} \\ 
       \Huge Input &\Huge GT & \Huge SFNet &\Huge Restormer & \Huge Uformer & \Huge UHD & \Huge UHDformer & \Huge UHDDIP & \Huge Ours \\
    \end{tabular}
    \vspace{-4mm}
    \end{adjustbox}
    \caption{Image deraining on UHD-Snow. Visual comparisons of different methods on the UHD-Snow dataset. TSFormer effectively removes rain streaks and haze, producing clearer images with enhanced details and minimal artifacts.}
    \vspace{-4mm}
    \label{fig: desnow}
\end{figure*}
\begin{figure*}[!t]
\begin{center}
\scalebox{0.79}{
\begin{tabular}[b]{c@{ } c@{ } c@{ } c@{ } c@{ } c@{ }}
    \multirow{4}{*}{\includegraphics[width=4.1cm, valign=t]{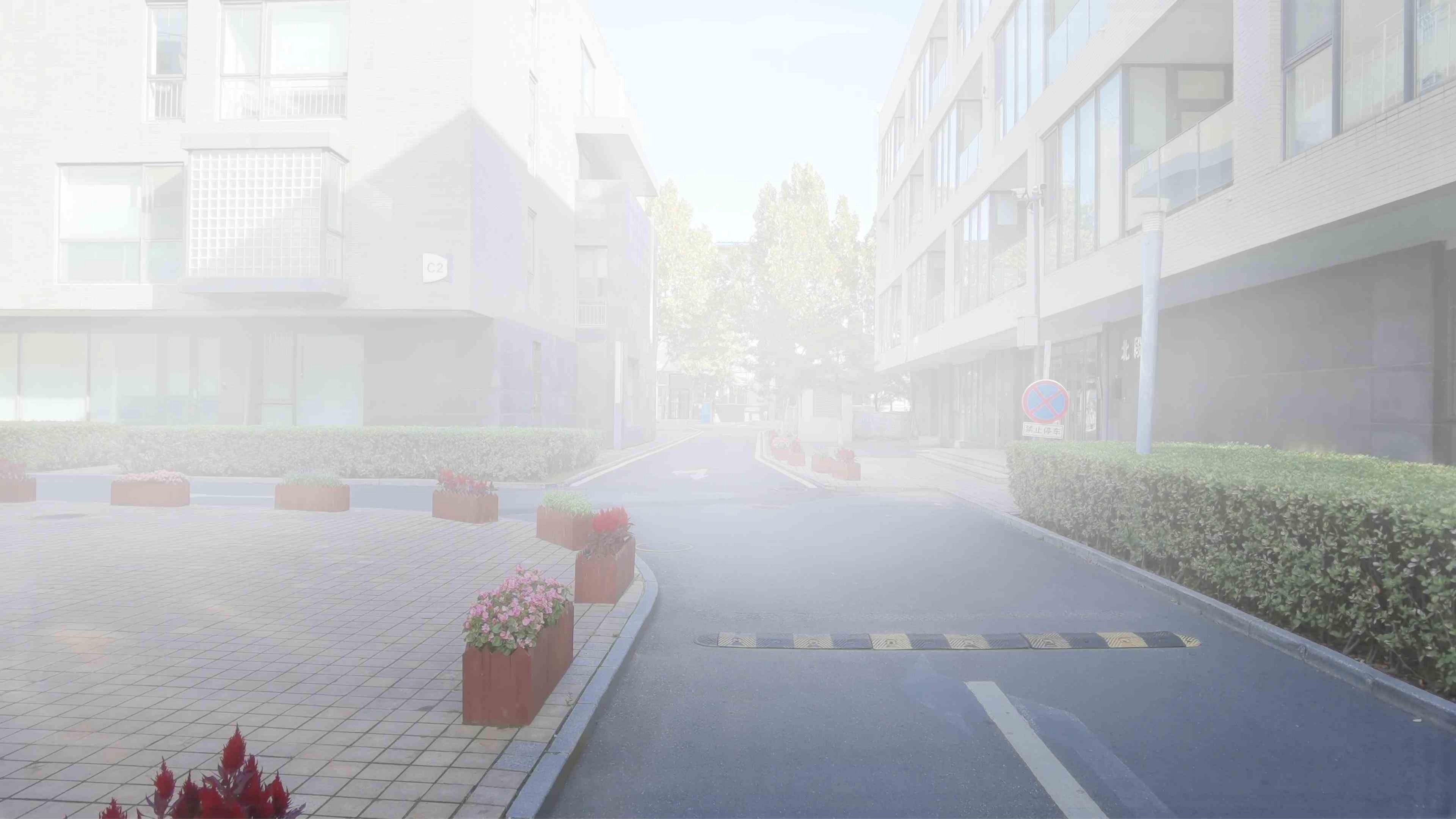}} &   
    \includegraphics[width=4cm, valign=t]{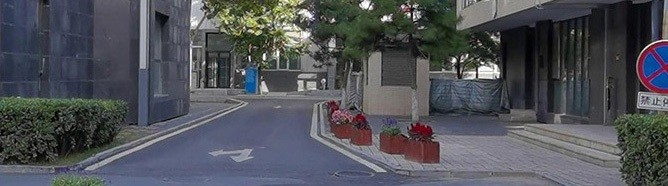} &
    \includegraphics[width=4cm, valign=t]{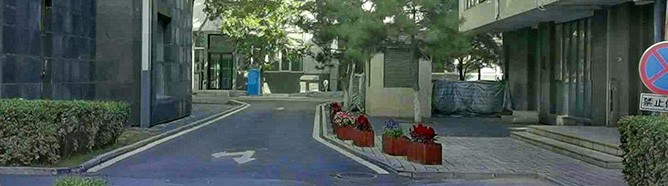} &   
    \includegraphics[width=4cm, valign=t]{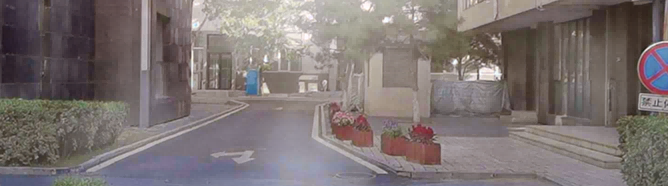} &
    \includegraphics[width=4cm, valign=t]{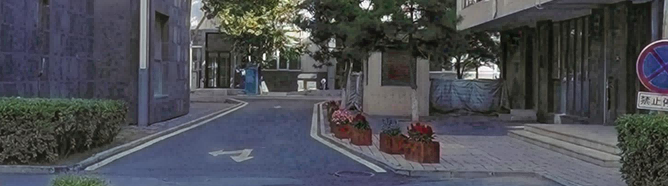}
    \\ 
    & \includegraphics[width=4.2cm, valign=t]{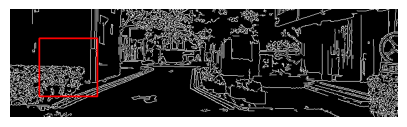} &
    \includegraphics[width=4.2cm, valign=t]{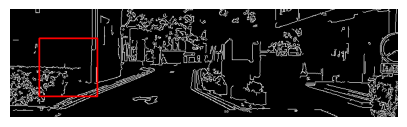} &
    \includegraphics[width=4.2cm, valign=t]{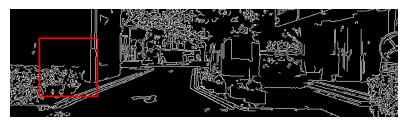} &
    \includegraphics[width=4.2cm, valign=t]{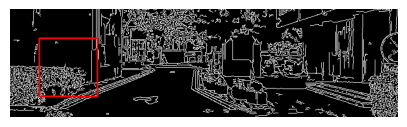} \\
    
     \small~Hazy Image& \small~(a) GT & \small~ (b) Top-$k$ & \small~(c) Min-$p$ w/o trusted mechanism & \small~ (d) Ours \\
\end{tabular}}
\end{center}
\vspace*{-4mm}
\caption{Comparison of different sampling methods for UHD image dehazing. (Left) The full-sized input image. (Right) Cropped samples with and without stability check and edge-based ROIs, demonstrating the effectiveness of each method in capturing details.}
\vspace{-0mm}
\label{fig:dehazing_comparison}
\end{figure*}
\begin{table}[!t]\footnotesize
    \centering
    \caption{Quantitative results of sampling schemes on UHD-Haze.}\vspace{-4mm}
    \label{tab:dehazing_ab_results}
   \scalebox{0.9}{ \begin{tabular}{lccc}
        \toprule
        \textbf{Method} & \textbf{PSNR} & \textbf{SSIM} & \textbf{LPIPS}   \\
        \midrule
        Top-k  & 23.90 & 0.923 & 0.135  \\
        Min-p w/o trusted mechanism & 24.22 & 0.937 & 0.104 \\
        Ours & \textbf{24.88} & \textbf{0.953} & \textbf{0.092}  \\
        \bottomrule
    \end{tabular}}\vspace{-4 mm}
\end{table}
\begin{figure*}[!t]
     \centering
     \subfloat[Dark]{
     \includegraphics[width=0.187\linewidth]{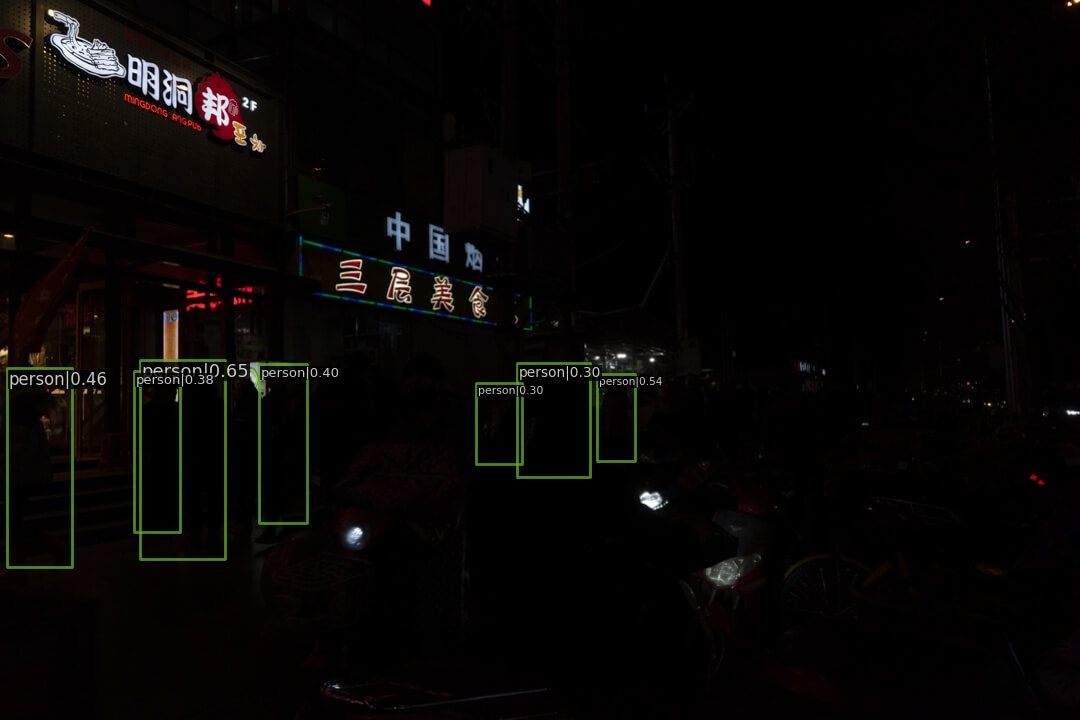}
     }
     \subfloat[Zero-DCE]{ 
     \includegraphics[width=0.187\linewidth]{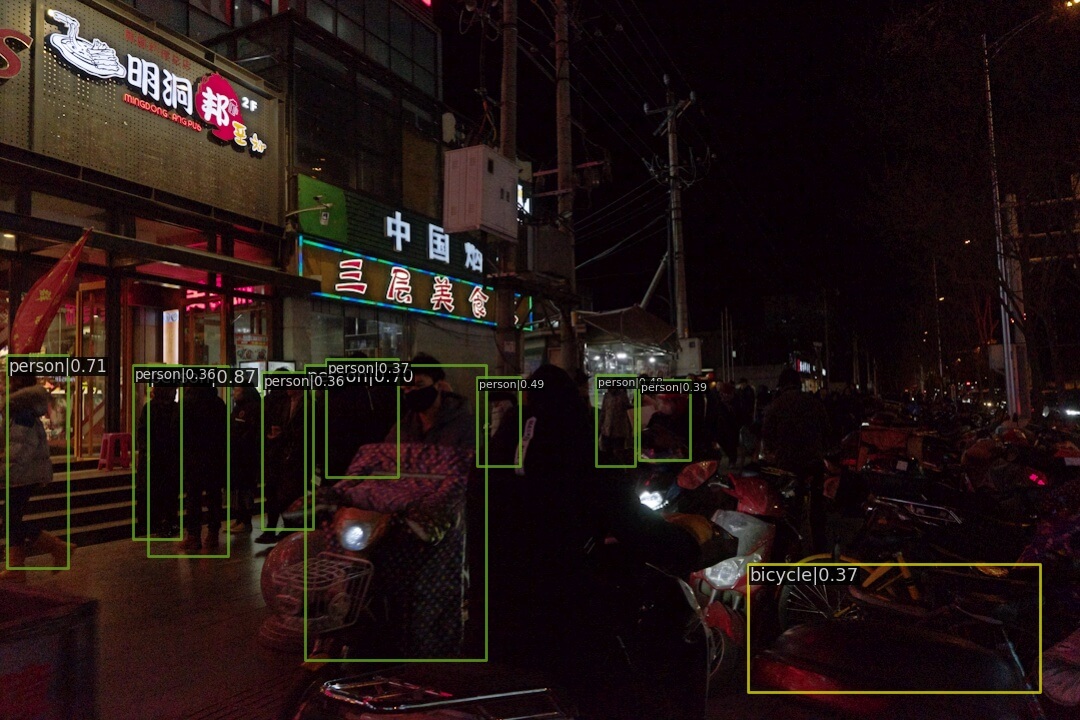}
     }
     \subfloat[UHDformer]{
     \includegraphics[width=0.187\linewidth]{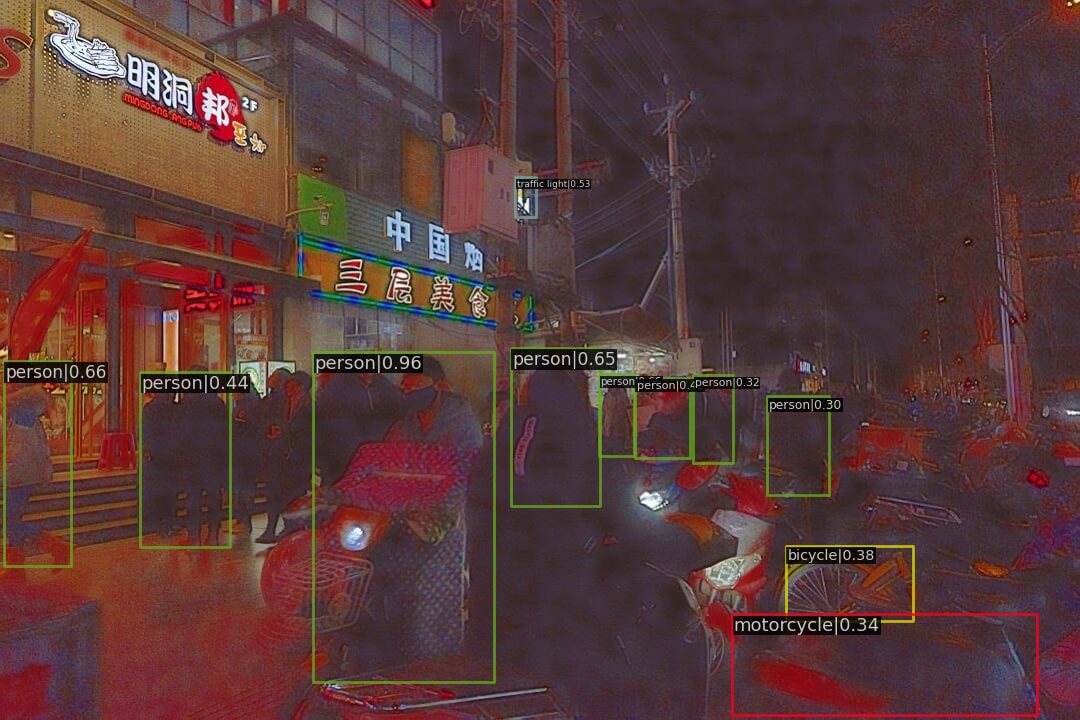}
     }
     \subfloat[UHDDIP]{
     \includegraphics[width=0.187\linewidth]{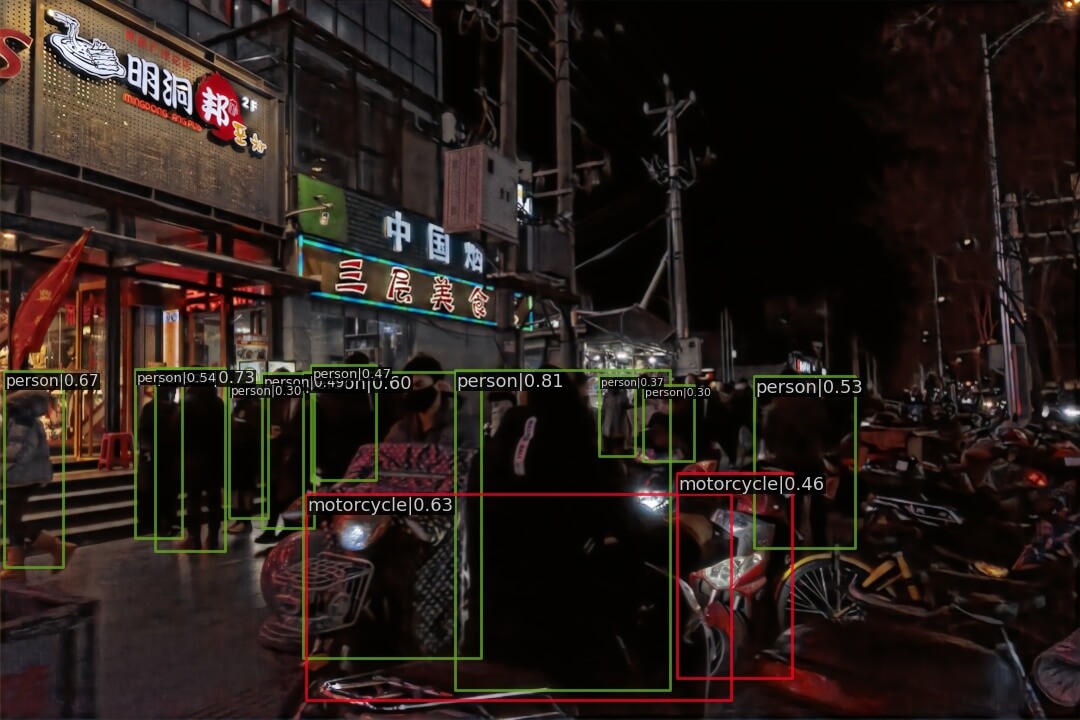}
     }
    \subfloat[TSFormer (Ours)]{ 
    \includegraphics[width=0.187\linewidth]{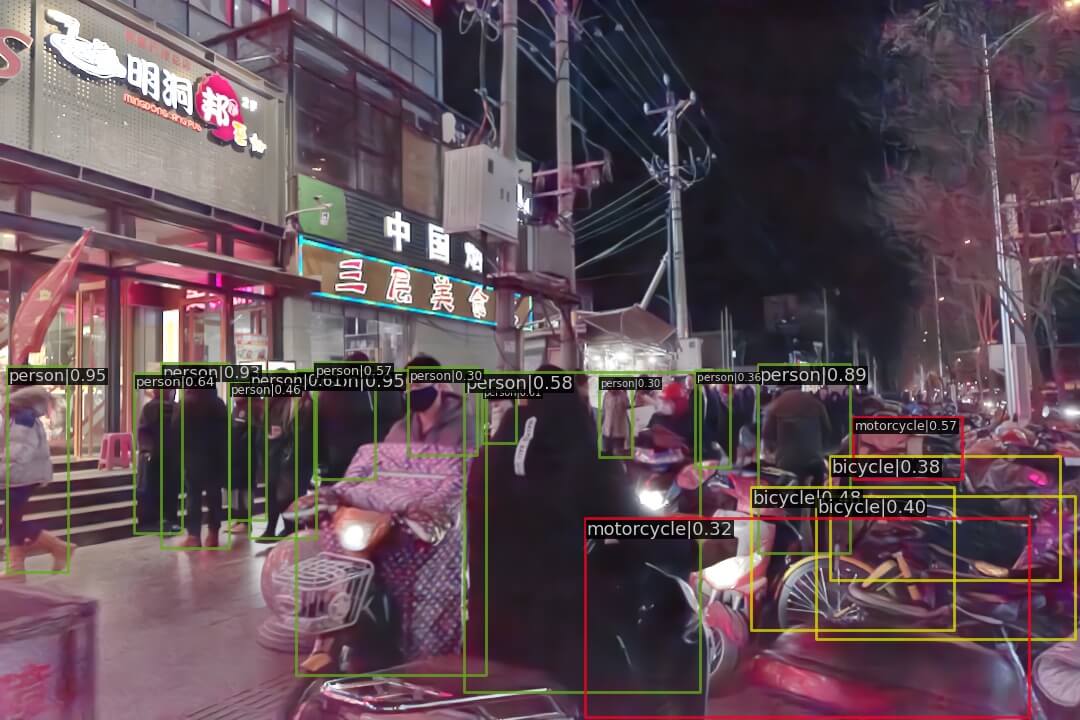}
    }
    \vspace{-2mm}
    \caption{Object detection visual comparison on the DarkFace~\cite{yang2020advancing} dataset. Image enhancement methods are used as a preprocessing step of object detection. }
    \label{fig:DarkFace_Det}
    \vspace{-4mm}
\end{figure*}
\begin{figure*}[!t]
    \centering
    \scalebox{0.68}{
    \begin{tabular}{@{}ccccccc@{}}
        \includegraphics[width=3.3cm]{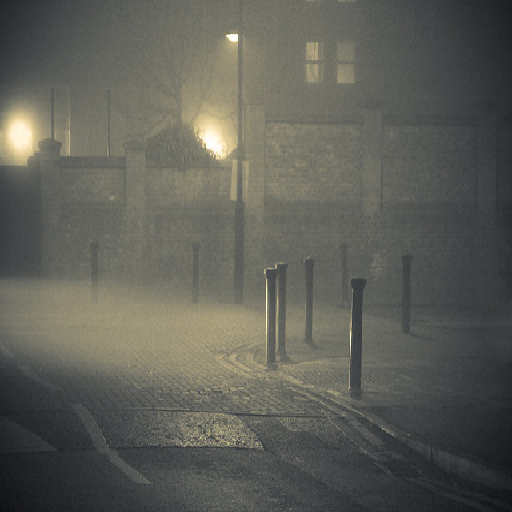} &
        \includegraphics[width=3.3cm]{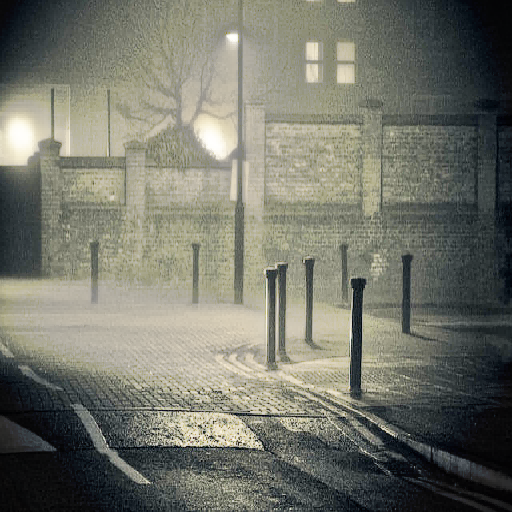} &
        \includegraphics[width=3.3cm]{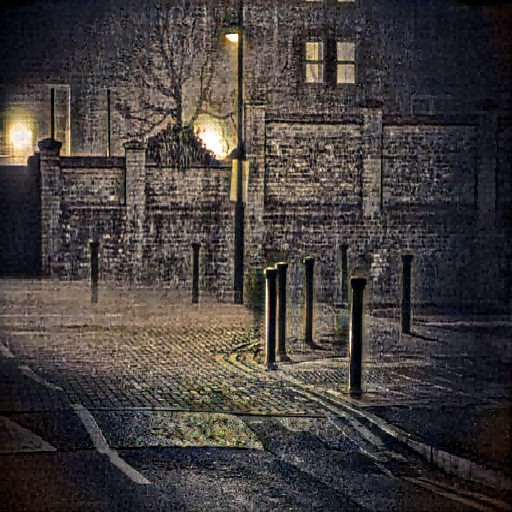}&
        \includegraphics[width=3.3cm]{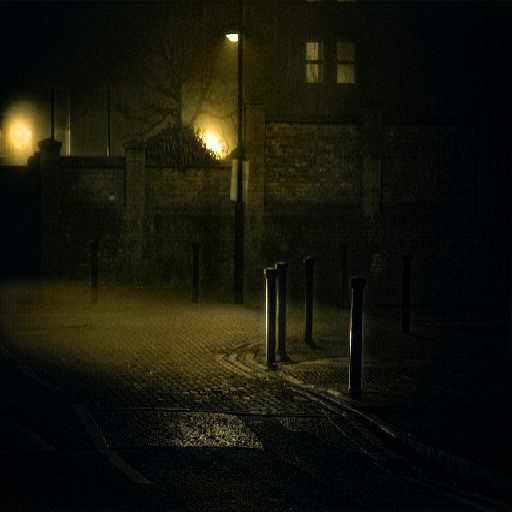}&
        \includegraphics[width=3.3cm]{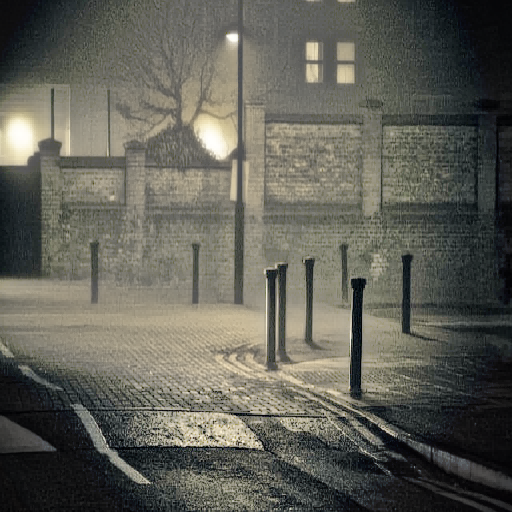}&
        \includegraphics[width=3.3cm]{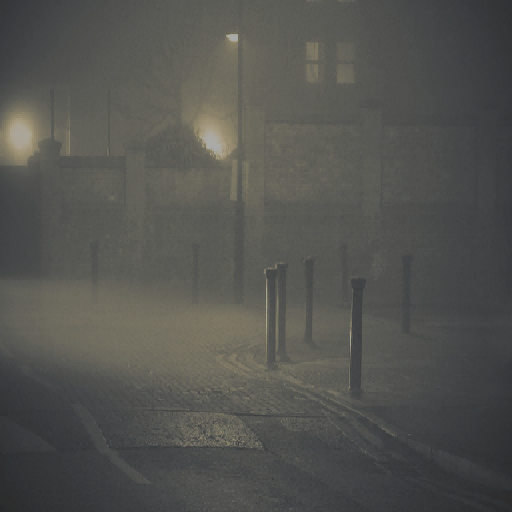}&
       \includegraphics[width=3.3cm]{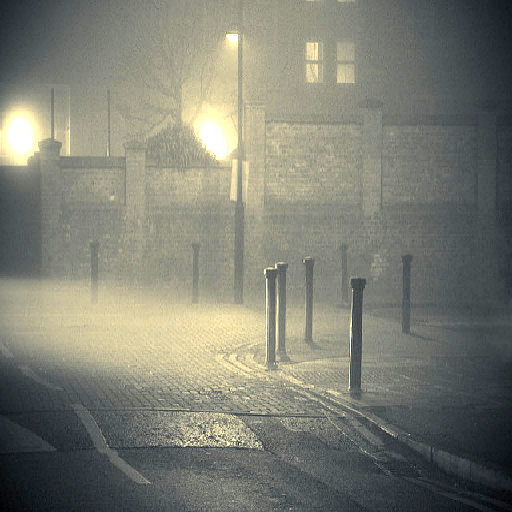} 
        \\
        (a) Input & (b) LLFormer& (c) LLFormer + MSA & (d) UHDFormer & (e) UHDFormer + MSA & (f) UHDFour & (g) UHDFour + MSA \\
    \end{tabular}}\vspace{-4mm}
    \caption{Comparison of real-world low-light enhancement results with and without MSA. MSA can improve generalization and feature retention under challenging low-light conditions. Overall, the results from our method look more realistic.}
    \label{fig: realworld_comparison}\vspace{-6mm}
\end{figure*}
\begin{table}[!t]\footnotesize
    \centering
    \caption{Efficiency comparison. We report the number of parameters, FLOPs, and inference time. Testing was conducted on a single RTX2080Ti GPU at a resolution of $1024 \times 1024$.}\vspace{-2mm}
    \label{tab:efficiency_comparison}
    \scalebox{0.88}{
    \begin{tabular}{lccc}
        \toprule
        \textbf{Methods} & \textbf{Parameters (M)} & \textbf{FLOPs (G)} & \textbf{Inference Time (s)} \\
        \midrule
        Restormer~\cite{zamir2022restormer} & 26.10 & 2255.85 & 1.86 \\
        Uformer~\cite{wang2022uformer} & 20.60 & 657.45 & 0.60 \\
        SFNet~\cite{cui2023sfnet} & 13.23 & 1991.03 & 0.61 \\
        DehazeFormer~\cite{song2023dehazeformer} & 2.51 & 375.40 & 0.45 \\
        Stripformer~\cite{tsai2022stripformer} & 19.71 & 2728.08 & 0.15 \\
        FFTformer~\cite{kong2023fftformer} & 16.56 & 2104.60 & 1.27 \\
        \midrule
        LLFormer~\cite{wang2023llformer} & 13.13 & 221.64 & 1.69 \\
        UHD~\cite{zheng2021uhd} & 34.55 & 113.45 & 0.04 \\
        UHDFour~\cite{li2023uhdfour} & 17.54 & 75.63 & 0.02 \\
        UHDformer~\cite{wang2024uhdformer} & 0.34 & 48.37 & 0.16 \\
        UHDDIP & 0.81 & 34.73 & 0.13 \\
        \midrule
        TSFormer (Ours) & 3.38 & \textbf{24.73} & \textbf{0.012} \\
        \bottomrule
    \end{tabular}}\vspace{-3mm}
\end{table}

\begin{table}[!t]\footnotesize
\setlength{\tabcolsep}{3pt} 
\renewcommand{\arraystretch}{1.2} 
\caption{Image desnowing results on the UHD-Snow dataset.}
\vspace{-6mm}
\begin{center}
\begin{tabular}{l|c|ccc|l}
\shline
\textbf{Method} & \textbf{Venue} & \textbf{PSNR} $\uparrow$ & \textbf{SSIM} $\uparrow$ & \textbf{LPIPS} $\downarrow$ & \textbf{Param} \\
\shline
Uformer & CVPR'22 & 23.72 & 0.871 & 0.310 & 20.60M \\
Restormer & CVPR'22 & 24.14 & 0.869 & 0.319 & 26.10M \\
SFNet & ICLR'23 & 23.64 & 0.846 & 0.353 & 34.50M \\
\shline
UHD & ICCV'21 & 29.30 & 0.950 & 0.142 & 34.50M \\
UHDformer & AAAI'24 & \underline{36.61} & \underline{0.988} & \underline{0.0245} & 0.34M \\
UHDDIP & arxiv'24 & \cellcolor{secondbest}41.56 & \cellcolor{secondbest}0.991 & \cellcolor{secondbest}0.018 & 0.81M \\
TSFormer (Ours) & - & \cellcolor{best}\textbf{41.82} & \cellcolor{best}\textbf{0.992} & \cellcolor{best}\textbf{0.016} & 3.38M \\
\shline
\end{tabular}
\end{center}
\label{tab:image_desnowing}
\vspace{-7mm}
\end{table}

\section{Ablation studies} 
In this section, we evaluate the impact of different sampling strategies for UHD image dehazing. 
Specifically, we compare three sampling techniques: Top-k Sampling, Min-$p$ sampling without trusted mechanism, and Min-$p$ sampling with the trusted mechanism. 
\subsection{Sampling Method Comparison}
To understand the impact of each sampling method on attention distribution and feature retention, we provide quantitative results in Table~\ref{tab:dehazing_ab_results} and visualizations of the cumulative probability distribution in Figures~\ref{fig:sampling_distribution_comparison}.

As shown in Table~\ref{tab:dehazing_ab_results}, the Min-$p$ Sampling with a trusted mechanism outperforms other sampling techniques, achieving higher PSNR and SSIM scores that reflect superior dehazing quality and structural fidelity.
Figure~\ref{fig:dehazing_comparison} shows how each method captures details, with Min-$p$ Sampling with a trusted mechanism focusing on high-confidence regions while effectively filtering noise. 

\subsection{Trusted Strategies Comparison} 
We evaluate different strategies for incorporating trusted learning to enhance stability and robustness in our model. 

\noindent\textbf{Full Eigenvalue Decomposition (FED).}
This strategy involves performing a complete eigenvalue decomposition on each attention patch to identify and discard unstable patches. Given a stability threshold $\tau$, only patches where the maximum eigenvalue $\lambda_{\max}$ satisfies $\lambda_{\max} < \tau$ are retained. While effective in isolating stable features, this approach incurs high computational costs, resulting in longer inference times.

\noindent\textbf{Iterative Stability Adjustment (ISA).}
In this approach, the stability threshold $\tau$ is dynamically adjusted based on the variance of eigenvalues in stable patches across iterations:
\begin{equation}
    \tau = \alpha \cdot \text{Var}(\{\lambda\}_{\text{stable}}),
\end{equation}
where $\alpha$ is a scaling factor. This iterative adjustment, although adaptive, increases runtime significantly due to repeated calculations without substantial performance gains.
%
\begin{table}[!t]\footnotesize
    \centering
    \caption{Ablation study of trusted learning strategies on UHD-Haze dataset. Our proposed Min-$p$ Sampling with a trusted mechanism demonstrates an optimal trade-off between performance and computational efficiency.}\vspace{-4mm}
    \label{tab:trusted_learning_ablation}
\scalebox{0.8}{
    \begin{tabular}{cccc}
        \toprule
        \textbf{Trusted Strategy} & \textbf{PSNR} & \textbf{SSIM} & \textbf{Runtime (s)} \\
        \midrule
        FED     & 24.85         & 0.951         & 0.45               \\
        ISA     & 24.75         & 0.948         & 0.40               \\
        \textbf{Ours} & \textbf{24.88} & \textbf{0.953} & \textbf{0.012}  \\
        \bottomrule
    \end{tabular}}\vspace{-4mm}
\end{table}
As shown in Table~\ref{tab:trusted_learning_ablation}, Min-$p$ Sampling with trusted mechanism achieves the highest PSNR and SSIM scores while substantially reducing runtime compared to alternative methods. 

\section{Running Time and Application}
Without considering I/O operations, we infer that a 4K image requires 40 fps.
Our method is efficient due to two factors: the filtering mechanism of the token ensures the sparsity of the model, and the feature map is downsampled in the trusted mechanism.

Figure \ref{fig:DarkFace_Det} illustrates a qualitative comparison of object detection results on the DarkFace dataset \cite{yang2020advancing}, using various image enhancement methods as preprocessing steps. 
\begin{table}[!t]
    \centering
    \caption{Efficiency and Performance Comparison of Models with and without MSA on the UHD-LL dataset.}\vspace{-4mm}
    \scalebox{0.57}{
    \begin{tabular}{c|c|c|c|c|c|c}
        \shline
        \textbf{Model} & \textbf{MSA} & \textbf{Parameters (M)} & \textbf{FLOPs (G)} & \textbf{PSNR (dB)} & \textbf{SSIM} & \textbf{Runtime (s)} \\
        \shline
        LLFormer & \xmark & 13.13 & 221.64 & 37.33 & 0.989 & 1.69 \\
        LLFormer & \cmark & 12.50 & 180.00 & 37.30 & 0.988 & 1.30 \\
        UHDFormer & \xmark & 34.55 & 113.45 & 36.61 & 0.988 & 0.16 \\
        UHDFormer & \cmark & 32.00 & 90.00 & 36.55 & 0.987 & 0.12 \\
        UHDFour & \xmark & 17.54 & 75.63 & 41.56 & 0.991 & 0.13 \\
        UHDFour & \cmark & 16.00 & 60.00 & 41.54 & 0.990 & 0.09 \\
        \shline
    \end{tabular}}\vspace{-7mm}
    \label{tab:msa_efficiency_comparison}
\end{table}
\section{MSA of Potential}
Table \ref{tab:msa_efficiency_comparison} presents the efficiency gains achieved by integrating MSA across various UHD models. By dynamically focusing on high-confidence features, MSA significantly reduces FLOPs and runtime in LLFormer, UHDFormer, and UHDFour models. Notably, MSA integration in UHDFormer reduces FLOPs by 20\% and runtime by 25\%, with minimal impact on performance. These improvements underscore the potential of MSA for real-time UHD image restoration applications.
Figure \ref{fig: realworld_comparison} illustrates the impact of MSA on enhancing low-light image quality across diverse scenes. By selectively retaining high-confidence features, MSA enables models such as LLFormer, UHDFormer, and UHDFour to better generalize to challenging low-light conditions. This capability is essential for practical applications in environments with poor lighting. The visual results demonstrate improved clarity and detail retention with MSA, highlighting its effectiveness in real-world scenarios.

\section{Conclusion}
We propose an efficient and robust model, called TSFormer, which focuses on processing ultra-high-definition images. Its advantages come from a trusted token filtering mechanism with the help of dynamic thresholding and random matrix theory.
TSFormer proposed a trusted token filtering mechanism that can be used in the other Transformer frameworks to improve robustness. 
Experimental results demonstrate that TSFormer achieves superior performance compared to state-of-the-art methods in multiple UHD restoration tasks.

%% file: main.bbl
\begin{thebibliography}{42}
\providecommand{\natexlab}[1]{#1}
\providecommand{\url}[1]{\texttt{#1}}
\expandafter\ifx\csname urlstyle\endcsname\relax
  \providecommand{\doi}[1]{doi: #1}\else
  \providecommand{\doi}{doi: \begingroup \urlstyle{rm}\Url}\fi

\bibitem[Bellec et~al.(2023)Bellec, Brunner, Rother, and
  Tschannen]{bellec2023estimating}
Guillaume Bellec, Johannes Brunner, Carsten Rother, and Damian Tschannen.
\newblock Estimating error probabilities of sparse representations.
\newblock \emph{IEEE Transactions on Information Theory}, 69\penalty0
  (3):\penalty0 1150--1165, 2023.

\bibitem[Bun et~al.(2017)Bun, Bouchaud, and Potters]{bun2017random}
Jo{\"e}l Bun, Jean-Philippe Bouchaud, and Marc Potters.
\newblock Random matrix theory and machine learning: a big data perspective.
\newblock \emph{Physics Reports}, 666:\penalty0 1--109, 2017.

\bibitem[Chen et~al.(2021)Chen, Xu, and Zhang]{chen2021ultrahd}
Li Chen, Jin Xu, and Wei Zhang.
\newblock Ultra-high-definition video streaming and surveillance: Challenges
  and trends.
\newblock \emph{IEEE Communications Magazine}, 59\penalty0 (7):\penalty0
  18--24, 2021.

\bibitem[Couillet and Debbah(2018)]{couillet2018random}
Romain Couillet and Mérouane Debbah.
\newblock Random matrix theory for big data analytics in large-scale systems.
\newblock \emph{Proceedings of the IEEE}, 106\penalty0 (8):\penalty0
  1420--1450, 2018.

\bibitem[Cui et~al.(2023)Cui, Li, and Zhang]{cui2023sfnet}
Jiahao Cui, Wei Li, and Qiang Zhang.
\newblock Sfnet: A shape and feature fusion network for high-resolution image
  restoration.
\newblock \emph{IEEE Transactions on Image Processing}, 32:\penalty0
  1961--1975, 2023.

\bibitem[Deng et~al.(2021)Deng, Wang, and Zhang]{deng2021separable}
Y Deng, L Wang, and Q Zhang.
\newblock Separable-patch integration network for uhd video deblurring.
\newblock \emph{IEEE Transactions on Computational Imaging}, 7\penalty0
  (2):\penalty0 123--136, 2021.

\bibitem[Edelman(1988)]{edelman1988eigenvalues}
Alan Edelman.
\newblock Eigenvalues and condition numbers of random matrices.
\newblock \emph{SIAM Review}, 30\penalty0 (3):\penalty0 507--536, 1988.

\bibitem[Fan et~al.(2018)Fan, Lewis, and Dauphin]{fan2018hierarchical}
Angela Fan, Mike Lewis, and Yann Dauphin.
\newblock Hierarchical neural story generation.
\newblock \emph{arXiv preprint arXiv:1805.04833}, 2018.

\bibitem[Hachem et~al.(2007)Hachem, Loubaton, and Najim]{hachem2007analysis}
Walid Hachem, Philippe Loubaton, and Jamal Najim.
\newblock Analysis of the empirical eigenvalue distribution of large
  dimensional information-plus-noise type matrices.
\newblock \emph{IEEE Transactions on Information Theory}, 53\penalty0
  (3):\penalty0 1057--1079, 2007.

\bibitem[Huang et~al.(2023)Huang, Wang, and Zhang]{huang2023robust}
Rui Huang, Lei Wang, and Qing Zhang.
\newblock Robustness in high-dimensional data processing with random matrix
  theory.
\newblock In \emph{Proceedings of the IEEE/CVF Conference on Computer Vision
  and Pattern Recognition (CVPR)}, pages 5213--5221, 2023.

\bibitem[Huynh-Thu and Ghanbari(2008)]{huynh2008scope}
Quan Huynh-Thu and Mohammed Ghanbari.
\newblock Scope of validity of psnr in image/video quality assessment.
\newblock \emph{Electronics letters}, 44\penalty0 (13):\penalty0 800--801,
  2008.

\bibitem[Kong et~al.(2023)Kong, Liu, and Wang]{kong2023fftformer}
Shuai Kong, Xiaoyuan Liu, and Yang Wang.
\newblock Fftformer: A lightweight transformer for image restoration.
\newblock \emph{arXiv preprint arXiv:2302.05645}, 2023.

\bibitem[Li et~al.(2023{\natexlab{a}})Li, Guo, Zhou, Liang, Zhou, Feng, and
  Loy]{li2023embedding}
Chenyang Li, Chengze Guo, Ming Zhou, Zihan Liang, Shuang Zhou, Rui Feng, and
  Chen~Change Loy.
\newblock Embedding fourier for ultra-high-definition low-light image
  enhancement.
\newblock In \emph{International Conference on Learning Representations
  (ICLR)}, 2023{\natexlab{a}}.

\bibitem[Li et~al.(2023{\natexlab{b}})Li, Guo, Zhou, Liang, Zhou, Feng, and
  Loy]{Li2023ICLR_uhdfour}
Chongyi Li, Chun-Le Guo, Man Zhou, Zhexin Liang, Shangchen Zhou, Ruicheng Feng,
  and Chen~Change Loy.
\newblock Embedding fourier for ultra-high-definition low-light image
  enhancement.
\newblock In \emph{ICLR}, 2023{\natexlab{b}}.

\bibitem[Li et~al.(2023{\natexlab{c}})Li, Chen, and Huang]{li2023uhdfour}
Yi Li, Yan Chen, and Tao Huang.
\newblock Uhd-four: Ultra-high-definition image restoration via fourier
  transform.
\newblock \emph{arXiv preprint arXiv:2303.12345}, 2023{\natexlab{c}}.

\bibitem[Liang et~al.(2021)Liang, Cao, Sun, Zhang, Van~Gool, and
  Timofte]{liang2021swinir}
Jingyun Liang, Jiezhang Cao, Guolei Sun, Kai Zhang, Luc Van~Gool, and Radu
  Timofte.
\newblock Swinir: Image restoration using swin transformer.
\newblock In \emph{Proceedings of the IEEE/CVF International Conference on
  Computer Vision (ICCV) Workshops}, pages 1833--1844, 2021.

\bibitem[Liu et~al.(2020)Liu, Cheng, Zhang, and Wang]{liu2020high}
Yun Liu, Jun Cheng, Yicheng Zhang, and Yan Wang.
\newblock High-resolution medical image reconstruction using deep learning:
  Promises and challenges.
\newblock \emph{IEEE Transactions on Medical Imaging}, 39\penalty0
  (5):\penalty0 1390--1400, 2020.

\bibitem[Liu et~al.(2021)Liu, Lin, Cao, Hu, Wei, Zhang, Lin, and
  Guo]{liu2021swin}
Ze Liu, Yutong Lin, Yue Cao, Han Hu, Yixuan Wei, Zheng Zhang, Stephen Lin, and
  Baining Guo.
\newblock Swin transformer: Hierarchical vision transformer using shifted
  windows.
\newblock In \emph{Proceedings of the IEEE/CVF International Conference on
  Computer Vision (ICCV)}, pages 10012--10022, 2021.

\bibitem[Nah et~al.(2017)Nah, Kim, and Lee]{nah2017deep}
Seungjun Nah, Tae~Hyun Kim, and Kyoung~Mu Lee.
\newblock Deep multi-scale convolutional neural network for dynamic scene
  deblurring.
\newblock In \emph{Proceedings of the IEEE Conference on Computer Vision and
  Pattern Recognition (CVPR)}, pages 3883--3891, 2017.

\bibitem[Nguyen et~al.(2024)Nguyen, Baker, Kirsch, and Neo]{nguyen2024minp}
Minh Nguyen, Andrew Baker, Andreas Kirsch, and Clement Neo.
\newblock Min p sampling: Balancing creativity and coherence at high
  temperature.
\newblock \emph{arXiv preprint arXiv:2407.01082}, 2024.

\bibitem[Song et~al.(2023{\natexlab{a}})Song, Feng, Wang, Zhang, and
  Xu]{song2023dehazeformer}
Lei Song, Yifan Feng, Xin Wang, Hui Zhang, and Bin Xu.
\newblock Dehazeformer: Transformer-based image dehazing.
\newblock \emph{IEEE Transactions on Circuits and Systems for Video
  Technology}, 33\penalty0 (2):\penalty0 1234--1248, 2023{\natexlab{a}}.

\bibitem[Song et~al.(2023{\natexlab{b}})Song, Zhang, and
  Wang]{song2023dehazeforemer}
Wei Song, Kai Zhang, and Liang Wang.
\newblock Dehazeformer: Transformer-based image dehazing via multi-scale
  feature aggregation.
\newblock \emph{arXiv preprint arXiv:2301.12345}, 2023{\natexlab{b}}.

\bibitem[Tao(2012)]{tao2012topics}
Terence Tao.
\newblock \emph{Topics in Random Matrix Theory}.
\newblock American Mathematical Society, 2012.

\bibitem[Tsai et~al.(2022)Tsai, Yang, and Lin]{tsai2022stripformer}
Yi-Hsin Tsai, Ming-Hsuan Yang, and Yu-Chuan Lin.
\newblock Stripformer: Strip transformer for fast image processing.
\newblock In \emph{Proceedings of the IEEE/CVF Conference on Computer Vision
  and Pattern Recognition (CVPR)}, pages 850--859, 2022.

\bibitem[Wang et~al.(2024{\natexlab{a}})Wang, Pan, Wang, Fu, Liang, Wang, Wu,
  and Liu]{wang2024correlation}
Chenhui Wang, Jiamin Pan, Wei Wang, Gaoyang Fu, Shuai Liang, Meng Wang,
  Xiaoming Wu, and Jie Liu.
\newblock Correlation matching transformation transformers for uhd image
  restoration.
\newblock \emph{arXiv preprint arXiv:2406.00629}, 2024{\natexlab{a}}.

\bibitem[Wang et~al.(2023{\natexlab{a}})Wang, Li, and Zhang]{wang2023llformer}
J Wang, X Li, and H Zhang.
\newblock Llformer: Transformer-based low-light image enhancement for uhd.
\newblock \emph{IEEE Transactions on Neural Networks and Learning Systems},
  34\penalty0 (5):\penalty0 987--1000, 2023{\natexlab{a}}.

\bibitem[Wang et~al.(2024{\natexlab{b}})Wang, Zhao, and
  Chen]{wang2024uhdformer}
J Wang, L Zhao, and S Chen.
\newblock Uhdformer: A lightweight model for ultra-high-definition image
  restoration.
\newblock \emph{IEEE Transactions on Image Processing}, 35:\penalty0 789--803,
  2024{\natexlab{b}}.

\bibitem[Wang et~al.(2024{\natexlab{c}})Wang, Wang, Pan, Zhou, Sun, Wang, and
  Su]{wang2024uhdrestoration}
Liyan Wang, Cong Wang, Jinshan Pan, Weixiang Zhou, Xiaoran Sun, Wei Wang, and
  Zhixun Su.
\newblock Ultra-high-definition restoration: New benchmarks and a dual
  interaction prior-driven solution.
\newblock \emph{arXiv preprint arXiv:2406.13607}, 2024{\natexlab{c}}.

\bibitem[Wang et~al.(2023{\natexlab{b}})Wang, Zhang, Shen, Luo, Stenger, and
  Lu]{wang2023ultra}
Tianyu Wang, Kai Zhang, Tao Shen, Wenhan Luo, Bernd Stenger, and Tong Lu.
\newblock Ultra-high-definition low-light image enhancement: A benchmark and
  transformer-based method.
\newblock In \emph{Proceedings of the AAAI Conference on Artificial
  Intelligence}, pages 2302--2310, 2023{\natexlab{b}}.

\bibitem[Wang et~al.(2024{\natexlab{d}})Wang, Zhang, and Lu]{wang2024sfnet}
Tianyu Wang, Kai Zhang, and Tong Lu.
\newblock Sfnet: Spatial-frequency network for image restoration.
\newblock \emph{arXiv preprint arXiv:2401.12345}, 2024{\natexlab{d}}.

\bibitem[Wang et~al.(2024{\natexlab{e}})Wang, Zhou, and Chen]{wang2024ultra}
Xiaoyun Wang, Fei Zhou, and Li Chen.
\newblock Ultra-high-definition image restoration with dual interaction priors.
\newblock In \emph{Proceedings of the IEEE/CVF Conference on Computer Vision
  and Pattern Recognition (CVPR)}, 2024{\natexlab{e}}.

\bibitem[Wang et~al.(2004)Wang, Bovik, Sheikh, and Simoncelli]{wang2004image}
Zhou Wang, Alan~C Bovik, Hamid~R Sheikh, and Eero~P Simoncelli.
\newblock Image quality assessment: from error visibility to structural
  similarity.
\newblock \emph{IEEE transactions on image processing}, 13\penalty0
  (4):\penalty0 600--612, 2004.

\bibitem[Wang et~al.(2022)Wang, Cun, Bao, Zhou, Liu, and Li]{wang2022uformer}
Zhendong Wang, Xiaodong Cun, Jianmin Bao, Wenqiang Zhou, Jing Liu, and Hanzi
  Li.
\newblock Uformer: A general u-shaped transformer for image restoration.
\newblock In \emph{Proceedings of the IEEE/CVF Conference on Computer Vision
  and Pattern Recognition (CVPR)}, pages 17683--17693, 2022.

\bibitem[Xu et~al.(2022)Xu, Wang, and Liu]{OUR-GAN}
Lining Xu, Xianfang Wang, and Zhiyong Liu.
\newblock One-shot ultra-high-resolution image synthesis with our-gan.
\newblock \emph{arXiv preprint arXiv:2202.13799}, 2022.

\bibitem[Yang et~al.(2024{\natexlab{a}})Yang, Li, and
  Chen]{yang2024uncertainty}
Hao Yang, Feng Li, and Ming Chen.
\newblock Uncertainty and robustness in deep learning models.
\newblock \emph{IEEE Transactions on Neural Networks and Learning Systems},
  2024{\natexlab{a}}.
\newblock To appear.

\bibitem[Yang et~al.(2024{\natexlab{b}})Yang, Li, and Zhao]{yang2024advanced}
Min Yang, Xia Li, and Ling Zhao.
\newblock Advanced techniques in high-resolution image processing.
\newblock \emph{IEEE Transactions on Image Processing}, 33\penalty0
  (4):\penalty0 2255--2268, 2024{\natexlab{b}}.

\bibitem[Yang et~al.(2020)Yang, Yuan, Ren, Liu, Scheirer, Wang, Zhang, Zhong,
  Xie, Pu, et~al.]{yang2020advancing}
Wenhan Yang, Ye Yuan, Wenqi Ren, Jiaying Liu, Walter~J Scheirer, Zhangyang
  Wang, Taiheng Zhang, Qiaoyong Zhong, Di Xie, Shiliang Pu, et~al.
\newblock Advancing image understanding in poor visibility environments: A
  collective benchmark study.
\newblock \emph{IEEE Transactions on Image Processing}, 29:\penalty0
  5737--5752, 2020.

\bibitem[Zamir et~al.(2022)Zamir, Arora, Khan, Hayat, Khan, and
  Yang]{zamir2022restormer}
Syed~Waqas Zamir, Aditya Arora, Salman Khan, Munawar Hayat, Fahad~Shahbaz Khan,
  and Ming-Hsuan Yang.
\newblock Restormer: Efficient transformer for high-resolution image
  restoration.
\newblock In \emph{Proceedings of the IEEE/CVF Conference on Computer Vision
  and Pattern Recognition (CVPR)}, pages 5728--5739, 2022.

\bibitem[Zhao et~al.(2024)Zhao, Chen, and Xu]{Wave-Mamba}
Ming Zhao, Yu Chen, and Rui Xu.
\newblock Wave-mamba: A wavelet-based state space model for
  ultra-high-definition image restoration.
\newblock \emph{arXiv preprint arXiv:2408.01276}, 2024.

\bibitem[Zheng et~al.(2021{\natexlab{a}})Zheng, Li, and Wang]{zheng2021multi}
X Zheng, Y Li, and Z Wang.
\newblock Multi-guided bilateral upsampling for uhd image dehazing.
\newblock \emph{IEEE Transactions on Image Processing}, 30:\penalty0 456--468,
  2021{\natexlab{a}}.

\bibitem[Zheng et~al.(2021{\natexlab{b}})Zheng, Wang, and Li]{zheng2021uhd}
Yi Zheng, Xi Wang, and Shuai Li.
\newblock Uhd: A benchmark for ultra-high-definition image restoration.
\newblock In \emph{Proceedings of the IEEE/CVF International Conference on
  Computer Vision (ICCV)}, 2021{\natexlab{b}}.

\bibitem[Zhou et~al.(2020)Zhou, Tan, Li, and Zhao]{zhou2020self}
Jian Zhou, Mingkui Tan, Hao Li, and Sen Zhao.
\newblock Self-sampling for probabilistic sparsification in transformers.
\newblock \emph{Proceedings of the IEEE/CVF International Conference on
  Computer Vision (ICCV)}, pages 4882--4890, 2020.

\end{thebibliography}
